\documentclass[doutorado,pos-defesa]{ppgee}

\usepackage{rotating}           
\usepackage[all,knot,arc,import,poly]{xy}   
\usepackage{pgfplots}
\pgfplotsset{compat=1.18}
\usepackage{tikz}
\usetikzlibrary{external}
\usepackage{import}
\usepackage{multirow}
\usepackage{tabularx}
\usepackage{pifont}
\usepackage[section,below]{placeins}

\usepackage[subpreambles=true]{standalone}
\usepackage{subfiles} 

\newcommand{\spi}[1]{\textsuperscript{\textit{#1}}} 

\tituloPT{Percepção Visual para Veículos Autônomos em Ambiente Fora de Estrada usando Aprendizado Profundo}
\tituloEN{Visual-Based Perception for Autonomous Vehicles in Off-Road Environment Using Deep Learning}
\autor[Alves, N.]{Nelson Alves Ferreira Neto}
\genero{M} 
\orientador[Orientador]{Prof.\ Dr.}{Eduardo Furtado de Simas Filho}
\coorientador{Prof.\ Dr.}{Wagner Luiz Alves de Oliveira}
\curso{EE}
\data{11}{07}{2022} 
\idioma{EN} 

\textoresumo[brazil]{
    Sistemas inteligentes de alto desempenho e com baixa latência computacional são necessários para condução autônoma em terrenos não uniformes comumente encontrados em países em desenvolvimento e minas a céu aberto.\@ Para ajudar a diminuir a lacuna nesse tipo de aplicação, este trabalho propõe um sistema de percepção para veículos autônomos e assistência avançada ao motorista especializado em estradas não pavimentadas e ambientes \textit{off-road} capazes de navegar em terrenos acidentados sem trilha pré-definida.\@ Como parte deste sistema, o \textit{framework Configurable Modular Segmentation Network} (CMSNet) foi proposto para facilitar a criação de diferentes arranjos de arquiteturas.\@ Algumas configurações do CMSNet foram portadas e treinadas para segmentar obstáculos e terrenos transitáveis em uma nova coleção de imagens de estradas não pavimentadas e cenários \textit{off-road} contendo condições adversas como noite, chuva e poeira.\@ Também foi realizada: uma investigação sobre a viabilidade de aplicação de \textit{deep learning} para detectar regiões por onde o veículo pode passar quando não há limite de pista explícito; um estudo de como o algoritmos de segmentação propostos se comportam em diferentes níveis de comprometimento da visibilidade; e uma avaliação de testes de campo realizados com arquiteturas de segmentação semântica aprimoradas para inferência em tempo real.\@ Um novo \textit{dataset} (chamado Kamino) foi apresentado tendo quase 12.000 novas imagens coletadas de um veículo operado com vários sensores, incluindo oito câmeras capturando sequências sincronizadas de diferentes pontos de vista.\@ O conjunto de dados tem um número alto de pixels rotulados em comparação com \textit{datasets} semelhantes disponíveis publicamente.\@ Inclui imagens coletadas de um campo de provas \textit{off-road} montado exclusivamente para testar o sistema que emula um cenário de mina a céu aberto sob diferentes condições adversas de visibilidade.\@ Para alcançar inferência em tempo real no sistema embarcado e permitir testes de campo, muitas camadas das redes geradas pela CMSNet foram removidas metodicamente e fundidas usando TensorRT, C++ e CUDA.\@ Experimentos empíricos em dois conjuntos de dados indicaram um desempenho satisfatório do sistema proposto.}{Percepção visual, \textit{Deep learning} (DL), Redes Neurais Artificiais, Veículos Autônomos, Redes Neurais Convolucionais (CNN), Inteligência Artificial (IA), ADAS, Segmentação em Tempo Real, Ambiente \textit{Off-Road}}

\textoresumo[english]{
     \textit{High-performance intelligent systems with low computational latency are required for autonomous driving on non-uniform terrain commonly found in open-pit mines and developing countries.\@ This work proposes a perception system for autonomous vehicles and advanced driver assistance specialized on unpaved roads and off-road environments capable of navigating by rough terrain without a predefined trail to help narrow the gap in this kind of application.\@ As part of this system, the Configurable Modular Segmentation Network (CMSNet) framework has been proposed facilitating the creation of different architectures arrangements.\@ Some CMSNet configurations has been ported and trained to segment obstacles and trafficable ground on a new collection of images from unpaved roads and off-road scenarios containing adverse conditions such as night, rain, and dust.\@ It was also performed: an investigation regarding the feasibility of applying deep learning to detect regions where the vehicle can pass through when there is no explicit track boundary; a study of how our proposed segmentation algorithms behave in different severity levels of visibility impairment; and an evaluation of field tests carried out with semantic segmentation architectures prepared for real-time inference.\@ A new dataset (named Kamino) has been presented.\@ It has almost 12,000 new images collected from an operated vehicle with various sensors, including eight cameras capturing synchronized sequences from different points of view.\@ The Kamino dataset has high number of labeled pixels compared to similar publicly available collections.\@ It includes images collected from an off-road proving ground exclusively assembled for testing the system that emulates an open-pit mine scenario under different adverse visibility conditions.\@ To achieve embedded real-time inference and allows field tests, many layers of the CMSNet CNN networks were methodically removed and fused using TensorRT, C++, and CUDA.\@ Empirical experiments on two datasets validated the effectiveness of the proposed system.\@ 
}
    }{\textit{Visual-based perception, Deep learning (DL), Artificial Neural Network (ANN), Autonomous Vehicle (AV), Convolutional Neural Network (CNN), Artificial Intelligence (AI), Advanced Driver-Assistance System (ADAS), Real-time Segmentation, Off-Road Environment}}

\incluifichacatalografica{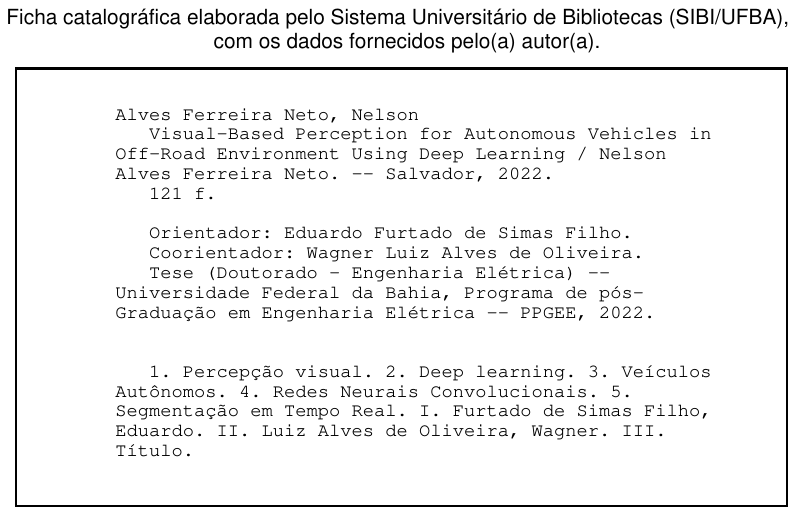}
\textofolhadeaprovacaoassinada{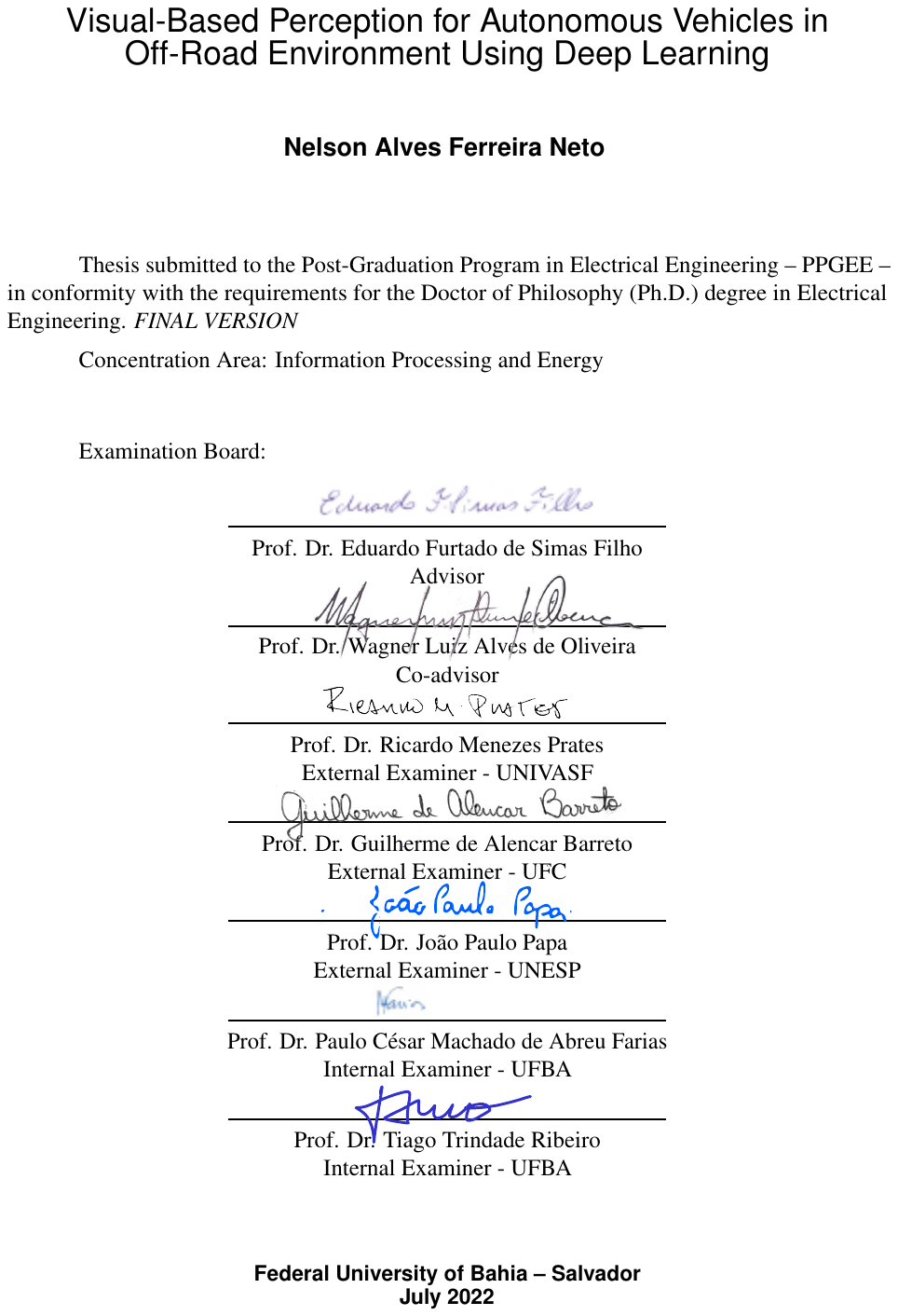}

\textodedicatoria*{dedicatoria}
\textoagradecimentos*{agradecimentos}
\textoepigrafe*{epigrafe}

\incluilistadefiguras%

\incluilistadetabelas%

\incluilistadecodigos%

\incluilistadesiglas%

\incluilistadesimbolos%

\begin{document}

\textual%

\chapter{Introduction}%
\label{chapter:introduction}

The searching for more comfortable and safe cars has guided the development of the automotive industry. Modern vehicles have incorporated various driving assistance features with the ultimate goal of being autonomous, providing safety and comfort to passengers. Even in the past decade, the Grand~\cite{darpa:2005:grad_challenge} and Urban~\cite{darpa:2007:urban_challenge} challenge competitions, promoted by the \sigla{DARPA}{Defense Advanced Research Projects Agency} and won by~\citeonline{thrun:2006:stanley_darpa_grand_won} and~\citeonline{urmson:2008:boss_darpa_grand_won}, respectively, showed that it was possible to build cars with skills to perceive the environment around them and navigate autonomously. In addition, other challenges such as \sigla{AVC}{Autonomous Vehicle Competition} also helped foster research that developed methodologies for autonomous cars based on modularized architectures for distributed systems aimed at reducing computational complexity and fault tolerance~\cite{jo:2014:car_i_distributed_arch,jo:2015:car_ii_distributed_arch}.

\begin{figure}[htb]
    \begin{subfigure}[b]{0.48\linewidth}
        \includegraphics[width=\linewidth]{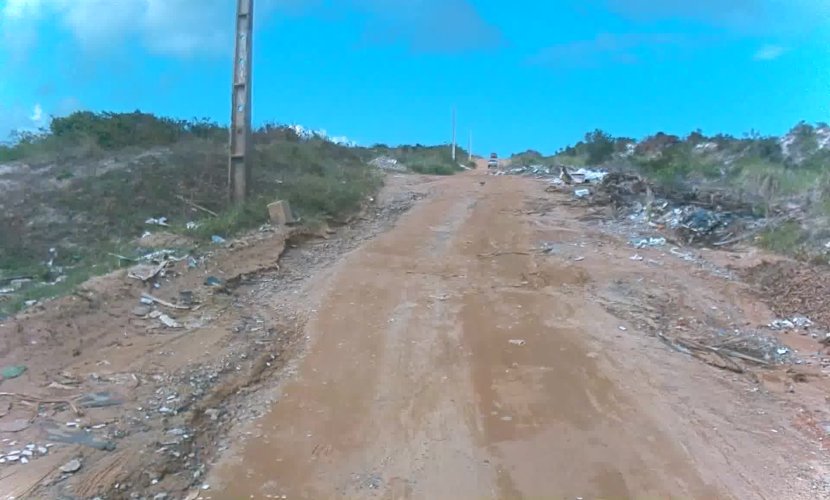}
        \caption{Unpaved road in Brazil}
        \fautor%
    \end{subfigure}%
    \hfill%
    \begin{subfigure}[b]{0.48\linewidth}
        \includegraphics[width=\linewidth]{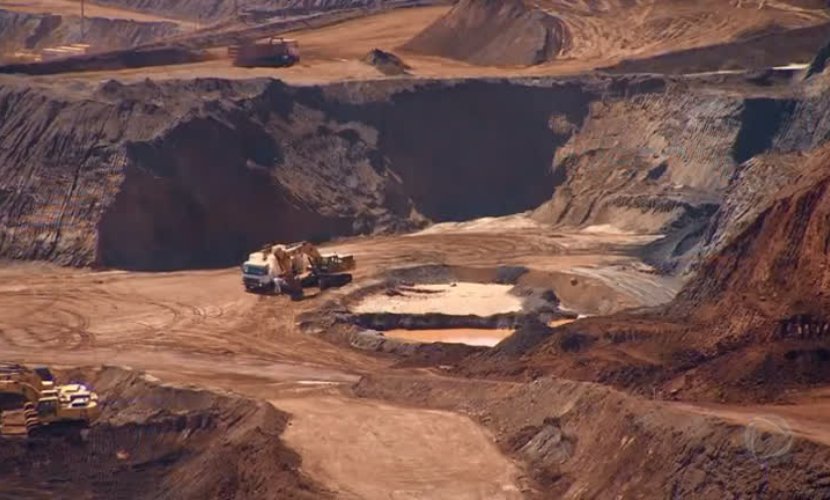}
        \caption{Mining environment}
        \fdireta{R7:2018:mineradora}
    \end{subfigure}
    \begin{subfigure}[b]{0.48\linewidth}
        \includegraphics[width=\linewidth]{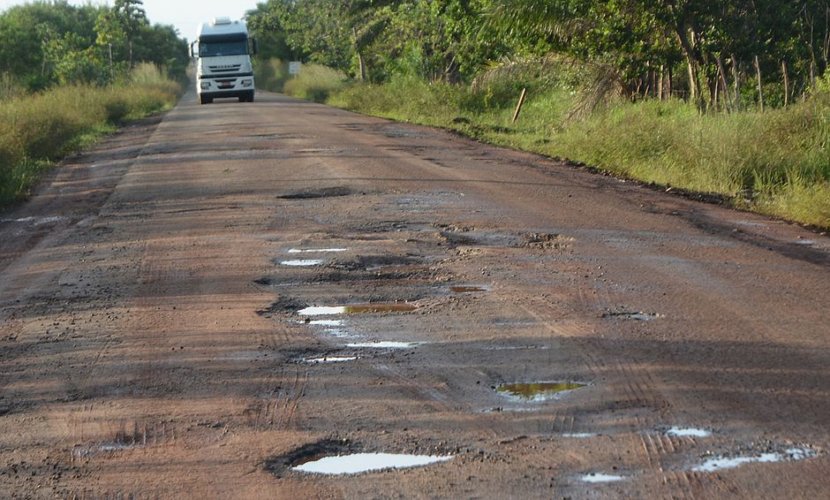}
        \caption{Bumpy intercity road}
        \fdireta{Correios:2017:palma}
    \end{subfigure}%
    \hfill%
    \begin{subfigure}[b]{0.48\linewidth}
        \includegraphics[width=\linewidth]{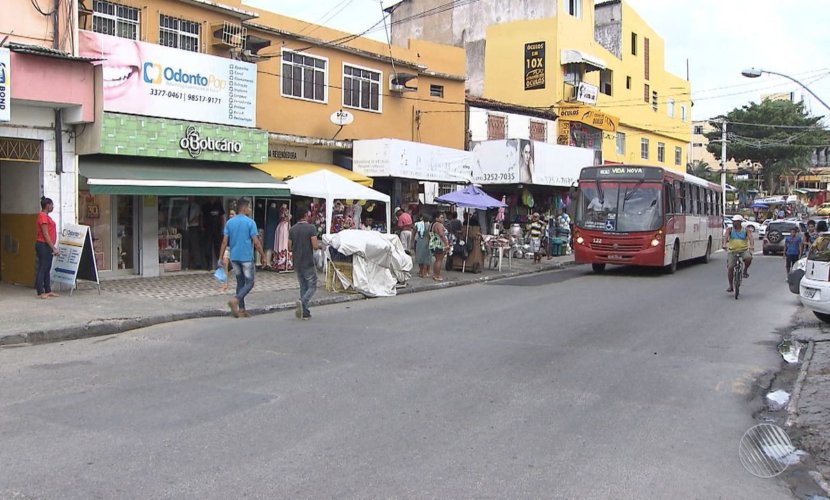}
        \caption{Road poorly signposted}
        \fdireta{G1:2018:lauro-de-freitas:limites}
    \end{subfigure}
    \caption{Common Road and Highway Situations in Brazil.}%
    \label{fig:fig1}
    \fautor%
\end{figure}

Although those challenging results helped boost research by universities and technology companies, \sigla{AV}{Autonomous Vehicles} and \sigla{ADAS}{Advanced Driver Assistant Systems} currently being designed for well-paved urban environments will likely face problems when operating on unpaved roads and some off-road environments. Nevertheless, there are many unpaved roads in developing countries, such as Brazil (\autoref{fig:fig1}). According to National Transport Confederation~\cite{cnt:2019:PesquisaCNTderodovias,dnit:2018:relatorio-de-gestao-tematica}, Brazil has only 12.4\% of its national road network paved. In addition, many open-pit industries are operating in off-road environment where vehicles should work in harsh conditions to improve production and logistics efficiency. Activities such as mining often need to interrupt the production process due to low visibility problems~\cite{Vale:2014:Report,Wheaton:2019:Salobo,Glebov_2021,Chatterjee:2019:MRE}.

In conditions including off-road environments with low visibility in mine and agricultural regions, ADAS should help mitigate risks by assisting drivers in the production and transport of workers. As well as, the AVs should help increase production efficiency and improve the logistics flow in rural areas and unpaved roads. Additionally, according to \citeonline{Kukkala:2018:IEEECEM}, one of the biggest challenges for vision-based ADAS is the susceptibility to environmental and visibility conditions such as rain and blinding glare in the late afternoon. Also, according to SAE J3016~\cite{SocietyforAutomotiveEngineers2021}, to reach level 5 of autonomy\footnote{Level 5 full-driving automation system must be capable of operating the vehicle on-road anywhere that a typically skilled human driver can reasonably do. The systems at this level are not affected by weather and can transport humans comfortably, safely, and efficiently.}, AVs should operate in adverse conditions like tracks covered by snow or dust. Then researches supporting AVs and ADAS development in such scenarios are necessary to democratize and allow the insertion of these technologies in developing and continental dimensions countries. Considering those requirements before mentioned, it is also relevant to evaluate the impact of different visibility conditions severity on the system's ability to perceive the environment correctly.

\begin{figure}[htb]
	\begin{subfigure}[b]{0.45\linewidth}
		\includegraphics[width=\linewidth]{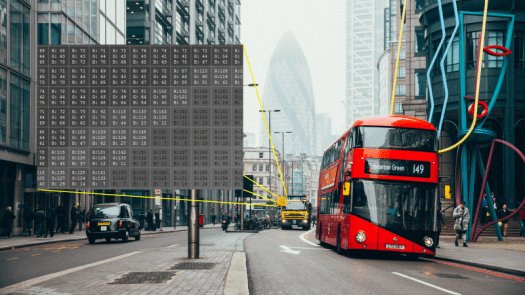}
		\caption{Camera sensor RAW information.}
	\end{subfigure}%
	\hfill%
	\begin{subfigure}[b]{0.45\linewidth}
		\includegraphics[width=\linewidth]{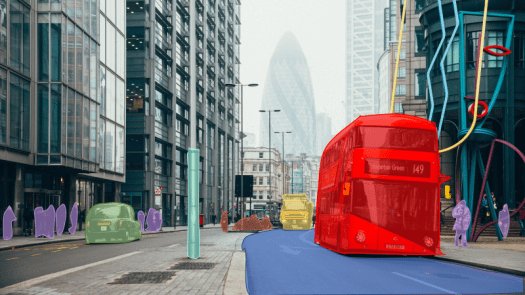}
		\caption{Scene understanding.}
	\end{subfigure}
	\caption{Camera sensor RAW information and scene understanding.}%
	\label{fig:raw_pixels-scene_understanding}
	\fadaptada{propelmee:2018:what-is-perception-vehicles}
\end{figure}

When people are driving, they can perceive scene elements almost instantly from visual information.\ That is crucial to traveling safely in traffic.\ The human can perceive traffic areas, pedestrians, animals, cars, crosswalks, traffic signals, and several other elements. In autonomous vehicles development, the perception subsystem does this understanding of the scene in real-time.\ It is a crucial step and allows them to travel in complex and dynamic environments.\ Perception is a common step to AVs and ADAS.\ It is one of the most critical tasks in developing an autonomous car~\cite{brummelen:2018:autonomous-vehicle-perception}.\ It is responsible for receiving information from various sensors and interpreting them by carrying out the recognition process on the data~\cite{BADUE:2021:IARA}, as illustrated in \autoref{fig:raw_pixels-scene_understanding}.

\begin{figure}[htb]
	\includegraphics[width=\linewidth]{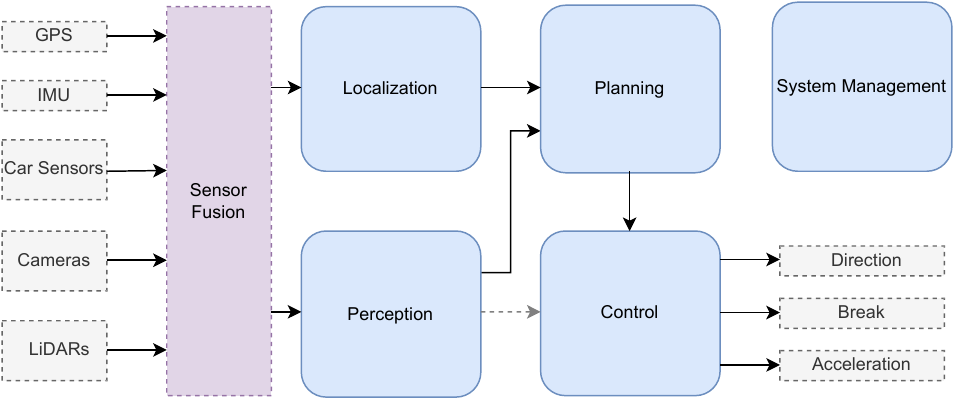}%
	\caption{Diagram with the main modules for an autonomous vehicle.}%
	\label{fig:diagrama-modulos-carro-autonomo}%
	\fautor%
\end{figure}

Considering a modularized architecture for AVs, such as shown in \autoref{fig:diagrama-modulos-carro-autonomo}, the Perception is the subsystem responsible for perceiving the environment around the vehicle or, in other words, for carrying out the visual recognition process based on data from various sensors, such as \sigla{RGB}{Red Green Blue} cameras, \sigla{IR}{Infra Red} cameras and \sigla{LiDAR}{Light Detection and Ranging}~\cite{BADUE:2021:IARA}. The Localization subsystem determines the vehicle position based on \sigla{GPS}{Global Positioning System}, \sigla{IMU}{Inertial Measurement Unit}, dead reckoning\footnote{In navigation, dead reckoning is the process of calculating the current position of some moving object by using a previously determined position and then incorporating estimates of speed, heading direction, and course over elapsed time.}, visual odometry, and road maps. The Planning module uses Perception and Localization information to decide how the car should behave and move within the lanes. The Control module commands the actuators of the car's steering, brake, and accelerator following the Planning decision. And finally, the System Management module is responsible for supervising the system, performing fault management, and providing the \sigla{HMI}{Human-Machine-Interface}.

Regarding perception, there are three paradigms in the literature: Mediated Perception, Direct Perception, and Behavioral Cloning (Behavior Reflex Perception or End-to-End Driving). Our work follows the classical paradigm (mediated perception), which is the most commonly used in autonomous cars nowadays~\cite{brummelen:2018:autonomous-vehicle-perception}. This approach uses algorithms to recognize relevant elements within the scene and combine them into a unified representation (world model). This representation is the input source for the Planning and Control modules (\autoref{fig:diagrama-modulos-carro-autonomo}) to decide the trajectory of the vehicle~\cite{chen:2015:deepdriving}. 

\begin{figure}[htb]
	\begin{subfigure}[b]{0.475\linewidth}
		\includegraphics[width=\linewidth,trim={0cm 0cm 0cm 0cm},clip]{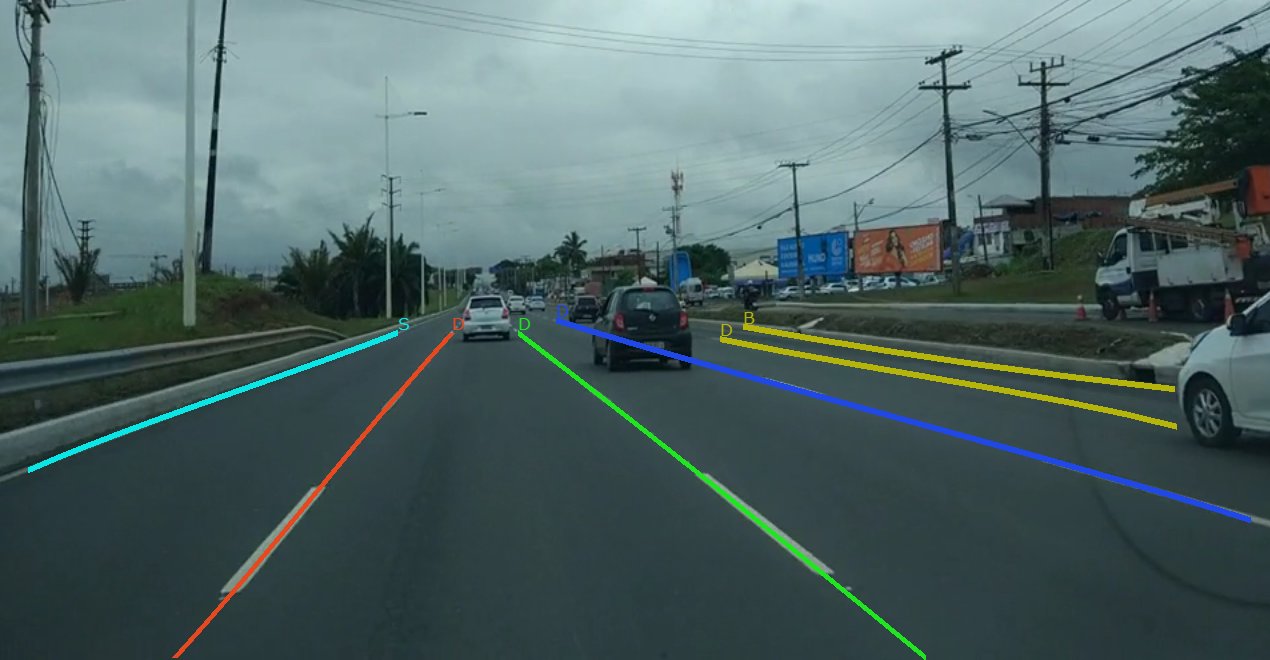}
		\caption{Line detection.}%
		\label{subfig:detectando-faixas-na-venida-paralela}
	\end{subfigure}%
	\hfill%
	\begin{subfigure}[b]{0.475\linewidth}
		\includegraphics[width=\linewidth,trim={0cm 0cm 0cm 0cm},clip]{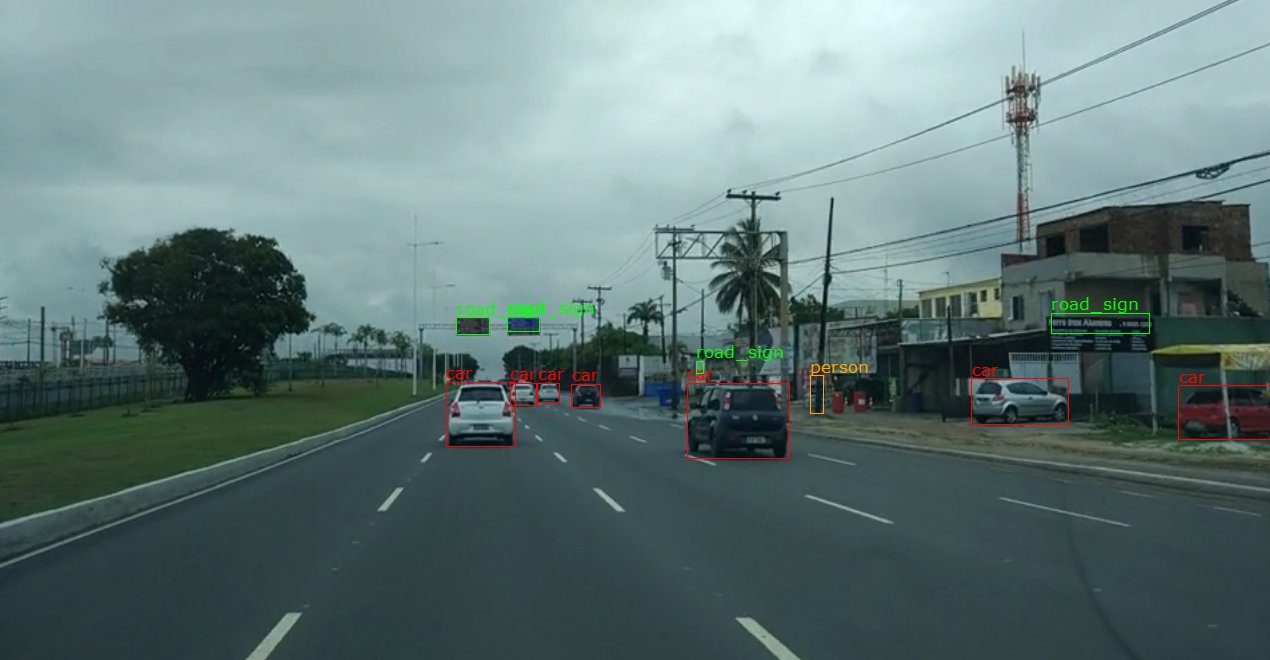}
		\caption{Object detection.}%
		\label{sunfig:detectando-objetos-na-venida-paralela}
	\end{subfigure}
	\begin{subfigure}[b]{0.475\linewidth}
		\includegraphics[width=\linewidth,trim={0cm 0cm 0cm 0cm},clip]{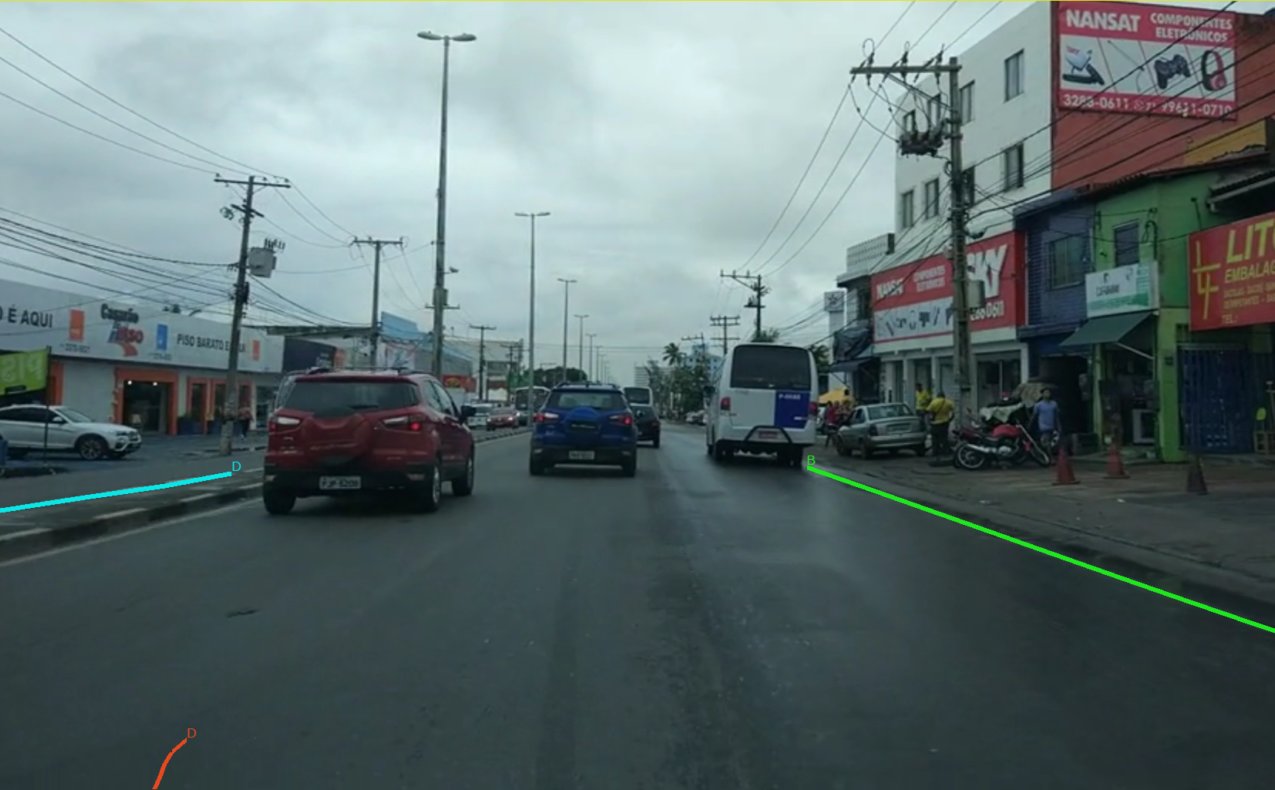}
		\caption{Line detection on unmarked road.}%
		\label{subfig:tentando-detectar-faixas-estrada-do-coco}%
	\end{subfigure}%
	\hfill%
	\begin{subfigure}[b]{0.475\linewidth}
		\includegraphics[width=\linewidth,trim={0cm 0cm 0cm 0cm},clip]{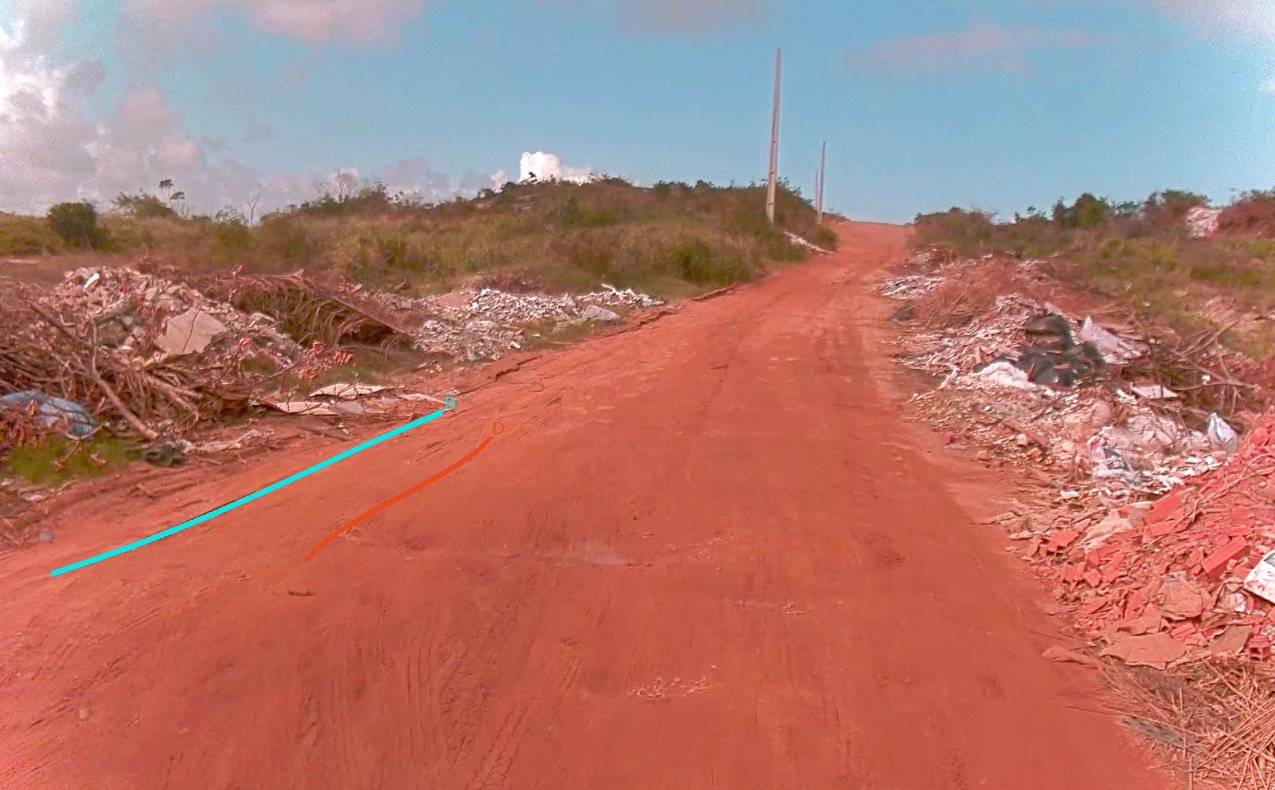}
		\caption{Line detection on unpaved road.}%
		\label{subfig:tentando-detectar-faixas-jaua}
	\end{subfigure}
	\caption{Line and object detection on paved and unpaved roads.}%
	\label{fig:detectando-faixas-e-objetos-na-venida-paralela}%
	\fautor%
\end{figure}

Detecting traffic areas is not a simple task. The vehicle needs to perceive the environment, recognize the road, and identify whether certain regions are obstructed by cars, animals, pedestrians, or other obstacles. One of the possible approaches to solve this problem is to use line detection algorithms~\cite{Lim:2009:ICIHMSC:Lane-Detection-Kalman, Deng:2012:ICCP:Deng:Robust-lane-detection, Ozgunalp:2017:Procedia-Computer-Science:Lane-detection-by-estimating, Deng:2018:DCABES:Double-Lane-Detection, Narote:2018:Pattern-Recognition:review-lane-detection, Nguyen:2018:ESTIJ:Study-rt-Detection-Lane}
(\autoref{subfig:detectando-faixas-na-venida-paralela}), associated with some method to object detection~\cite{Ren:2015:faster-rcnn:2969239.2969250, Redmon:2016:YOLO:ieeecvpr} (\autoref{sunfig:detectando-objetos-na-venida-paralela}). However, this approach depends on properly paved roads and lane markings (Figures~\ref{subfig:tentando-detectar-faixas-estrada-do-coco} and~\ref{subfig:tentando-detectar-faixas-jaua}), which is not the case in an off-road environment.%

Another possible strategy is to use panoptic\footnote{Panoptic segmentation unifies the typically distinct tasks of semantic segmentation (assigns a class label to each pixel) and instance segmentation (detects and segments each object instance)} or semantic segmentation to find the traffic area~\cite{Long:2015:FCN:ieeecvpr, Zhao:2017:PSPNet:ieeecvpr, Chen:2018:deeplab:ieeeTPAMI, Alexander:2018:CoRR:PanopticSegmentation, Costea:2018:ITSC:Fusion:panotic}, in which the traffic region can be segmented independently of the road markings. The advantage of this approach is that in addition to segmenting the road limits, it can also discover and segment obstacles on the road simultaneously. This thesis is particularly interested in the behavior of visual perception as a Deep Supervised Learning problem for semantic segmentation~\cite{Liu:2018:AIR:Recent-progress-in-semantic-s,YUAN:2021:eswa} in off-road environments and unpaved roads, with no signs or road marks to indicate where the vehicle should travel. Therefore, to perform the visual scene perception, this work employs computer vision algorithms that use data-guided modeling with \sigla{CNN}{Convolutional Neural Networks}~\cite{lecun:1998:ieee-gradient-based-learning}. In such a solution, the researchers annotate the data to generate the ground truth and train the deep networks to learn which characteristics are relevant to achieve their objective (detection, classification, identification, segmentation).

\newword{Convolutional Neural Networks}{is a class of artificial neural networks and Deep Learning algorithms, most commonly used in image recognition. They are especially good at processing data with time or spatial connections like pixel data}



Visual perception is still challenging for machines, even more in low-visibility conditions, such as those found in open-pit mines and off-road environments. Moreover, real-time processing is a concern in this kind of decision system where is crucial to building perception able to be accurate, fast, and stable over different adverse conditions. However, in consequence of the search for more accurate algorithms, there has been a trend towards more complex and deeper network architectures that reach up to hundreds of millions of parameters and tens of billions of multiply-accumulate (MAC) operations~\cite{EfficientNet:pmlr:2019:tan19a}. So there is no guarantee that such algorithms are fast and computationally efficient to be embedded in real-time applications with limited computational capacity and power restrictions such as autonomous cars and robots.

\sigla*{MAC}{Multiply-accumulate operations}



On the other hand, fast and computationally efficient inference has also been a concern of other works, which propose network architectures capable of keeping a reduced size and performing well in the benchmarks~\cite{Sandler:2018:MobileNetV2:IEEE-CVF,Howard:2019:ICCV:MobileNetV3,pmlr-v139-d-ascoli21a} or aiming to facilitate the re-implementation of inference algorithms in real-time~\cite{Jacob:2018:Quantization:IEEE-CVF}. These works manage to keep the computational cost in the millions of parameters and hundreds of millions of MACs. 

\begin{figure}[htb]
    \captionsetup[subfigure]{font=scriptsize,labelformat=empty}
    \begin{subfigure}[b]{0.245\linewidth}
        \caption{Image}
        \includegraphics[width=\linewidth]{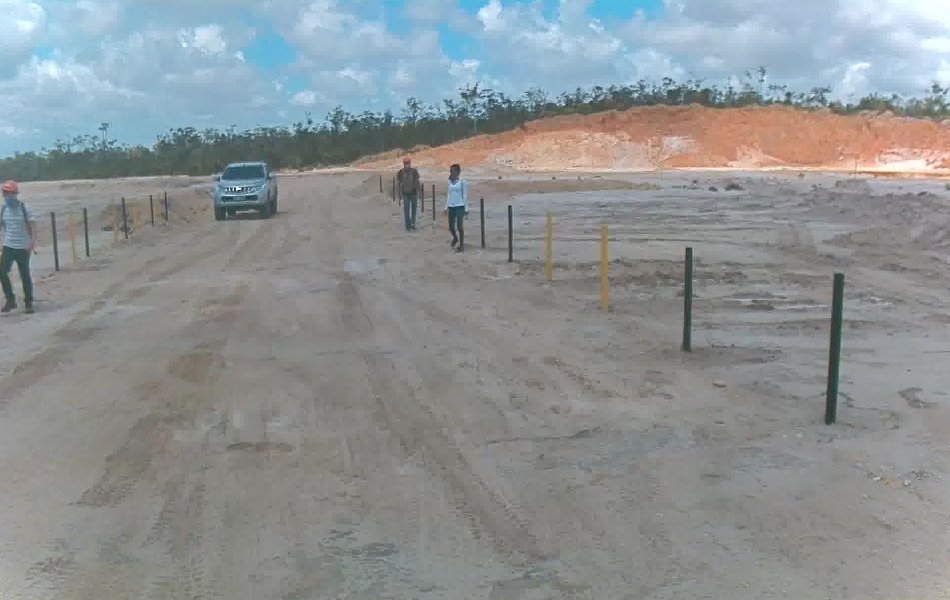}
    \end{subfigure}%
    \hfill
    \begin{subfigure}[b]{0.245\linewidth}
        \caption{Expected}
        \includegraphics[width=\linewidth]{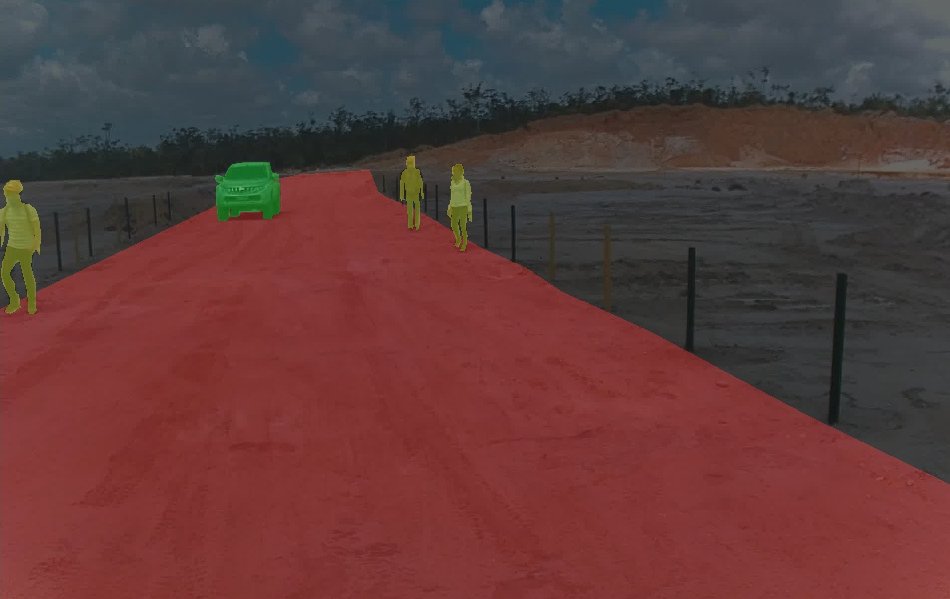}
    \end{subfigure}%
    \hfill
    \begin{subfigure}[b]{0.245\linewidth}
        \caption{PSPNet}
        \includegraphics[width=\linewidth]{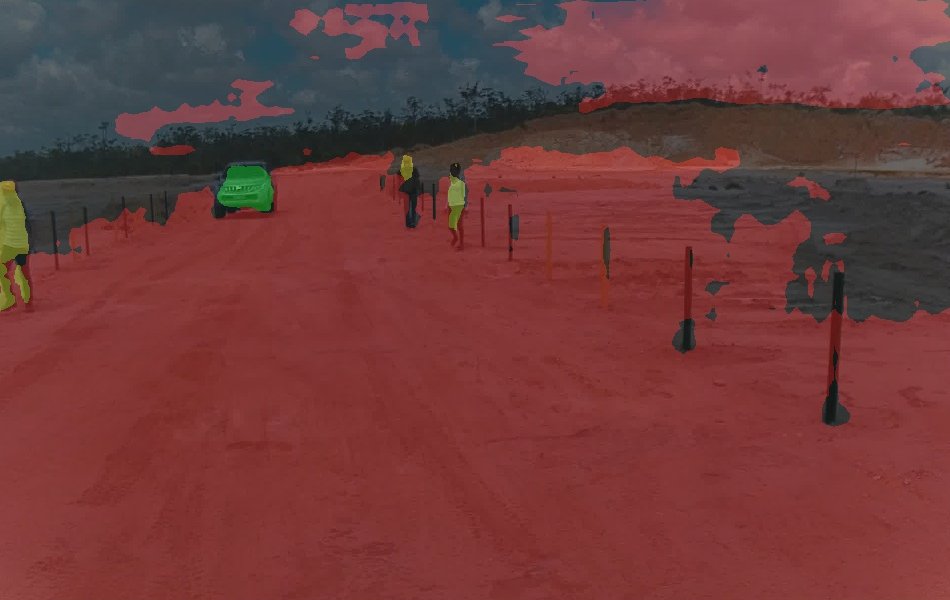}
    \end{subfigure}%
    \hfill
    \begin{subfigure}[b]{0.245\linewidth}
        \caption{DeepLab}
        \includegraphics[width=\linewidth]{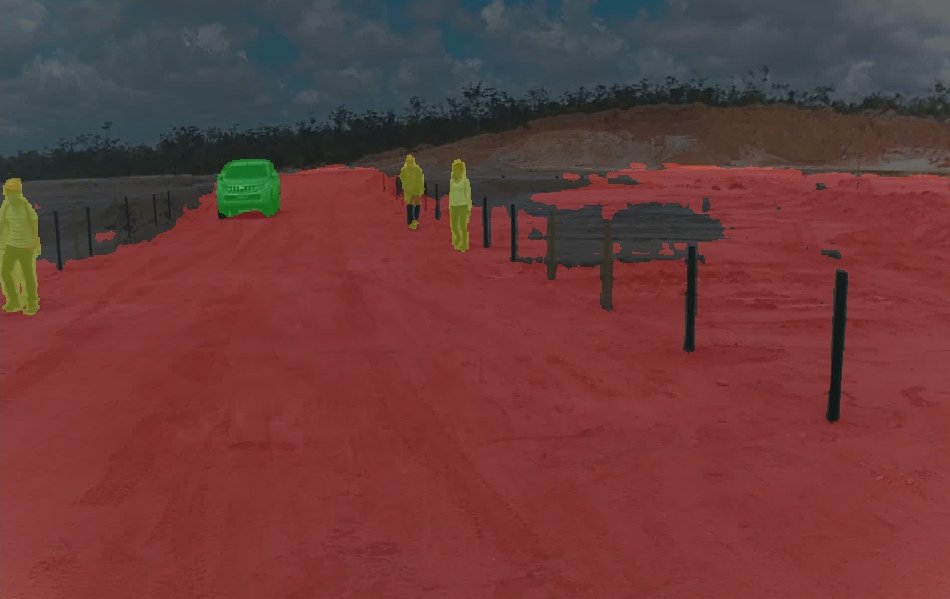}
    \end{subfigure}%
    \vspace{0.003\linewidth}
    
    \begin{subfigure}[b]{0.245\linewidth}
        \includegraphics[width=\linewidth]{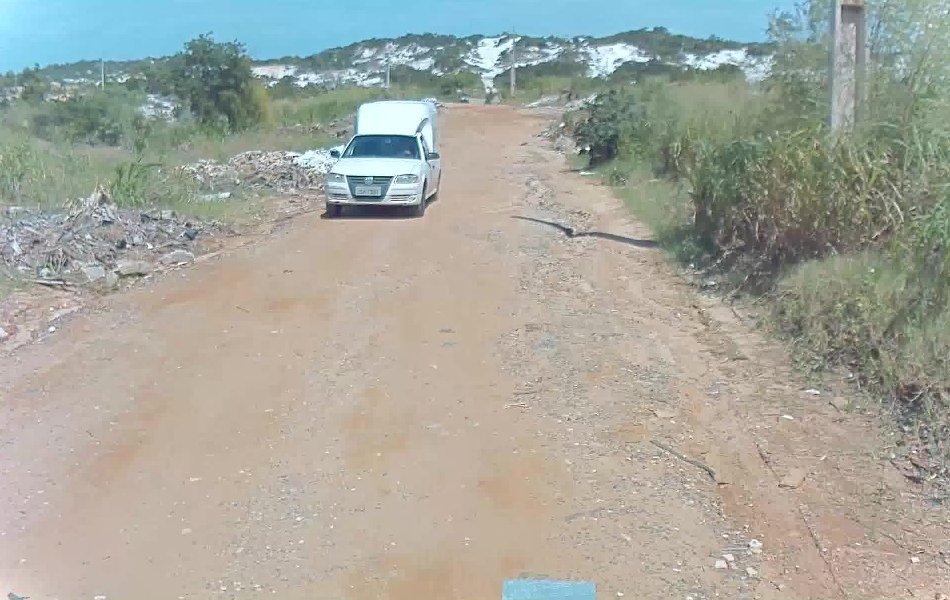}
    \end{subfigure}%
    \hfill
    \begin{subfigure}[b]{0.245\linewidth}
        \includegraphics[width=\linewidth]{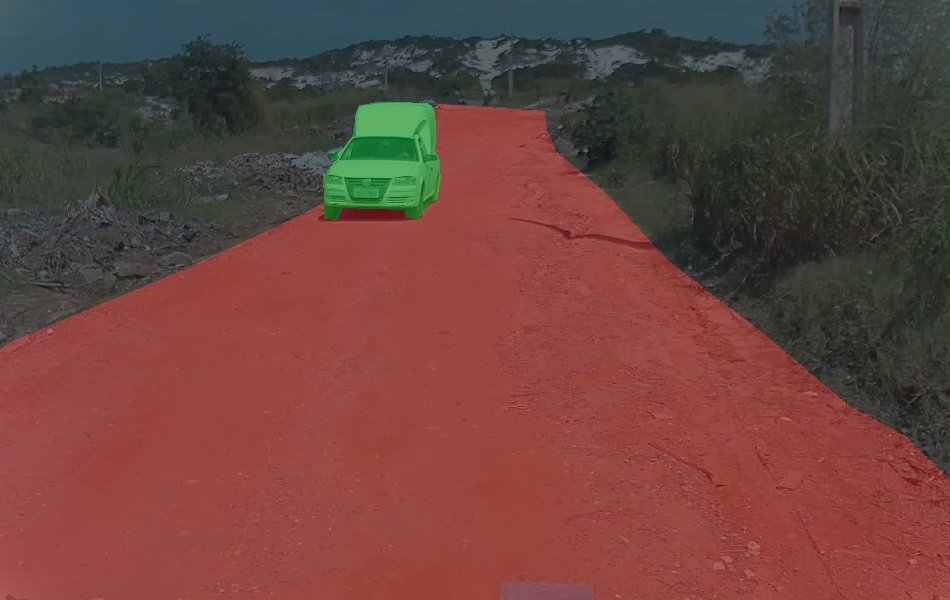}
    \end{subfigure}%
    \hfill
    \begin{subfigure}[b]{0.245\linewidth}
        \includegraphics[width=\linewidth]{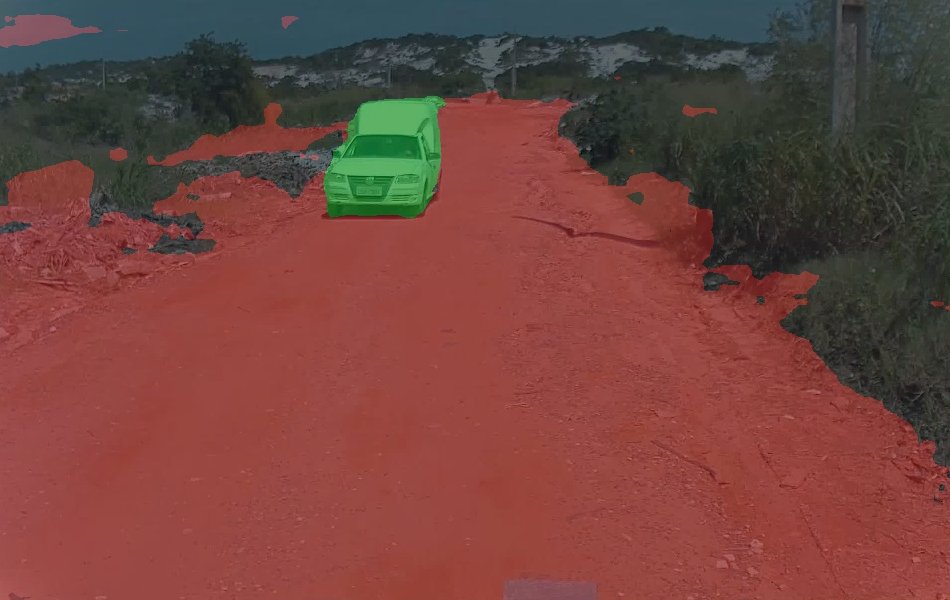}
    \end{subfigure}%
    \hfill
    \begin{subfigure}[b]{0.245\linewidth}
        \includegraphics[width=\linewidth]{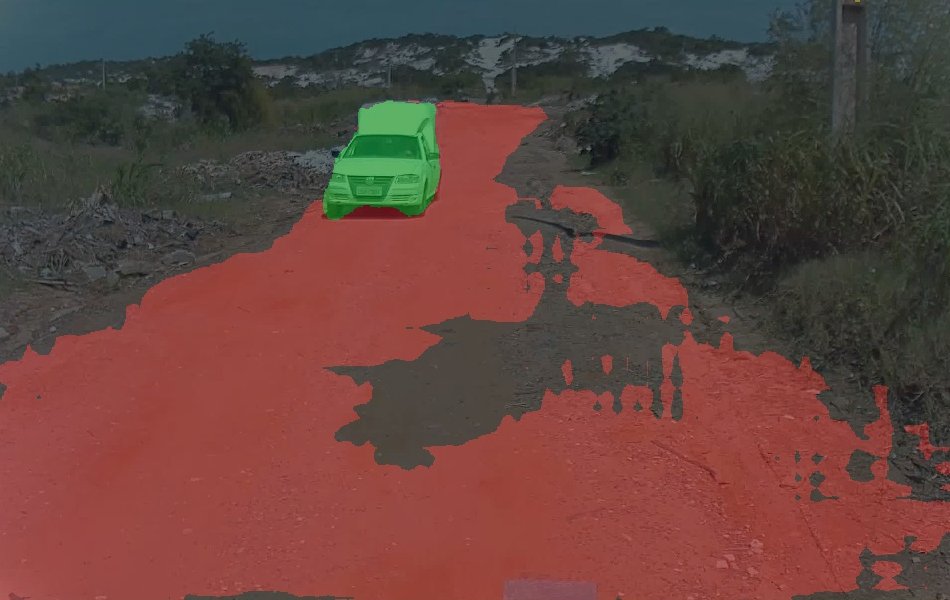}
    \end{subfigure}%
    \vspace{0.003\linewidth}
    
    \begin{subfigure}[b]{0.245\linewidth}
        \includegraphics[width=\linewidth]{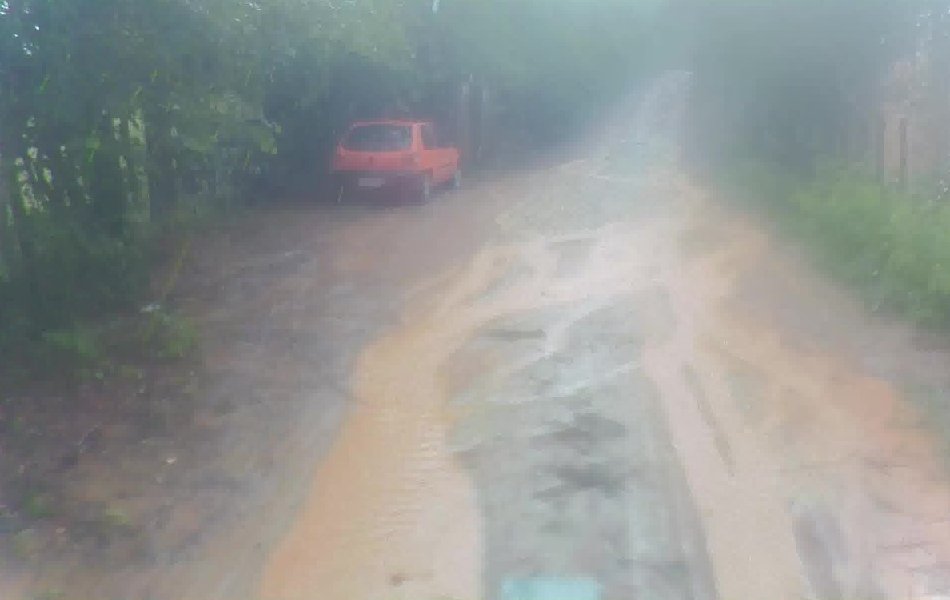}
    \end{subfigure}%
    \hfill
    \begin{subfigure}[b]{0.245\linewidth}
        \includegraphics[width=\linewidth]{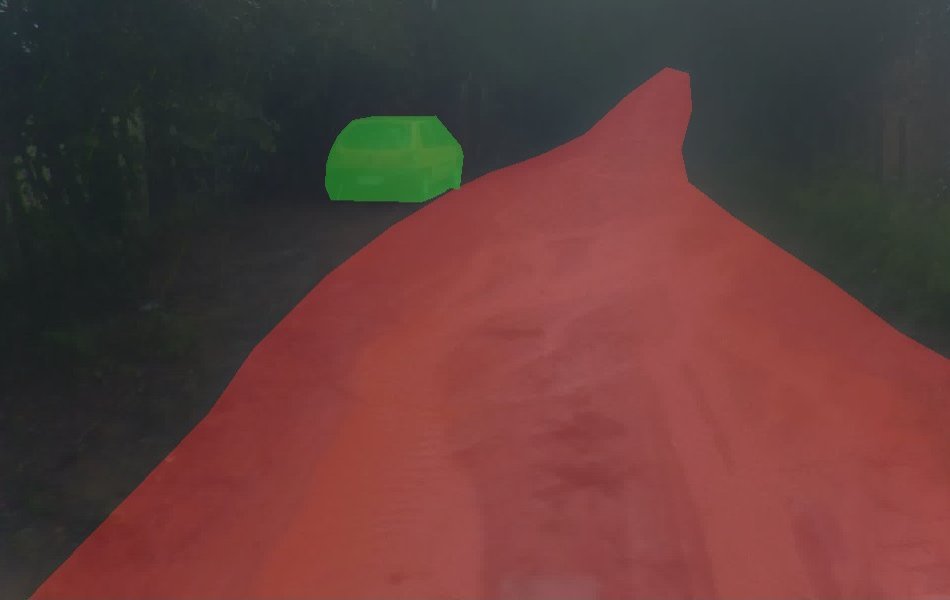}
    \end{subfigure}%
    \hfill
    \begin{subfigure}[b]{0.245\linewidth}
        \includegraphics[width=\linewidth]{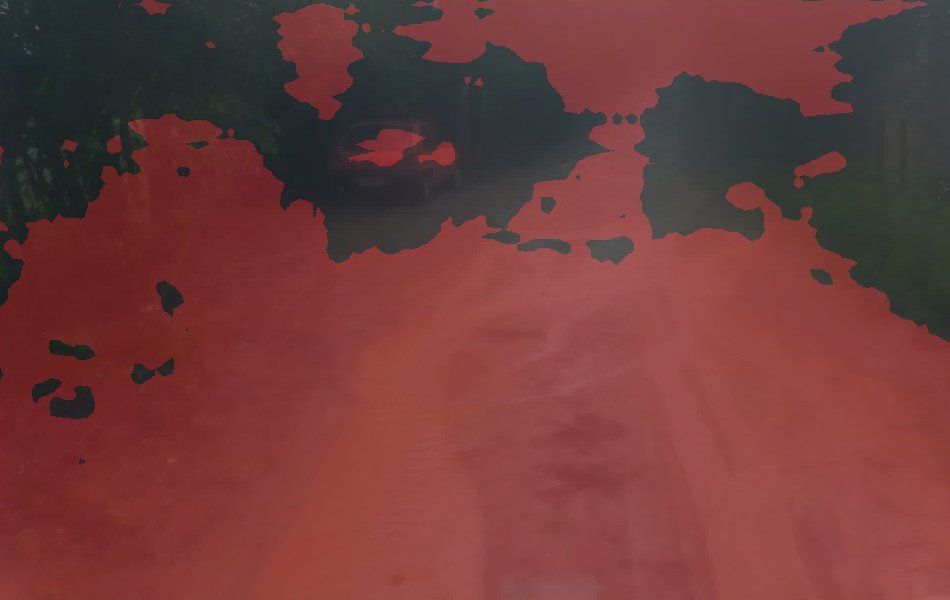}
    \end{subfigure}%
    \hfill
    \begin{subfigure}[b]{0.245\linewidth}
        \includegraphics[width=\linewidth]{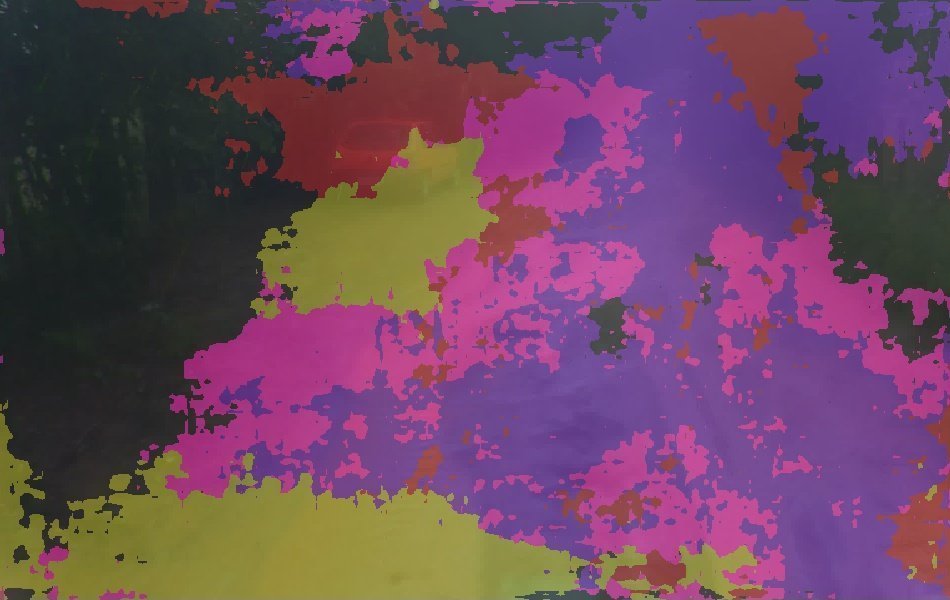}
    \end{subfigure}%
    
    \caption{Test of PSPNet and DeepLab with Cityscape on unpaved roads.}%
    \label{fig:test_pspnet_deeplab_cityscape_unpaved_roads}%
    \fautor%
\end{figure}

Datasets are another challenge in the development of visual perception systems suitable for off-road environments. Although autonomous cars research is advancing fast, most of the datasets available for training visual perception are focused on urban environments~\cite{cordts:2016:cityscapes,complex-urban-dataset-jjeong-2019-ijrr,Sun:2020:CVPR:WaymoOpenDataset}. To build this thesis, the researcher carried out some tests with PSPNet~\cite{Zhao:2017:PSPNet:ieeecvpr} and DeepLabV3~\cite{Chen:2018:deeplab:ieeeTPAMI} networks trained with the Cityspace~\cite{cordts:2016:cityscapes} urban dataset to check the possibility of using these pre-trained networks in visual perception in off-road environments and unpaved roads (\autoref{fig:test_pspnet_deeplab_cityscape_unpaved_roads}). It was possible to see that the before-mentioned systems currently being developed for autonomous vehicles may not be suitable for developing countries remaining restricted to a small set of roads in urban centers. This restriction limits even the implementation of autonomous systems in cargo vehicles, such as buses and trucks.

Summarizing the thesis hypotheses are:
\begin{enumerate}
\item Deep Supervised Learning algorithms are suitable for building a visual perception on unpaved roads and off-road environments, even in low-visibility conditions; and
\item Systems currently being developed for autonomous vehicles relying only on a well-paved urban dataset are not suitable for developing countries or will limit their implementation in cargo vehicles on unpaved roads and off-road environments.
\end{enumerate}

Given such a scenario, this research aims to contribute to the understanding of how visual perception as a problem of \sigla{SL}{Supervised Learning} and \sigla{DL}{Deep Learning} behave on unpaved roads commonly found in developing countries and off-road industrial environments spotted in farming or open-pit mining. Besides the main research question, this work also tries to answer some underlying questions like: 
\begin{enumerate}
\item How do those DL algorithms have their segmentation capability affected by variations in visibility conditions such as rain, night, dust, fog, and noise? 
\item How can different architectural arrangements composed of some modules proposed in the literature for semantic segmentation manage to segment the traffic zone and obstacles in this kind of environment? 
\item Since such algorithms have a high computational cost, how to embed those solutions for field testing? 
\item Besides, how efficient are the algorithms trained with datasets for well-paved urban environments when applied to unpaved roads to off-road environments? 
\end{enumerate}



In order to help answer such questions, this work presents a series of contributions, namely: 
\begin{itemize}
    \item The proposal for a new dataset for unpaved roads and off-road environments containing several adverse visibility situations, such as rain, dust, and poor lighting (night condition) to fill the gap of such kind of a dataset and making possible to carry out this research;  
    \item An investigation of the feasibility of applying Deep Supervised Learning to detect track limits where there is no clear delimitation between what is a road and what is not a road, as is the case with sandy off-road environments; 
    \item The proposition of a \sigla{CMSNet}{Configurable Modular Semantic Segmentation Neural Network} framework; 
    \item A study of how our proposed segmentation algorithms behave in different levels of visibility impairment severity; and 
    \item The evaluation of semantic segmentation architectures ported\footnote{Port (past participle ported) in computing means the transferring process of a software or application from one system architecture or machine to another.} to embedded field applications capable of real-time inference.
\end{itemize}

\newword{Port \- past participle ported}{in computing means the transferring process of a software or application from one system architecture or machine to another}

This work is organized with \autoref{chapter:literature-and-fundamentals} having the theoretical fundamentals and the literature review.\ \autoref{chapter:design-and-methodology} presents the research methodology and the system design.~\autoref{chapter:experiments-and-finds} presents the research findings and experiments results. Finally, Chapter~\autoref{chapter:conclusion} presents the conclusions.

\chapter{Literature and Fundamentals}
\label{chapter:literature-and-fundamentals}
This chapter shows the theoretical framework (\autoref{sec:theoretical-framework}) behind the research. The fundamentals explained are Artificial Intelligence (\autoref{subsec:artificial_intelligence}), Machine Learning (\autoref{subsec:machine-learning}), Supervised Learning (\autoref{subsec:supervised-learning}), Artificial Neural Networks (\autoref{subsec:artificial_neural-networks}), Deep Learning (\autoref{subsec:deep-learning}), and Convolutional Neural Networks (\autoref{subsec:convolutional-neural-networks}).

This chapter also shows the relevant works in the literature related to this research, directly or indirectly (\autoref{sec:related-works}). Those works are grouped by theme, and their relationships with this research are explored. The topics covered in this state-of-the-art study are system architecture for autonomous vehicles and perception categorization (\autoref{subsec:perception-architecture}), networks for classification and backbones 
for features extraction (\autoref{subsec:backbones-and-classification}), architectures for segmentation (\autoref{subsec:semantic-segmentation}), segmentation of track areas on unpaved roads and off-road environments (\autoref{subsec:off-road-segmentation}) and datasets (\autoref{subsec:av-adas-datasets}).

\newword{Backbone}{refers to the section of the Deep Neural Network responsible for doing the feature extraction of the inputs. It is a term used by some authors to refer to the part of networks that compute features from the input image. These features can be used with some decoder structure to provide a specific vision function like classification, detection, or segmentation.}

\section{Theoretical Framework}
\label{sec:theoretical-framework}  

In this work, the authors propose a Visual Perception subsystem modeled as a problem of Deep Supervised Learning applied to semantic segmentation. Supervised Learning refers to the theory encompassing the whole class of \sigla{ML}{Machine Learning} algorithms capable of learning from data~\cite{Caruana:2006:SLComp}. In this kind of approach, the human or another algorithm is responsible for labeling the reference data that comprise the training and test examples (dataset) used to instruct the learning system. On the other hand, Deep Learning (DL) refers to the class of ML algorithms that allows deep computational models (with multiple layers) to learn representations from large datasets in different abstraction levels~\cite{LeCun:2015:Deeplearning}. DL acts by applying optimization algorithms and backpropagation to tune the model parameters. Nevertheless, the researcher is responsible for modeling the hyper-parameters based on empirical experiments. This class of algorithms has improved the state-of-the-art for computer vision applications such as semantic segmentation \cite{YUAN:2021:eswa} and many other domains. 

\newword{Supervised Learning}{refers to the theory encompassing the whole class of Machine Learning algorithms capable of learning from data}

\subsection{Artificial Intelligence}
\label{subsec:artificial_intelligence}

\sigla{AI}{Artificial intelligence} is a branch of computer science, engineering, and electronics concerned with using computer systems to build machines able to perform tasks commonly associated with intelligent beings. The idea is to use computer systems to simulate the skill of reasoning, discovery mining, generalization, and learning from experiences. Such intelligent software fits for expert systems, natural language processing, speech recognition, and machine vision, among other possibilities. The solutions coming from AI are suitable for automating routine labor, understanding speech or images, making diagnoses in medicine, and supporting basic scientific research \cite{10.5555/1796343,AI_XU2021100179}. 

\newword{Artificial intelligence}{refers to a branch of computer science, engineering, and electronics concerned with using computer systems to build machines able to perform tasks commonly associated with intelligent beings. The term AI applies to intelligence demonstrated by machines oppositely to natural intelligence manifested by animals.}


Initially grounded in mathematical logic and philosophy, the AI aimed at formal descriptions of human thinking and the application of symbolic information processing. It takes this representative information as input, manipulating it according to a set of rules, and in so doing can solve problems, formulate judgments, and make decisions \cite{newell1972human, Dick2019Artificial}. In the beginning, the AI quickly starts to solve problems that are intellectually hard for people but straightforward for computers like the ones possible of being described by a list of formal mathematical rules. However, the real challenge was to solve tasks that are easy for humans to solve intuitively but hard to describe formally \cite{FATHI2018229}.

While some tasks like recognizing objects are easily done by animals, they are hard to achieve by applying hard-code knowledge about the world in a formal language. Such a difficulty faced by formal hard-coded knowledge indicates that AI machines should be able to build their own knowledge by learning from the data. This capability has been named Machine Learning \cite{Goodfellow-et-al-2016}.

\subsection{Machine Learning}
\label{subsec:machine-learning}

\begin{figure}[htb]
	\centering
	\includegraphics[width=.8\linewidth]{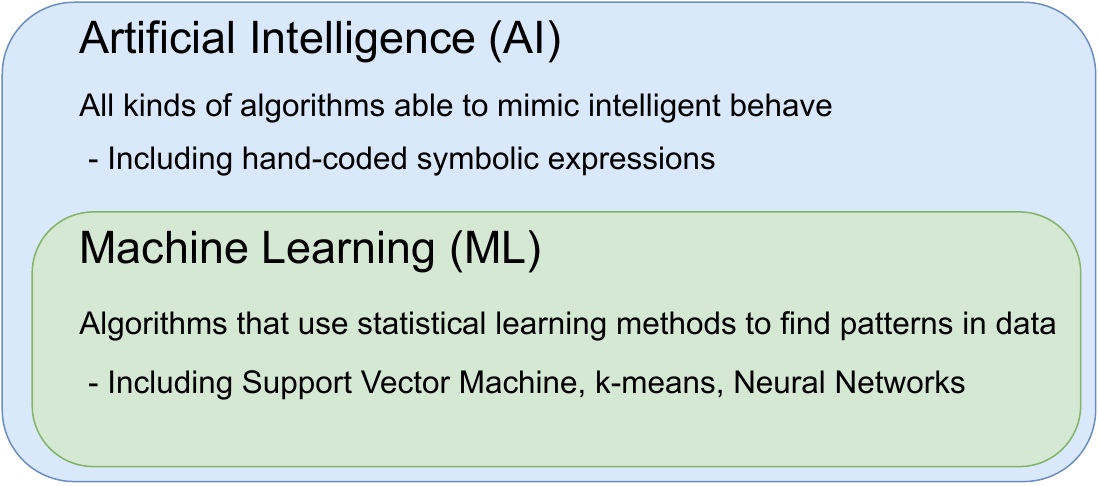}
	\caption{Machine Learning as subset of Artificial intelligence.}
	\fautor
	\label{fig:ai-ml}
\end{figure}

Machine Learning is a branch of AI (\autoref{fig:ai-ml}) that encompasses the algorithms able to learn their model rules from the data. They try to imitate how natural intelligence learns from examples and experiences, improving themselves automatically without being explicitly programmed. The introduction of machine learning allowed computers to deal with problems involving knowledge from the real world and make decisions that seem subjective~\cite{Goodfellow-et-al-2016}.

The performance of ML algorithms depends on the representation of the data used to train them. The majority of the traditional ML algorithms do not infer directly from the RAW data. Instead, a specialist is responsible for analyzing the data and extracting the pieces of information considered relevant (i.e., the appropriate features), so the algorithms can learn and make predictions from that pieces of information~\cite{Goodfellow-et-al-2016}. 

The ML algorithms can learn from a supervised process by using a pre-labeled set of data. Generally, a specialist classifies this set of examples to teach the algorithm to adjust its prediction model to make inferences adequately~\cite{LeCun:2015:Deeplearning}. Otherwise, the algorithms also can learn from unlabeled data through an unsupervised process~\cite{10.1162/neco.1989.1.3.295,Ghahramani2004}. Besides, the ML algorithms can learn by a semi-supervised process ~\cite{ZHOU20141239}, mixing labeled and unlabeled data, or through the reinforcement learning~\cite{sutton2018reinforcement} by using a rewards/punishments system.

\subsection{Supervised Learning}
\label{subsec:supervised-learning}

Supervised Learning is the most common Machine Learning method. In this approach, the first step is to collect or build a data set of samples (images) and label them. So, in the training process, the machine receives data samples as input to generate output values. Such values are compared with expected labels, and an error is calculated to help tune the model parameters~\cite{LeCun:2015:Deeplearning}. 


In classification problems, the algorithm usually receives input data and predicts a discrete value identifying the inputs as belonging to a specific class or group. Such outputs may come as a vector of scores indicating the most probable group. Even semantic segmentation is a kind of classification task, although at the level of the pixel. Each pixel is associated with a specific class. Differently, regression problems seek to predict continuous data such as the price of apartments or stocks given the features related to them.

\begin{figure}[htb]
	\centering
	\includegraphics[width=\linewidth]{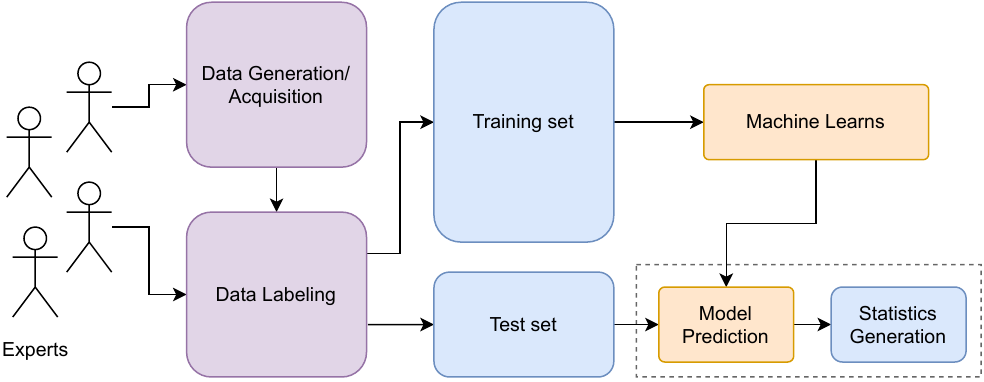}
	\caption{Supervised Learning process flow where the experts collect or generate the data, annotate them and create the labeled dataset depicted by the purple boxes. The hand-coded process, represented by blue boxes, uses training and test subsets as input to the learning system and colets output to generate statistics regarding the accuracy of the results. So, the yellow boxes indicate the machine learning and prediction model process.}
	\fautor
	\label{fig:supervised-learning}
\end{figure}

\autoref{fig:supervised-learning} shows a diagram presenting the steps involved in the SL model. The experts are responsible for collecting or generating the data, annotating them, and creating the labeled dataset. The research should split the dataset into training and test subsets, so they should use the train the model with training data and use the model to predict on test data. Finally, the research should gather statistical about the prediction quality to evaluate the performance of the algorithm.

There are many SL algorithms, such as \sigla{SVM}{Support Vector Machine}, Neural Networks, Logistic Regression, Naive Bayes, Memory-Based Learning, Random Forests, Decision Trees, Bagged Trees, Boosted Trees, and Boosted Stumps on binary classification problems ~\cite{Caruana:2006:SLComp}. However, the research object of this thesis is related to Supervised Deep Learning (SL and Deep Neural Networks) applied for semantic segmentation.

\subsection{Artificial Neural Networks}
\label{subsec:artificial_neural-networks}

\begin{figure}[htb]
	\centering
	\includegraphics[width=.8\linewidth]{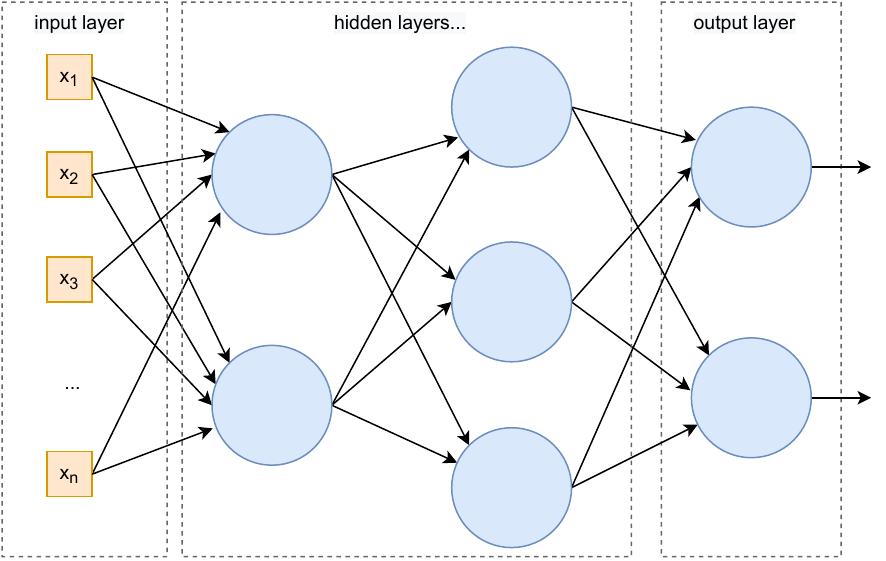}
	\caption{Multilayer perceptron.}
	\fautor
	\label{fig:multilayer-perceptron}
\end{figure}

The \sigla{ANN}{Artificial Neural Networks}, also known as Neural Networks, are biologically inspired computational network algorithms ~\cite{PARK2016123}. These algorithms use a reduced set of concepts from biological neural systems and are composed of connected processing elements, also known as either a artificial neuron or perceptron. The simplified neurons model has been introduced by \citeonline{McCulloch1943} and the perceptron mathematical model by \citeonline{Rosenblatt_1957_6098, Rosenblatt1958}. Such nodes are arranged in layers having outputs of each layer connected to a node input in the following layers. Each input is multiplied by weight in the nodes to mimic a synaptic connection, and those weights are real numbers that are adjusted during the training phase ~\cite{WALCZAK2003631}. \autoref{fig:multilayer-perceptron} shows an example of a multi-layer perceptron -- a popular class of ANNs.

\begin{figure}[htb]
	\centering
	\includegraphics[width=.8\linewidth]{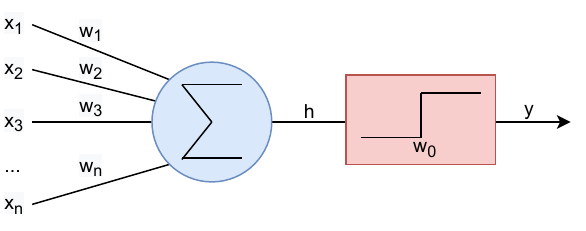}
	\caption{Artificial neuron mathematical model where $x_{i}, i=1,2...,n$ represents the inputs signal, $w_{i}$ is the synapse weight associated with the $i_{th}$ input, $\theta(.)$ is a unit step function at 0, and $w_{0}$ is associated with an input value -1 to serve as an extra bias to control the threshold.}
	\fautor
	\label{fig:perceptron-model}
\end{figure}

\autoref{fig:perceptron-model} shows an artificial neuron model, where $\theta(.)$ is a unit step function at 0, $w_{i}$ is the synapse weight associated with the $i_{th}$ input, and $w_{0}$ is associated with an input value -1 to serve as an extra bias to control the threshold. \autoref{eq:mcculloch-pitts} describes the basic weighted sum computation of its inputs signal, $x_{i}, i=1,2...,n$, passing the sum result by an activation function. In this case, it is a binary threshold unit function~\cite{Jain1996}. 

\simbolo{\sum}{Indicates summation and is used to denote a sum of multiple terms.}
\simbolo{\theta(.)}{Indicates a unit step function at 0.}

\begin{equation} \label{eq:mcculloch-pitts} 
y = \theta(\sum_{i=1}^{n} w_{i}x_{i} - w_{0})
\end{equation}

The Backpropagation was developed in practice and popularized by the work \citeonline{Rumelhart1986LearningRB}, who was responsible for the renovation of interest in Artificial Neural Networks \cite{Widrow199030years}. However, the first propose of using the ideas behind such algorithm intended for ANNs was published in \citeonline{werbos1982applications} and rediscovered independently by \citeonline{cun1986learning}. Using a smooth quadratic cost function as \sigla{MSE}{Mean Square Error}, differentiable with respect to the weights and inputs, it can calculate the gradients~\cite{LeCun:2015:Deeplearning}. The backpropagation can be applied repeatedly to backpropagate the error and calculate the gradients with respect to the all weights, i.e., how the error is affected by the change of the weights in the multiple layers of an ANNs. So, the training process can apply an optimization technic such as \sigla{SGD}{Stochastic Gradient Descent} to minimize the cost function~\cite{LeCun:2015:Deeplearning}, i.e., uses an optimization algorithm to find the better values of weight or parameters to take the learning error toward zero.

Even with the advances in training shallow networks using backpropagation, the ANNs had been largely rejected by the ML community and ignored by the computer-vision developers in the late 1990s. They believed it was infeasible to learn deep multistage feature extractors. It also was generally thought that gradient descent would get trapped in poor local minima ~\cite{LeCun:2015:Deeplearning}. Recent theoretical and empirical results suggest that local minima are not an issue in general, and poor local minima are rarely a problem with large networks~\cite{LeCun:2015:Deeplearning}.

\subsection{Deep Learning}
\label{subsec:deep-learning}
%
%

The Deep Learning term was introduced to the ML context by \citeonline{Dechter86DLTERM}. However, the interest in Deep Neural Networks was refreshed by Geoff Hinton and others researchers in \sigla{CIFAR}{Canadian Institute for Advanced Research}~\cite{Hinton2006DLFirst, Hinton2005DLFirst, Bengio2006DLFirst, Ranzato2006DLFirst}. The researchers showed how to solve DNNs by training individual layers firsts using an unsupervised technic. Then they added the output units and used backpropagation to fine-tune the model applying supervised learning~\cite{LeCun:2015:Deeplearning}.

The performance of classical ML algorithms depends on the representation of the data (high-level features) passed to them. However, several aspects of the environment influence individual pieces of data that can be observed. For example, a pixel of an off-road test track can have the same value as a no-track pixel, or the group of them can present the same texture, and the changes in visibility conditions or the angle of view can make them change the values completely~\cite{Goodfellow-et-al-2016}. In such a case, the process of analytically determining an optimal algorithm to extract high-level features of RAW data could not be hard to achieve, and it is not viable for a specialist to do it in real-time. Different from traditional ML methods, where a specialist is responsible for analyzing the RAW data and extracting the pieces of information (features) considered relevant, DL algorithms are capable of automatically learning how to extract high-level information directly from RAW data ~\cite{Goodfellow-et-al-2016,LiangSSG17}.

\begin{figure}[htb]
	\centering
	\includegraphics[width=\linewidth]{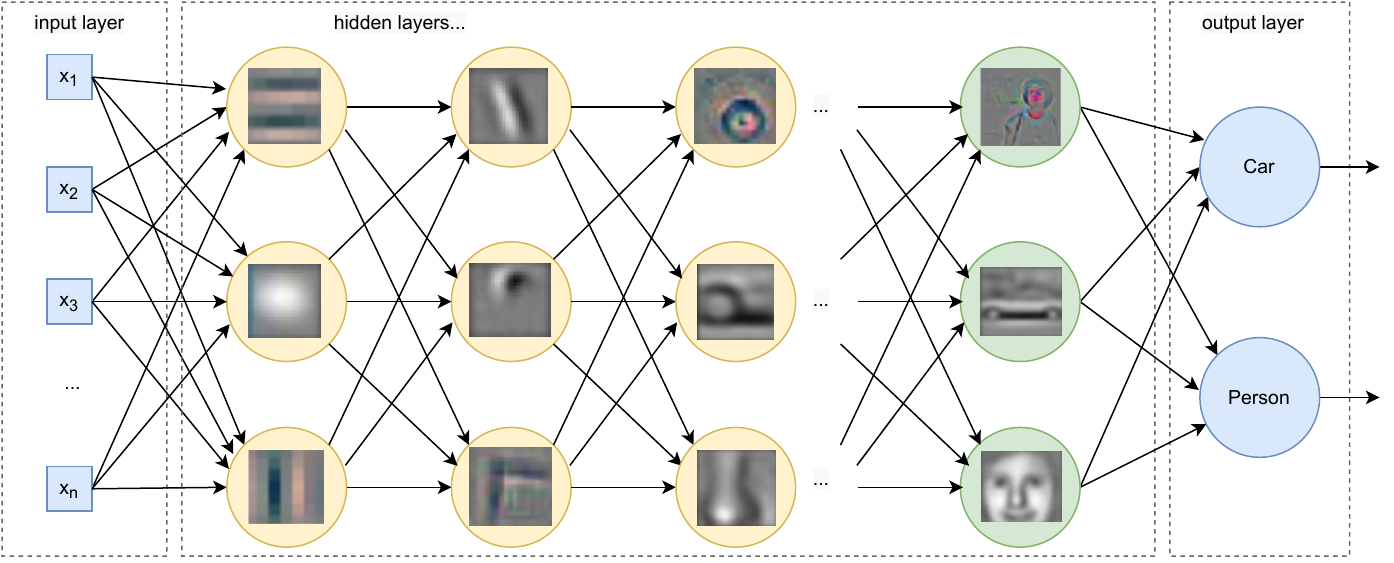}
	\caption{Graphic visualization of Deep Learning hierarchical feature representation, going from the most simple features until the most abstract representation and the classification process. The illustration shows how a deeper layer builds its concepts based on more simple ones extracted by the first layers.}
	\fautor
	\label{fig:dl_model_representation}
\end{figure}

Deep learning solves the problem of extracting high-level information directly from RAW data by introducing the idea of complex representations expressed in terms of other simpler representations~\cite{Goodfellow-et-al-2016} -- e.i., it learns features representation hierarchically. Besides, it also enables to take advantage of big data, which would be hard to do in an application relying on a priori knowledge of designers~\cite{LiangSSG17}. \autoref{fig:dl_model_representation} shows the idea of hierarchical features representation, going from most complex features (concepts relevant for a human) such as classes defined in terms of contours until the low-level motif expressed in terms of simple edge features~\cite{matthew:2014:zfnet:win-imagenet2013-classification}.

\begin{figure}[htb]
	\centering
	\includegraphics[width=.6\linewidth]{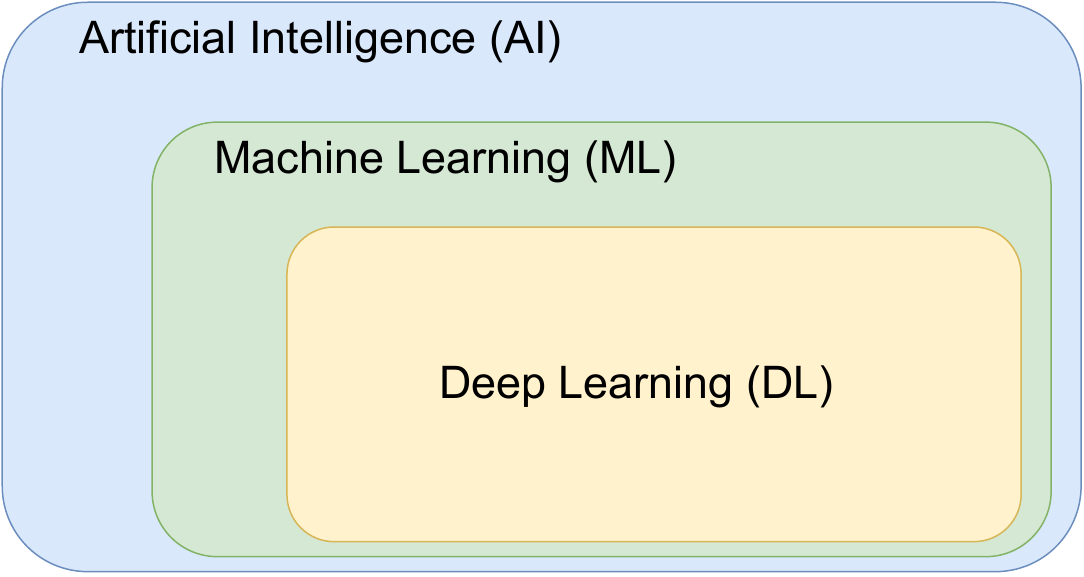}
	\caption{Diagram showing Deep Learning as a subset of Machine Learning algorithms, which is, in turn, a subset of Artificial Intelligence algorithms.}
	\fautor
	\label{fig:ai_ml_dl}
\end{figure}

\begin{figure}[htb]
	\centering
	\includegraphics[width=.8\linewidth]{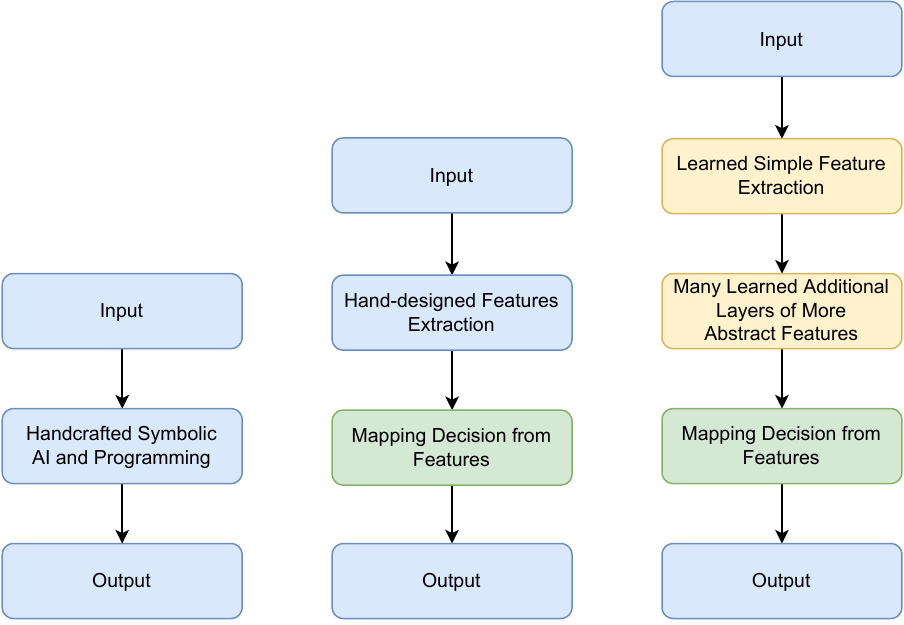}
	\caption{Flowchart showing a structured relationship between different AI subdivisions. The blue box indicates the hand-coded algorithms, the green ones indicate common parts between Deep Learning and Machine Learning, and the Yellow denotes components able to learn exclusive for Deep Learning.}
	\fautor
	\label{fig:flowchart_ia_parts}
\end{figure}

\citeonline{Goodfellow-et-al-2016} define Deep Learning as a type of Machine Learning technique that enables computer systems to improve with experience and data, building out a hierarchy of concepts on top of each other. Most complex concepts are deﬁned through their relation to simpler ones, represented as a deep graph with many layers. They also argue that machine learning is the only viable approach to building AI systems that can operate in complicated real-world environments. \autoref{fig:ai_ml_dl} shows Deep Learning as a subset of Machine Learning algorithms, which is, in turn, a subset of Artificial Intelligence algorithms, and \autoref{fig:flowchart_ia_parts} shows a structured relationship between different AI components.

There is one particular variation of \sigla{DFNN}{Deep Feedforward Neural Network} that has shown easier to train and generalize than fully connected networks \sigla{MLP}{Multilayer Perceptron}. It is the Convolutional Neural Network (CNN). They have achieved many practical successes, including computer-vision applications~\cite{LeCun:2015:Deeplearning}. Deep learning and CNNs have demonstrated applicable in many areas as visual perception, speech and audio processing, natural language processing, robotics, bioinformatics and chemistry, video games, search engines, online advertising, and ﬁnance~\cite{Goodfellow-et-al-2016}.

\subsection{Convolutional Neural Network}
\label{subsec:convolutional-neural-networks}

\begin{figure}[htb]
	\begin{subfigure}[b]{0.32\linewidth}
		\includegraphics[width=\linewidth]{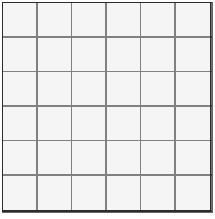}
		\caption{2D drid-like array.}
	\end{subfigure}%
	\hfill%
	\begin{subfigure}[b]{0.32\linewidth}
		\includegraphics[width=\linewidth]{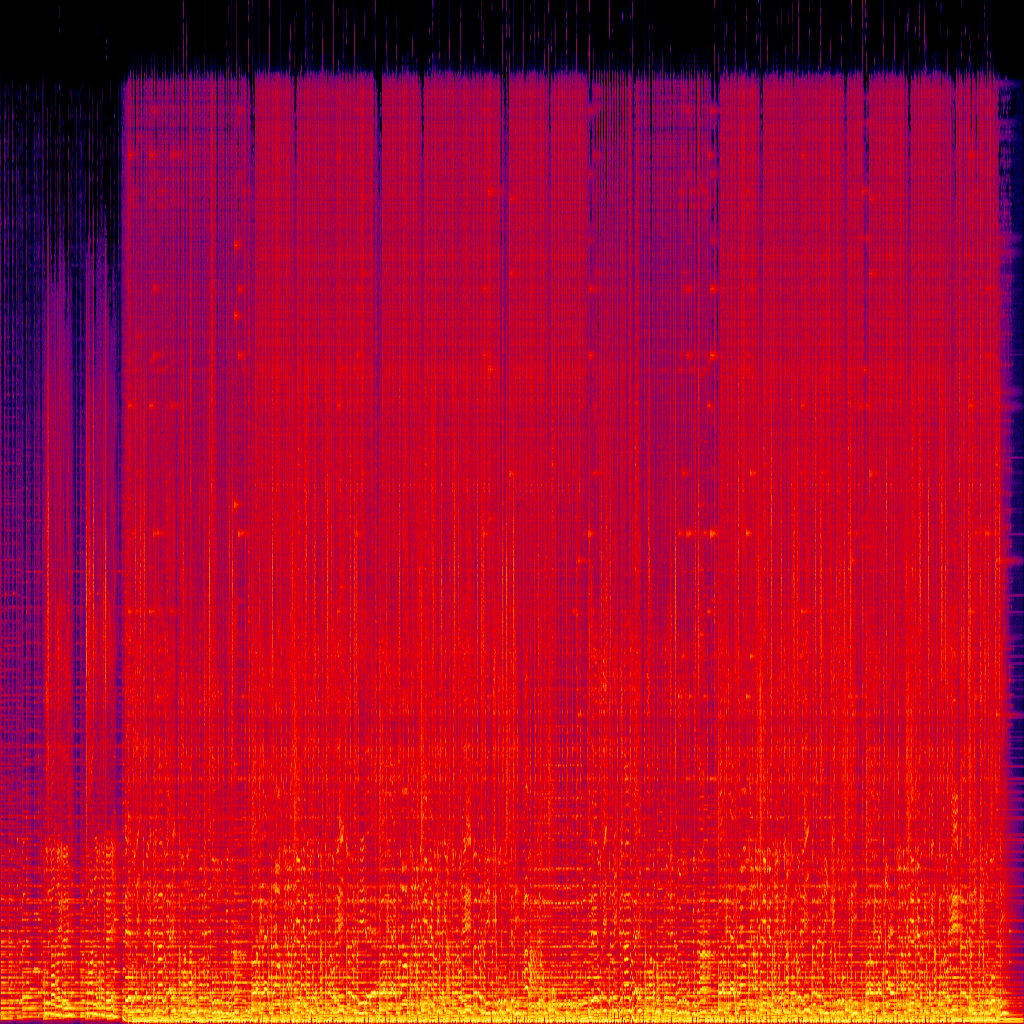}
		\caption{Audio Spectrogram.}
	\end{subfigure}
\hfill%
	\begin{subfigure}[b]{0.32\linewidth}
		\includegraphics[width=\linewidth]{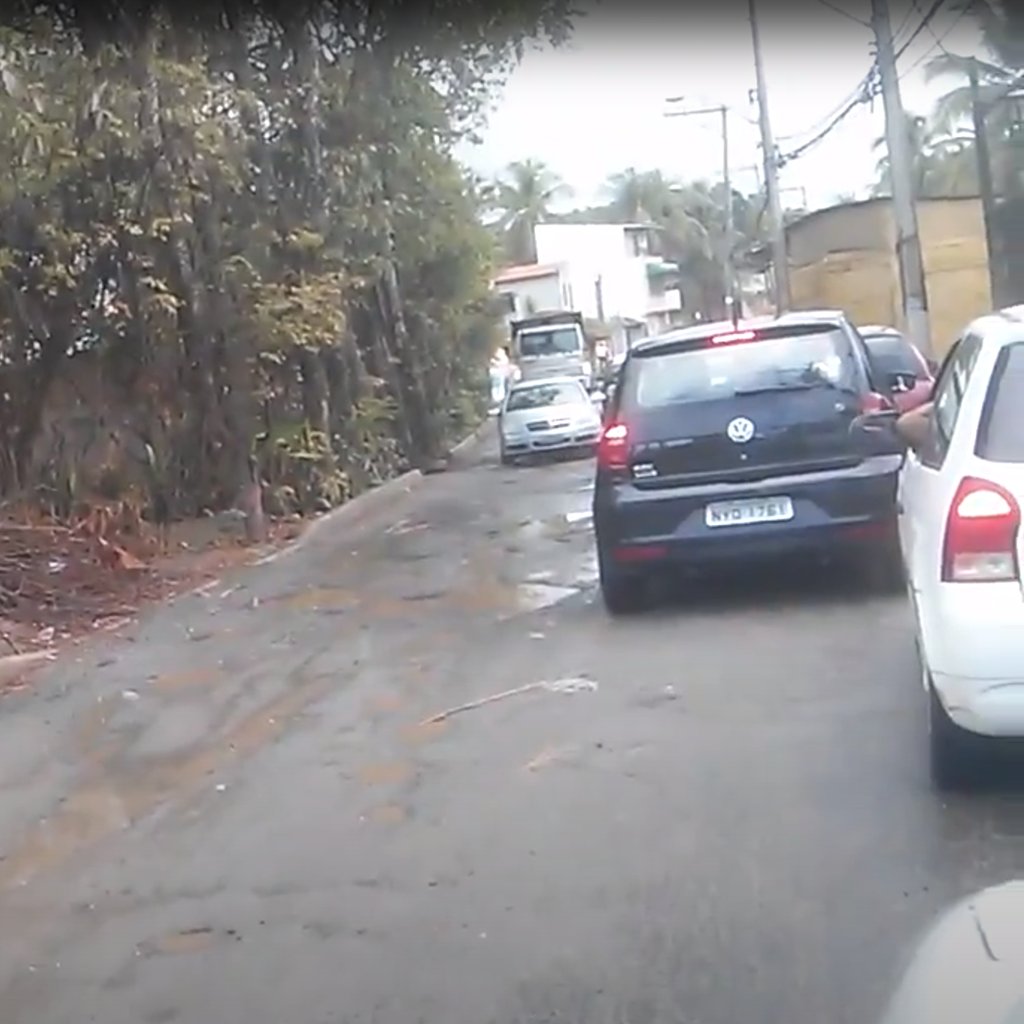}
		\caption{Image.}
	\end{subfigure}%
	\hfill%
	\begin{subfigure}[b]{\linewidth}
		\includegraphics[width=\linewidth]{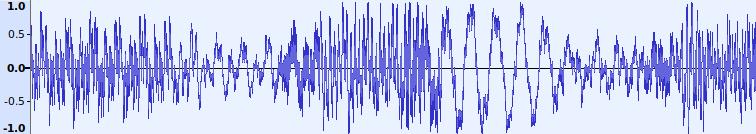}
		\caption{1D grid-like audio signal.}
	\end{subfigure}
	\caption{Grid-like input signals for Constitutional Neural Networks: (a) shows a generic 2D grid-like array, (b) presents a audio spectrogram display a spectro variation with the time for a audio signal, (c) shows a 2D image with the pixel values organized as matrix of values, and (d) brings a 1D audio signal.}
	\label{fig:cnn_inputs}
	\fautor
\end{figure}

The initial idea of convolution was proposed in Neocognitron by \citeonline{Fukushima1980}. However, the modern concept of CNN was presented in \citeonline{lecun:1998:ieee-gradient-based-learning} and after popularized in the DL context by \citeonline{Krizhevsky:2012:ICD:alexnet} with the architecture AlexNet. The CNNs are Neural Network architectures intended to process data in the form of multiple arrays, having a grid-like topology. Those arrays can be a 1-D grid, taking samples at regular time intervals like signals and sequences, including audio; a 2-D grid of pixels for images or audio spectrograms; or a 3-D for video or volumetric images~\cite{LeCun:2015:Deeplearning, Goodfellow-et-al-2016}. \autoref{fig:cnn_inputs} shows some examples of those input signals.

\begin{figure}[htb]
	\centering
	\includegraphics[width=\linewidth]{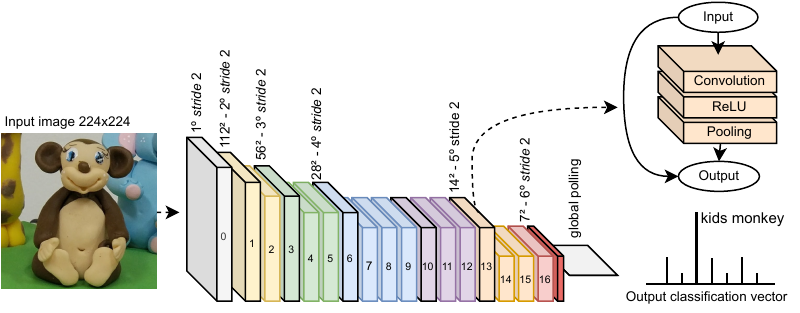}
	\caption{Typical CNN architecture with many stages composed of convolution, nonlinear activation ReLU, and pooling layers. 
}
	\fautor
	\label{fig:tipical_cnn_structure}
\end{figure}

The Convolutional Neural Network name indicates the discrete mathematical convolution operation employed by those networks. So CNN is the name of ANN that uses convolution operation in at least one of their layers~\cite{Goodfellow-et-al-2016}. The main idea behind CNNs is to take advantage of shared weights, pooling, use of many layers, and natural signals properties such as local connections~\cite{LeCun:2015:Deeplearning}. A typical CNN architecture is arranged as a sequence of stages composed of convolutional layers, a nonlinear activation function like ReLU, and pooling layers (\autoref{fig:tipical_cnn_structure}) -- in some cases also batch normalization and other variations of filter organization. In this example, a grid array of $224 \times 224 \times 3$ having three channels with RGB information is the 2-D image input. In this architecture, the initial input passes by many transformations being filtered (convolution layers) and nonlinear transformed (ReLu layers). Besides, it also suffers polling with strides 2 in some stages. The strides stand for the downsample level applied in the pooling stage.

\begin{figure}[htb]
	\centering
	\includegraphics[width=\linewidth]{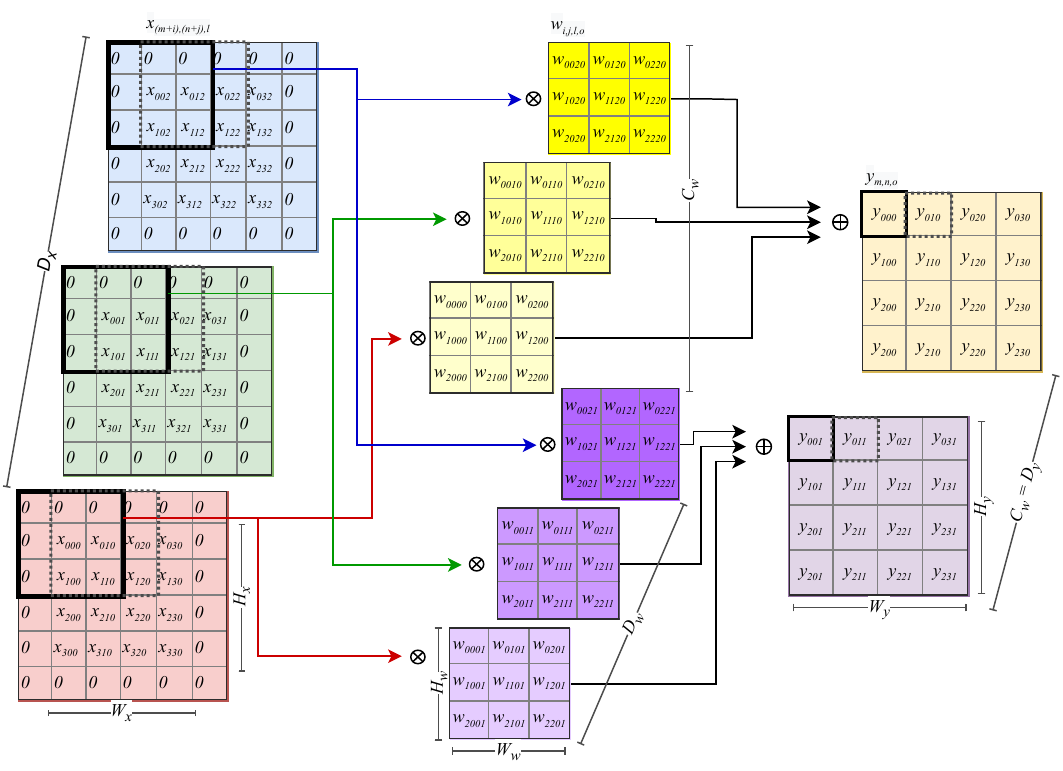}
	\caption{Convolution layer core operation mechanism. An input tensor $ X $ is convoluted per a tensor filter bank $ W $ (kernel) and generates an output feature map tensor $ Y $. The output  feature map (tensor $ Y $) is composed of data units $ y_{m,n,o} $ mapped through the weights data units $ w_{i,j,l,o} $ (tensor  $ W $) to the input feature map (tensor $ X $) data units $ x_{(m+i),(n+j),l} $.}
	\fautor
	\label{fig:cnn_core_operation}
\end{figure}

The convolutional layers organize the data units as feature maps where each unit is connected to both, previous and current layers through a set of weights. The weights are called filter banks or kernels. The resulting sum of filtering process is passed to the activation function to insert non-linearity to the transformation. In this process, all data units in a specific feature map share the same kernel, and different layers (different feature maps) have distinct filter banks~\cite{LeCun:2015:Deeplearning}. \autoref{fig:cnn_core_operation} shows a convolution operation example with a tensor\footnote{In the context of machine learning tensors are multi-dimensional arrays (a data structure) with a uniform data type. Their attributes are rank, shape, and data type.} $ X $ as an input being convoluted per a tensor filter bank $ W $ (kernel) and resulting in an output feature map as tensor $ Y $. In this process, the output tensor $ Y $ is composed of data units $ y_{m,n,o} $ mapped through the weights tensor units $ w_{i,j,l,o} $ to the previous feature map units $ x_{(m+i),(n+j),l} $ in the tensor $ X $. The process depicted in the \autoref{fig:cnn_core_operation} is described by the \autoref{eq:2d_conv}. 
In this exemple the input tensor $ X $ has dimensions $ H_{x} \times W_{x} \times D_{x} $, the weights tensor $ W $ has dimensions $ H_{w} \times W_{w} \times D_{w} \times C_{w} $, and the output tensor $ Y $ has dimensions $ H_{y} \times W_{y} \times D_{y} $, where the output depth $ D_{y} $ is directly related with the amount of kernel channels $ C_{w} $. The terms $ H_{w} $, $ W_{w} $, and $ D_{w} $ represent the dimensions (height, width and depth) of each kernel $ W $ channel.
In reality, the convolutional layer is implemented as a cross-correlation operation. It does not flip the kernels. With this implementation strategy, it loses the cumulative property, but this does not affect the result since the machine will learn the kernel values during the training process~\cite{Goodfellow-et-al-2016}.

\begin{equation} \label{eq:2d_conv}
Y(m,n,o)=(W*X)(m,n,o) = \sum_{i}^{H_{w}}\sum_{j}^{W_{w}}\sum_{l}^{D_{w}} X(m+i,n+j,l)W(i,j,l,o)
\end{equation}

\begin{figure}[htb]
	\begin{subfigure}[b]{0.45\linewidth}
		\includegraphics[width=\linewidth]{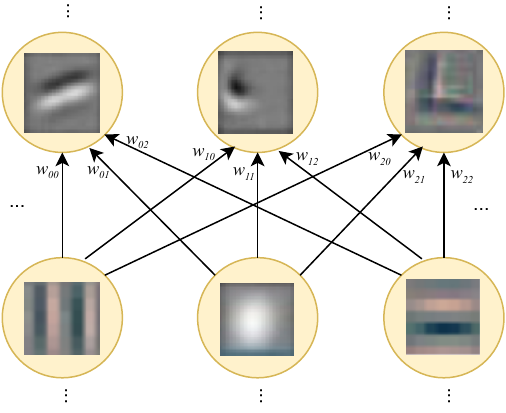}
		\caption{Fully connected.}
	\end{subfigure}%
	\hfill%
	\begin{subfigure}[b]{0.45\linewidth}
		\includegraphics[width=\linewidth]{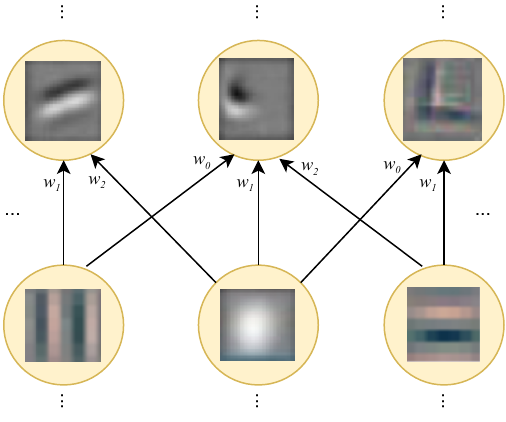}
		\caption{Locally connected.}
	\end{subfigure}
	\caption{Comparison of fully and locally connected layers.}
	\label{fig:fully_locally_connected_exemple}
	\fautor
\end{figure}

Comparing a fully connected layer with a convolutional layer is possible observe the second one acts as locally connected layer sharing the weights between them  (\autoref{fig:fully_locally_connected_exemple}). Share weight in this context is possible because in a specific feature map as a 2-D grid array like a image or 1-D like a audio signal, the set of values are correlated and presents local patterns easy detectable. Besides, the local statistics in those kinds of signals are invariant to location. Therefore, if a pattern can appear in one part of the image, it could appear anywhere~\cite{LeCun:2015:Deeplearning}.

Regarding the pooling layers, they are responsible for merging similar features into one -- doing the downsampling. It replaces the output of the convolutional layers at a location with a summary statistic of the nearby data units in the feature map. A typical pooling layer reports the maximum output within a rectangular neighborhood of the feature map. It also can take the average of a rectangular area or a weighted average
based on the distance from the central pixel. This process helps make the representation approximately invariant to small translations of the input data. This makes it easier to know if some input properties are there rather than worrying about its position ~\cite{LeCun:2015:Deeplearning,Goodfellow-et-al-2016}. Besides, it helps the subsequent convolutional layers make a global analysis of the context (large field of view) without increasing the kernel size.

\section{Related Works}
\label{sec:related-works}

This section shows the relevant works in the literature, grouped by theme. It also shows their relationships with this thesis.

\subsection{Perception Categorization and Architecture Strategies}
\label{subsec:perception-architecture}
\begin{figure}[htb]
	\begin{subfigure}[b]{0.48\linewidth}
		\includegraphics[width=\linewidth]{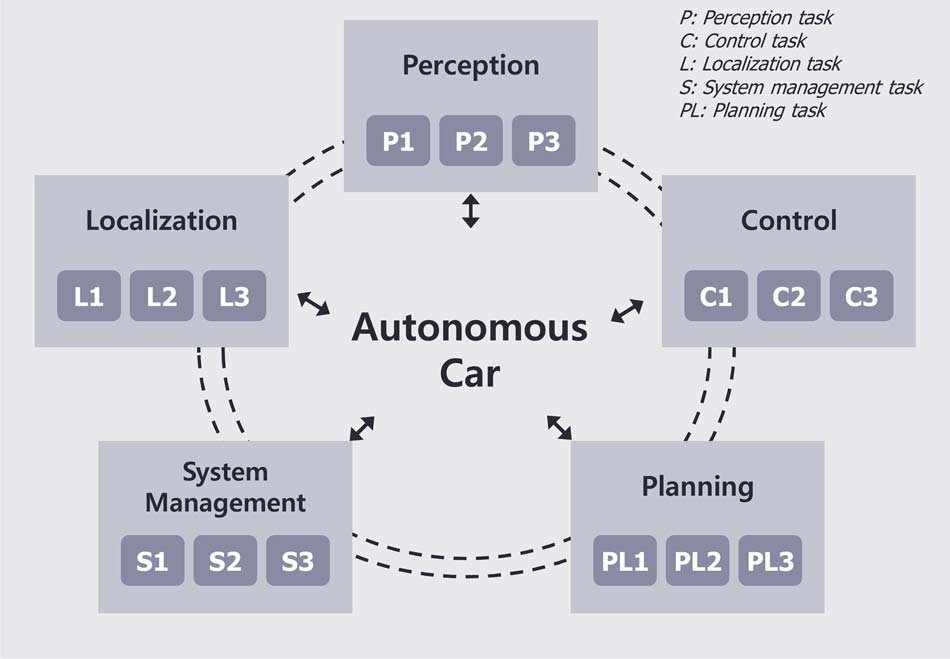}
		\caption{Components for AVs and ADAS.}
	\end{subfigure}%
	\hfill%
	\begin{subfigure}[b]{0.48\linewidth}
		\includegraphics[width=\linewidth]{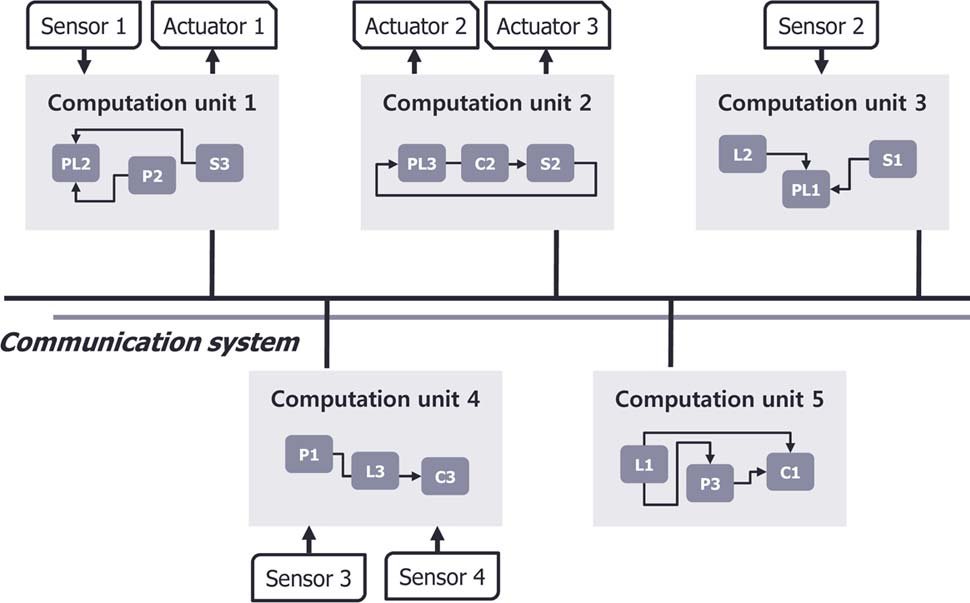}
		\caption{Distributed system architecture.}
	\end{subfigure}
	\caption{Distributed Components for AVs and ADAS.}
	\fdireta{jo:2014:car_i_distributed_arch}
	\label{fig:distributed-components-for-autonomous-vehicles}
\end{figure}

\citeonline{jo:2014:car_i_distributed_arch} and \citeonline{jo:2015:car_ii_distributed_arch} show possible strategies for AVs and ADAS architecture. They propose a methodology based on architectures for distributed systems containing the modules of perception, localization, planning, control, and system management. \autoref{fig:distributed-components-for-autonomous-vehicles} show this approach that aims at modularization, better fault tolerance, and reducing computational complexity. This thesis focuses on the study of the perception module.

\begin{figure}[htb]
	\includegraphics[width=\linewidth]{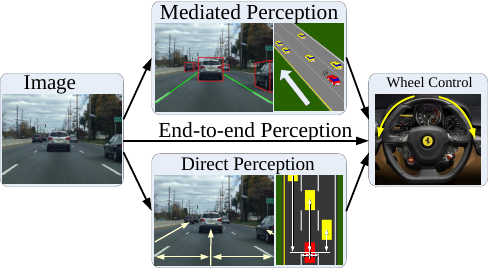}
	\caption{Perception paradigms.}
	\label{fig:paradigmas-de-percepcao}
	\fadaptada{chen:2015:deepdriving}
\end{figure}

Visual perception is a fundamental challenge to build Autonomous Vehicles or ADAS. Many works have studied that theme \cite{Kukkala:2018:IEEECEM,BADUE:2021:IARA}. \citeonline{brummelen:2018:autonomous-vehicle-perception} present a review of the state-of-the-art concerning perception in autonomous vehicles and \cite{chen:2015:deepdriving} categorized them in three categories as shown in \autoref{fig:paradigmas-de-percepcao}. Among them, Mediated Perception is the most common in researches. It only interprets sensor data to understand the scene while other modules such as planning and control perform the remaining system functionalities. Another category is the End-to-End Perception. It generates the control information to the vehicle straight from the data provided by the sensors. Finally, the Direct Perception~\cite{chen:2015:deepdriving}  approach maps the information received from the sensors into a set of key indicators related to the driving possibilities, given the current state of the track or traffic at the moment. The study presented in this thesis follows the Mediated Perception approach. This strategy provides observability for AVs and ADAS internal processing steps instead of delegating the whole task to a black-box algorithm.

\subsection{Backbones for Feature Extraction and Classification}
\label{subsec:backbones-and-classification}
\begin{figure}[htb]
	\includegraphics[width=\linewidth]{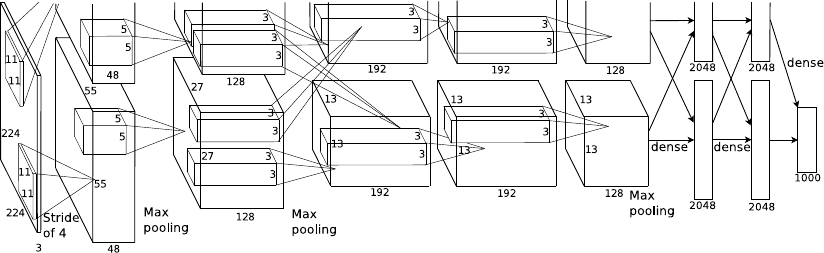}
	\caption{Convolutional Neural network AlexNet.}
	\fdireta{Krizhevsky:2012:ICD:alexnet}
	\label{fig:alexnet-architecture}
\end{figure}

Backbone refers to the section of the \sigla{DNN}{Deep Neural Network} responsible for doing the feature extraction of the inputs. 
Network backbones for feature extraction are the basis for building segmentation and detection architectures used in mediated perception systems. Most of them derive from the architectures developed for image classification. In this area, the AlexNet (\autoref{fig:alexnet-architecture}) proposed in \citeonline{Krizhevsky:2012:ICD:alexnet} was the main responsible for attracting the attention of the computer vision field to Deep Learning algorithms. This phenomenon has occurred after they won the \sigla{ILSVRC}{ImageNet Large Scale Visual Recognition Challenge} \cite{russakovsky:2015:imagenet}, establishing a new benchmark threshold for the competition.

\begin{figure}[htb]
	\centering
	\includegraphics[width=.81\linewidth]{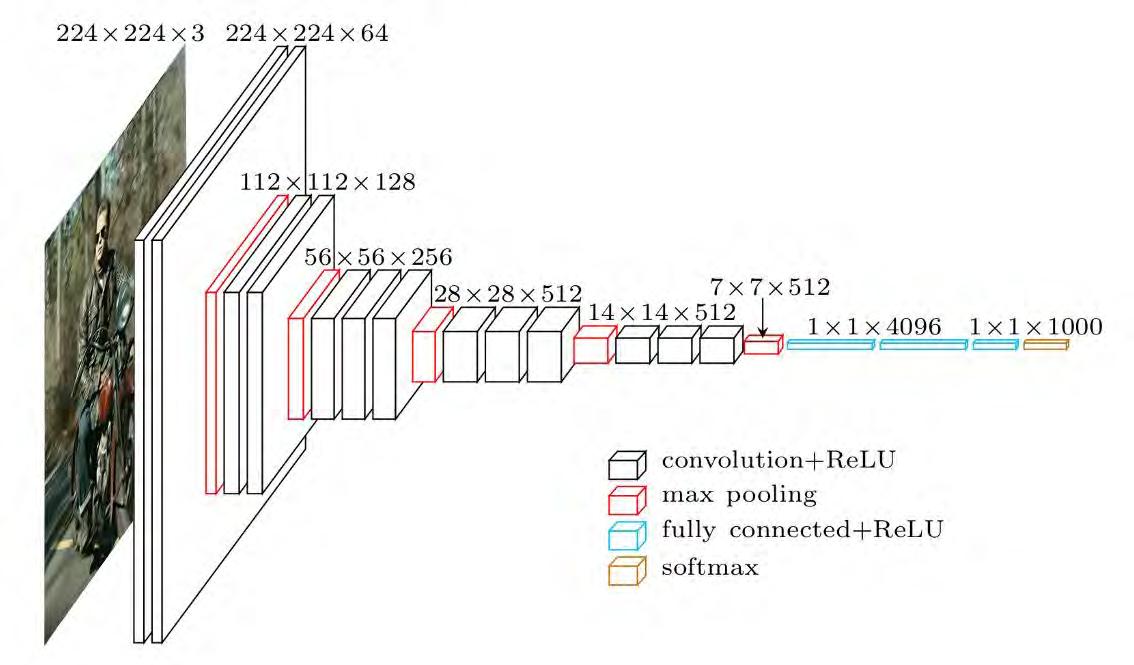}
	\caption{Convolutional Neural network VGG.}
	\fdireta{Chen:2018:vggimage}
	\label{fig:vgg-architecture}
\end{figure}

The ImageNet challenge became a reference to measure the capacity of CNNs and, consequently, of backbones for feature extraction. In 2014, VGG~\cite{simonyan:2015:vgg} was the champion in localization and second in classification task at ILSVRC2014. \autoref{fig:vgg-architecture} shows the VGG architecture. This network uses regular structures with a filter size of $3\times3$ with the number of filters layers doubled for each level. They carried out a study to characterize the relationship between the depth of the network and its accuracy considering a fixed filter dimension. They presented two models, one with 16 and others with 19 layers. While ZFNet~\cite{matthew:2014:zfnet:win-imagenet2013-classification} and OverFeat~\cite{sermanet:2014:overfeat:win-imagenet2013-localization} used different filter sizes to improve network performance, VGG set it to $3\times3$ to exploit the impact of depth in the performance. This idea influenced several subsequent works, and various activities such as segmentation and detection have used this backbone for features extraction. Therefore, our proposed CMSNet has integrated support for this backbone. 

\begin{figure}[htb]
	\begin{subfigure}[b]{0.48\linewidth}
		\includegraphics[width=\linewidth]{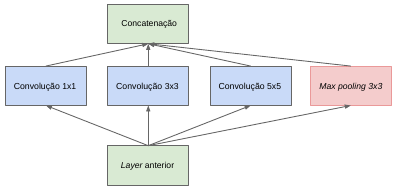}
		\caption{Inception block.}
	\end{subfigure}%
	\hfill%
	\begin{subfigure}[b]{0.48\linewidth}
		\includegraphics[width=\linewidth]{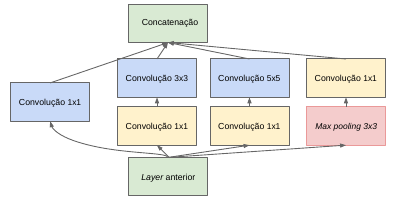}
		\caption{Inception block with channel compression.}
	\end{subfigure}
	\caption{Inception blocks.}
	\fadaptada{Szegedy:2015:inception-win-imagenet2014}
	\label{fig:inceptionv1-architecture}
\end{figure}

GoogLeNet \cite{Szegedy:2015:inception-win-imagenet2014} was the network winner of the ILSVRC2014 classification challenge. Their architecture improved the computational resources efficiency against predecessors. Although it had 22 layers, more than twice as much as AlexNet, the number of parameters was 12 times less, even though they achieved performance close to the human on the classification task. They improved accuracy by implementing modules composed of filters with different sizes ($1\times1$, $3\times3$ and $5\times5$) to operate in parallel, increasing the ability to evaluate distinct fields of view. \autoref{fig:inceptionv1-architecture} shows these blocks named Inception. Although larger filters are computationally costly, they achieved efficiency by inserting $1\times1$ convolutions to reduce the number of channels before the largest convolutions. They have optimized the model, including some innovations such as residual connections and Batch Normalization to improve accuracy, reduce the demand for computational resources \cite{Szegedy:2016:inceptionv2} and training time \cite{Szegedy:2017:III:Inceptionv4}.

ResNet \cite{He:2016:IEEECVPR-resnet} is another relevant architecture. It was the first network to beat human performance in the ImageNet Challenge. They have built the architecture with 152 layers (8 times deeper than the VGG-19). Before their work, there were severy degradation problems when trying to train deeper networks. The increased number of layers had used to degrade network performance. To solve this, the ResNet authors proposed residual blocks with shortcuts linking inputs to outputs through addition. This solution allowed ResNet to achieve an error rate of 3.57\% on the Top-5. Due to the relevance of sch architecture, its backbone also was included in our CMSNet.

\begin{figure}[htb]
	\begin{subfigure}[b]{0.48\linewidth}
		\includegraphics[width=\linewidth]{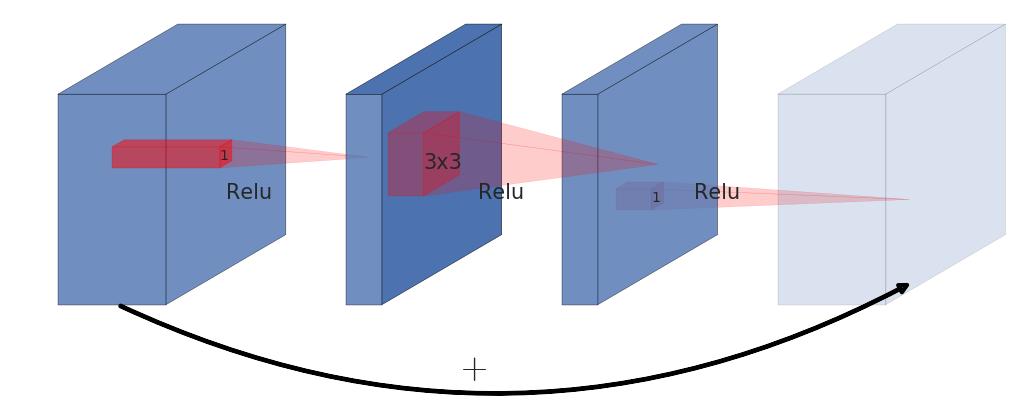}
		\caption{Residual block.}
	\end{subfigure}%
	\hfill%
	\begin{subfigure}[b]{0.48\linewidth}
		\includegraphics[width=\linewidth]{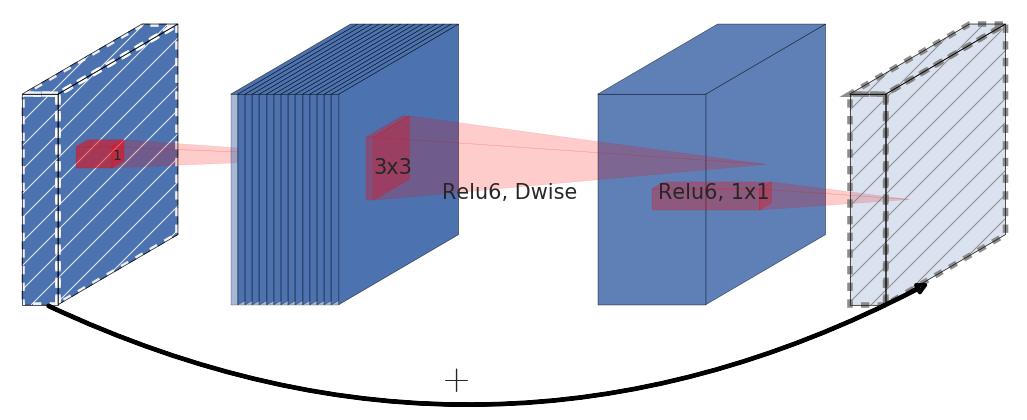}
		\caption{Inverted residual block.}
	\end{subfigure}
	\caption{Difference between residual and inverted residual block.}
	\label{fig:residual-vs-inverted}
	\fadaptada{Sandler:2018:MobileNetV2:IEEE-CVF}
\end{figure}

In addition to accuracy, in applications with real-time execution demand, the number of parameters and operations performed during inference is relevant when choosing the backbone for extracting features that will compose the solution. In this sense, the MobileNetV2 architecture \cite{Sandler:2018:MobileNetV2:IEEE-CVF} presents great efficiency. Like its predecessor \cite{Andrew:2017:MobileNets:MobileNets}, MobileNetV2 uses convolution factored in depthwise (1-channel depth convolution) and pointwise ($1\times1$ convolution). In addition, they propose a scalable solution both in input resolution and in the number of channels per layer. Furthermore, it adopts a structure based on residual blocks plus compression of the number of channels. In the inverted residual blocks, or bottleneck, the inputs have their number of channels expanded through pointwise convolution ($1\times1$), data pass by a depthwise convolution, and the number of channels is compressed again through a pointwise convolution. This block also adds the inputs to the outputs, performing a data shortcut as in \citeonline{He:2016:IEEECVPR-resnet}. \autoref{fig:residual-vs-inverted} shows the two types of blocks. Under similar conditions and the same image resolution, MobileNetV2 achieves similar accuracy results as VGG-19, requiring only 3.4 million parameters compared to 144 million parameters used by VGG. This backbone is present in CMSNet and is the main one used for feature extraction in the development of this thesis.

As well as MobileNet, other works such as ShuffleNet~\cite{Zhang:2018:ShuffleNet:IEEE-CVF}, NasNet-A~\cite{Zoph:2018:NasNet:IEEE-CVF}, MNasNet~\cite{MNasNet:Tan:2019:CVPR}, and EfficientNet~\cite{EfficientNet:pmlr:2019:tan19a}, also explore aspects related to the construction of architectures aimed at real-time applications. In addition, NasNet, MNasNet, EfficientNet, and SENet~\cite{Hu:2018:SENet:IEEE-CVF} have performance in line with the state of the art when configured to achieve maximum accuracy. Besides, recent works such as \citeonline{Dosovitskiy:2021:iclr} and  \citeonline{pmlr-v139-d-ascoli21a} have successfully introduced the use of transforms architecture for vision problem.

\subsection{Semantic Segmentation}
\label{subsec:semantic-segmentation}
Regarding semantic segmentation, the \citeonline{Long:2015:FCN:ieeecvpr} proposed the FCN architecture showing how to convert classification networks \cite{Krizhevsky:2012:ICD:alexnet,Szegedy:2015:inception-win-imagenet2014} into segmentation ones. They had achieved 20\% improvement over previous work in the PASCAL VOC benchmark. 

\begin{figure}[htb]
	\includegraphics[width=\linewidth]{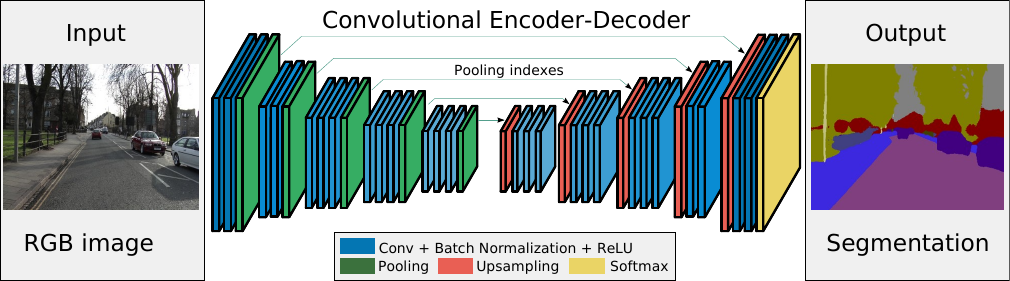}
	\caption{SegNet segmentation architecture.}
	\fadaptada{Badrinarayanan:2017:segnet:ieee-TPAMI:}
	\label{fig:segnet-architecture}
\end{figure}

There is also an encoder-decoder architecture proposed in \citeonline{Badrinarayanan:2017:segnet:ieee-TPAMI:} that applies fully convolutional networks for pixel-level classification. In that architecture, the encoder extracts the features while the decoder generates the segmentation masks. \autoref{fig:segnet-architecture} shows the SegNet architecture. They have used the VGG-13 backbone to feature extraction. SegNet's main innovation was how the decoder expands the low-resolution feature using the max-pooling\footnote{Max-pooling is a network layer responsible for decreasing the resolution of feature blocks.} index to guide the resampling of data at the generation of segmentation mask, eliminating the need for the filter to learn the best way to resample the data. On the other hand, the PSPNet, proposed by \citeonline{Zhao:2017:PSPNet:ieeecvpr}, was responsible for applying the spatial pyramid pooling module in semantic segmentation to explore the global and regional context of the information contained in the images. That work was responsible for reaching the state-of-the-art accuracy of 85.4 \% in the PASCAL benchmark. The spatial pyramid pooling is another module available in CSMNet.

\newword{Max-pooling}{refers to a network layer responsible for decreasing the resolution of feature blocks.}

\begin{figure}[htb]
	\includegraphics[width=\linewidth]{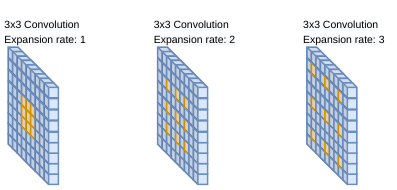}
	\caption{Dilated filter with different expansion rates over a feature map.}
	\label{fig:atrous}
	\fautor
\end{figure}

Furthermore, the work proposed by \citeonline{Chen:2018:deeplab:ieeeTPAMI} applied atrous convolution (\autoref{fig:atrous}) on pixel-level classification, so allowing to increase the feature processing resolution (the field of view) and to keep the size of the filters stable. Such work was also responsible for proposing Atrous Spatial Pyramid Pooling (ASPP) to help perceive the context in images at different scales. This architecture managed to reach the mark of 79.7 \% $mIoU$\footnote{Mean Intersection over Union.} in the PASCAL dataset for semantic segmentation, and was updated to improve its accuracy in \citeonline{DeepLabV3:DBLP:journals/corr/ChenPSA17,Chen:2018:EncoderDecoderWA:ECCV:deeplabv3Plus}. Spatial Pyramid Pooling (SPP) and Atrous Spatial Pyramid Pooling (ASPP) are both technics used on our proposed CMSNet. Besides such works, some recent researches have also presented innovations to try to improve accuracy by either maintains high-resolution representations through the CNN processing \cite{Yu:2021:CVPR:lite-hrnet,WU:2021:HRNet} or use transformer encoder-decoder architecture \cite{Carion:2020:ECCV:DETR}.

\subsection{Off-Roads Segmentation}
\label{subsec:off-road-segmentation}
Although semantic segmentation is a hot topic on visual perception and scene understanding for AVs and ADAS, there is still a gap in off-road environments and adverse visibility conditions. The work proposed by \citeonline{Real-Time-Semantic-Off-Road:Maturana:2018} is the one that most resembles the proposal of this research. In the same way as this research, they used RGB cameras and convolutional networks to distinguish the road limits and obstacles. They also built a dataset, but it is not available, avoiding any comparisons. Another similar aspect was the inference time concern for embedding the application with the segmentation occurring in real-time. There is also the work \citeonline{Semantic-Forested:Valada:2016} who proposed architecture and a dataset for the forest environment. Both \citeonline{Semantic-Forested:Valada:2016} and \citeonline{Real-Time-Semantic-Off-Road:Maturana:2018} evaluated their systems only in environments where there is a relative difference in texture/color on the track limits -- green vegetation with a sand track. Even considering those works, there are gaps with off-road environments where both parts are composed of sand having the same color, and there is a lighter difference between what is or not the region in which the car may pass through. There is also a gap of works in the literature investigating the before-mentioned condition mixed with unpaved roads and visibility adversity, including night, rainy and dusty. So, the study presented in this Ph.D. thesis intends to cover that investigation gap, and to do that, it proposes the CMSNet framework.

\subsection{AV and ADAS Datasets}
\label{subsec:av-adas-datasets}
Additionally, datasets are a key step to help the system learning how to solve the problem in Supervised Learning theory. However, most of the open datasets published in the literature aim for urban environments. One of the first published datasets for ADAS and AVs vision perception was CamVid \cite{BROSTOW200988,brostowSFC:2008:ECCV08-camvid}. It has 32 classes, images captured in a well-paved urban environment from the driver's perspective with more than 10 minutes of video collected at 30Hz and 700 high-quality images manually labeled at 1 Hz. Another one is the Kitti dataset \cite{kitti-geiger:2012:cvpr, kitti-geiger:2013:ijrr, kitti-fritsch:2013:itsc, kitti-menze:2015:cvpr} that contains several benchmarks, including semantic segmentation for urban roads and 3D object detection. Also, regarding paved urban environments, one of the most important datasets for semantic segmentation is Cityscapes \cite{cordts:2015:cvprw-cityscapes}. It contains stereo video sequences captured in 50 different cities with pixel-level labeling. Altogether there are 5,000 images precisely labeled. There are also have others recent datasets published for such conditions as \citeonline{Sun:2020:CVPR:WaymoOpenDataset}. To address that data gap for the unpaved and off-road environments in adverse visibility conditions, the researchers have built an off-road test track and proposed a new dataset covering this kind of environment in such visibility conditions to support the research in those circumstances.


\chapter{System Design and Methodology}
\label{chapter:design-and-methodology}

This chapter presents the methodology and the design of the thesis research. Section \ref{sec:methodology} describes the methods followed by this research, and \autoref{sec:perception} describes the design, going from the proposed CMSNet framework (\autoref{subsec:cmsnet})  until the development of the dataset used (\autoref{subsection:kamino_dataset}). Finally, \autoref{sec:experimental_setup} includes the experimental setup.

\section{Methodology}
\label{sec:methodology}
The main objective of this research is to contribute to the comprehension of how visual perception formulated as a Deep Supervised Learning problem for semantic segmentation can behave on unpaved roads and off-road environments, open-pit mines, and agriculture industries, even under adverse visibility conditions such as rainy, dusty, and night. 

\begin{figure}[!htb]
	\includegraphics[width=\linewidth]{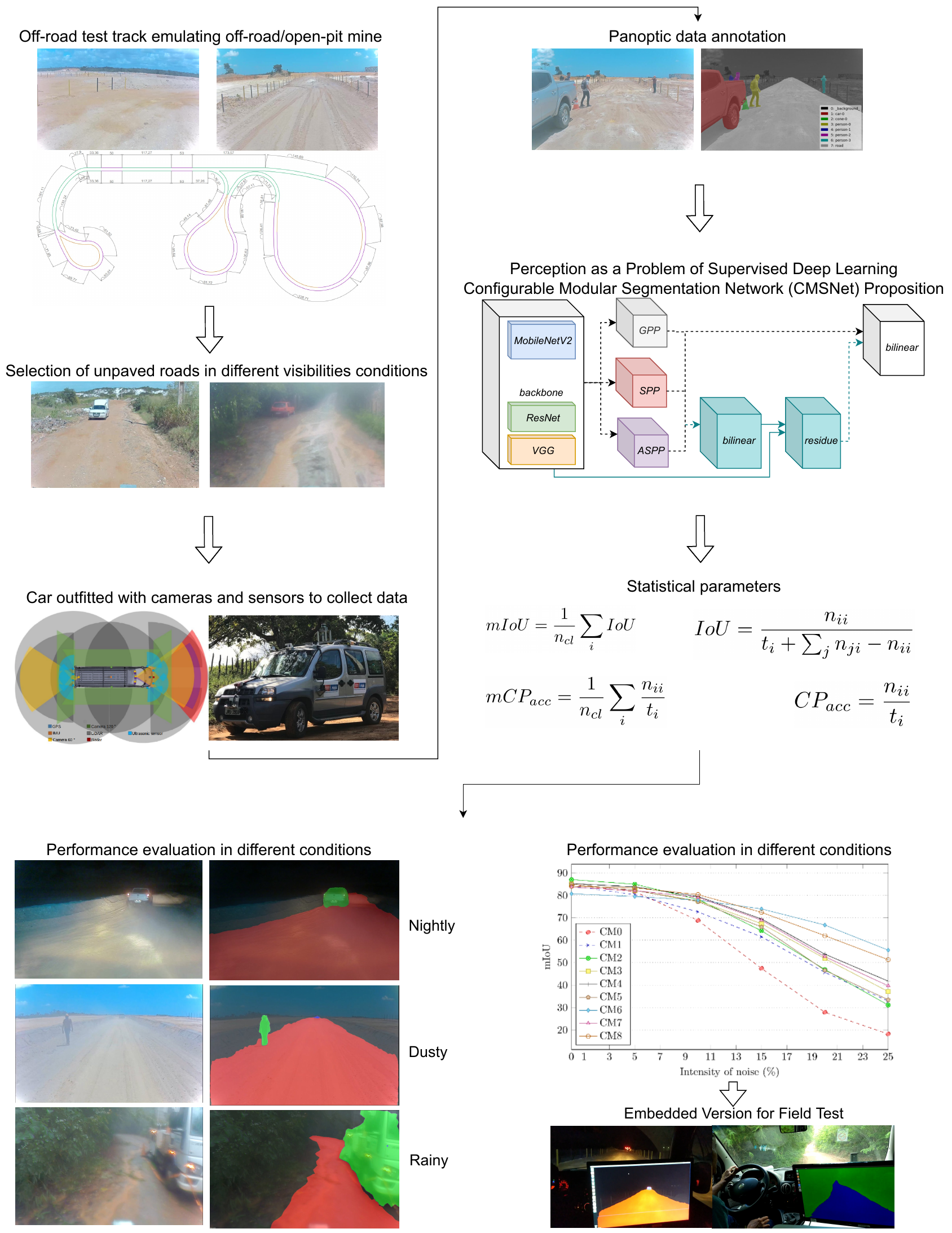}
	\caption{Graphical illustration of the methodology followed in the research.}
	\label{fig:graphical_methodology}
	\fautor
\end{figure}

The proposed methodology is illustrated in the \autoref{fig:graphical_methodology}. The researchers developed an off-road test track emulating open-pit mine environments and agricultural zones, where trucks and buses traffic to transport the industrial production and workers. Together with the test track, unpaved urban and rural roads also were selected and used as scenarios to collect data. A car was outfitted with cameras and sensors, so driven by those selected environments filming in different conditions, including night, day, sunset, rainy, and dusty. Then the recorded video was subsampled to generate fewer images per second, so they were carefully panoptic annotated. The resulting dataset has been clustered into a few subsets according to the environment and visibility condition. The subsections~\ref{subsec:kamino_project} and~\ref{subsection:kamino_dataset} detail the dataset construction.

To accomplish the research objective regarding Deep Learning models and conduct the experiments, the researcher has created a framework (CMSNet) encompassing the main segmentation structures to make it possible to generate and test different segmentation architecture arranges (\autoref{subsec:cmsnet}). Those architectures were trained with our train set and evaluated with the test set to verify their efficiency under several visibility conditions on detecting traffic zones and obstacles in off-road environments and see how different architecture strategies impact the performance. The architectures generated by CMSNet also have their accuracy evaluated under progressive severity on visibilities to estimate how the impairments affect the perception modules. Besides, this work tested the performance of systems developed for urban environments and compared our results with systems designed for forest environments.

The main method used to calculate the segmentation accuracy was the Jaccard similarity coefficient~\cite{jaccard1908nouvelles,YU201882}. It is one of the most used statistic in semantic segmentation state-of-the-art to quantify similarities between sample sets~\cite{Taghanaki2022}. This work has used the average of the similarities between different elements in the image and between all images. On the other hand, 
to compare the models computational performance, the study used the frames per second average and standard deviation. The \autoref{sec:experimental_setup} details the experimental setup.

\section{Visual-based perception in off-road environments}
\label{sec:perception}

Within the scope of this research was used semantic segmentation to find obstacles and the track limits where the car can pass through on unpaved roads and off-road environments in different visibility conditions. Semantic segmentation is the task that assigns classification at the pixel level by grouping them as belonging to the same object. The advantage of this approach is that in addition to segmenting the road limits, it can also discover and segment obstacles on the road, eliminating in some cases the use of a second network for object detection.

\subsection{CMSNet}
\label{subsec:cmsnet}
\begin{figure}[htb]
	\begin{center} 
		\includegraphics[width=\linewidth]{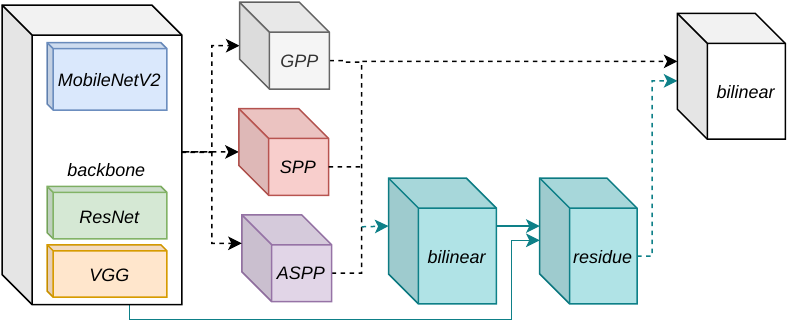} 
	\end{center} 
	\caption{CMSNet framework.} 
	\label{fig:modular-configurable-architecture} 
	\fautor
\end{figure}

The CMSNet is the framework proposed by this research to make it possible to configure different arrangements combining modules found in the state-of-the-art for semantic segmentation. It intended to compose different architecture variations, making it possible to test those innovations and compare the latency and accuracy results achieved for each arrangement.~\autoref{fig:modular-configurable-architecture} shows the components of proposed CMSNet framework. It is capable of operating with different backbones
for feature extraction. It can be configured by parameters to operate with the backbones MobileNetV2~\cite{Sandler:2018:MobileNetV2:IEEE-CVF}, ResNet \cite {He:2016:IEEECVPR-resnet} or VGG~\cite{simonyan:2015:vgg}. It also supports output strides of 8 or 16,  pyramid modules GPP, SPP, or ASPP, and shortcuts when the output stride is 16.

\subsubsection{Backbone}
Choosing the backbone suitable for the target application is an important step. There are architectures capable of achieving accuracy above 90\% in the Top-1 on the ImageNet benchmark~\cite{russakovsky:2015:imagenet}. However, when building a perception system based on Deep Learning for real-time inference, it is necessary to consider the latency of the backbone in addition to accuracy. Thus, the choice of network for extracting features must respect architectural aspects that offer a trade-off between accuracy and latency. The authors of this work have chosen the MobileNetv2 architecture~\cite{Sandler:2018:MobileNetV2:IEEE-CVF} as the main backbone for feature extraction as it demands low computational power compared to other architectures in the same level of accuracy. Besides, the CMSNet also supports ResNet and VGG as backbone~\cite{simonyan:2015:vgg, He:2016:IEEECVPR-resnet}.

The \citeonline{Sandler:2018:MobileNetV2:IEEE-CVF} architecture, in its standard version, has 3.5 million parameters and has a computational cost of 300 million of Multiply–accumulate (MAC) operation. It uses Depthwise Separable Convolutions and a residual block structure with a bottleneck. This thesis study has slightly modified it by removing the latest convolution and pooling layers. Such change decreased the total number of parameters from 3.5 million to 1.84 million — approximately 48\% fewer parameters.~\autoref{tab:mobilenet_architecture_os16} shows the final configuration for output strides of 16 and 8 (OS16 and OS8), where \textit{{h}} is the height, \textit{{w}} is the width, \textit{{c}} is the number of channels, \textit{{e}} is the expansion factor for each block, \textit{{d}} is the input dimension, \textit{{n}} indicates the block repetition, and \textit{{s}} defines the stride.

\begin{table}[htb]
    \centering
    \scriptsize
	\caption{Adapted MobilenetV2 architecture for OS16 and OS8, where \textit{{h}} is the height, \textit{{w}} is the width, \textit{{c}} is the number of channels, \textit{{e}} is the expansion factor for each block, \textit{{d}} is the input dimension, \textit{{n}} indicates the block repetition, and \textit{{s}} defines the stride.}
	\label{tab:mobilenet_architecture_os16}
	\begin{tabular*}{\textwidth}{@{\extracolsep{\fill} } r r r r r c c c c c}
	    \specialrule{.1em}{.05em}{.05em}
		\multicolumn{2}{c}{\bfseries OS16} & \multicolumn{2}{c}{\bfseries OS8} & \multicolumn{1}{c}{\multirow{2}*{\bfseries \textit{c}}} &  \multirow{2}*{\bfseries Operation} & \multirow{2}*{\bfseries \textit{e}} & \multirow{2}*{\bfseries \textit{d}} & \multirow{2}*{\textit{n}} & \multirow{2}*{\bfseries \textit{s}} \\ 
		\cline{0-3} 
		\multicolumn{1}{c}{\bfseries \textit{h}} & \multicolumn{1}{c}{\bfseries \textit{w}} &\multicolumn{1}{c}{\bfseries \textit{h}} & \multicolumn{1}{c}{\bfseries \textit{w}} &  &  &  &  &  &  \\
		\specialrule{.1em}{.05em}{.05em} 
		 483&769& 483&769&   4   & conv2d              & - & 32   & 1 & 2 \\ 
		 242&385& 242&385&  32  & \textit{bootleneck} & 1 & 16   & 1 & 1 \\ 
		 242&385& 242&385&  16  & \textit{bootleneck} & 6 & 24   & 2 & 2 \\ 
		 121&192& 121&192&  24  & \textit{bootleneck} & 6 & 32   & 3 & 2 \\ 
		  61& 97&  61& 97&  32  & \textit{bootleneck} & 6 & 64   & 4 & 2 \\ 
		  31& 49&  61& 97&  64  & \textit{bootleneck} & 6 & 96   & 3 & 1 \\ 
		  31& 49&  61& 97&  96  & \textit{bootleneck} & 6 & 160  & 3 & 1 \\ 
		  31& 49&  61& 97& 160  & \textit{bootleneck} & 6 & 320  & 1 & 1 \\ 
		\specialrule{.1em}{.05em}{.05em}
	\end{tabular*}
	\fautor
\end{table}

\subsubsection{Semantic segmentation architecture}
In addition to the backbone for extracting features, it is necessary to build structures responsible for performing the core activity — e.i., carrying out the pixel-level classification. There are several network architecture proposals for semantic segmentation~\cite{Long:2015:FCN:ieeecvpr, Zhao:2017:PSPNet:ieeecvpr, Chen:2018:EncoderDecoderWA:ECCV:deeplabv3Plus}. These architectures present significant and complementary contributions, and different solutions can be proposed and tested by combining them. Those arrangements were the basis for the configurable modular framework (CMSNet) proposed and developed in this research. The CMSNet allows several configurations by enabling or removing some structures as described in the following subsections.

\subsubsection{Shortcut}
\label{subsection:shotcut-strategy} 
\begin{figure}[htb] 
	\begin{center} 
		\includegraphics[width=0.75\linewidth]{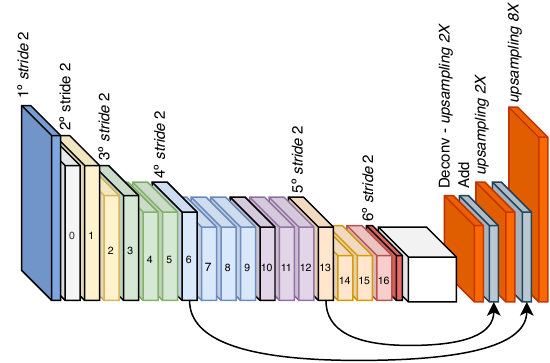} 
	\end{center} 
	\caption{Shortcut strategy.} 
	\label{fig:shortcut} 
	\fautor
\end{figure}

In architectures for segmentation, the latest step usually is responsible for generating the segmentation mask in an appropriate size. That result is achieved by upsampling the activation map output on the last layer of a network. Some works use transposed convolution (deconvolution) to perform interpolation and generate the output image. Instead of using linear interpolation with fixed parameters, this layer can learn the best way to interpolate the output producing the most suitable segmentation mask. The upsampling process can be done in a single step or using multiple ones to improve detailing. When performed in more than one stage, a shortcut adds the most external features to the result of the previous resizing (\autoref{fig:shortcut}). That strategy is responsible for improving the resolution details on the segmentation mask~\cite{Long:2015:FCN:ieeecvpr}. Shortcuts are one of the options available in the configurable architecture CMSNet proposed in our research. It can be enabled or disabled on its configurations.

\subsubsection{Scene analysis by Spatial Pyramid Pooling (SPP)}
Although fully convolutional networks~\cite{Long:2015:FCN:ieeecvpr} performed well in semantic segmentation, they have difficulty taking into account the global context during the analysis of each pixel. This difficulty can lead to incorrect classification as it does not consider the appropriate relationships between classes, e.i., confusing pixels of the track with the background since both contain sand of the same color. As a solution for that problem, the CMSNet framework implements a pyramid pooling structure. 

\begin{figure}[htb]
	\begin{center} 
		\includegraphics[width=0.75\linewidth]{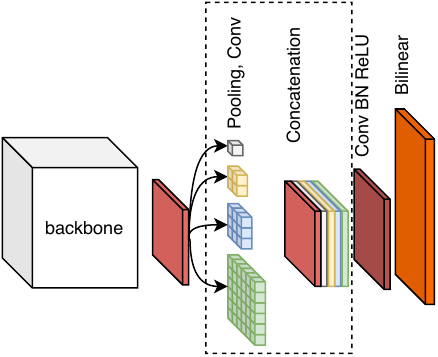} 
	\end{center} 
	\caption{Spatial Pyramid Pooling Module} 
	\label{fig:pyramid-pooling-module} 
	\fautor
\end{figure}

The spatial pyramid pooling is a module formed by a pyramid of pooling layers, followed by convolution and concatenation~\cite{Zhao:2017:PSPNet:ieeecvpr} (\autoref{fig:pyramid-pooling-module}). It is capable of providing scene analysis at different scales, allowing to infer the contribution of global or local context in the classification of each pixel and mitigating the consequences of the lack of context analysis found in \citeonline{Long:2015:FCN:ieeecvpr}. In this module, each pooling comes with a pointwise convolution having $d/N$ filters where N represents the size of the pooling, and $d$ denotes the input channels at the convolution. The CMSNet  SPP uses four average pooling with different compressions rates: the first is global pooling, the second is 1/2 of the resolution of the features, the third is 1/3 of the height of the features block, and the fourth is 1/6 of the resolution block. All of these values are concatenated with the original data to go through another convolution, so generates the segmentation map (\autoref{fig:pyramid-pooling-module}).

\subsubsection{Dilated convolution}
\label{subsection:atrous-convolution}
The standard convolution followed by pooling is a common building block in CNNs. It increases the output stride and reduces the size of the feature map on the output of the deepest layers of the network. That is interesting to enlarge the field of view of the filter and improve the ability to observe the context without the need for larger filters that increase the computational cost. However, narrowing the feature map through consecutive strides is harmful to semantic segmentation due to the loss of spatial information in the deeper layers of the backbone. A solution to this problem may be the use of atrous convolution (\autoref{fig:atrous}), which allows keeping the size of the feature map (stride) constant and arbitrarily control the field of view without increasing the number of network parameters or computational cost \cite{Chen:2018:deeplab:ieeeTPAMI,DeepLavV1:CRFs:2015:ChenPKMY14}. 

Our CMSNet always uses atrous convolution. However, it is possible to configure to start from the 4th or 5th stride pooling generating outputs stride of 16 or 8 respectively  — e.i., 1/16 or 1/8 of the size of the input image.

\subsubsection{Atrous Spatial Pyramid Pooling (ASPP)}
\label{subsection:atrous-pooling}
\begin{figure}[htb] 
	\begin{center} 
		\includegraphics[width=0.75\linewidth]{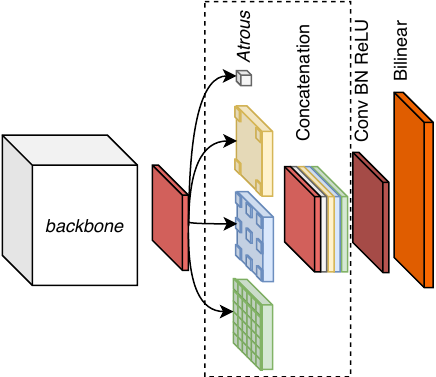} 
	\end{center} 
	\caption{Atrous Spatial Pyramid Pooling Module} 
	\label{fig:atrous-pyramid-pooling-module} 
	\fautor
\end{figure}

Just like the SPP module (\autoref{fig:pyramid-pooling-module}), it is also possible to improve the understanding of the global and local context of the scene through the application of a pyramid module formed by dilated convolution (\autoref{fig:atrous-pyramid-pooling-module}) -- atrous spatial pyramid pooling 
\cite{Chen:2018:deeplab:ieeeTPAMI, DeepLabV3:DBLP:journals/corr/ChenPSA17, Chen:2018:EncoderDecoderWA:ECCV:deeplabv3Plus}. The CMSNet framework presented in this work used the ASPP module with expansion rates of 1, 6, 12, or 18  for output stride of 16, and expansion rates of 1, 12, 24, or 36 for output stride 8.

\subsubsection{Global Pyramid Pooling (GPP)}
\label{subsection:global-pooling}
\begin{figure}[htb]
	\begin{center} 
		\includegraphics[width=0.75\linewidth]{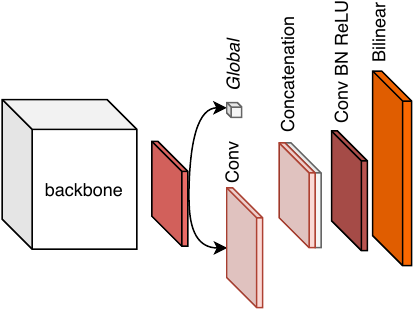} 
	\end{center} 
	\caption{Global Pyramid Pooling Module.} 
	\label{fig:global-pooling-module} 
	\fautor
\end{figure}

The segmentation framework CMSNet also supports global pooling. Even using separable convolution, pyramid pooling modules introduce a computational overhead. To deal with this limitation, the proposed framework supports a global pyramid pooling~\cite{Sandler:2018:MobileNetV2:IEEE-CVF,DeepLabV3:DBLP:journals/corr/ChenPSA17} to provide a cost-effective global context analysis for semantic segmentation. This solution uses just one global pooling concatenated with a pointwise convolution (\autoref{fig:global-pooling-module}).

\subsubsection{Bilinear Interpolation}
\label{subsection:bilinear-interpolation} 
The transposed convolution used in some segmentation architectures is not computationally efficient. A solution can be use convolution followed by bilinear interpolation that achieves equivalent results with less computational overhead. Both functions have the purpose of learning the best way to interpolate the segmentation maps and resize them to the image size. 

The CMSNet supports only the convolution followed by bilinear interpolation. The researcher chose not to use the transposed convolution to keep the computational cost consistent and make it possible to embed the solution to field tests. All the pyramids methods, SPP, ASPP, and GPP, are followed by convolution with bilinear interpolation instead of transposed convolution. 

\subsection{Hardware Platform}
\label{subsec:kamino_project}
Even using a backbone optimized for computational efficiency, CNNs for dense pixel classification demands high parallel processing power and memory bandwidth. These requirements create problems in the moment of embedding the perception subsystem for field tests with real-time inference. One possible way to do this would be to build dedicated hardware using FPGA or ASIC. However, such solutions are highly complex to implement and may not be flexible concerning eventual changes. The solution used in this research was porting the subsystem to the NVIDIA DrivePX 2 Autochauffeur. Nevertheless, once the network was developed on an x86\_64 platform, it was necessary to reimplement it with C++/CUDA merging several layers to make it possible to run it in real-time on the ARMv8-A.

\begin{figure}[htb]
	\centering
	\includegraphics[width=0.4\linewidth,trim={0cm 2cm 5cm 8cm},clip]{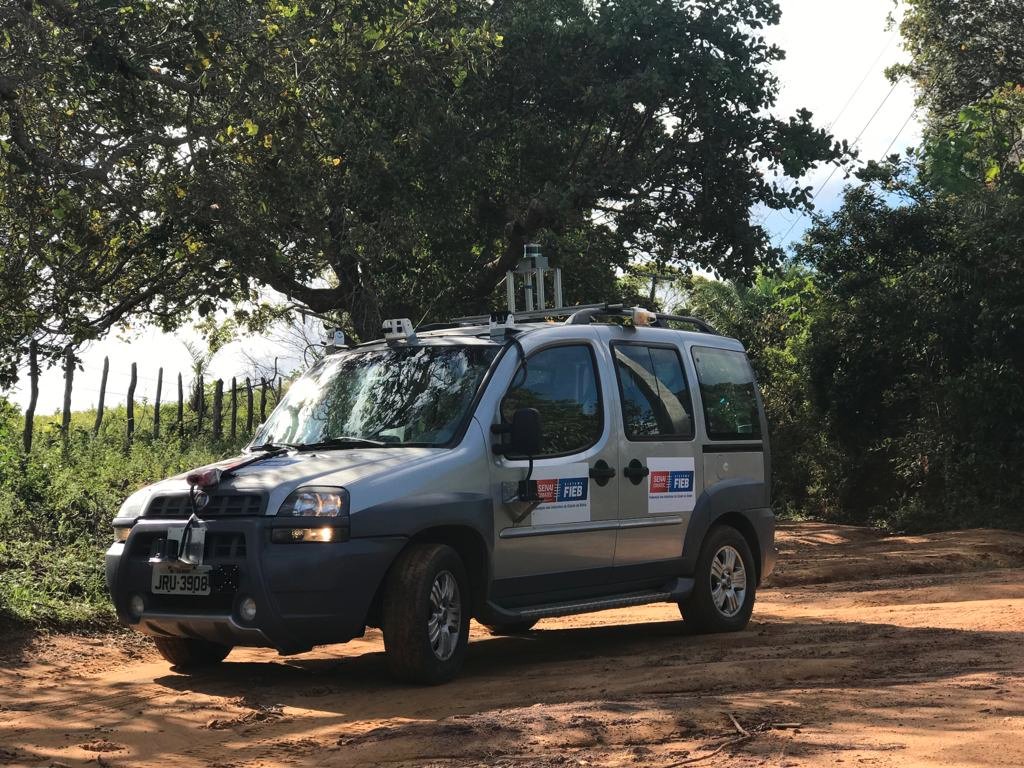}
	\caption{The vehicle used for data acquisition and validation of the proposed system.}
	\label{fig:carro-plataforma}
	\fautor
\end{figure}

\begin{figure}[htb]
	\includegraphics[width=\linewidth]{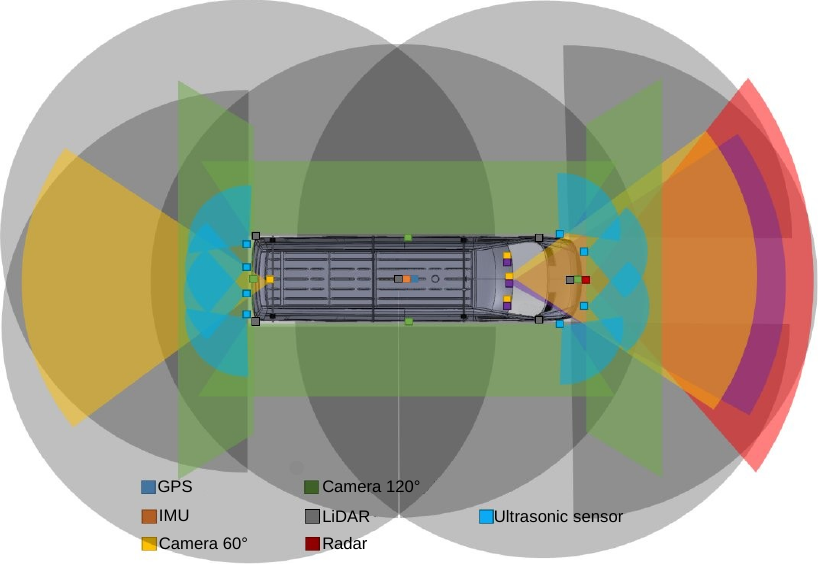}
	\caption{Sensor layout and operating region.}
	\label{fig:montagem-de-sensores}
	\fautor
\end{figure}

A utility van (\autoref{fig:carro-plataforma}) was used to mount the hardware for data acquisition and system validation. The system was composed of four RGB cameras with 60º Field of View (FOV), four RGB cameras with 120º FOV, one 16-beam LiDAR, four 8-beam LiDARs, eight ultrasonic sensors, one Inertial Measurement Unit (IMU), one GPS, one Radar, and one DrivePX 2 (\autoref{fig:montagem-de-sensores}).

\simbolo{^{\circ}}{Indicates angle or arc measure degree.}

\subsection{Off-road dataset}
\label{subsection:kamino_dataset}
Unpaved roads represent a relatively low explored scenario concerning the insertion of autonomous AVs and ADAS technology. For this research, the authors created a dataset for off-road and unpaved roads. This dataset is one of the contributions of this work. It has images collected in different places, including a test track built to emulate off-road environments and adverse conditions such as night, rainy, and dusty.

\subsubsection{Setup}
A hardware platform has been mounted with various sensors for collecting many hours of data. Subsequently, the most relevant videos pieces of information were selected and converted into frames at 1 or 5 Frames per Second (FPS). Further, the researchers accurately labeled the images resulting from that process. 

\begin{figure}[htb]
	\begin{subfigure}[b]{0.45\linewidth}
		\includegraphics[width=\linewidth,trim={0cm 0cm 0cm 10cm},clip]{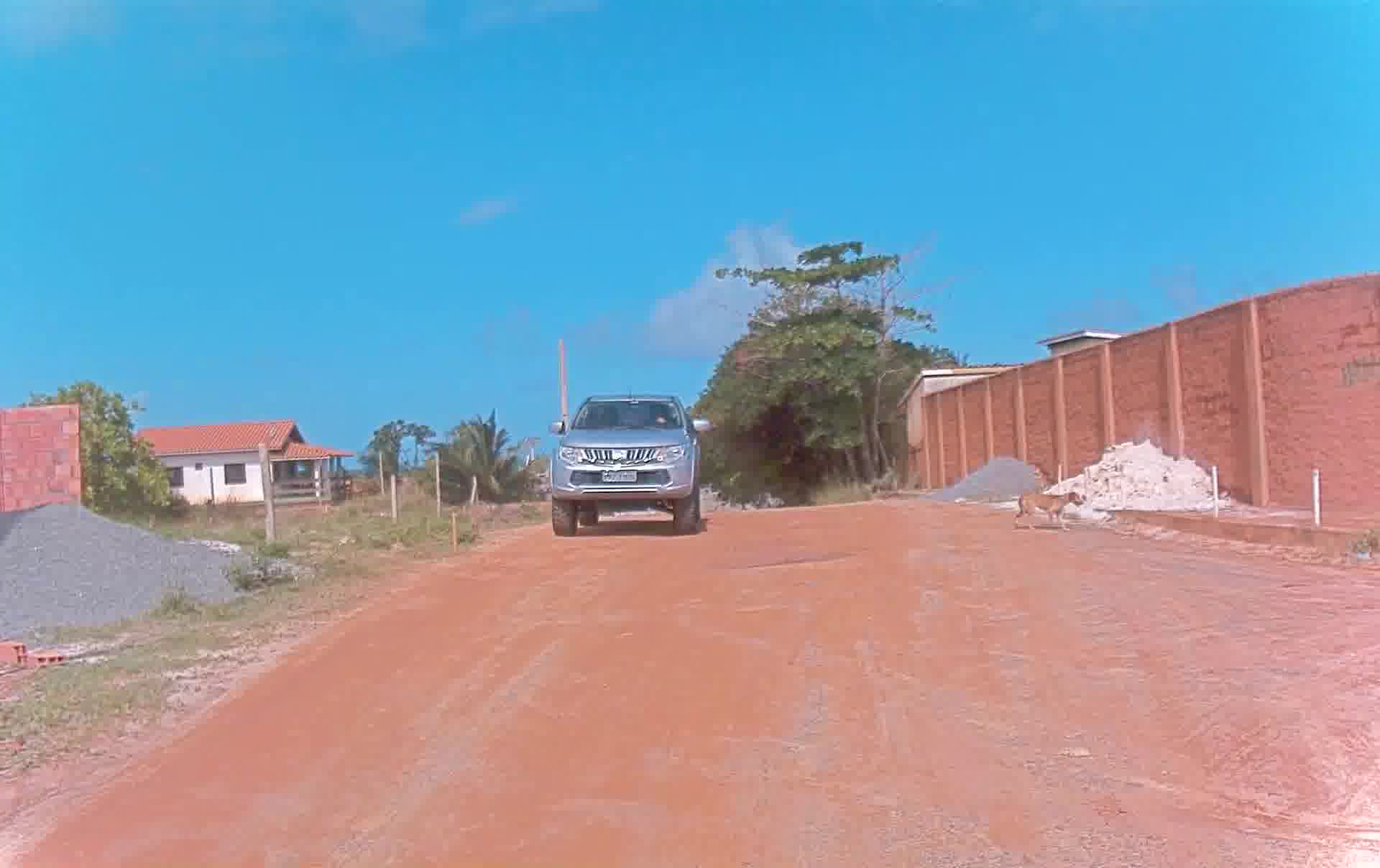}
		\caption{Jauá.}
	\end{subfigure}%
	\hfill%
	\begin{subfigure}[b]{0.45\linewidth}
		\includegraphics[width=\linewidth,trim={0cm 5cm 0cm 5cm},clip]{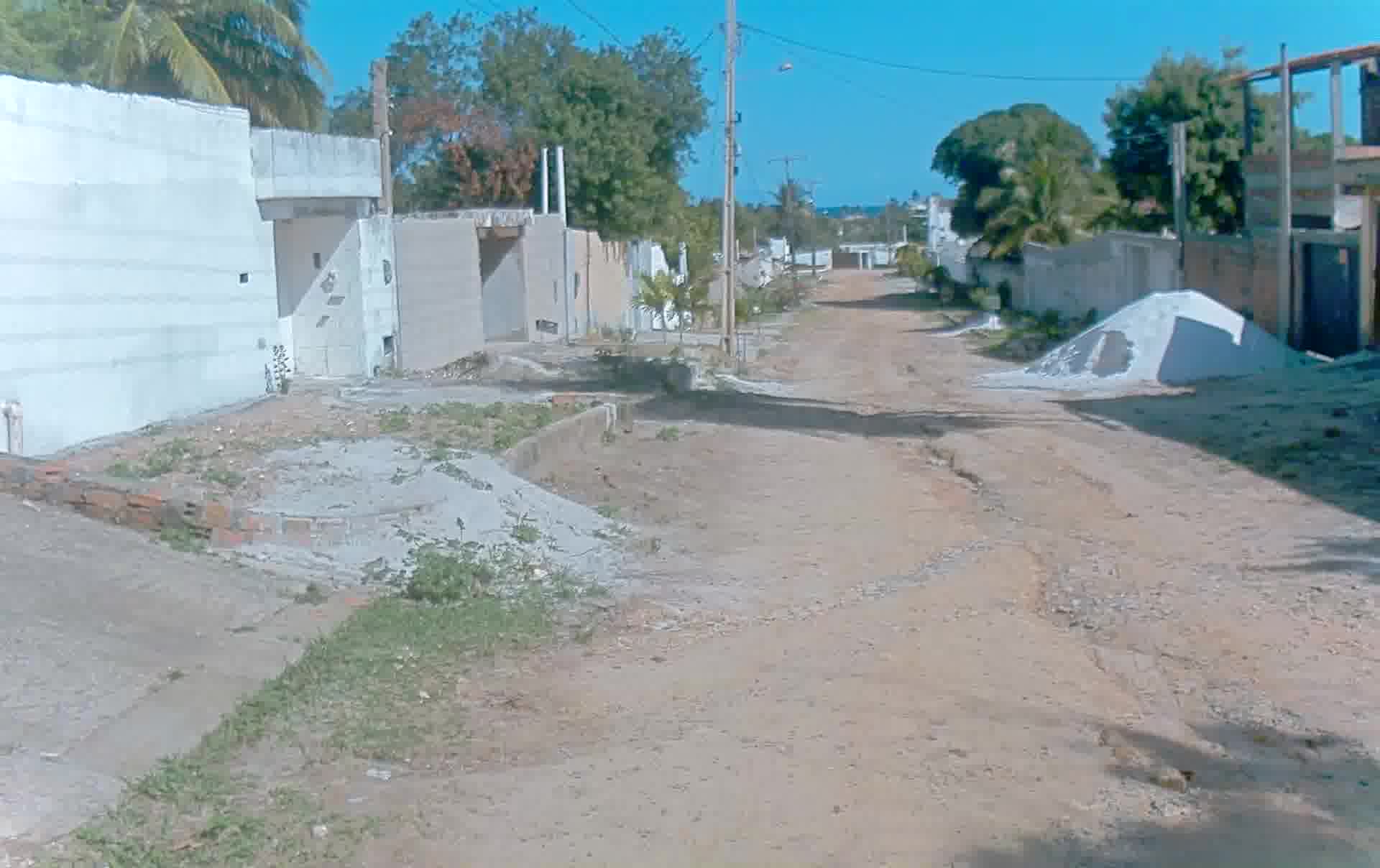}
		\caption{Jauá.}
	\end{subfigure}
	\begin{subfigure}[b]{0.45\linewidth}
		\includegraphics[width=\linewidth,trim={0cm 5cm 0cm 5cm},clip]{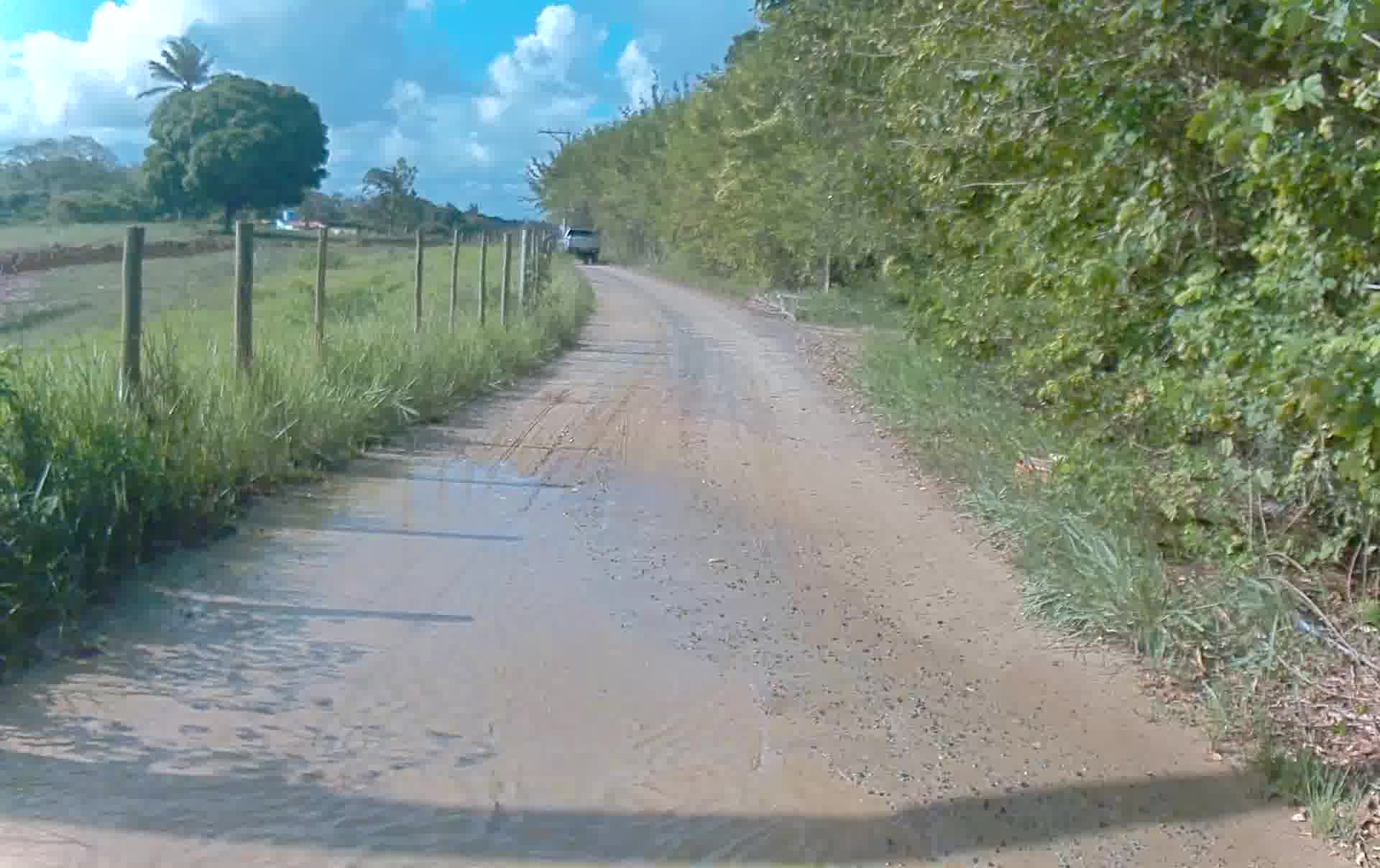}
		\caption{Estrada dos Tropeiros.}
	\end{subfigure}%
	\hfill%
	\begin{subfigure}[b]{0.45\linewidth}
		\includegraphics[width=\linewidth,trim={0cm 5cm 0cm 5cm},clip]{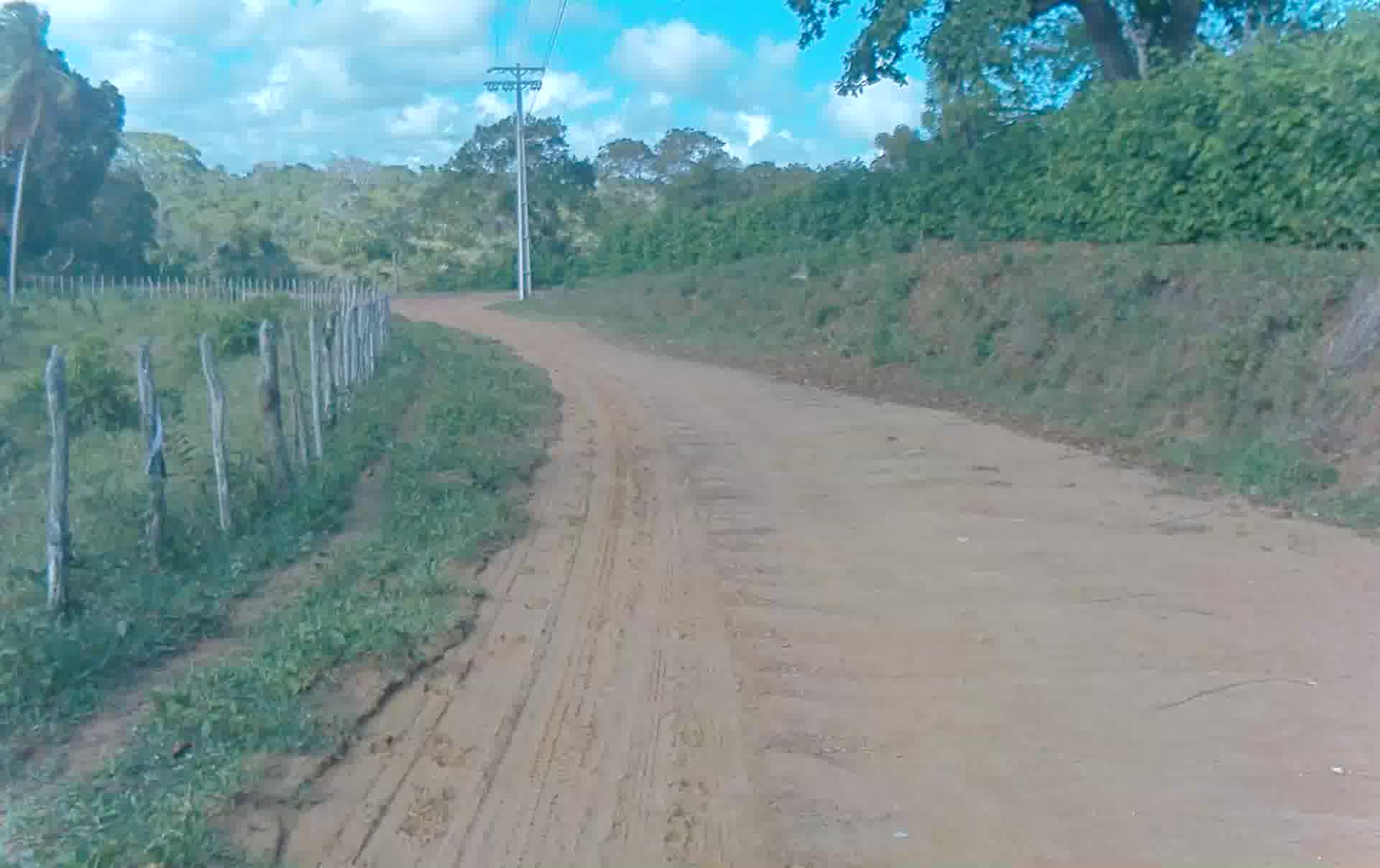}
		\caption{Estrada dos Tropeiros.}
	\end{subfigure}
	\begin{subfigure}[b]{0.45\linewidth}
		\includegraphics[width=\linewidth,trim={0cm 5cm 0cm 5cm},clip]{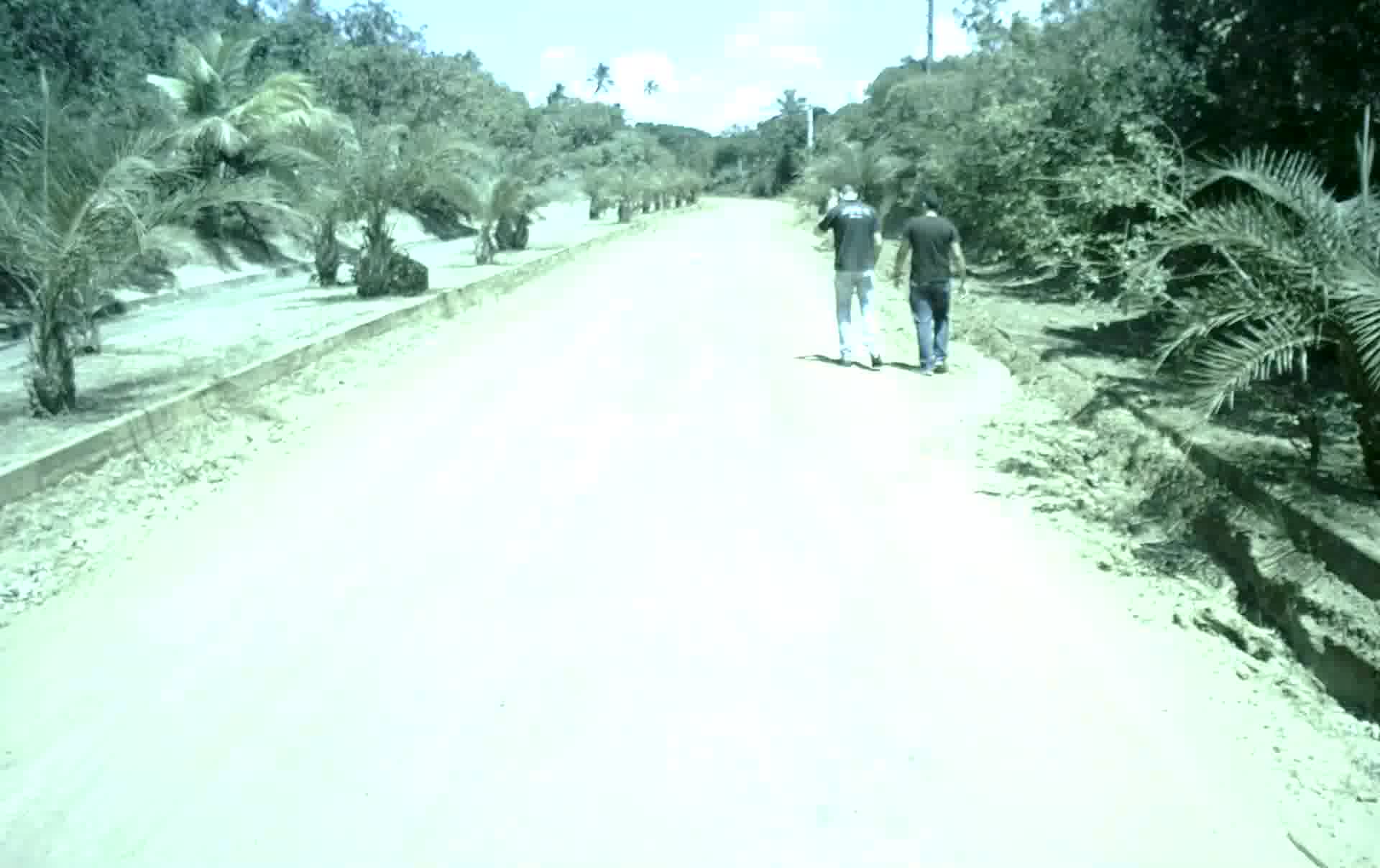}
		\caption{Praia do Forte.}
	\end{subfigure}%
	\hfill%
	\begin{subfigure}[b]{0.45\linewidth}
		\includegraphics[width=\linewidth,trim={0cm 7cm 0cm 3cm},clip]{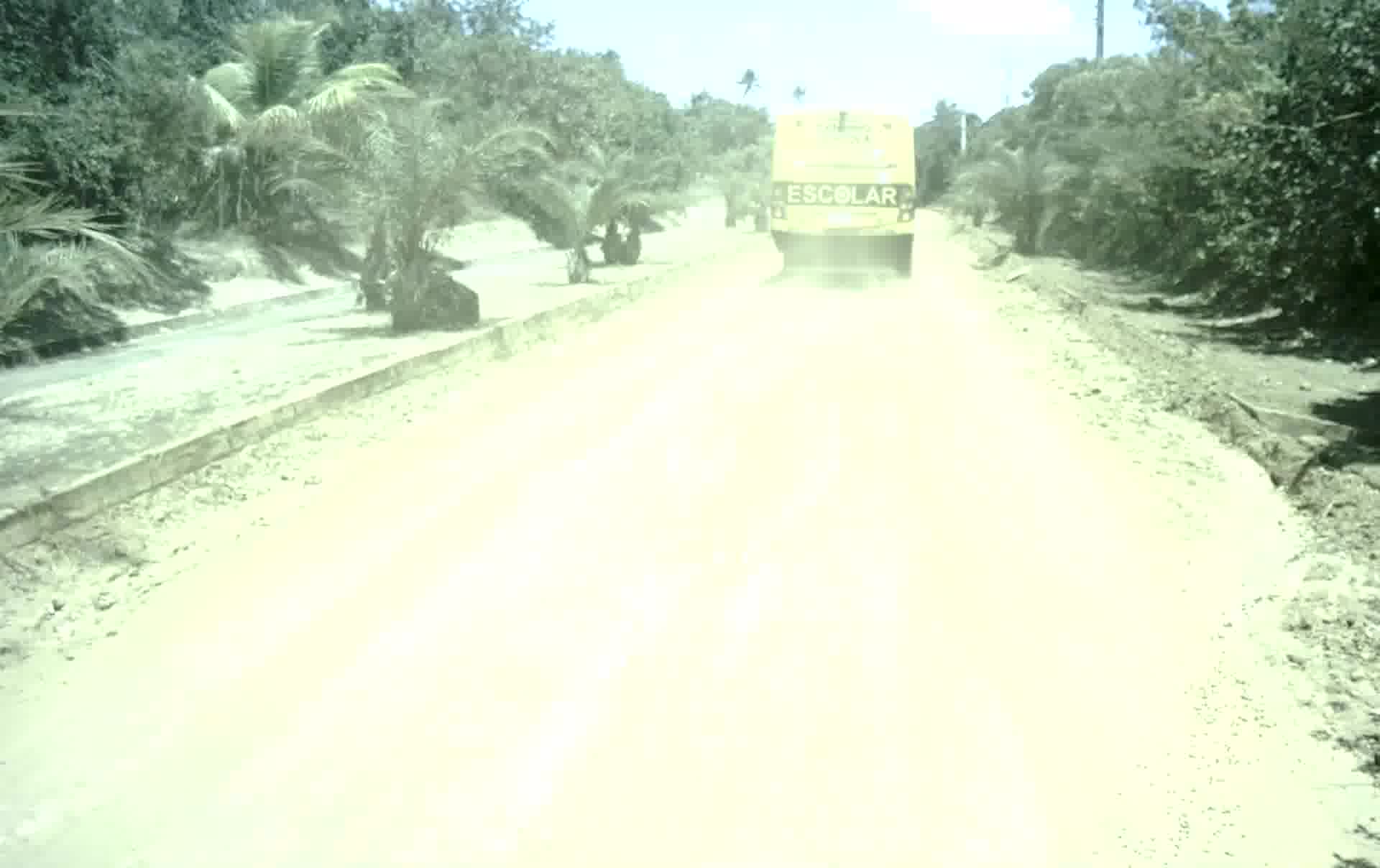}
		\caption{Praia do Forte.}
	\end{subfigure}
	\caption{Images collected in the metropolitan region of Salvador using the 60° FOV camera.}
	\label{fig:imagens-coletados-na-regiao-metropolitana-de-salvados-ba}
	\fautor
\end{figure}

\begin{figure}[htb]
	\begin{subfigure}[b]{0.32\linewidth}
		\includegraphics[width=\linewidth,trim={2cm 0cm 0cm 2cm},clip]{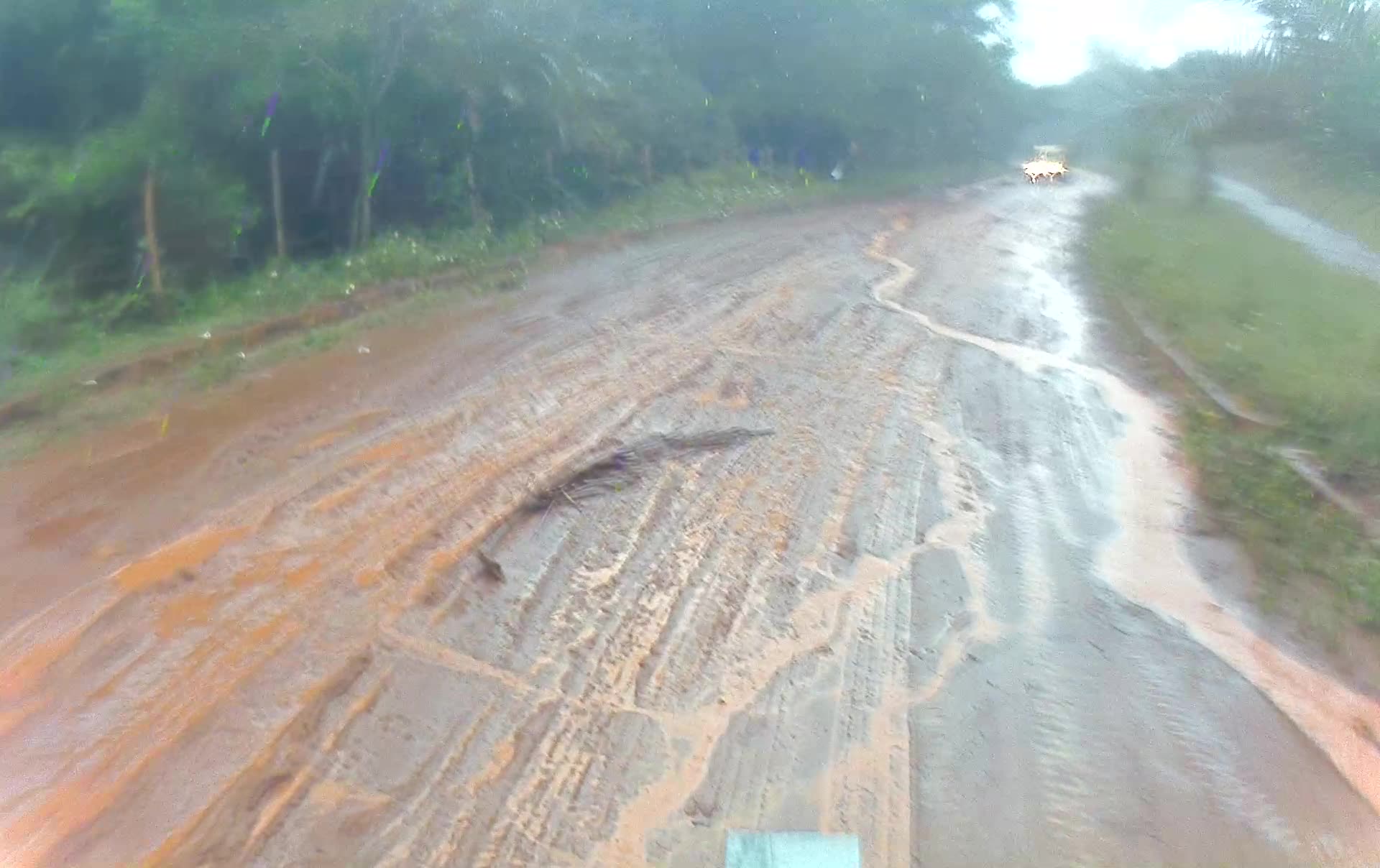}
		\caption{P. do Forte raining.}
	\end{subfigure}%
	\hfill
	\begin{subfigure}[b]{0.32\linewidth}
		\includegraphics[width=\linewidth,trim={2cm 0cm 0cm 2cm},clip]{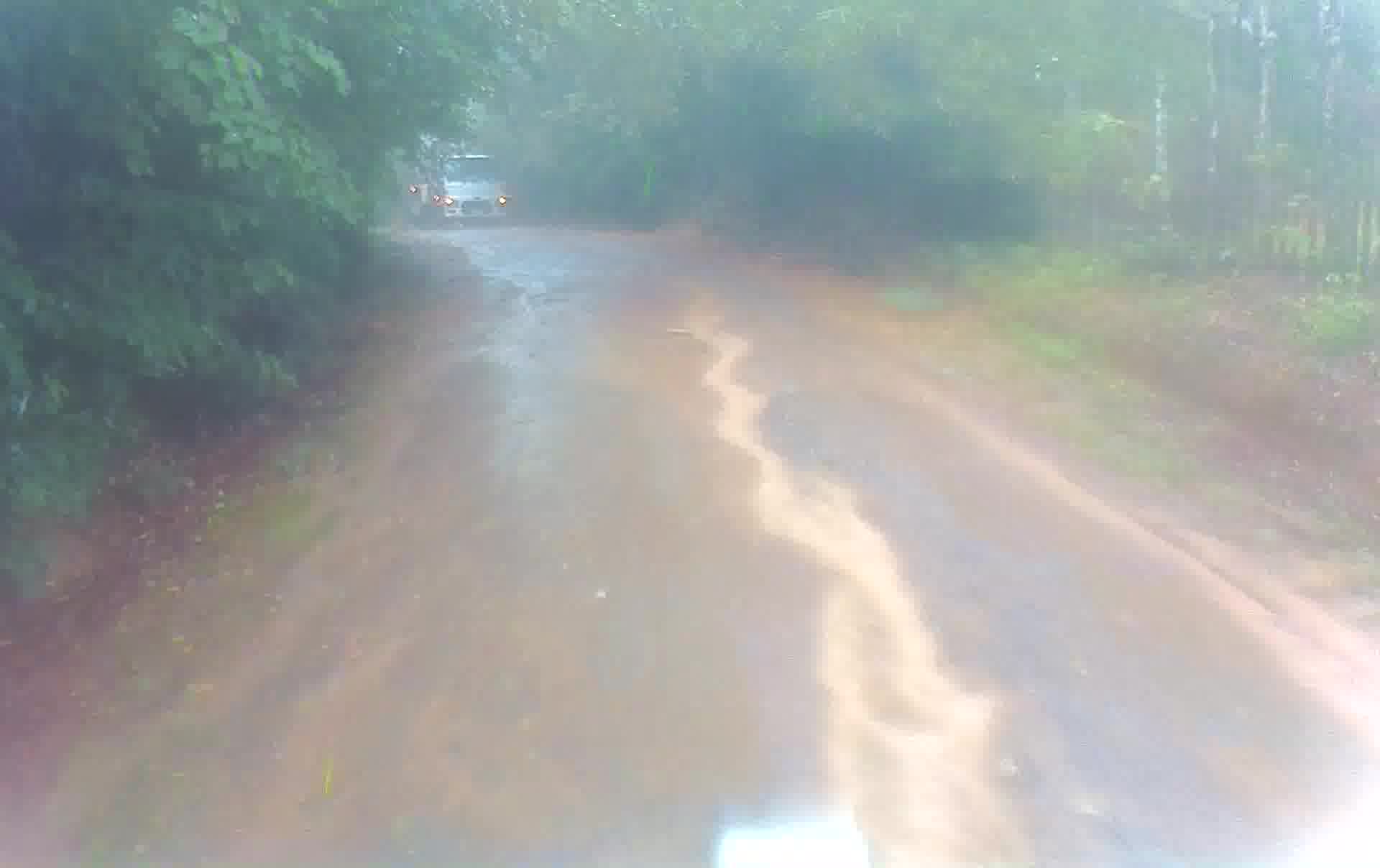}
		\caption{Tropeiros raining.}
	\end{subfigure}
	\hfill
	\begin{subfigure}[b]{0.32\linewidth}
		\includegraphics[width=\linewidth,trim={2cm 0cm 0cm 2cm},clip]{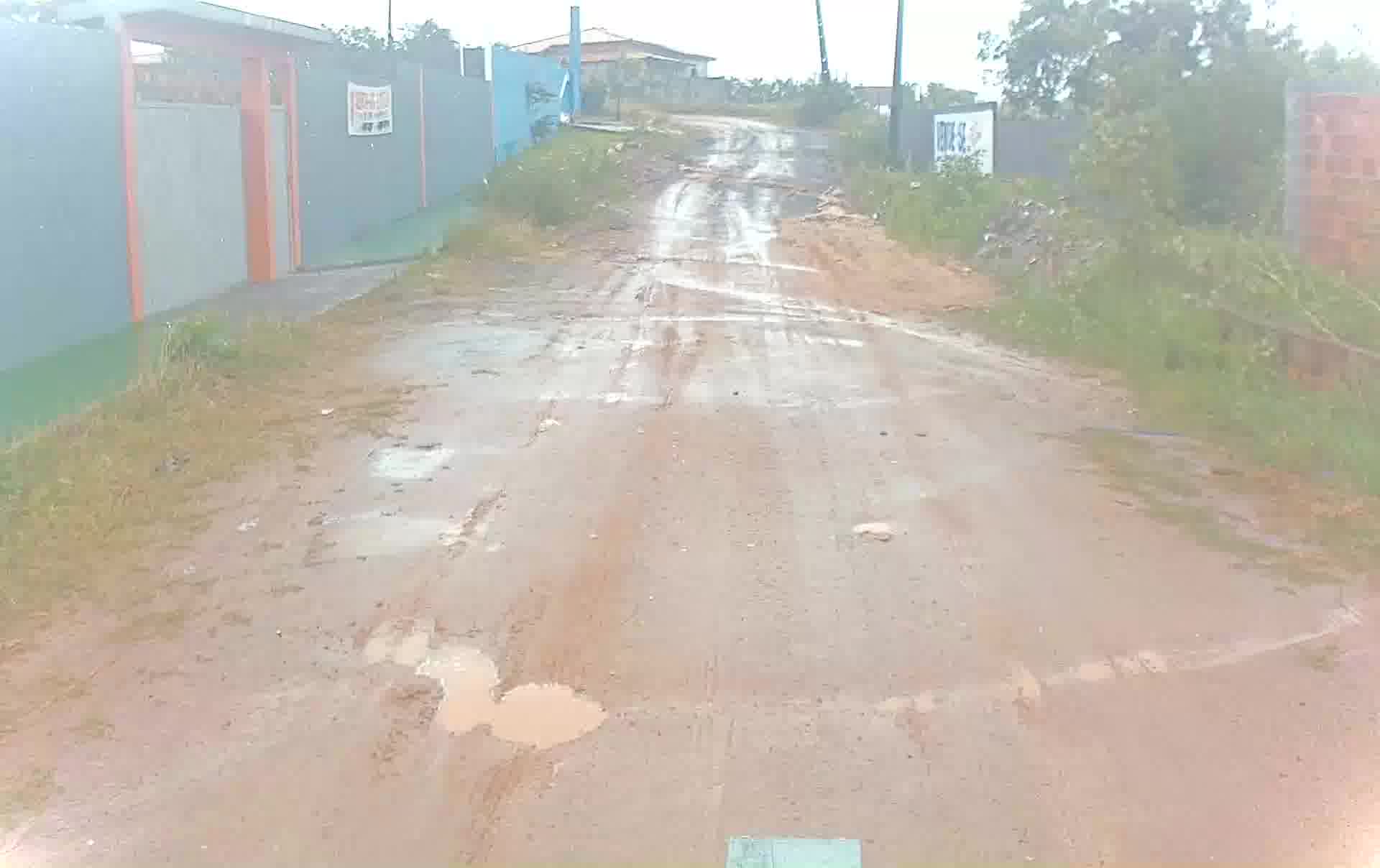}
		\caption{Jauá raining.}
	\end{subfigure}%
	
	\begin{subfigure}[b]{0.32\linewidth}
		\includegraphics[width=\linewidth,trim={2cm 0cm 0cm 2cm},clip]{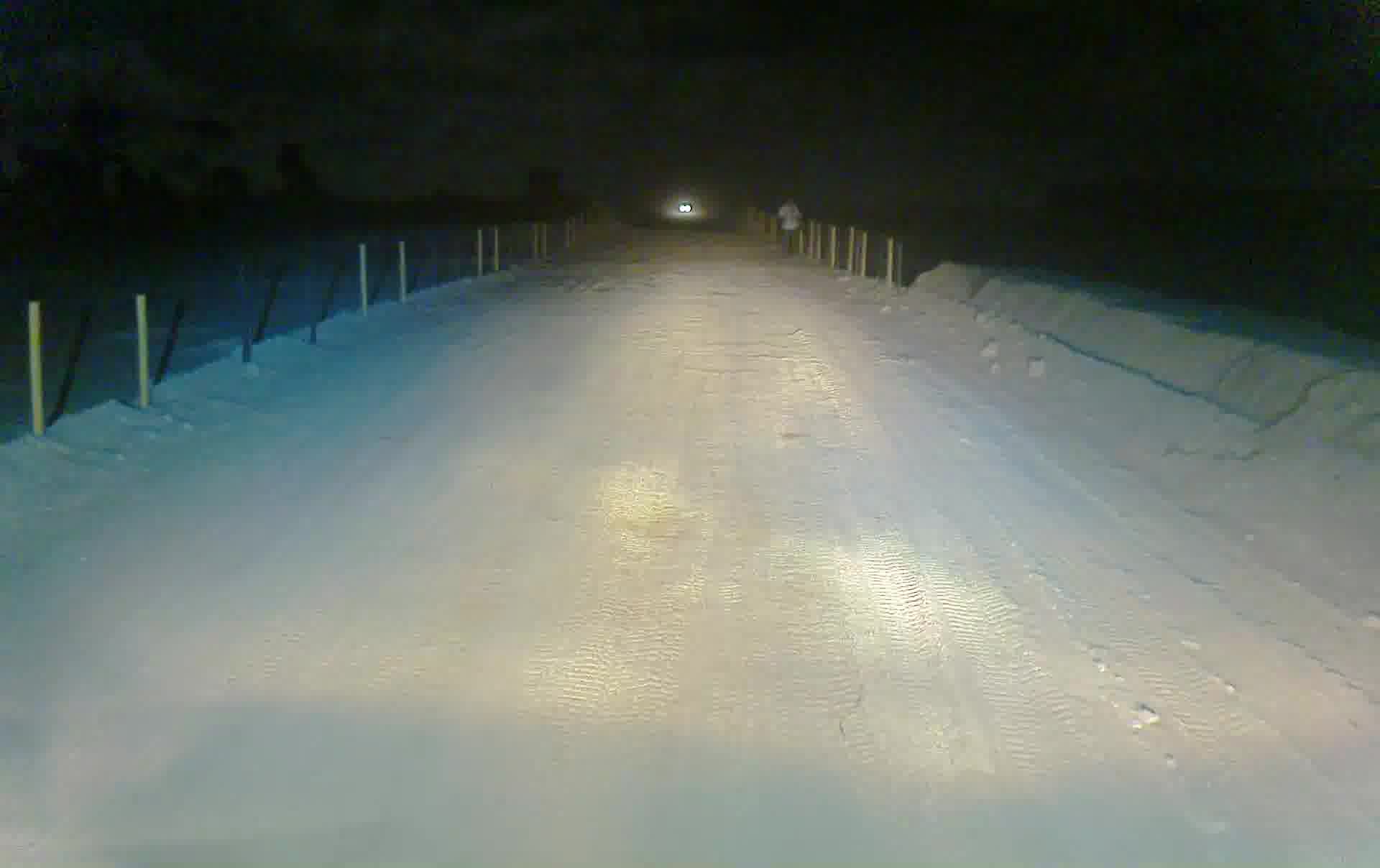}
		\caption{Test track at night.}
	\end{subfigure}%
	\hfill
	\begin{subfigure}[b]{0.32\linewidth}
		\includegraphics[width=\linewidth,trim={2cm 0cm 0cm 2cm},clip]{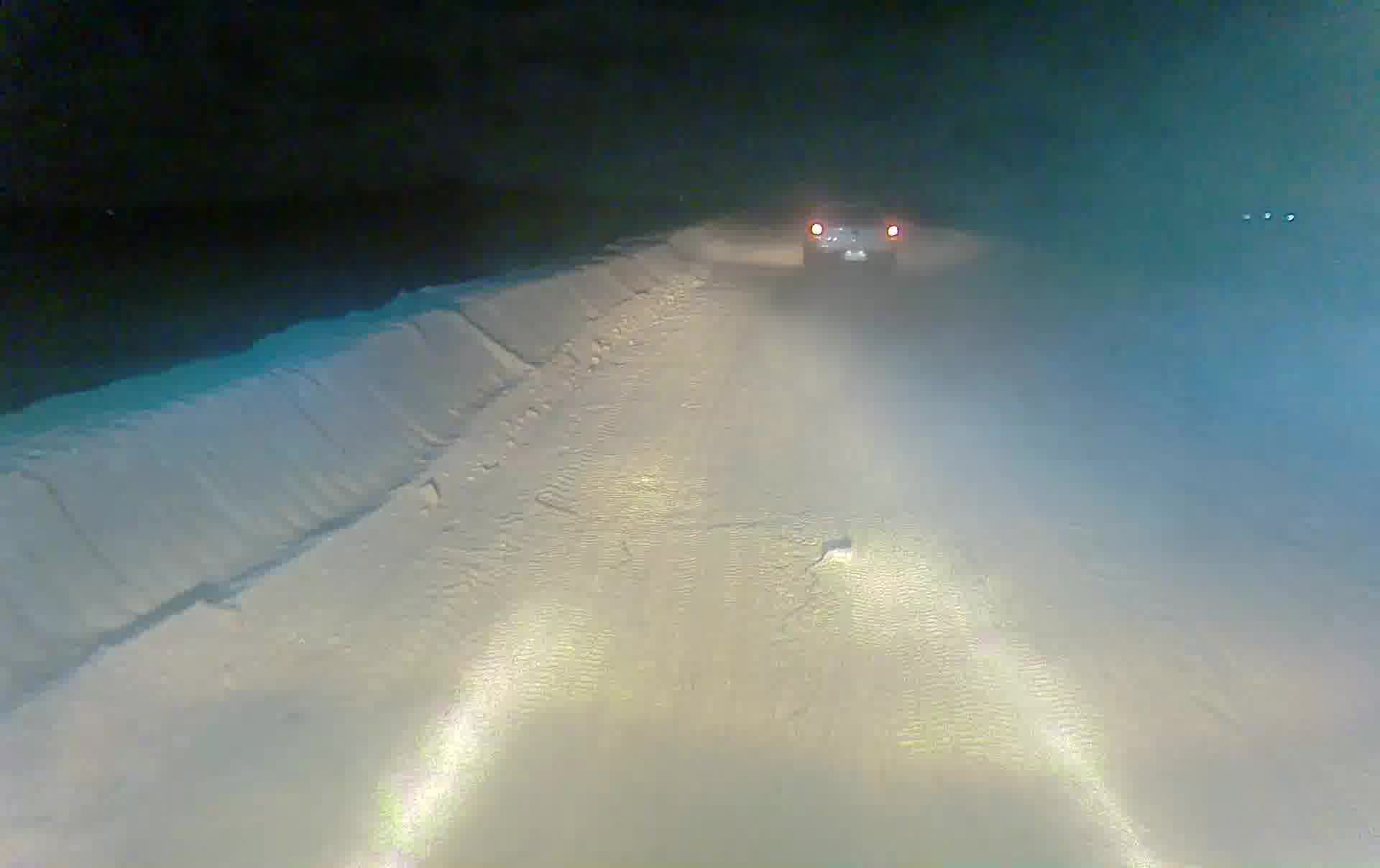}
		\caption{Track dusty/night.}
	\end{subfigure}
	\hfill
	\begin{subfigure}[b]{0.32\linewidth}
		\includegraphics[width=\linewidth,trim={2cm 0cm 0cm 2cm},clip]{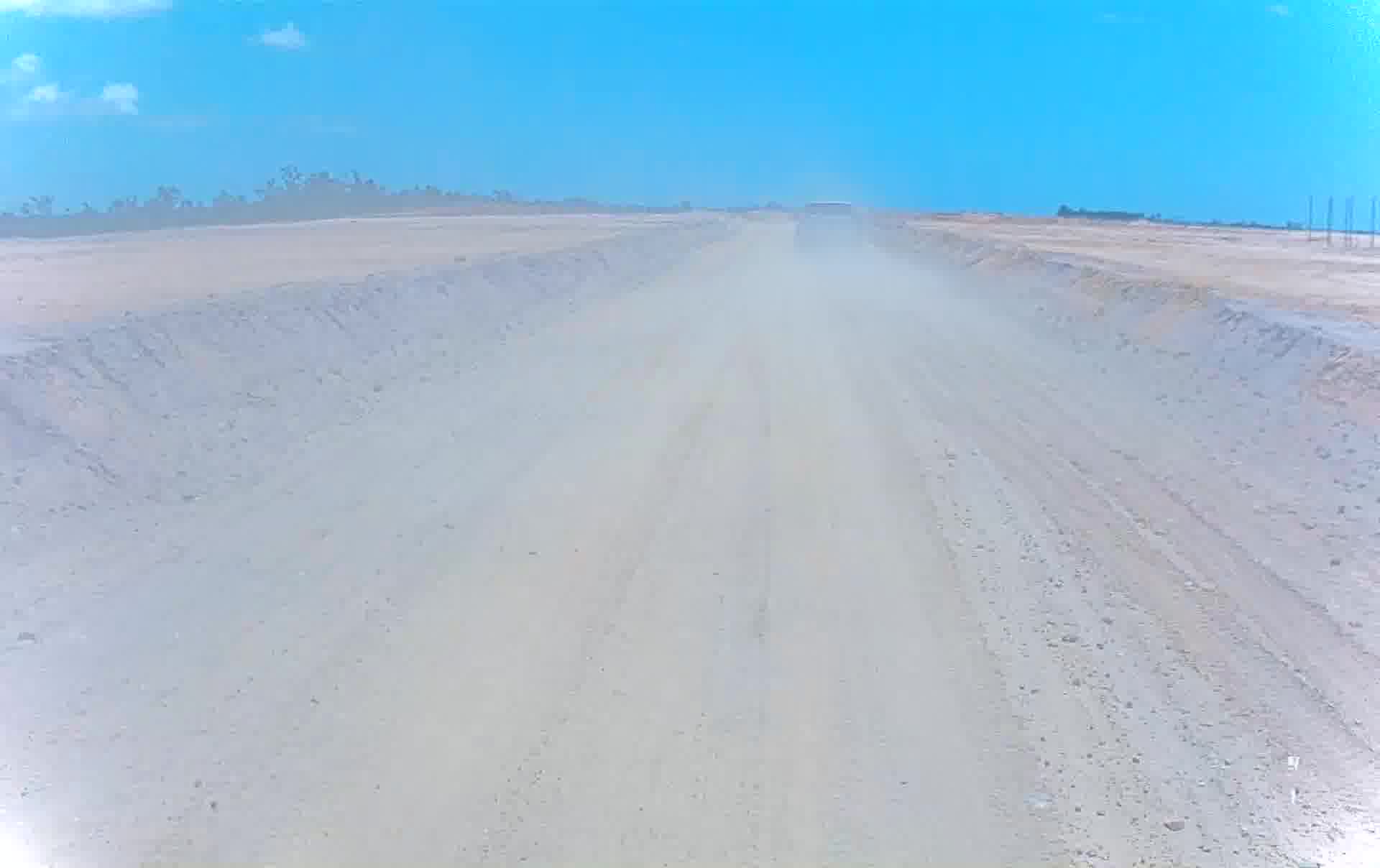}
		\caption{Track with dusty.}
	\end{subfigure}%
	\caption{Adverse conditions.}
	\label{fig:situacoes-adversas}
	\fautor
\end{figure}

Several unpaved roads at the Salvador city area also at the north coast of Bahia state were used together with the off-road test track as the scenario for data capturing (\autoref{fig:imagens-coletados-na-regiao-metropolitana-de-salvados-ba}). For technical reasons, it was not possible to acquire all adverse visibility conditions for all places. \autoref{tab:adverse-situation} and \autoref{fig:situacoes-adversas} show the list of the places and their respective visibility conditions.

\begin{table}[htb]
	\centering
    \scriptsize
	\caption{Adverse condition.}
	\label{tab:adverse-situation}
	\begin{tabular*}{\textwidth}{@{\extracolsep{\fill}} l l l}
	\specialrule{.1em}{.05em}{.05em}
		\bfseries Type 	& \bfseries Place & \bfseries Condition \\ 
		\specialrule{.1em}{.05em}{.05em}
		Off-road  & CIMATEC test track & Daytime, night, dirty  \\
		\hline
		\multirow{3}{*}{Unpaved roads}   	& Jauá            		   	& \multirow{3}{*}{Daytime, Raining}   \\
		& Praia do Forte     	   	&    \\
		& Estrada dos Tropeiros	   	&    \\
		\specialrule{.1em}{.05em}{.05em}
	\end{tabular*}
	\fautor
\end{table}

\subsubsection{Off-road test track}
\begin{figure}[htb]
	\begin{subfigure}[b]{0.45\linewidth}
		\includegraphics[width=\linewidth,trim={0cm 0cm 0cm 2cm},clip]{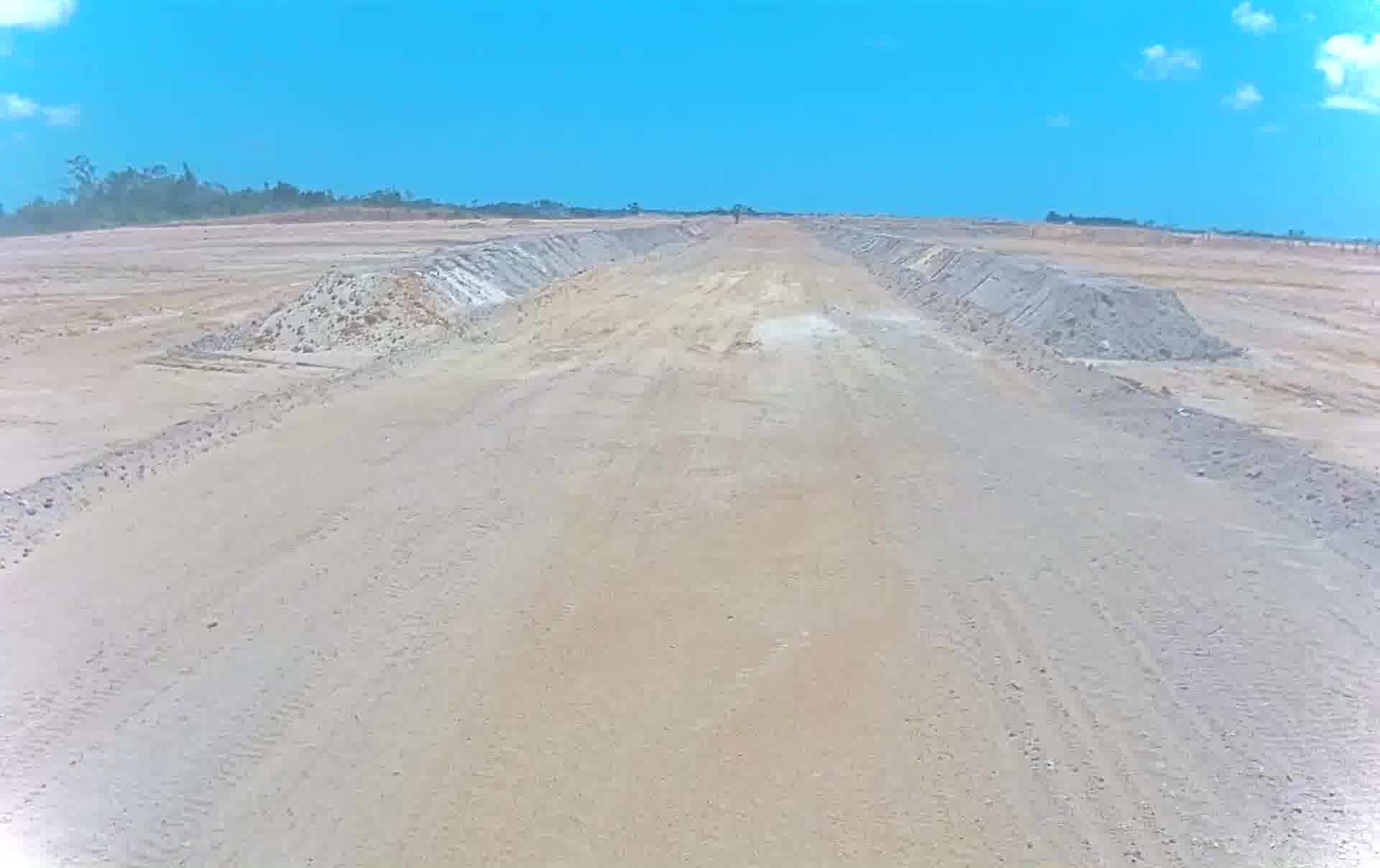}
		\caption{Slopes and open space.}
	\end{subfigure}%
	\hfill%
	\begin{subfigure}[b]{0.45\linewidth}
		\includegraphics[width=\linewidth,trim={0cm 0cm 0cm 2cm},clip]{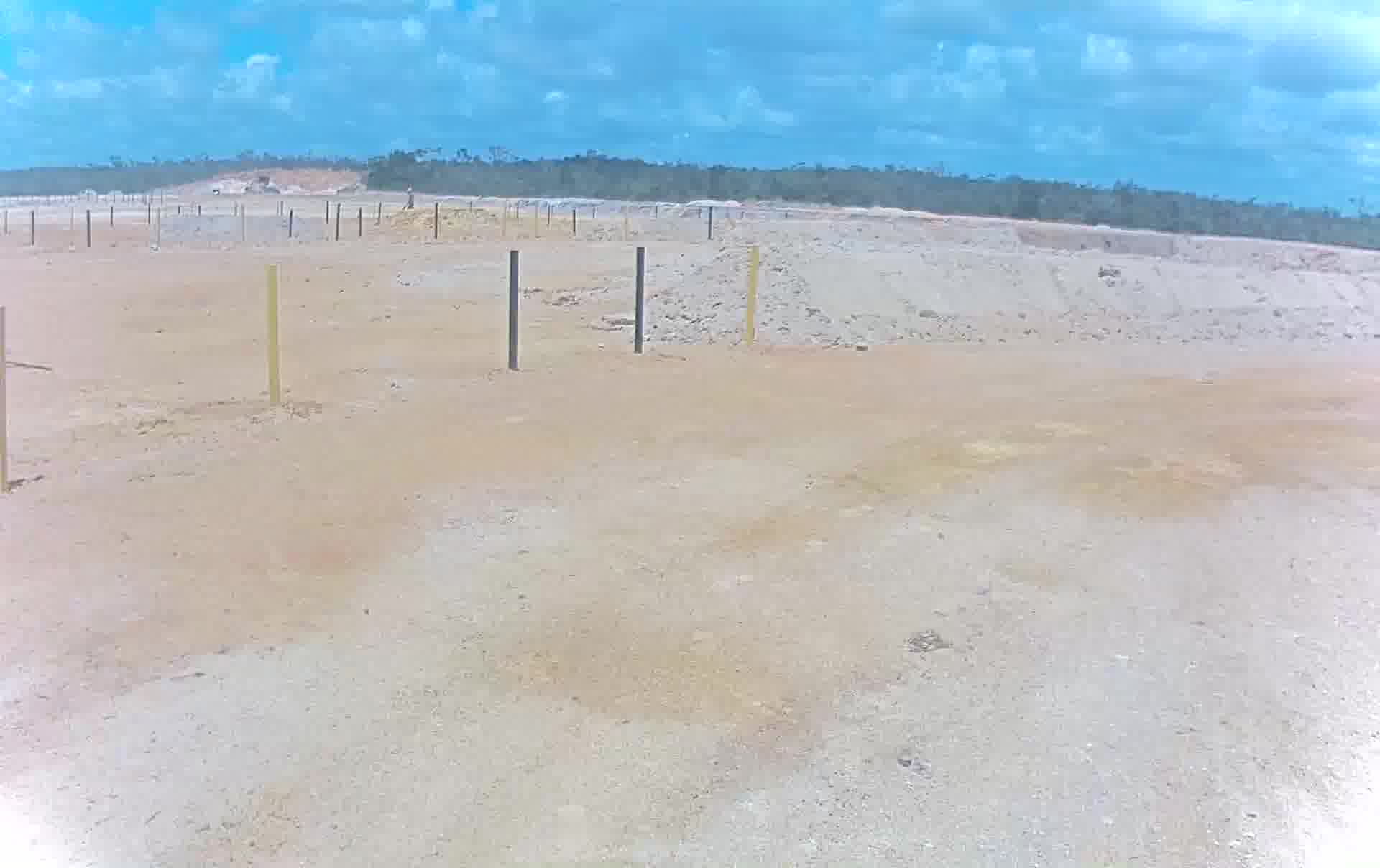}
		\caption{Pickets and slopes.}
	\end{subfigure}
	
	\begin{subfigure}[b]{0.45\linewidth}
		\includegraphics[width=\linewidth,trim={0cm 0cm 0cm 2cm},clip]{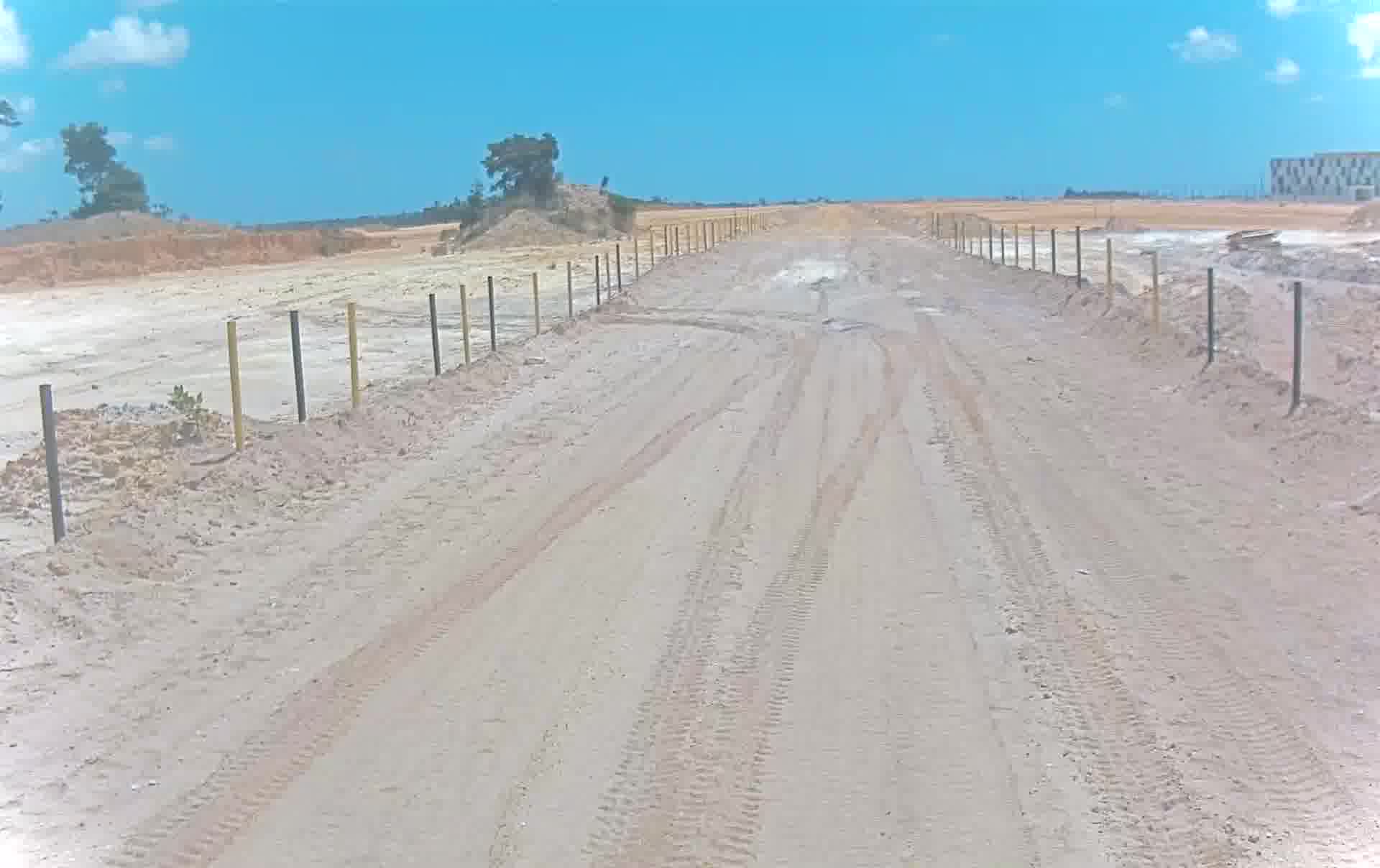}
		\caption{Pickets.}
	\end{subfigure}%
	\hfill%
	\begin{subfigure}[b]{0.45\linewidth}
		\includegraphics[width=\linewidth,trim={0cm 0cm 0cm 2cm},clip]{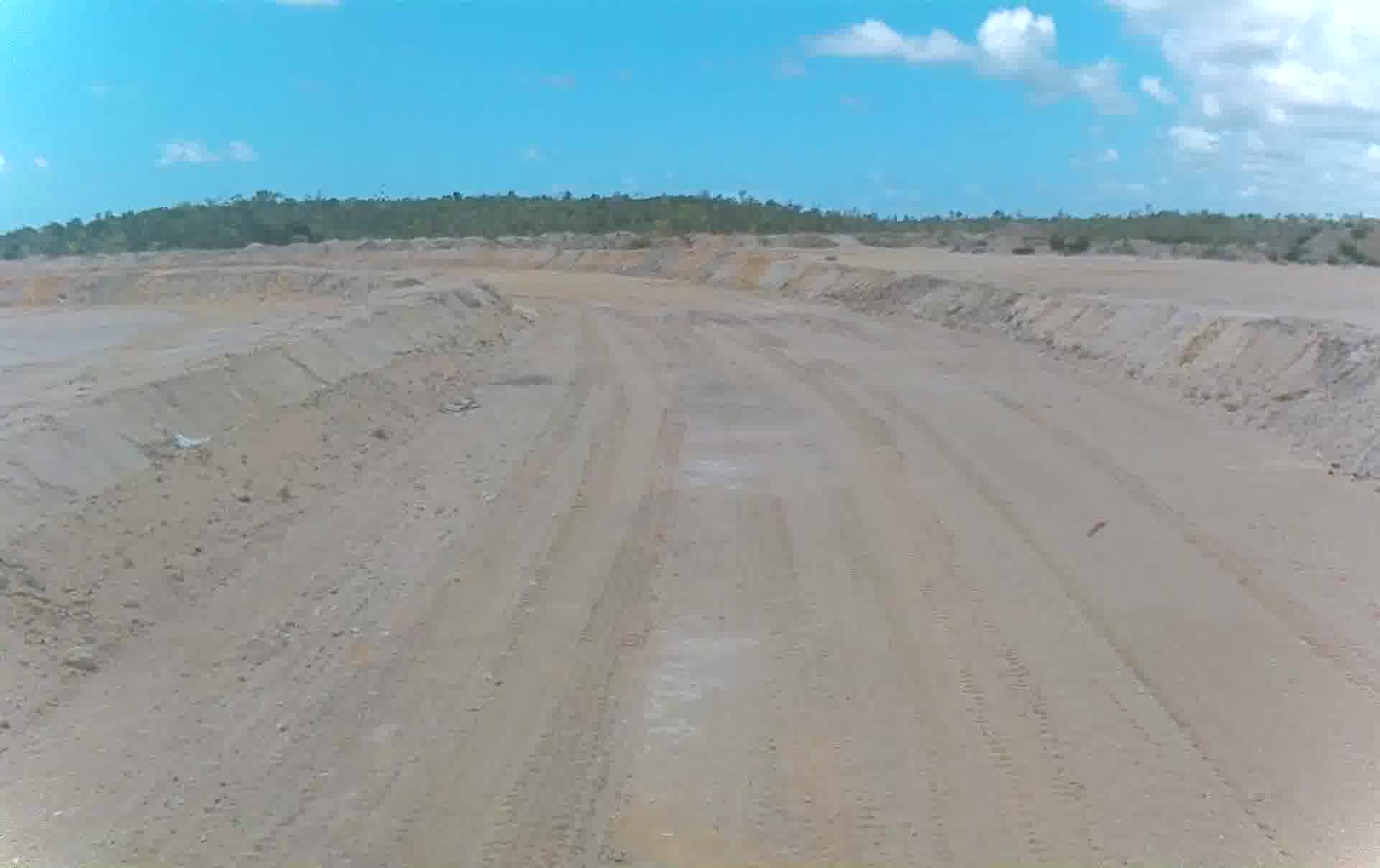}
		\caption{Slopes.}
	\end{subfigure}
	
	\caption{Different limits of the test track.}
	\label{fig:diferentes-limites-de-pista}
	\fautor
\end{figure}

Considering the application of vehicles for transporting cargo and passengers in industrial operation, the researchers have developed an off-road test track emulating environments such as open-pit mines where the difference in colors and textures are slight, making it difficult to segment the track area. \autoref{fig:diferentes-limites-de-pista} shows parts of the track and their different kinds of limiters.

\begin{figure}[htb]
	\includegraphics[width=1\textwidth]{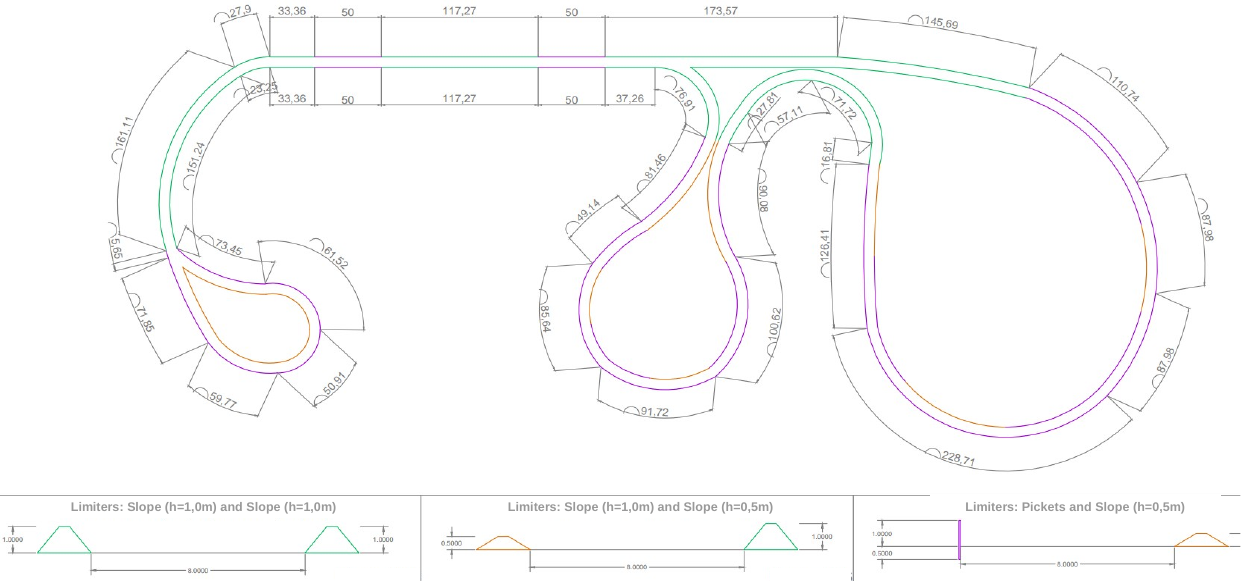}
	\caption{Map of the off-road test track.}
	\label{fig:projeto-da-pista-off-road}
\end{figure}

The test track is approximately 3,000 meters long. It is a closed circuit with straight sectors and open and closed curves to the right and the left. The researchers marked the test track limits with pickets and embankment slopes of different sizes. \autoref{fig:projeto-da-pista-off-road} shows the track design. It is possible to see the lines in green indicating slopes of 1 meter, yellow lines indicating slopes of 50 cm, and purple lines indicating pickets and empty spaces interspersed. In those scenarios (\autoref{fig:diferentes-limites-de-pista}), the lack of paving on the roads leads to the absence of well-defined edges delimiting correctly where the region of the traffic zone ends or begins. Besides, the weak variation in textures and colors on the off-road test track makes the segmentation task even more difficult. 

\subsubsection{Data acquisition}
The data was recorded in good and bad visibility conditions in different places of the Salvador metropolitan region. The researchers recorded images in a mix of dirt roads, urban environments with houses and buildings, and rural areas with farms and narrow tracks partially delimited by a curb surrounded by palm trees. Those data were collected during the morning and the afternoon within sunny and rainy conditions. 

\begin{figure}[htb]
	\begin{subfigure}[b]{0.45\linewidth}
		\includegraphics[width=\linewidth,trim={0cm 0cm 0cm 2cm},clip]{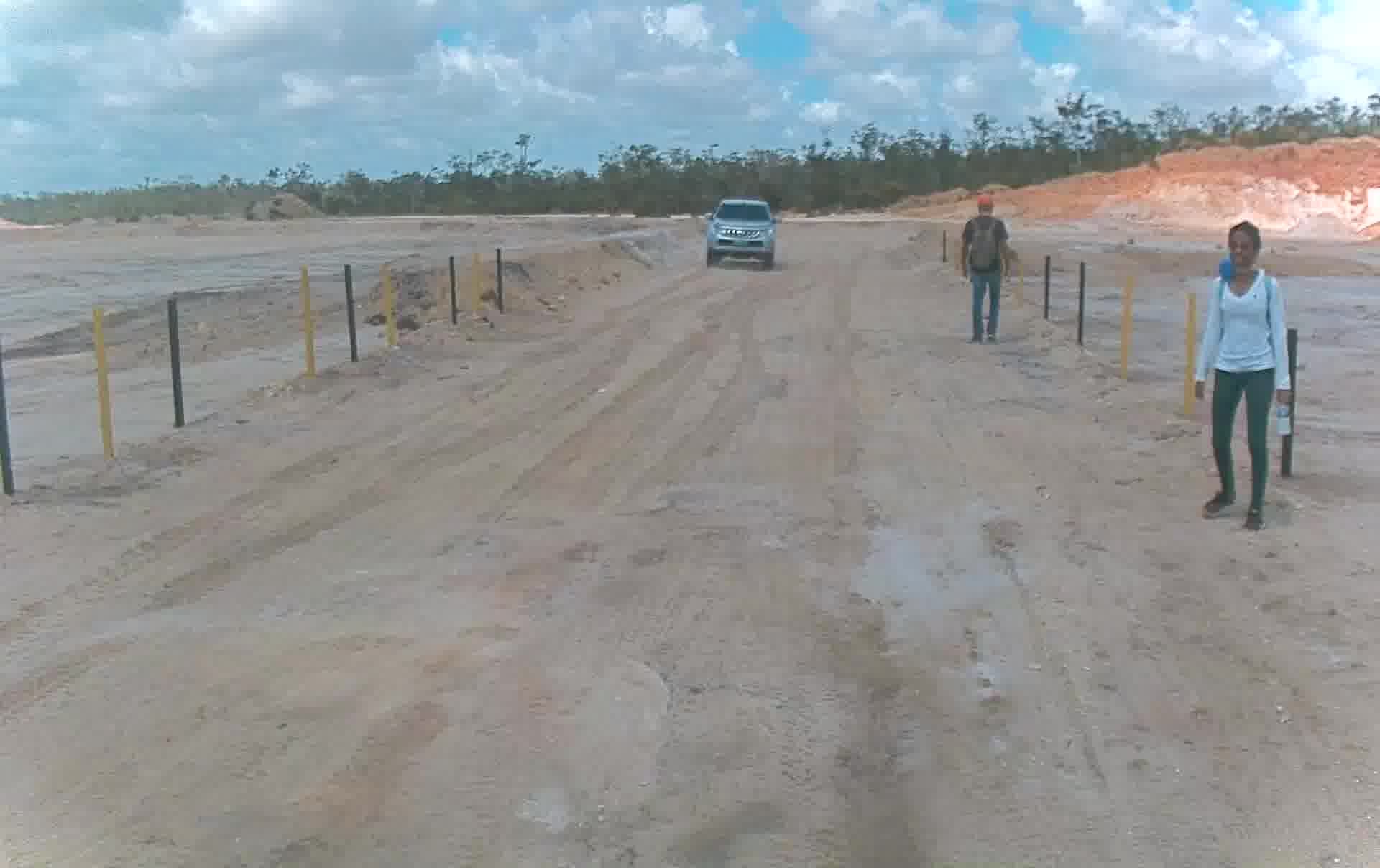}
		\caption{Original image.}
	\end{subfigure}%
	\hfill%
	\begin{subfigure}[b]{0.45\linewidth}
		\includegraphics[width=\linewidth,trim={0cm 0cm 0cm 2cm},clip]{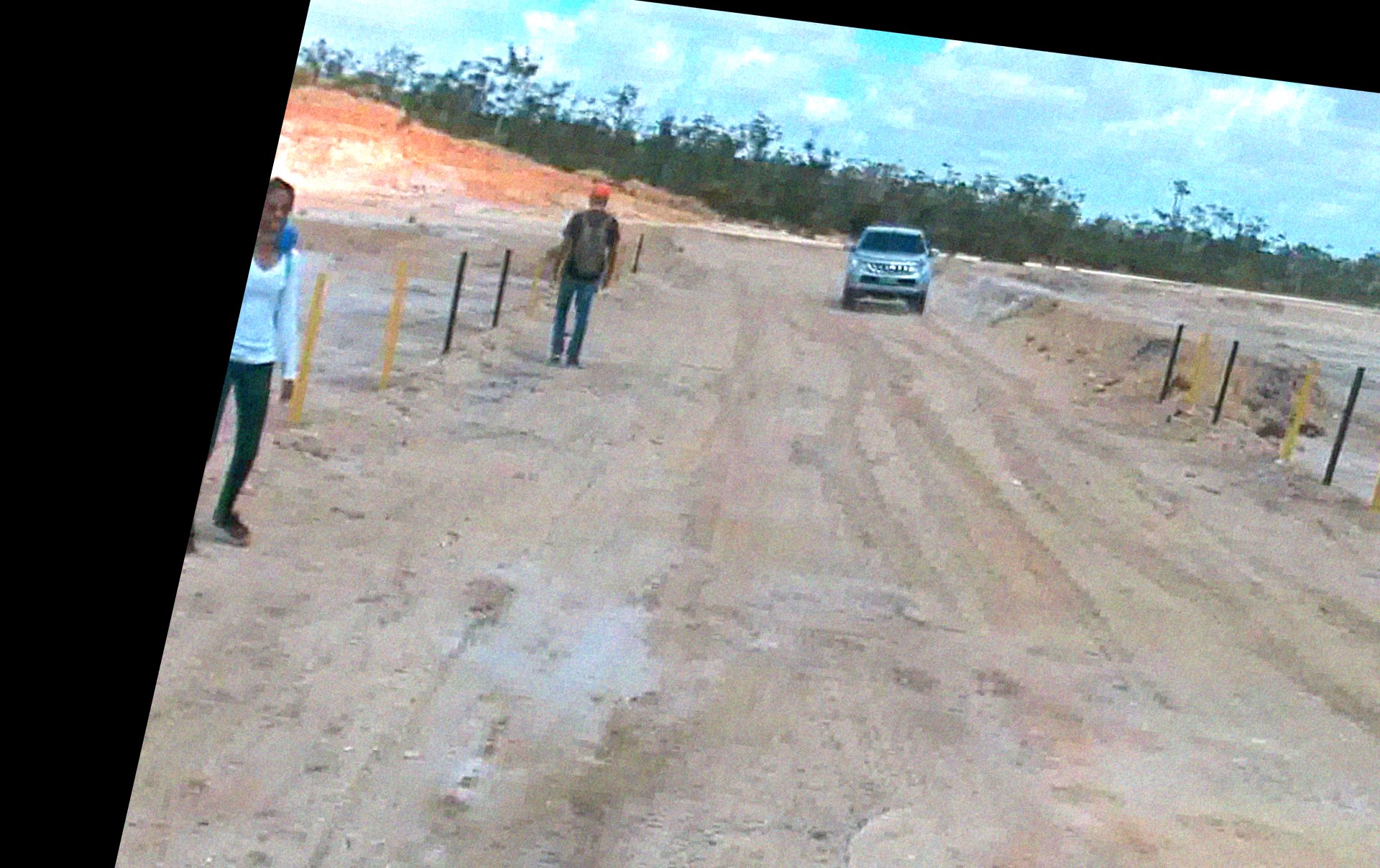}
		\caption{Noise, rotating and crop.}
	\end{subfigure}
	\begin{subfigure}[b]{0.45\linewidth}
		\includegraphics[width=\linewidth,trim={0cm 0cm 0cm 2cm},clip]{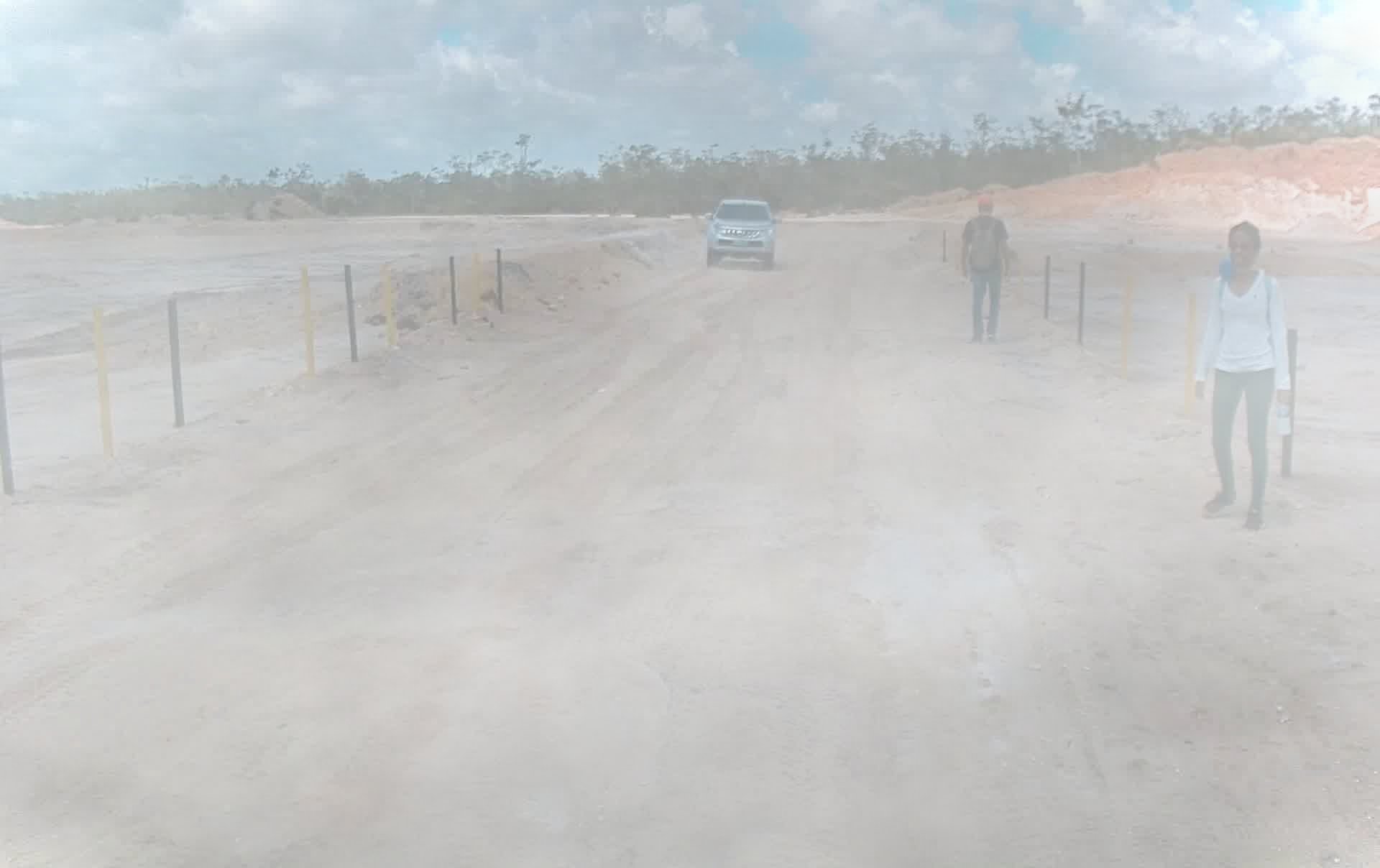}
		\caption{Fog.}
	\end{subfigure}%
	\hfill%
	\begin{subfigure}[b]{0.45\linewidth}
		\includegraphics[width=\linewidth,trim={0cm 0cm 0cm 2cm},clip]{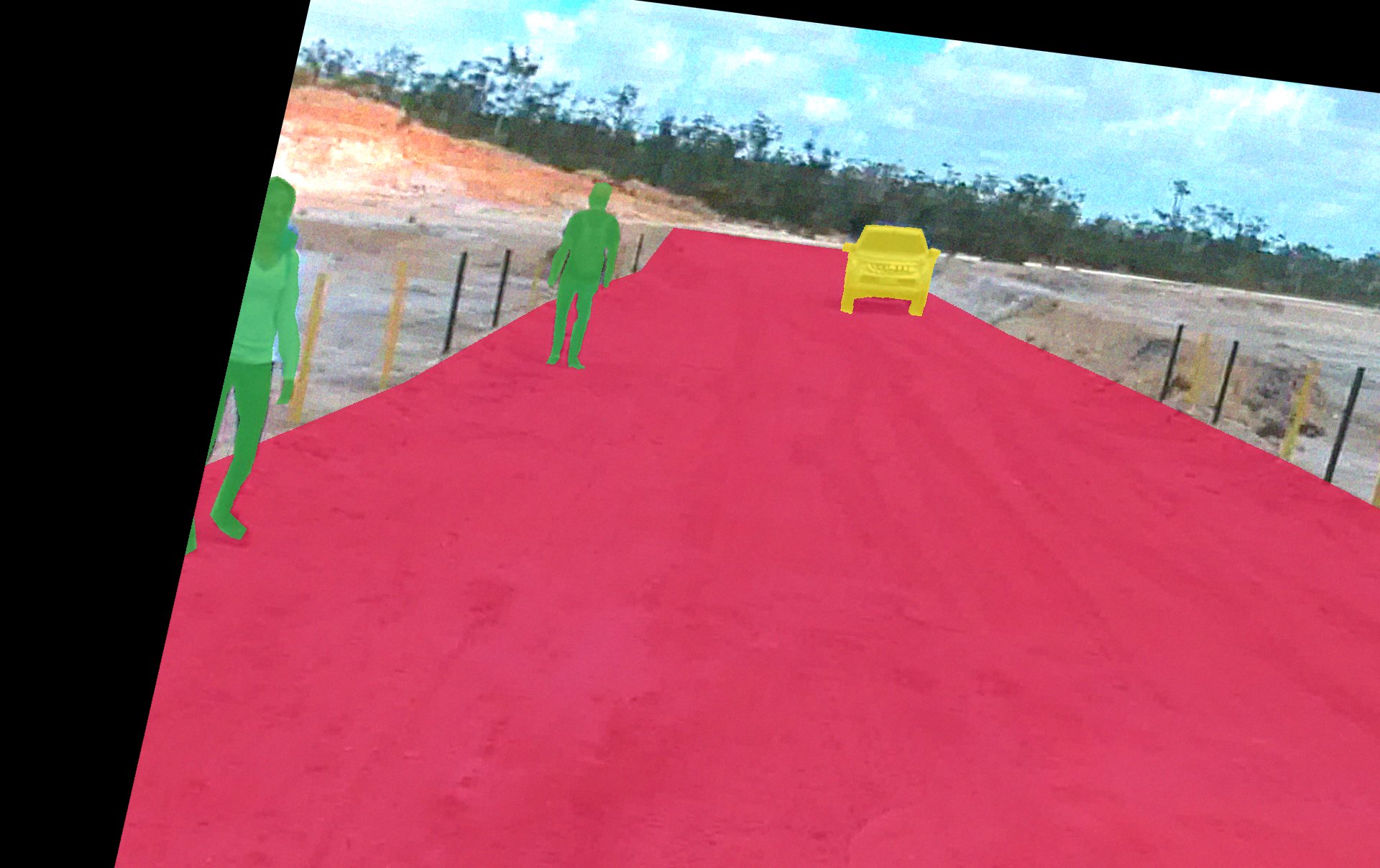}
		\caption{Image, segmentation mask.}
	\end{subfigure}
	\caption{Artificial data generation.}
	\label{fig:geracao-artificial-de-dados}
	\fautor
\end{figure}

In addition to the acquisitions on unpaved roads, the researchers have also recorded data in the controlled environment — the off-road test track built for the research (\autoref{fig:projeto-da-pista-off-road}). The data acquisition was carried out around noon, evening, and night. Images have been recorded in adverse situations such as low light and dust to increase the diversity of the dataset. Besides recording images at night, with dust and rain, the research also creates a script to allow synthetically increasing the dataset diversity by rendering fog, snow, and other impairments (\autoref{fig:geracao-artificial-de-dados}). Such scripts were developed with the help of the Imgaug library \cite{imgaug}.

\subsubsection{Annotation}
\begin{figure}[htb]
	\begin{subfigure}[b]{0.45\linewidth}
		\includegraphics[width=\linewidth,trim={2cm 0cm 0cm 2cm},clip]{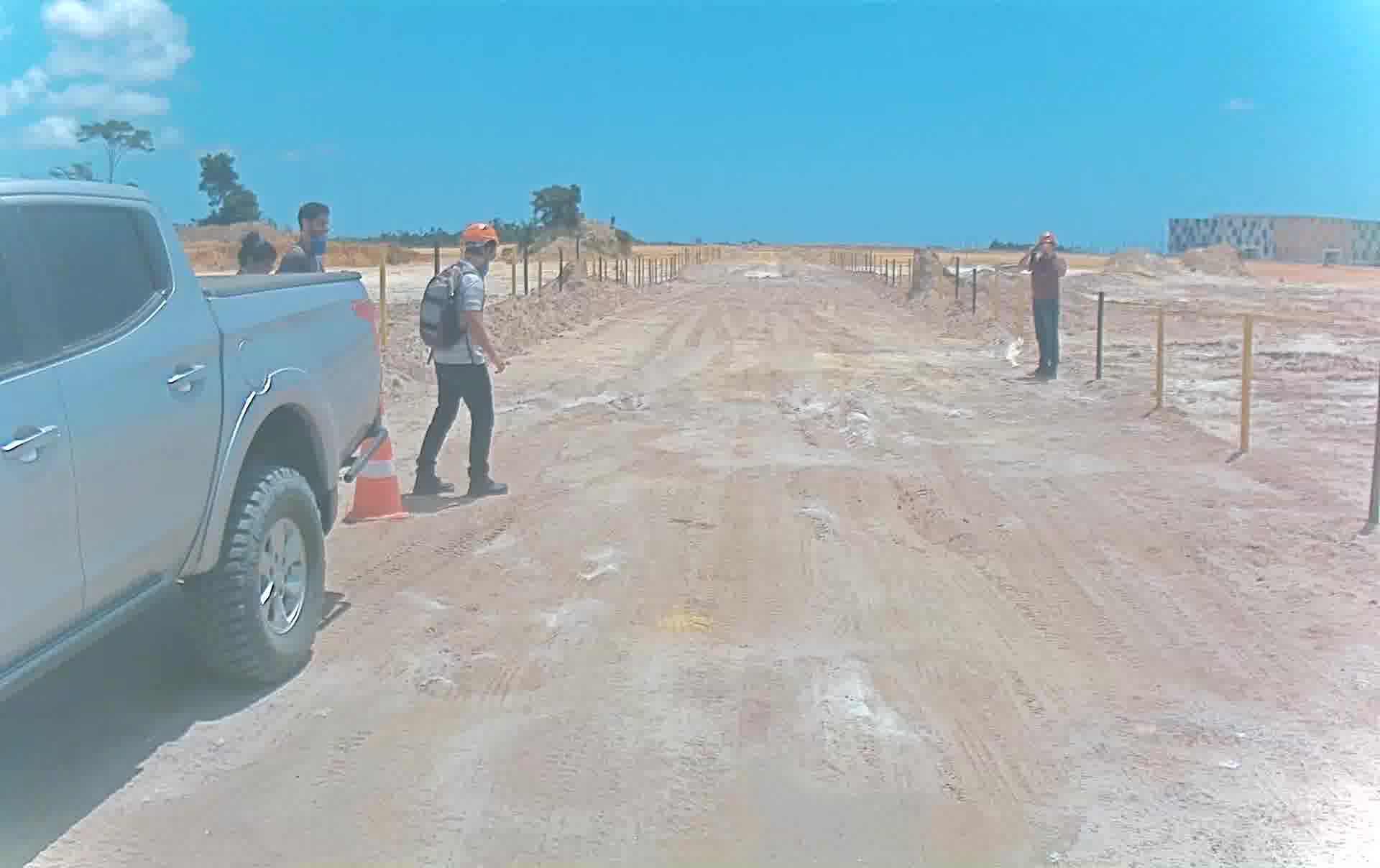}
		\caption{Original image.}
	\end{subfigure}%
	\hfill%
	\begin{subfigure}[b]{0.45\linewidth}
		\includegraphics[width=\linewidth,trim={2cm 0cm 0cm 2cm},clip]{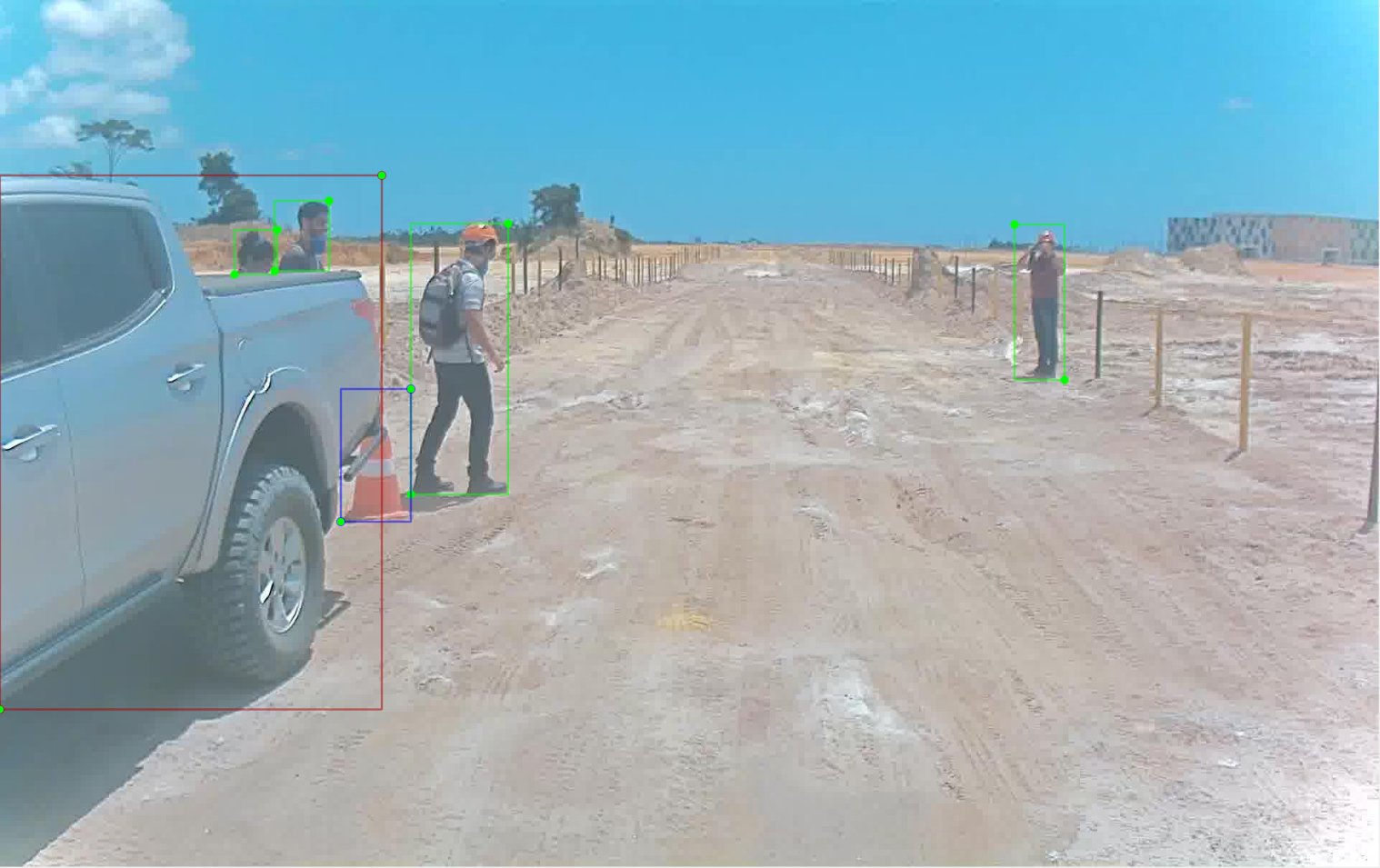}
		\caption{Detection and bounding box.}
	\end{subfigure}
	\hfill%
	\begin{subfigure}[b]{0.45\linewidth}
		\includegraphics[width=\linewidth,trim={2cm 0cm 0cm 2cm},clip]{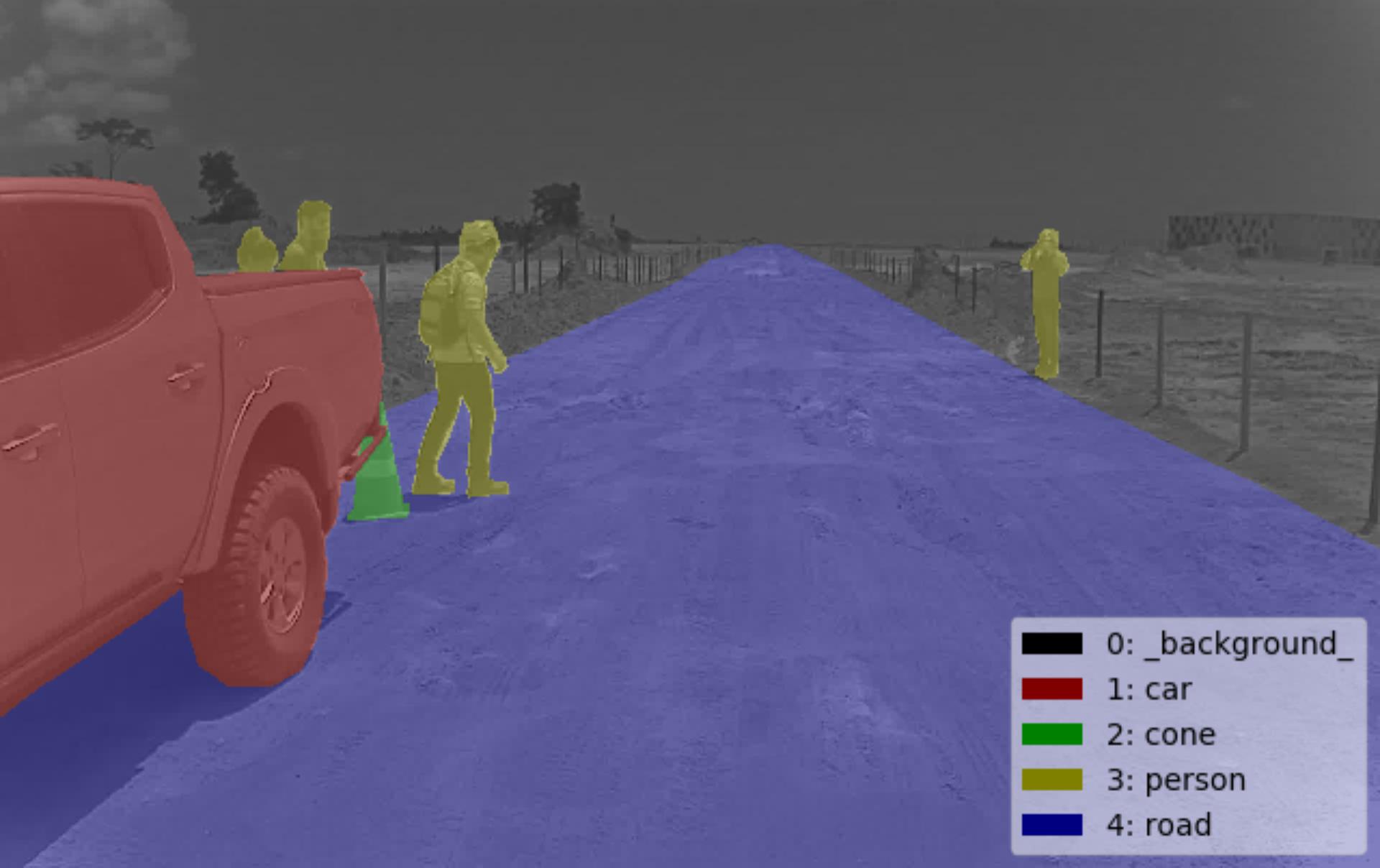}
		\caption{Semantic segmentation.}
	\end{subfigure}%
	\hfill%
	\begin{subfigure}[b]{0.45\linewidth}
		\includegraphics[width=\linewidth,trim={2cm 0cm 0cm 2cm},clip]{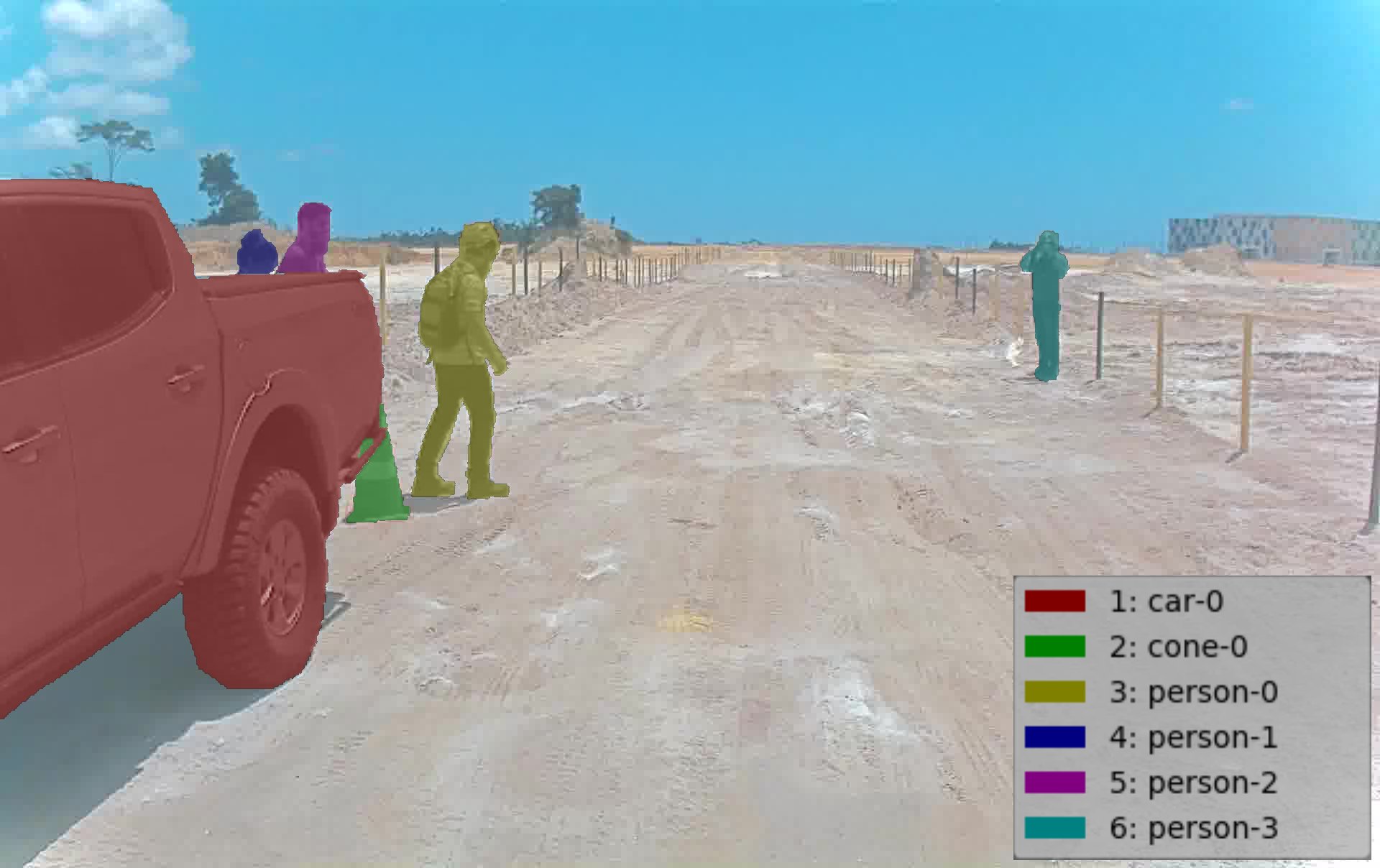}
		\caption{Instance segmentation.}
	\end{subfigure}
	\begin{subfigure}[b]{0.45\linewidth}
		\includegraphics[width=\linewidth,trim={2cm 0cm 0cm 2cm},clip]{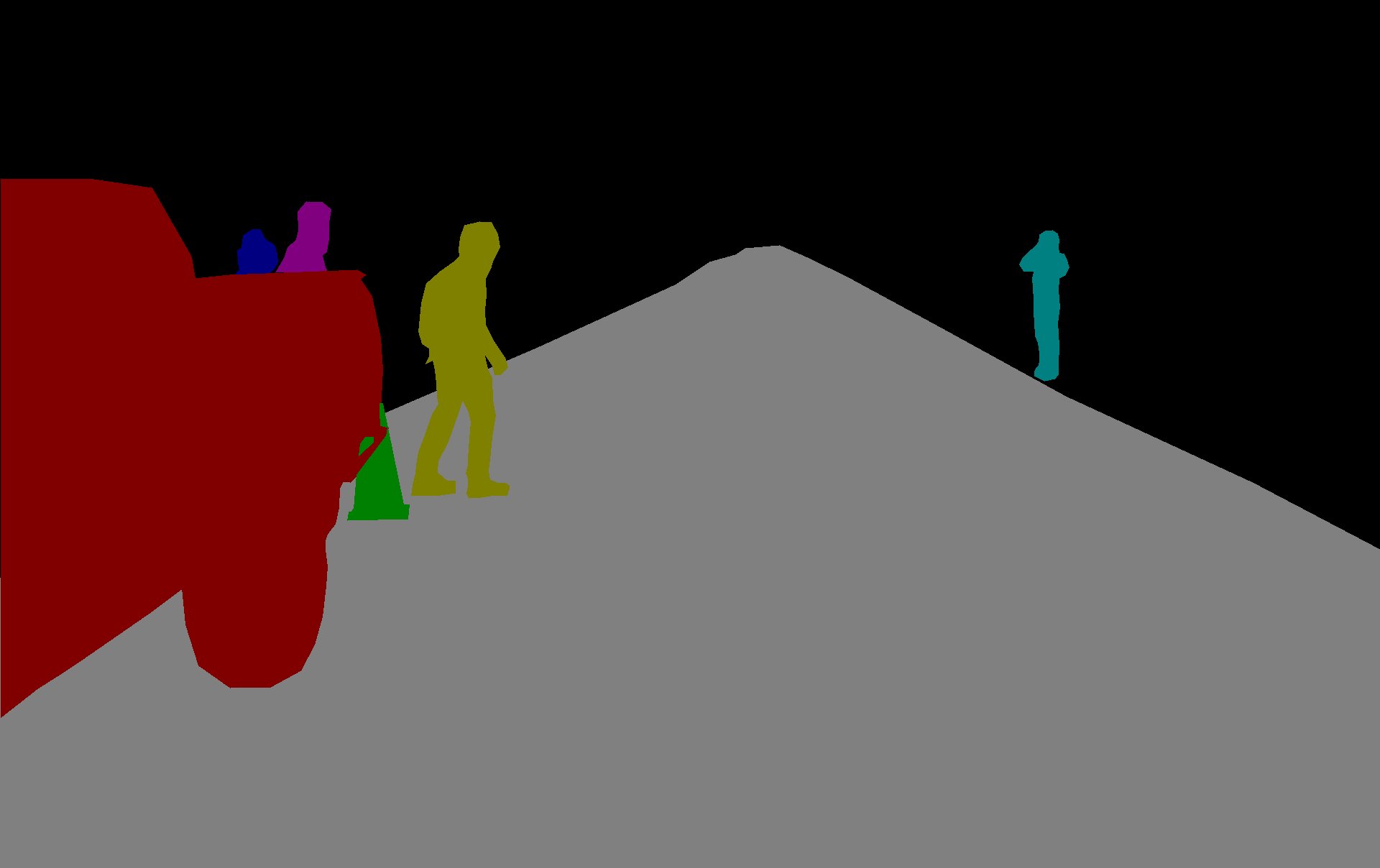}
		\caption{Panoptic Ground-truth.}
	\end{subfigure}%
	\hfill%
	\begin{subfigure}[b]{0.45\linewidth}
		\includegraphics[width=\linewidth,trim={2cm 0cm 0cm 2cm},clip]{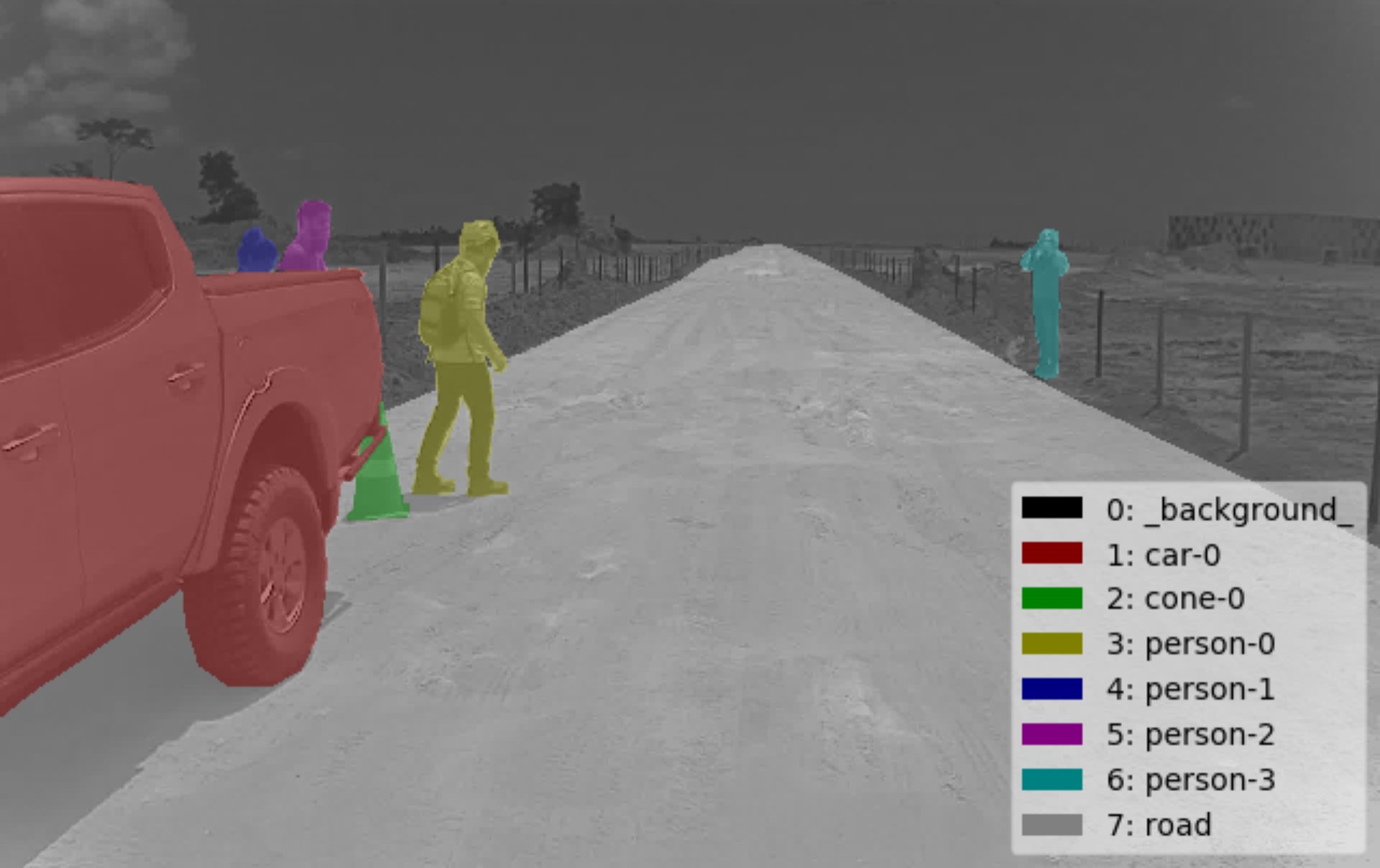}
		\caption{Panoptic Segmentation.}
	\end{subfigure}
	\caption{Types of annotations.}
	\label{fig:diferentes-tipos-de-anotacoes}
	\fautor
\end{figure}

The data were labeled suitable for panoptic segmentation.
Panoptic segmentation treats countable things like people and cars simultaneously with non-countable stuff such as roads and vegetation. This task unifies the semantic and instance segmentation (\autoref{fig:diferentes-tipos-de-anotacoes}) \cite{Alexander:2018:CoRR:PanopticSegmentation}. This study adopted this strategy because it allows generating ground truth to the instance segmentation and object detection in addition to semantic segmentation. Even though the focus of this work is the semantic segmentation of unpaved roads, this choice seemed to be prudent because it allows future research using the same dataset.

\newword{Panoptic}{is what allows showing or seeing the whole at one view}

\newword{Panoptic segmentation}{is an image segmentation method that unifies the concepts of semantic segmentation and instance segmentation. It treats countable things like people and cars simultaneously with non-countable stuff such as roads and vegetation}

This study has used the LabelMe \cite{Russell:2008:LabelMe} annotation style, applying polygons to outline the object. The results of each image annotation — groups of polygons and the respective classes associated with them — were written in a .json file (Annex \ref{anexos:annotation-code-style}). An identifier was attached to the label to ensure the correct annotation of different instances of the same class (e.g., \textit{person-0, person-1, ..., car-0, car-1, ..., car-n}).  On the other hand, in the labeling of non-countable stuff, such as the road, it was used only the label (e.g., \textit{road}).

\begin{figure}[htb]
	\begin{subfigure}[b]{0.32\linewidth}
		\includegraphics[width=\linewidth,trim={2cm 0cm 0cm 2cm},clip]{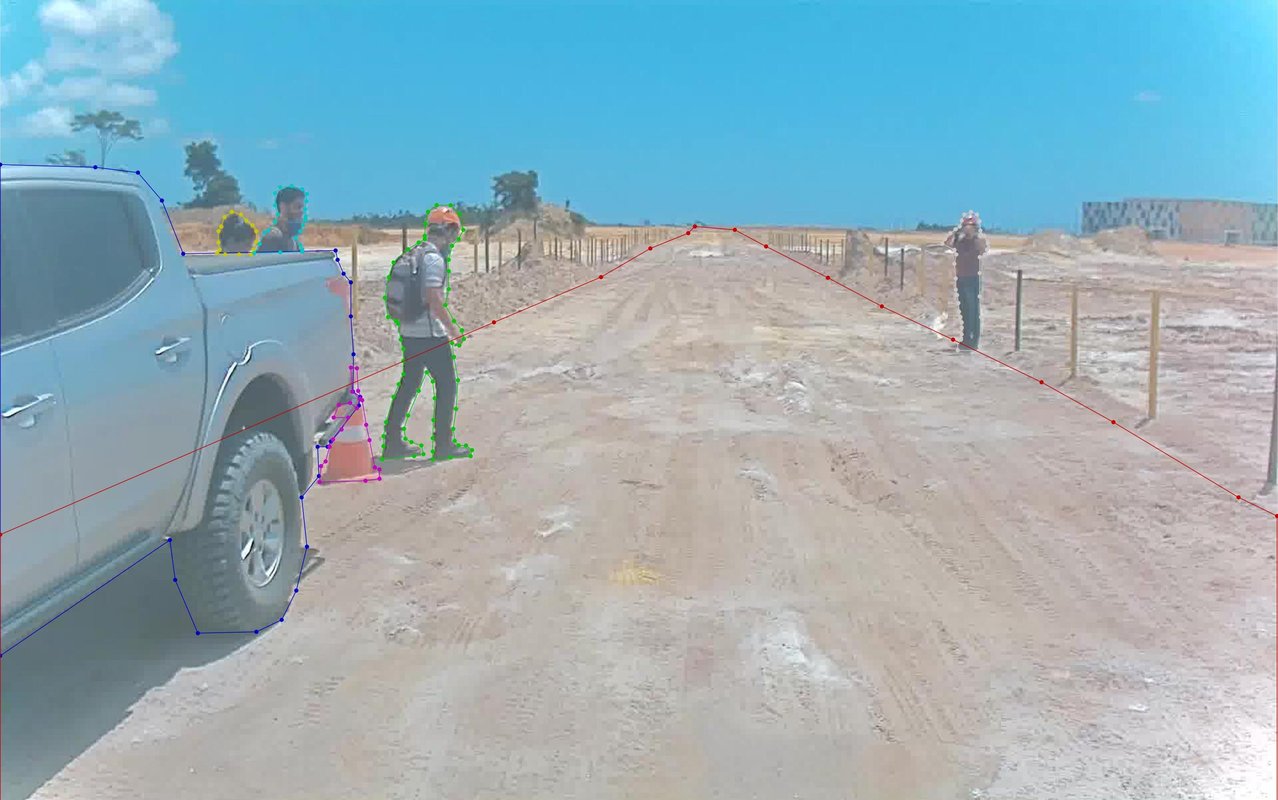}
	\end{subfigure}%
	\hfill%
	\begin{subfigure}[b]{0.32\linewidth}
		\includegraphics[width=\linewidth,trim={2cm 0cm 0cm 2cm},clip]{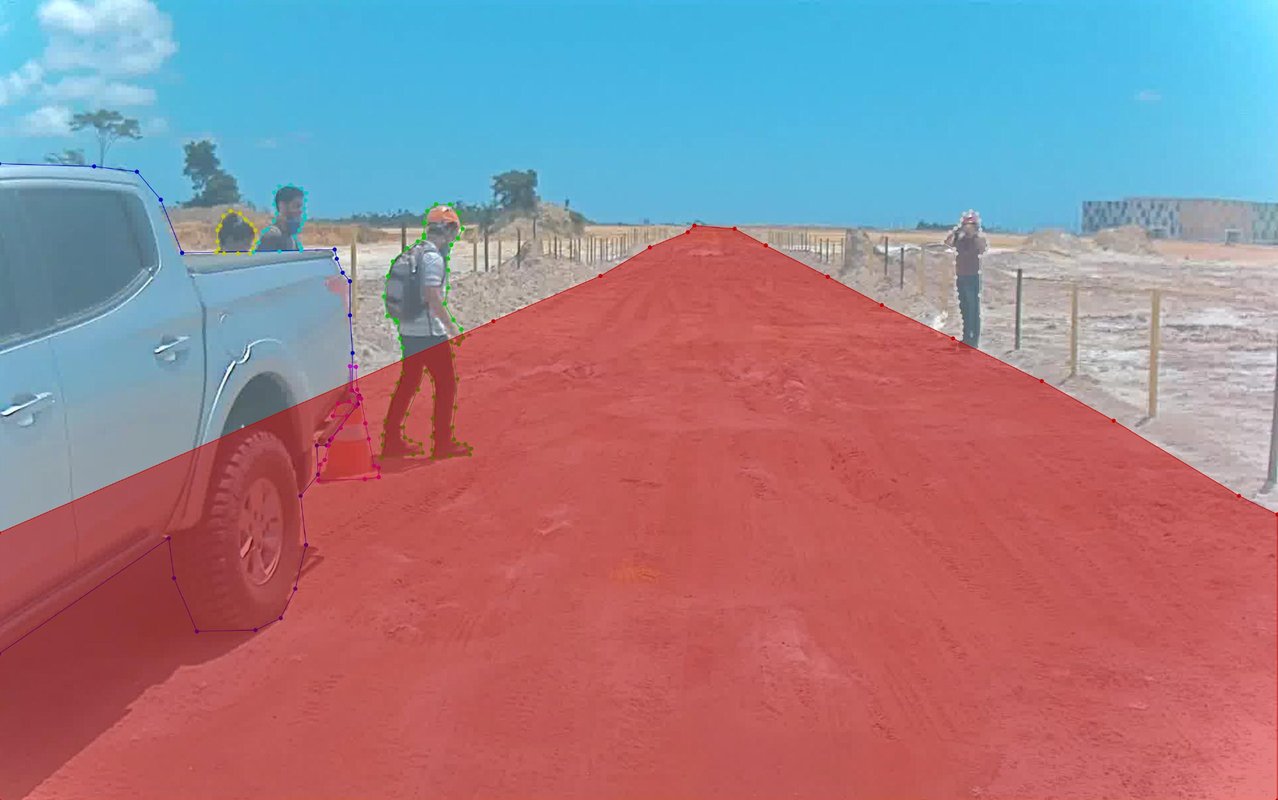}
	\end{subfigure}
	\hfill%
	\begin{subfigure}[b]{0.32\linewidth}
		\includegraphics[width=\linewidth,trim={2cm 0cm 0cm 2cm},clip]{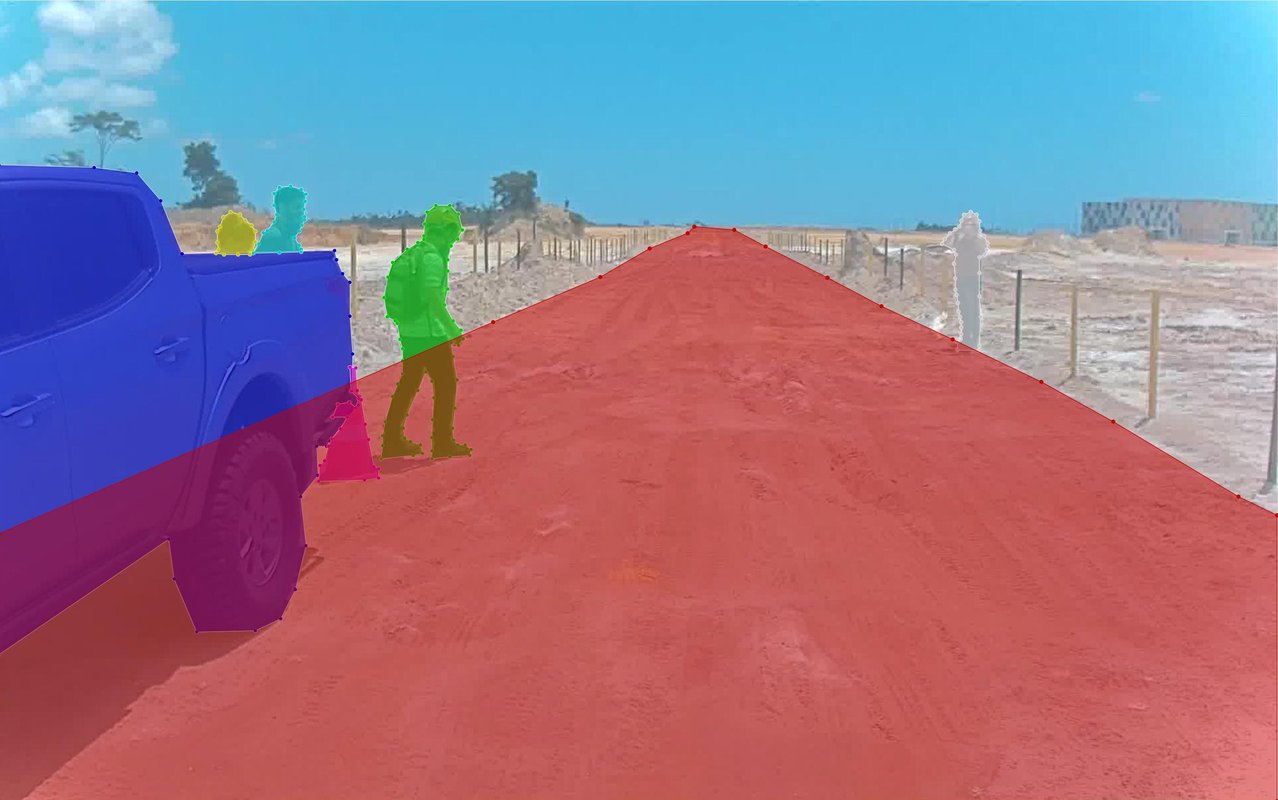}
	\end{subfigure}%
	\caption{Image annotation process.}
	\label{fig:processo-de-anotação-das-imagens}
	\fautor
\end{figure}

In the annotation process, the researchers annotated the road first and after all the elements over it so that the result was an annotation of layers over layers (\autoref{fig:processo-de-anotação-das-imagens}). This strategy has been used to speed up the creation of the dataset. To avoid overlapping the road over other classes like person or car, the script developed to convert the .json files into .png masks uses a pre-established order to render the information. 

This research considered only the segmentation of the traffic area and obstacles as being relevant to the visual perception. For this task, the segmentation of sky, buildings, and other elements not directly involved in the decision-making process to drive a vehicle was not required. Besides, the researchers found a limited number of relevant classes in those less dense off-road environments. So, they have opted for a limited number of annotated classes to decrease the effort and speed up the research development.

The strategy of focusing on a few groups has proved to be adequate to validate the concept. In total, eight classes were recorded, grouped into six distinct categories, prioritizing the traffic area (road) and obstacles encountered during several hours of data acquisition. The \autoref{tab:lista-declasses-e-categorias} shows the classes grouped in categories where only the road and background classes do not have multiple instances. The ground category has only the class road, the human group has just the person class, and the animal group has the animal class. On the other hand, the vehicle group has the classes car, motorcycle, truck, and bike. These elements are relevant to the research as it involves imminent obstacles and risks to driving. Besides that, there is also the cone class in the infrastructure group and the background class, including the elements not considered relevant.

As can be seen in \autoref{tab:lista-declasses-e-categorias}, our dataset has an imbalance. That happens due to the lack of some classes in unpaved environments. Those places are distant from downtown, where there are several cars and pedestrians. Also, there is a perceived scarcity regarding animals crossing the track.

Like the unpaved roads, the test track used to develop and validate the system also has a limited number of people and cars. Nevertheless, several datasets such as COCO~\cite{Lin:2014:ECCV:COCO}, Pascal VOC and Cityscapes~\cite{cordts:2016:cityscapes} have already covered the segmentation of people and animals. However, the segmentation of unpaved roads and traffic area in an off-road environment where the track has the same color and texture as the non-traffic area is our research contribution.

\begin{table}[htb]
	\centering
	\scriptsize
	\caption{List of classes and categories, average pixels occupied in all images, and the total number of occurrences.}
	\label{tab:lista-declasses-e-categorias}
	\begin{tabular*}{\textwidth}{@{\extracolsep{\fill}} l l c c} 
	   \specialrule{.1em}{.05em}{.05em}
		\bfseries Group &
		\bfseries Class &
		\bfseries Pixels Avg. (\%) &
		\bfseries Total instances \\ 
		\specialrule{.1em}{.05em}{.05em}
		Ground  & \textit{road}  & 47.20 & 11,508   \\
		Human  & \textit{person}  & 0.08 & 1,896  \\
		Animal & \textit{animal} & 0.001 & 27  \\
		Infrastructure   & cone  & 0.002 & 129   \\ 
		Void   & \textit{background} & 52.34 & 11,512   \\
		Vehicle  & \textit{car}   & 0.29 & 4,186   \\
		& \textit{moto}   & 0.006 & 114   \\
		& \textit{truck}  & 0.03 & 154  \\
		& \textit{bus}    & 0.03 & 101   \\
		& \textit{bike}   & 0.001 & 41   \\
		\specialrule{.1em}{.05em}{.05em}
	\end{tabular*}
	\fautor
\end{table}

\subsubsection{Data description}
In total, this dataset has 11,479 annotated images recorded in different places (\autoref{tab:adverse-situation}). It has been collected 823 images in rainy conditions and 5,135 in the daytime on unpaved roads, 1,556 on the off-road test track during the day, 1,546 in the late afternoon, and 1,953 at night (\autoref{tab:dataset}). Furthermore, there is possible to generate additional synthetic data through a script to increase the dataset. When activated on the training, this script produces dynamic images applying filters and random cuts from real images annotated.

\begin{table}[htb]
	\centering
	\scriptsize
	\caption{Annotated images.}
	\label{tab:dataset}
	\begin{tabular*}{\textwidth}{@{\extracolsep{\fill}} l l c c c c}
		\specialrule{.1em}{.05em}{.05em}
		\bfseries Type    &
		\bfseries Places          &
		\bfseries Day   &
		\bfseries Evening&
		\bfseries Night&
		\bfseries Rain \\ 
		\specialrule{.1em}{.05em}{.05em}
		Paved   & Salvador Metropolitan Area & 202   & --     & --     & 209 \\ 
		Unpaved & Salvador Metropolitan Area & 5,135 & --     & --     &  823 \\
		Off-road& Test Track      & 1,556 & 1,546 & 1,953  &  55 \\
		\specialrule{.1em}{.05em}{.05em}
		Total	&                 & 6,893 & 1,546 &  1,953  & 1,087  \\
		\specialrule{.1em}{.05em}{.05em}
	\end{tabular*}
\end{table}

Unlike other datasets such as Cityscapes~\cite{cordts:2016:cityscapes} and KITTI~\cite{kitti-menze:2015:cvpr} that annotate walls, buildings, sky, tree, and sidewalk, our study have only annotated the traffic zone and non-traffic area, in addition to dynamic obstacles such as cars, people, and animals. This approach facilitates the annotation task and keeps the algorithm focused on segmenting what is relevant for the vehicle on the road. \autoref{tab:lista-declasses-e-categorias} shows the classes annotated in the scene.

As far as the authors know, the presented dataset in this research and the Mapillary \cite{Neuhold:2017:Mapillary} are the only ones that cover paved, non-paved off-road, and adverse conditions altogether, as shown in \autoref{tab:comparison_datasets}. However, Mapillary has few off-road and unpaved image samples compared to the one presented here. Besides, our dataset has a high number of pixels labeled and an annotated pixel density of 47.66\%, even not considering the background label.

\simbolo{\%}{Indicates the percent sign.}

\begin{table}[htb]
	\centering
	\scriptsize
\caption{Comparison between ours Kamino dataset and other ones.}
\label{tab:comparison_datasets}
\begin{tabular*}{\textwidth}{@{\extracolsep{\fill}} l c c c c c c c c}
\specialrule{.1em}{.05em}{.05em}
\multicolumn{1}{c}{\bfseries Dataset} &
  \multicolumn{1}{c}{\bfseries \begin{tabular}[c]{@{}c@{}}\# \\ images\end{tabular}} &
  \multicolumn{1}{c}{\bfseries \begin{tabular}[c]{@{}c@{}}\# \\ classes\end{tabular}} &
  \multicolumn{1}{c}{\bfseries Paved} &
  \multicolumn{1}{c}{\bfseries \begin{tabular}[c]{@{}c@{}}Non-\\ paved\end{tabular}} &
  \multicolumn{1}{c}{\bfseries \begin{tabular}[c]{@{}c@{}}Off-\\ road\end{tabular}} &
  \multicolumn{1}{c}{\bfseries \begin{tabular}[c]{@{}c@{}}Adv. \\ cond.\end{tabular}} &
  \multicolumn{1}{c}{\bfseries Semantic} &
  \multicolumn{1}{c}{\bfseries Instance}
   \\
\specialrule{.1em}{.05em}{.05em}

A2D2             & 41k  & 38  & \checkmark & \ding{55}                        & \ding{55}                        & \checkmark & \checkmark & \checkmark \\
Mapillary & 25k  & 152 & \checkmark & {\color[HTML]{FE0000} \checkmark} & {\color[HTML]{FE0000} \checkmark} & \checkmark & \checkmark & \checkmark\\
CamVid           & 700  & 32  & \checkmark & \ding{55}                        & \ding{55}                        & \ding{55} & \checkmark & \ding{55} \\
Cityscapes       & 5k   & 30  & \checkmark & \ding{55}                        & \ding{55}                        & \ding{55} & \checkmark & \checkmark \\
KITTI            & 5k   & 30  & \checkmark & \ding{55}                        & \ding{55}                        & \ding{55} & \checkmark & \checkmark  \\
YCOR             & 1k   & 8   & \ding{55} & \ding{55}                        & \checkmark                        & \checkmark & \checkmark & \ding{55}  \\
DeepScene        & 372  & 6   & \ding{55} & \ding{55}                        & \checkmark                        & \ding{55} & \checkmark & \ding{55} \\
\bfseries{Kamino}           & 11.5k & 10  & \checkmark & \checkmark                        & \checkmark                        & \checkmark & \checkmark & \checkmark \\

\specialrule{.1em}{.05em}{.05em}
\end{tabular*}
	\fautor
\end{table}

\autoref{tab:valores-absolutos-e-media-de-pessoas-veiculos-e-animais-por-imagem} presents a comparison regarding the number of vehicles, animals, and people between our dataset and Cityscapes~\citeonline{cordts:2016:cityscapes} or KITTI~\citeonline{kitti-menze:2015:cvpr}.  As expected, our dataset's total number of dynamic entities occurrences is smaller due to differences among the acquisition environments.

\begin{table}[htb]
	\centering
	\scriptsize
	\caption{Absolute and average values of instances per image.}
	\label{tab:valores-absolutos-e-media-de-pessoas-veiculos-e-animais-por-imagem}
	\begin{tabular*}{\textwidth}{@{\extracolsep{\fill}} l c c c c c c}
	\specialrule{.1em}{.05em}{.05em}
		\bfseries Dataset & 
		\bfseries Person &
		\bfseries Vehicle &
		\bfseries Animal &
		\bfseries P\%  &
		\bfseries V\%  &
		\bfseries A\%\\ 
		\specialrule{.1em}{.05em}{.05em}
		CamVid        & --      & --    & 0  & --  &  --   & 0.0  \\
		Cityscapes   & 24.4k     & 41.0k    & 0      & 7.0     & 11.8      & 0.0     \\ 
		KITTI        & 6.1k      & 30.3k    & 0      & 0.8     &  4.1      & 0.0     \\ 
		YCOR        & --      & -- & --  & --  &  --  & --  \\
		DeepScene        & 0k      & 0k    & -- & 0.0  &  0.0  & -- \\
		\bfseries{Kamino}               & 1.9k      & 4.56k    & 27     & 0.08      & 0.37    & 0.001 \\ 
		\specialrule{.1em}{.05em}{.05em}
	\end{tabular*}
	\fautor
\end{table}

In addition to the annotated data, this dataset has several videos and LiDARs point cloud collected during the development. Altogether there were four LiDARs, two Velodyne VLP-16, and two Quanergy M8. It was also recorded data from 4 SEKONIX cameras with 120º FOV and 3 SEKONIX cameras with 60º FOV.

\section{Experimental setup}
\label{sec:experimental_setup}
This section describes some aspects related to the training setup (\autoref{subsec:dataset}). Subsection \ref{performance-evaluation} presents the statistical parameters used in the performance evaluation of CMSNet, \autoref{training-setup} describes the training setup, \autoref{inference-time-evaluation} shows the process for measuring and evaluating the inference time, and finally, \autoref{cmsnet-arrangements} shows the CMSNet architecture arrangments.

\subsection{Dataset}
\label{subsec:dataset}
The dataset presented in this work has 11,479 labeled images. However, the study experiments used just a data subset to speed up the training process for the CMSNet arrangements. Such a subset has been named Kamino-Small. It has a total of 5,523 images in several situations, as shown in \autoref{tab:subset_composition}. Altogether, they are 4,026 samples for training, 449 for validation, and 1,048 for testing. These data are distributed between daytime, raining, night, and evening. Furthermore, in the off-road test track, some images also have dust.

In addition to a reduced set of images, it also chose not to include some classes in training and testing. Groups such as bus, motorcycle, animal, and bike have not been considered. These classes are rare in the proposed dataset and, in some cases, are not sufficient for the test step. A similar approach of merging or ignoring some groups in tests is also used in other datasets such as Cityscapes~ \cite{cordts:2015:cvprw-cityscapes} and  \citeonline{Semantic-Forested:Valada:2016}.

\begin{table}[htb]
	\centering
	\scriptsize
	\caption{Distribution of data in training, validation and testing sets.}
	\label{tab:subset_composition}
	\begin{tabular*}{\textwidth}{@{\extracolsep{\fill}} l c  c  c  c}
		\specialrule{.1em}{.05em}{.05em}
		\bfseries Condition &
		\bfseries Training &
		\bfseries Validation &
		\bfseries Testing &
		\bfseries All \\
		\specialrule{.1em}{.05em}{.05em}
		Daytime        & 1,471 ($73.5\%$) & 164 ($8.2\%$) &   367 ($18.3\%$) & 2,002 \\
		Daytime\spi{1} &   666 ($73.1\%$) &  74 ($8.1\%$) &   171 ($18.8\%$) &   911 \\
		Raining        &   539 ($72.2\%$) &  60 ($8.0\%$) &   148 ($19.8\%$) &   747 \\
		Night  \spi{1} &   751 ($71.9\%$) &  84 ($8.0\%$) &   209 ($20.0\%$) & 1,044 \\
		Evening\spi{1} &   599 ($73.1\%$) &  67 ($8.2\%$) &   153 ($18.7\%$) &   819 \\
		Total          & 4,026 ($72.9\%$) & 449 ($8.1\%$) & 1,048 ($19.0\%$) & 5,523 \\
		\specialrule{.1em}{.05em}{.05em}
	\end{tabular*}
	
	{\textit{1} -- denotes that data have frames with dust condition.}
	\fautor
\end{table}



\subsection{Performance Evaluation}
\label{performance-evaluation}
It was needed to quantify the similarities between expected and inferred results regarding the test sets to calculate the performance of the study proposal. The primary statistical parameter used to do that was the Jaccard similarity coefficient. The average of the similarities for image elements and between all images has also been used in the evaluation, as well as other attributes commonly found for semantic segmentation analysis:

\begin{itemize}
	\item \textbf{Pixel accuracy} ($P_{acc}$). It is a simple accuracy metric that tells us the percentage of pixels in the image that are correctly classified. \autoref{eq:pixel-acuracia} shows how this indicator is calculated, with $t_{i}$ representing the total number of pixels for class $i$ and $\sum_{i}t_{i}$ representing the sum of all pixels belonging to all classes — the total amount of pixels in the image. Furthermore, $n_{ii}$ represents the number of pixels of the class $i$ correctly inferred as belonging to the class $i$. 
	This parameter can also be expressed by class ($CP_{acc}$, Eq. \ref{eq:per-classe-pixel-acuracia}) instead of a global way. In this case, the pixel accuracy is calculated per an individual class instead of for the whole image.
	
	\begin{equation} \label{eq:pixel-acuracia}
	P_{acc}= \frac{\sum_{i}n_{ii}}{\sum_{i}t_{i}}
	\end{equation}

	\begin{equation} \label{eq:per-classe-pixel-acuracia}
	CP_{acc}= \frac{n_{ii}}{t_{i}}
	\end{equation}
	
	\item \textbf{Mean accuracy} ($mCP_{acc}$). The average accuracy among all classes can be calculated as shown by \autoref{eq:acuracia-media}. This equation shows the sum of the accuracy calculated for each class  $\sum_{i}\frac{n_{ii}}{t_{i}}$ divided by the number of classes $n_{cl}$.
	
	\begin{equation} \label{eq:acuracia-media} 
	mCP_{acc}= \frac{1}{n_{cl}}\sum_{i}\frac{n_{ii}}{t_{i}}
	\end{equation}

	\item \textbf{Jaccard Similarity Coefficient or Intersection over Union} ($IoU$). It is a parameter used to measure the diversity and similarity of sample sets. In the context of this research, it is a metric that quantifies the percentage of overlap between ground truth and the segmentation mask inferred by the algorithm (see Eq. \ref{eq:intersecao-uniao}) 
	in which $\sum_{j}n_{ji}$ representing the number of pixels in all classes $j$ that are inferred as belonging to the class $i$. 
	
	\begin{equation} \label{eq:intersecao-uniao}
	IoU= \frac{n_{ii}}{t_{i}+\sum_{j}n_{ji}-n_{ii}}
	\end{equation}
	
	\item \textbf{Mean Intersection over Union} ($mIoU$). The mean of intersection over the union between classes is very similar to the previous metric. However, it calculates the average of $IoU$ between classes. The \autoref{eq:media-da-intersecao-uniao} shows how this metric is calculated
	\cite{Long:2015:FCN:ieeecvpr,Liu:2018:AIR:Recent-progress-in-semantic-s}.
	
	\begin{equation} \label{eq:media-da-intersecao-uniao}
	mIoU = \frac{1}{n_{cl}}\sum_{i}IoU
	\end{equation}
	
	\item \textbf{Frequency Weighted Intersection over Union} (FWIoU). Refers to the average of the intersection over union between classes weighted by the frequency of occurrence as in \autoref{eq:frequencia-ponderada}  \cite{Long:2015:FCN:ieeecvpr, Liu:2018:AIR:Recent-progress-in-semantic-s}.
	
	\begin{equation} \label{eq:frequencia-ponderada}
	FWIoU = \frac{1}{\sum_{k}t_{k}}\sum_{i}t_{i}IoU
	\end{equation}
\end{itemize}

\subsection{Inference time evaluation}
\label{inference-time-evaluation}
\simbolo{\sigma}{Standard deviation.}
The thesis study has used the mean and the standard deviation $ \sigma $ (Std.) over a sequence of 500  measurement iterations to evaluate the inference time and ensure the results' reliability. It also has calculated the boxplot parameters and displayed them graphically.

\subsection{CMSNet arrangements}
\label{cmsnet-arrangements}
This work has presented the CMSNet framework. 
Their different modules can be configured to build several architectures solutions. It can use an output stride of 8 or 16 by choosing where the dilated convolution starts in the backbone pipeline. Furthermore, the architecture can have either spatial pyramid pooling (SPP), atrous spatial pyramid pooling (ASPP), or global pyramid pooling (GPP). Besides, it may have a shortcut with high-resolution features. \autoref{tab:cmsnet-arrangements} shows the different arrangements and their configuration considered in the experiments. From here on, this work use only the names defined in \autoref{tab:cmsnet-arrangements} to refer to each of the arrangements.

\begin{table}[htb]
	\centering
	\scriptsize
	\caption{Different arrangements for CMSNet.}
	\label{tab:cmsnet-arrangements}
	\begin{tabular*}{\textwidth}{@{\extracolsep{\fill}} c c c c c} 
		\specialrule{.1em}{.05em}{.05em}
		\bfseries Name &
		\bfseries Abbr. &
		\bfseries Output Stride & 
		\bfseries Pyramid  & 
		\bfseries Shortcut \\
		\specialrule{.1em}{.05em}{.05em}
		CMSNet-M0 & CM0 & 8    & GPP   &  No   \\
		CMSNet-M1 & CM1 & 8    & SPP      &  No   \\
		CMSNet-M2 & \textbf{CM2} & \textbf{8}    & \textbf{ASPP}     &  No   \\
		CMSNet-M3 & CM3 & 16   & GPP      & No    \\
		CMSNet-M4 & CM4 & 16   & SPP      & No    \\
		CMSNet-M5 & \textbf{CM5} & \textbf{16}   & \textbf{ASPP}     & No \\
		CMSNet-M6 & CM6 & 16   & GPP      & Yes \\
		CMSNet-M7 & CM7 & 16   & SPP      & Yes \\
		CMSNet-M8 & CM8 & 16   & ASPP     & Yes \\
		\specialrule{.1em}{.05em}{.05em}
	\end{tabular*}
	\fautor
\end{table}

\subsection{Training setup}
\label{training-setup} 
The training was performed in a computer with a GPU RTX 2060 with 6 GB and a 9th generation i7 processor, having six cores and capable of running 12 threads. After tunning the hyperparameters, the study has included the validation set in the training processes to increase diversity. Altogether, it was used 4.475 images for training, randomly distributed between all conditions and places. The research trained each scenario for 200 epochs using a batch of 4 images using a learning rate of 0.007 with the first-order polynomial decaying until 0. The study also has used artificial data augmentation techniques to help avoid over-fitting and increase training performance. 


\chapter{Experiments and Findings}
\label{chapter:experiments-and-finds}

This chapter includes the results of the experiments carried out during the study (\autoref{sec:results_and_discussion}), and the discussion (\autoref{sec:discussion}) where is elaborated and describes the main findings of the research compared to other works strengthening the significance of the study -- how it contributes to the area.

\section{Results}
\label{sec:results_and_discussion}

This section displays the results of the tests carried out in this study. Subsection \ref{subsec:ablation-study-for-cmsnet} shows an ablation study for arrangement composition with different parts of CMSNet. Subsection \ref{subsec:results-on-kamino-dataset} compares the solution results for the urban environment in unpaved and off-road scenarios against the system designed for this proposal. Subsection \ref{subsection:cmsnet-on-adverse-environmental-conditions} and \ref{subsection:cmsnet-on-synthetic-impairments} show the performance of the proposed solution on adverse visibilities condition such as night, rain, dust, noise, and fog, and \autoref{subsection:comparing-the-adverse-conditions-results} compare the results in all adverse conditions. Subsection \ref{subsection:results-on-deepscene-dataset} shows the comparison of the solution in other datasets.

\subsection{Ablation study for CMSNet}
\label{subsec:ablation-study-for-cmsnet}
\autoref{tab:summary-testes-com-configuracoes-para-backbone-mobilenetv2} shows the results of an investigation carried out with the architecture arrangements defined in \autoref{tab:cmsnet-arrangements}. In this ablation study was investigate how the different arrangements perform concerning $mIoU$, $FWIoU$, $mCP_{acc}$, and $CP_{acc}$. Besides, \autoref{tab:summary-testes-com-configuracoes-para-backbone-mobilenetv2} also shows the number of parameters demanded by each arrangement combination.

This study allows us to observe that the output stride of 8 makes a positive effect of 1\% in the ASPP $mIoU$ (CMSNet-M2 and CMSNet-M5). However, lower values for output stride harm the inference time, as can be seen in \autoref{subsec:results-on-kamino-dataset}. Despite increasing the processing time, the researcher has not seen the positive effect on the $mIoU$ reflected in all arrangements using this configuration, as is the case of CMSNet-M0 and CMSNet-M1. The researcher also have noted that the metric $FWIoU$ suffers low variation independent of the CMSNet configuration, which means that the classes with the worst prediction accuracy are the ones that have small objects in the image.

In general, architectures with ASPP modules (M2, M5, and M8) have performed better than ones with SPP (M1, M4, and M7), which in turn have achieved better results than GPP ones (M0, M3, and M6). However, the arrangements with ASPP demand more parameters than others. The researcher also has noted that configurations with shortcuts (M6, M7, and M8) have not performed better, although they have more parameters than the other arrangements.

The study supposes that ASPP configurations have demanded more parameters because their implementation used standard 2D convolutions instead of the factored one used in the SPP module. The factored convolutions are composed of depthwise and pointwise convolutions and are computationally less expensive.

\begin{table}[htb]
  \centering
  \scriptsize
	\caption{Tests with settings for \textit{backbone} MobileNetV2.}
	\label{tab:summary-testes-com-configuracoes-para-backbone-mobilenetv2}
  \begin{tabular*}{\textwidth}{@{\extracolsep{\fill}} c c c c c c} 
	\specialrule{.1em}{.05em}{.05em}
		\bfseries Name & 
		\bfseries mIoU\% & 
		\bfseries FWIoU\% & 
		\bfseries mCP\textsubscript{acc}\% & 
		\bfseries P\textsubscript{acc}\% & 
		\bfseries Param.\\
		\specialrule{.1em}{.05em}{.05em}
	 	CM0 & 84.66 & 95.72 & 94.33 & 97.78 & 2,144 k \\
	 	CM1 & 84.15 & 95.97 & 91.91 & 97.91 & 2,033 k \\
	 	\textbf{CM2} & \textbf{86.98} & 96.51 & 92.11 & 98.21 & \textbf{4,408 k} \\
	 	CM3 & 85.02 & 96.21 & 91.89 & 98.05 & 2,144 k \\
	 	CM4 & 85.25 & 96.30 & 91.88 & 98.09 & 2,033 k \\
	 	\textbf{CM5} & \textbf{85.01} & 96.33 & 91.48 & 98.11 & \textbf{4,408 k} \\
	 	CM6 & 80.67 & 96.08 & 85.99 & 97.97 & 2,150 k \\
	 	CM7 & 83.62 & 96.27 & 89.21 & 98.07 & 2,039 k \\
	 	CM8 & 84.02 & 96.31 & 89.72 & 98.09 & 4,414 k \\
    \specialrule{.1em}{.05em}{.05em}
	\end{tabular*}
	\fautor
\end{table}

\subsection{Visual perception results on Kamino dataset}
\label{subsec:results-on-kamino-dataset}

\subsubsection{Comparison with pre-trained networks}
\autoref{tab:methods_eval} shows the results for different arrangements presented in \autoref{tab:cmsnet-arrangements} compared with other architectures trained for fully urban environments. The architectures used for comparing were PSPNet  \cite{Zhao:2017:PSPNet:ieeecvpr} and some variations of DeepLab -- MNV2, Xc65, and Xc71  \cite{Sandler:2018:MobileNetV2:IEEE-CVF,Chen:2018:deeplab:ieeeTPAMI,Chen:2018:EncoderDecoderWA:ECCV:deeplabv3Plus}. Cityscapes  \cite{cordts:2016:cityscapes} was the urban dataset used in those networks. The link for pre-trained networks used in this experiment are: PSPNet \footnote{PSPNet url: \url{https://drive.google.com/file/d/1vZkk9nLvM9NNBCVCuEjnoXms30OMcZ8K}}, DeepLab+MNV2 \footnote{DeepLab+MNV2 url: \url{http://download.tensorflow.org/models/deeplabv3\_mnv2\_cityscapes\_train\_2018\_02\_05.tar.gz}}, DeepLab+Xc6 \footnote{DeepLab+Xc6 url: \url{http://download.tensorflow.org/models/deeplabv3\_cityscapes\_train\_2018\_02\_06.tar.gz}}, and DeepLab+Xc7 \footnote{DeepLab+Xc7 url: \url{http://download.tensorflow.org/models/deeplab\_cityscapes\_xception71\_trainfine\_2018\_09\_08.tar.gz}}.

The researchers have included the most common classes between urban and off-road datasets in this experiment. The study has used the classes road, car, person, and background (everything else). It has used $mIoU$ for all used classes and $IoU$ for individual ones presented in \autoref{tab:methods_eval}.

As previously presented in \autoref{fig:test_pspnet_deeplab_cityscape_unpaved_roads}, this thesis also shows quantitatively in \autoref{tab:methods_eval} that pre-trained architectures with those urban datasets like Cityscapes do not perform so well in non-paved and off-road environments. Despite using more parameters, such architectures performed worst even in classes like car and person. 

The results in \autoref{tab:methods_eval} show that the DeepLab+MNV2 trained with Cityscape urban dataset has achieved worse results. It has reached 31.46\% of $mIoU$ (All) and has obtained only 3.57\% of $IoU$ for the class person. On the other hand, PSPNet and DeepLab+Xc65 have reached 57.83\% and 55.68\% of $mIoU$, respectively. Nevertheless, their results have been far from those achieved by our approach.

\begin{table}[htb]
	
  \centering
  \scriptsize
  \caption{Results of the semantic segmentation on the categories of the Kamino dataset.}
  \label{tab:methods_eval}
  \begin{tabular*}{\textwidth}{@{\extracolsep{\fill}} l c c c c c c c c c c}
    \specialrule{.1em}{.05em}{.05em}
    \multirow{2}*{\bfseries Name}  & \multicolumn{4}{c}{\bfseries IoU (\%)} & \multirow{2}*{\bfseries mIoU(\%)} &  \multicolumn{2}{c}{\bfseries Batch 1} &
    \multicolumn{2}{c}{\bfseries Batch 4} \\
    \cline{2-5}\cline{7-8}\cline{9-10}
     & 
    \bfseries Road &  
    \bfseries Car & 
    \bfseries Person &  
    \bfseries Bg & 
    & 
    \bfseries FPS & 
    \bfseries Std. $ \sigma $ (\%) &
    \bfseries FPS & 
    \bfseries Std. $ \sigma $ (\%) \\
    \specialrule{.1em}{.05em}{.05em}
	 CM0 & 95.72 & 75.92 & 71.06 & 95.96 & 84.66 & 19.16 & 4.92 & 20.59 & 8.77\\
	 CM1 & 95.96 & 74.16 & 70.22 & 96.27 & 84.15 & 19.37 & 4.08 & 20.52 & 5.88\\
	 CM2 & 96.51 & 78.74 & 75.89 & 96.78 & 86.98 & 16.46 & 3.24 & 17.03 & 8.37\\
	 CM3 & 96.23 & 76.71 & 70.63 & 96.49 & 85.02 & 28.87 & 5.38 & 32.65 & 8.95 \\
	 CM4 & 96.31 & 77.00 & 71.12 & 96.59 & 85.25 & 27.77 & 3.48 & 32.82 & 4.43 \\
	 CM5 & 96.35 & 75.43 & 71.63 & 96.62 & 85.01 & 27.14 & 3.78 & 30.24 & 3.88 \\
	 CM6 & 96.11 & 77.08 & 53.03 & 96.47 & 80.67 & 28.10 & 3.10 & 32.77 & 4.29 \\
	 CM7 & 96.27 & 75.57 & 66.05 & 96.60 & 83.62 & 27.10 & 4.79 & 32.81 & 3.44 \\
	 CM8 & 96.34 & 73.81 & 69.32 & 96.63 & 84.02 & 26.64 & 4.46 & 30.37 & 3.39 \\
	PSPNet       & 63.22 & 44.25 & 54.12 & 69.70 & 57.83 & 2.79 & 9.14 & -- & -- \\
	DMNV2 & 59.39 & 9.03 & 3.57 & 53.83 & 31.46 & 5.90 & 8.34 & -- & --\\
	DLX65 & 65.92 & 46.10 & 52.54 & 58.15 & 55.68 & 0.69 & 7.30 & -- & -- \\
	DLX71 & 63.09 & 55.30 & 8.66 & 64.93 & 47.99 & 2.32 & 8.00 & -- & -- \\
    \specialrule{.1em}{.05em}{.05em}
  \end{tabular*}
	\fautor
\end{table}

These results suggest that the perception subsystems developed for autonomous vehicles aiming at a well-paved urban environment may not be suitable for developing countries or could be restricted to the set of roads in urban centers. This restriction could limit the implementation of autonomous systems in cargo vehicles, such as buses and trucks.

\subsubsection{Inference time comparison}
The study also has calculated the mean and standard deviation $ \sigma $ (Std.) for frames per second (FPS) achieved for each one of the arrangements in \autoref{tab:cmsnet-arrangements} and compared them with PSPNet, DeepLab+MNV2 (DMNV2), DeepLab+Xc6 (DLX65), and DeepLab+Xc7 (DLX71). The \autoref{tab:methods_eval} shows the inference time in FPS and standard deviation $ \sigma $ in percentage for each one of these architectures. Such times were calculated using a GPU RTX 2060 and a CPU core i7. 

As it can be seen from \autoref{tab:methods_eval}, the proposed architecture arrangements have achieved higher FPS and lower standard deviation $ \sigma $ than PSPNet and DeepLabs variations. The  DLX65 produced the worst-case in FPS, and CMSNet-M4 achieved the best results. The proposed solutions achieved approximately 4\% of standard deviation $ \sigma $ while other architectures had obtained 8\%. In those tests, the best performance in accuracy (CMSNet-M2) was also our worst-case at inference time. 

The study also has calculated the inference time for a batch of four images. The arrangements with output stride 8 (CM0, CM1, and CM2) have achieved an average improvement of 1 FPS, and the architectures with output stride 16 (CM3, CM4, CM5, CM6, CM7, and CM8) were 4 FPS better.

\subsubsection{Inference on different hardware} 
\begin{figure*}[htb]
	\begin{center} 
		\includegraphics[width=\linewidth]{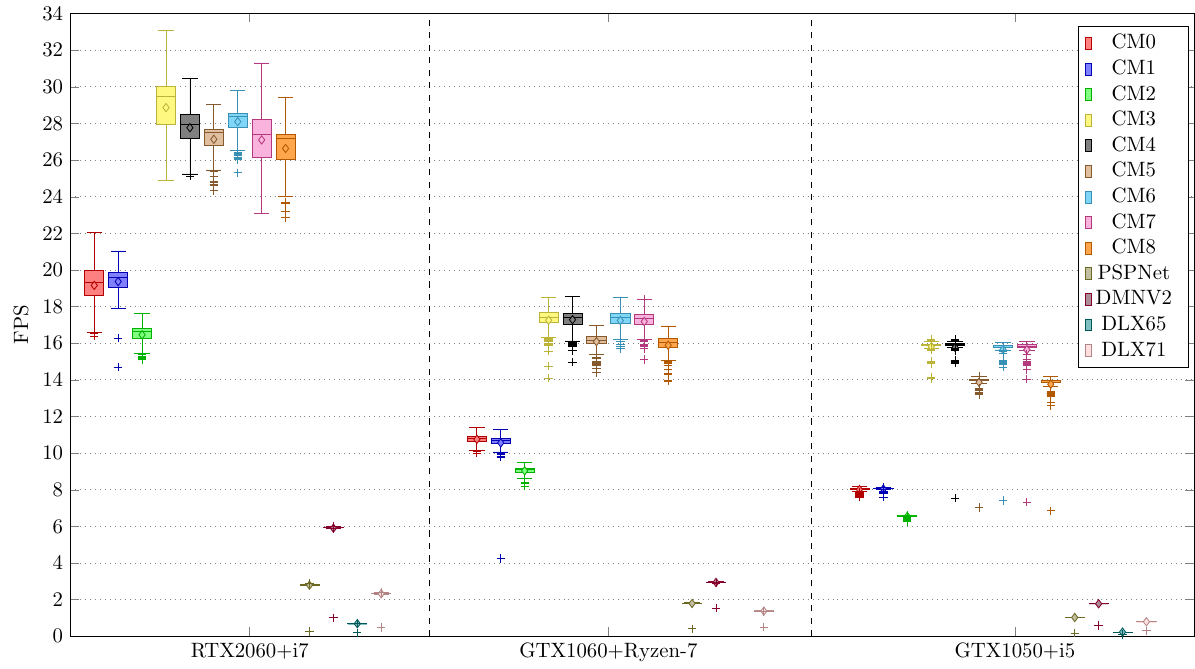}
	\end{center} 
	\caption{Inference time is shown in FPS (box-plot) as a function of the architecture model and hardware platform. The models were tested on three different platforms -- RTX2060+core-i7, GTX1060+Ryzen7, and GTX1050+core-i5.}
	\label{fig:inference_time}
\end{figure*}

The researcher has also made tests in other hardware configurations. \autoref{fig:inference_time} shows the results for the GPUs GTX 1050,  GTX 1060, and RTX 2060. The CMSNet-M3 on the GPU RTX2060 has achieved the best inference time result, and the DeepLab+Xc65 on GPU GTX 1050 produced the worst case. Among the architectures composed with the framework CMSNet, those using output stride (OS) 16 performed better regarding FPS than those using output stride 8. The Global Pyramid Pooling (GPP) module (CM0, CM3, and CM6) achieved the best FPS regarding their output stride and shortcut groups. On the other hand, the Atrous Spatial Pyramid Pooling (ASPP) module (CM2, CM5, and CM8) has achieved the worst FPS results considering those same groups.

From the observations, the researchers see that the inference time performance for those different architectures held the proportion regarding the computation power of each platform. Furthermore, the proposed architectures performed better than the other ones used in the comparison. Regarding all platforms, the CM2 achieved the worst FPS for the CMSNet arrangements, and DLX65 had the worst inference time between all tested architectures.

\subsection{CMSNet on adverse environmental conditions}
\label{subsection:cmsnet-on-adverse-environmental-conditions}
Another contribution of this research was evaluating how visibilities impairments affect the inference quality. To evaluate the system behavior in the adverse conditions of visibility, the researcher has separated the dataset into subsets (\autoref{tab:subset_composition}) and calculated the Jaccard similarity coefficient (IoU) to different proportions of images in an incremental transition between subsets. The study measured the IoU results with conditions ranging from 100\% clear daytime images to 100\% images in poor visibility. The poor conditions included in these analyses were rainy, dusty, nightly, and nightly with dust.

\subsubsection{Dusty condition}
\begin{figure}[htb]
	\captionsetup[subfigure]{font=scriptsize,labelformat=empty}
	\begin{subfigure}[b]{0.497\linewidth}
		\caption{Image}
		\includegraphics[width=\linewidth]{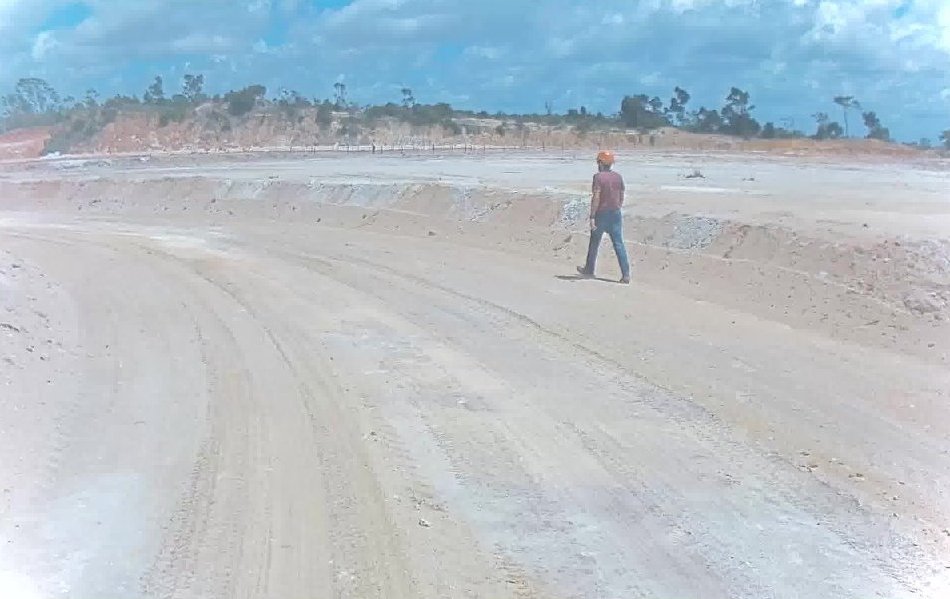}
	\end{subfigure}%
	\hfill
	\begin{subfigure}[b]{0.497\linewidth}
		\caption{Inference}
		\includegraphics[width=\linewidth]{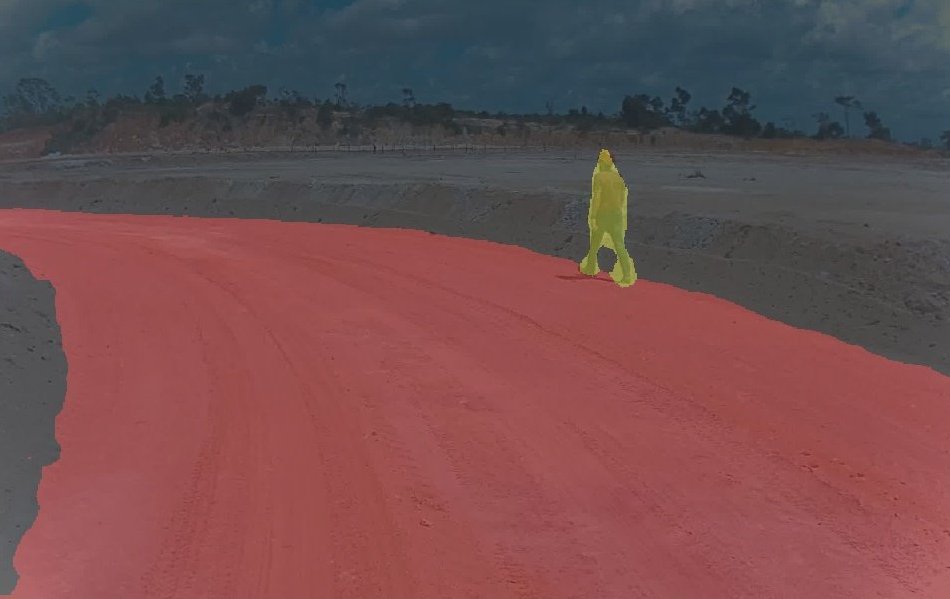}
	\end{subfigure}%
	\vspace{0.003\linewidth}
	\begin{subfigure}[b]{0.497\linewidth}
		\includegraphics[width=\linewidth]{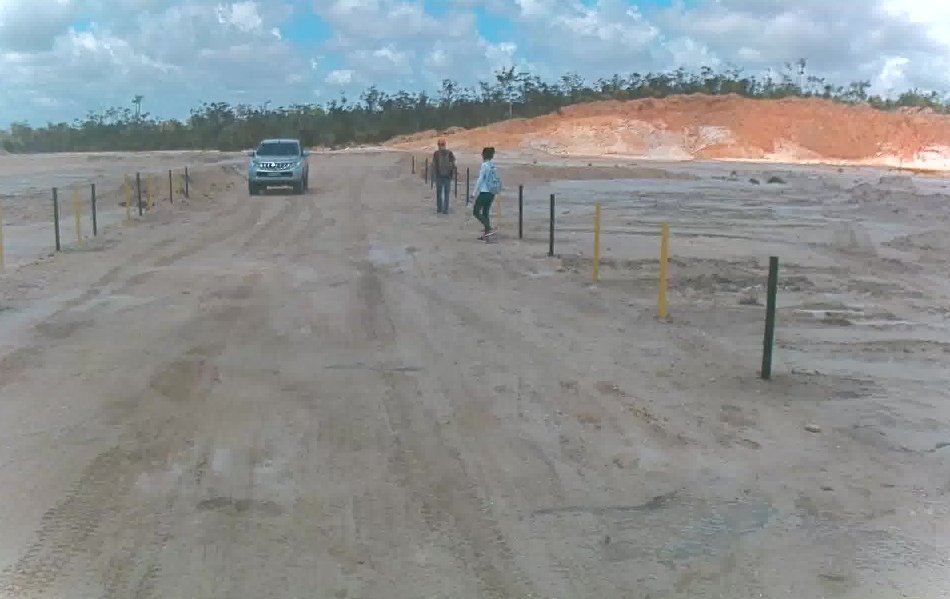}
	\end{subfigure}%
	\hfill
	\begin{subfigure}[b]{0.497\linewidth}
		\includegraphics[width=\linewidth]{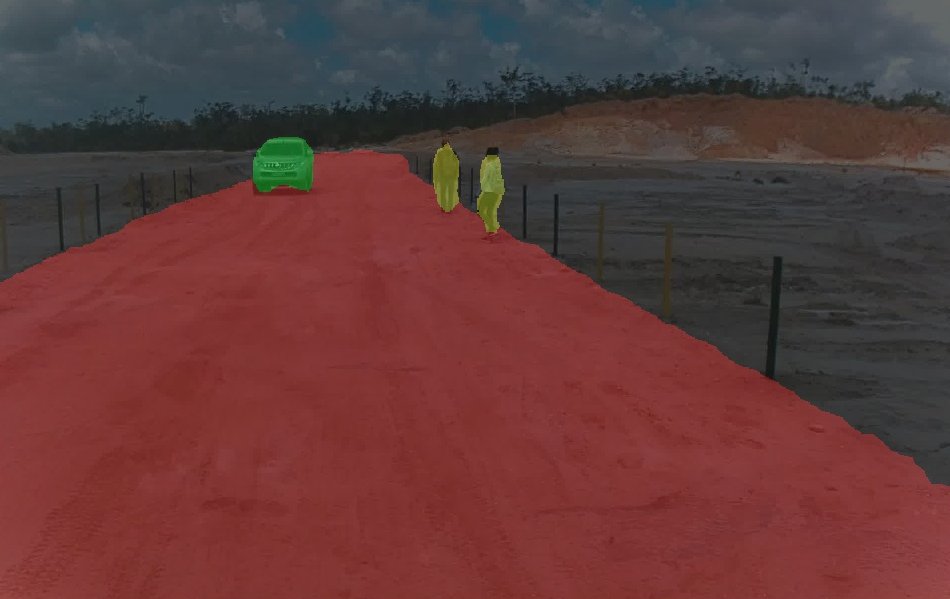}
	\end{subfigure}%
	\caption{Inference in daytime condition on the off-road track.}
	\label{fig:daytime}
	\fautor
\end{figure}

\begin{figure}[htb]
	\captionsetup[subfigure]{font=scriptsize,labelformat=empty}
	\begin{subfigure}[b]{0.497\linewidth}
		\caption{Image}
		\includegraphics[width=\linewidth]{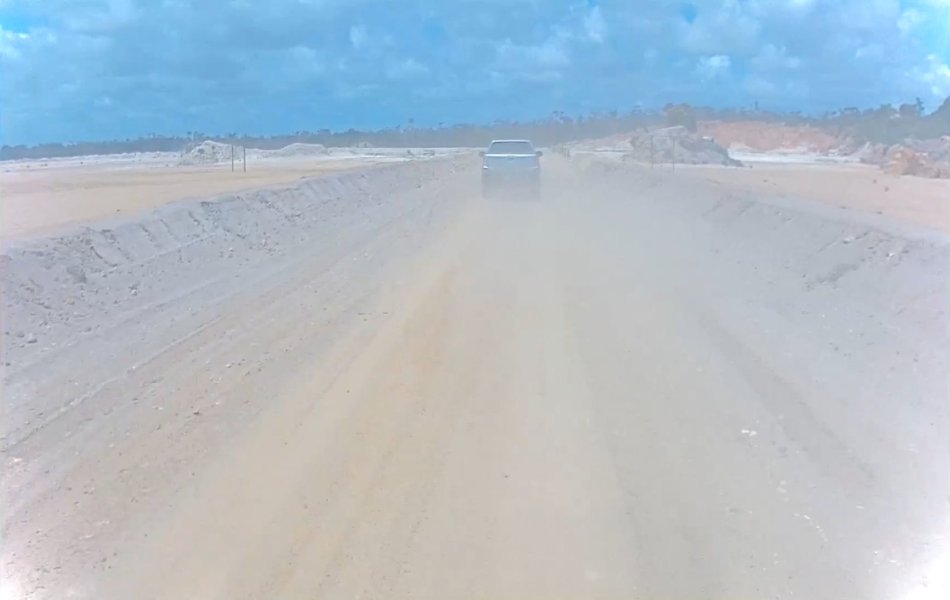}
	\end{subfigure}%
	\hfill
	\begin{subfigure}[b]{0.497\linewidth}
		\caption{Inference}
		\includegraphics[width=\linewidth]{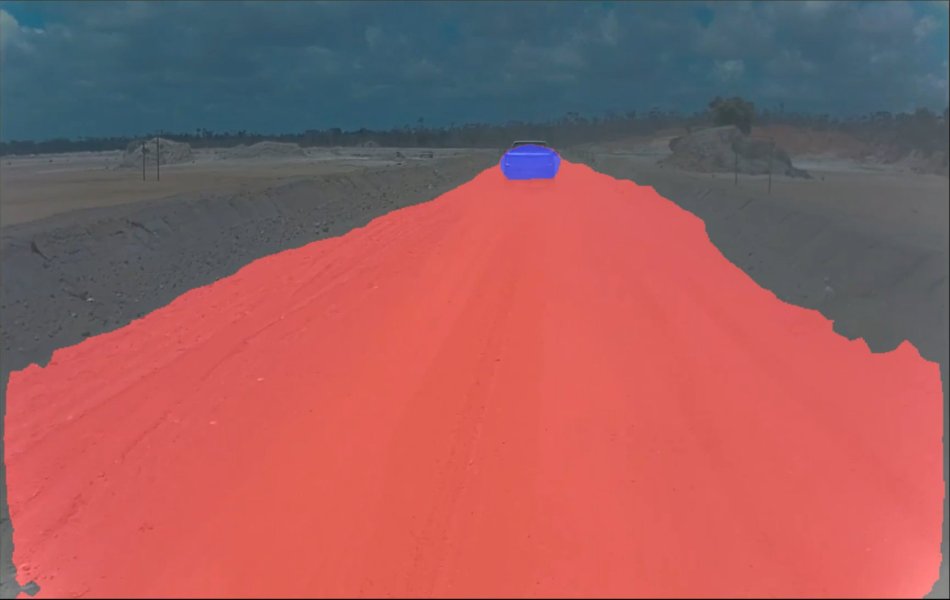}
	\end{subfigure}%
	\vspace{0.003\linewidth}
	\begin{subfigure}[b]{0.497\linewidth}
		\includegraphics[width=\linewidth]{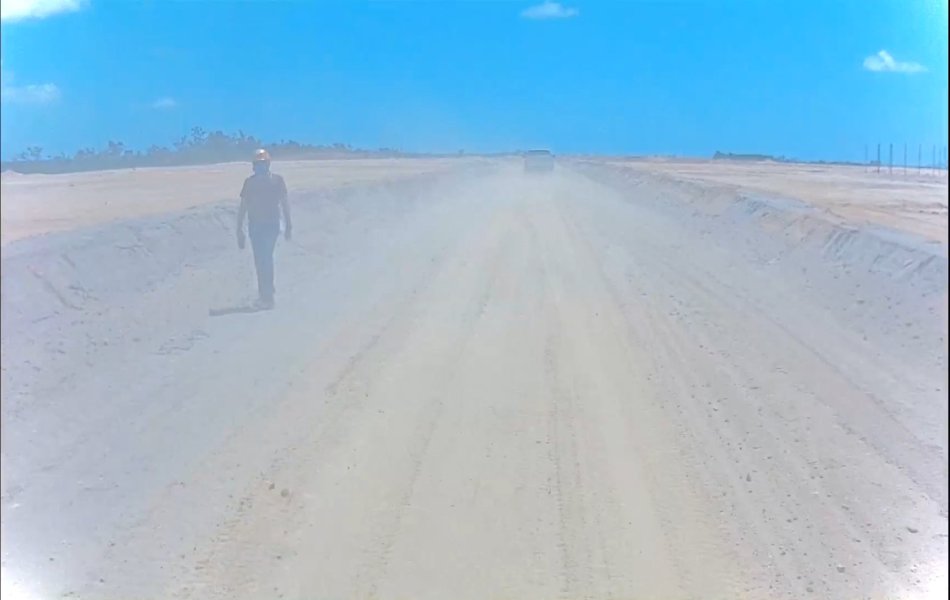}
	\end{subfigure}%
	\hfill
	\begin{subfigure}[b]{0.497\linewidth}
		\includegraphics[width=\linewidth]{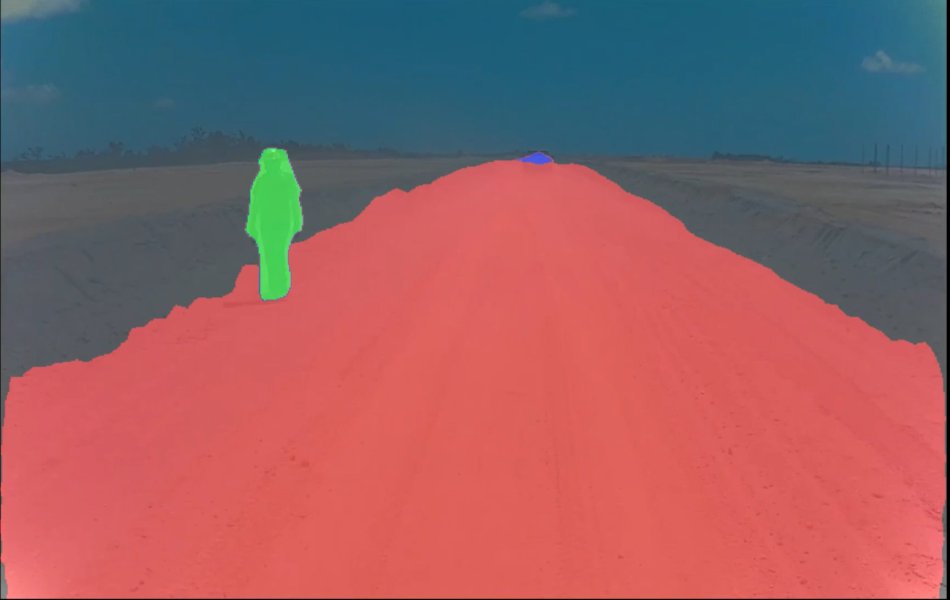}
	\end{subfigure}%
	\caption{Inference in dusty condition on the off-road track.}
	\label{fig:dusty}
	\fautor
\end{figure}

The subsets used in this test had images collected in the off-road test track. The researchers have used a pickup truck passing and crossing in front of the cameras to raise dust on the test track during the recording process. Figures \ref{fig:daytime} and  \ref{fig:dusty} show some pictures and their respective segmentation on the off-road test track, including day and dusty. The study has used daytime images recorded in the same place as a good visibility counterpart subset.

\begin{figure}[htb]
	\begin{center} 
		\includegraphics[width=\linewidth]{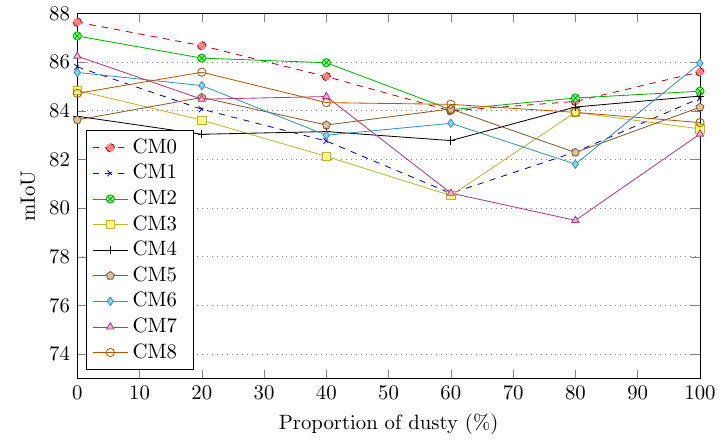}
	\end{center}
	\caption{Day vs. dusty condition evaluation. The axis x (\%) represents the proportion of dusty images in the evaluation, and the axis y ($mIoU$) represents the inference performance archived by each configuration of CMSNet.}
	\label{fig:day_dusty_condition}
\end{figure}

\autoref{fig:day_dusty_condition} shows a downward trend in inference quality (IoU) when good-quality images are replaced by dusty ones. That process has been achieved by changing the daytime and dusty subsets proportion in the test. The scenario went from 0\% of dusty images until 100\% of them, and reversely, from 100\% of daytime images to 0\%. As can be seen, the configurations with output stride 8 (CM0, CM1, and CM2) were less affected by increasing dusty condition images. The best result was 87.54\% of $mIoU$ in daytime circumstances, whereas it was 85.60\% in dusty conditions (all from the dusty subset).

\subsubsection{Nightly condition}
\begin{figure}[htb]
	\captionsetup[subfigure]{font=scriptsize,labelformat=empty}
	\begin{subfigure}[b]{0.497\linewidth}
		\caption{Image}
		\includegraphics[width=\linewidth]{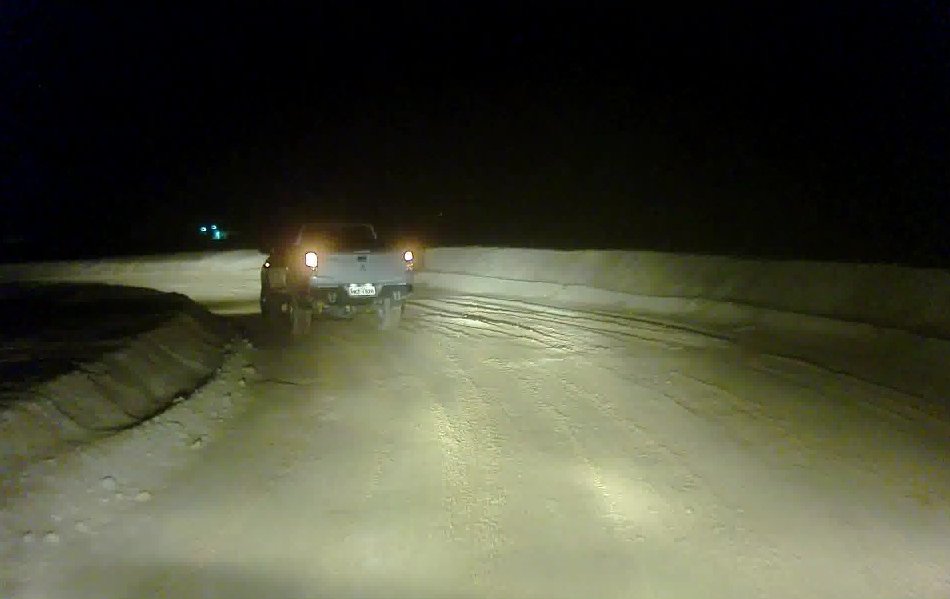}
	\end{subfigure}%
	\hfill
	\begin{subfigure}[b]{0.497\linewidth}
		\caption{Inference}
		\includegraphics[width=\linewidth]{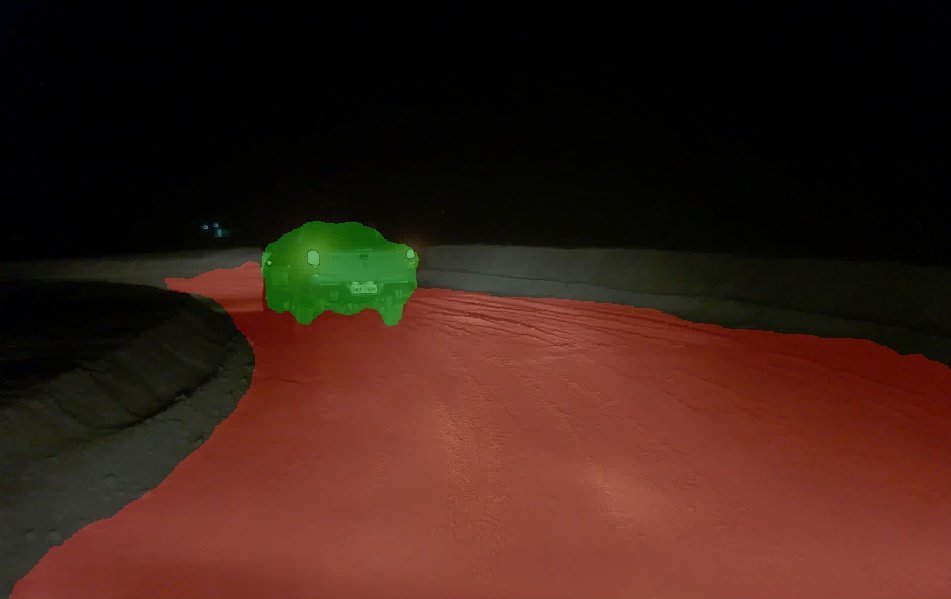}
	\end{subfigure}%
	\vspace{0.003\linewidth}
	\begin{subfigure}[b]{0.497\linewidth}
		\includegraphics[width=\linewidth]{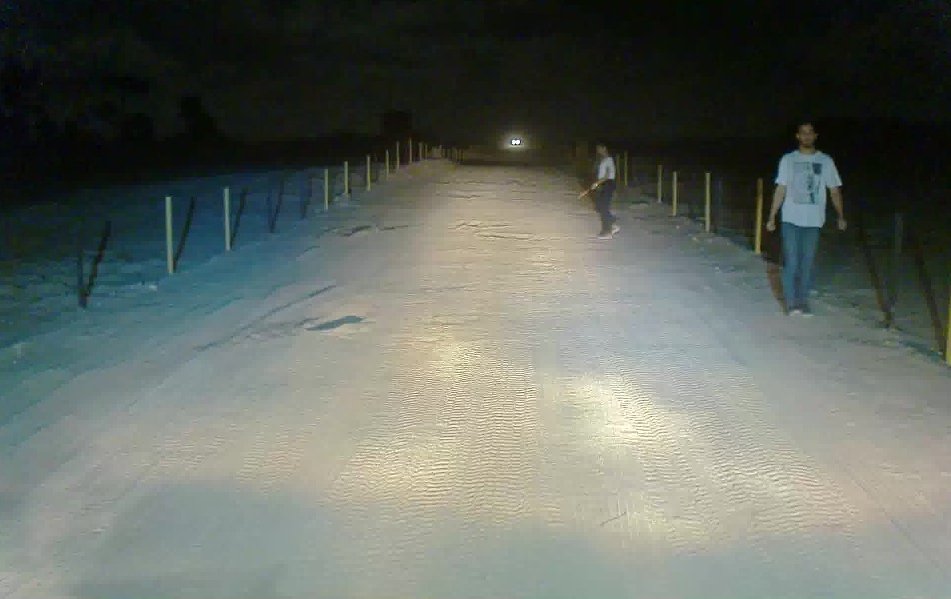}
	\end{subfigure}%
	\hfill
	\begin{subfigure}[b]{0.497\linewidth}
		\includegraphics[width=\linewidth]{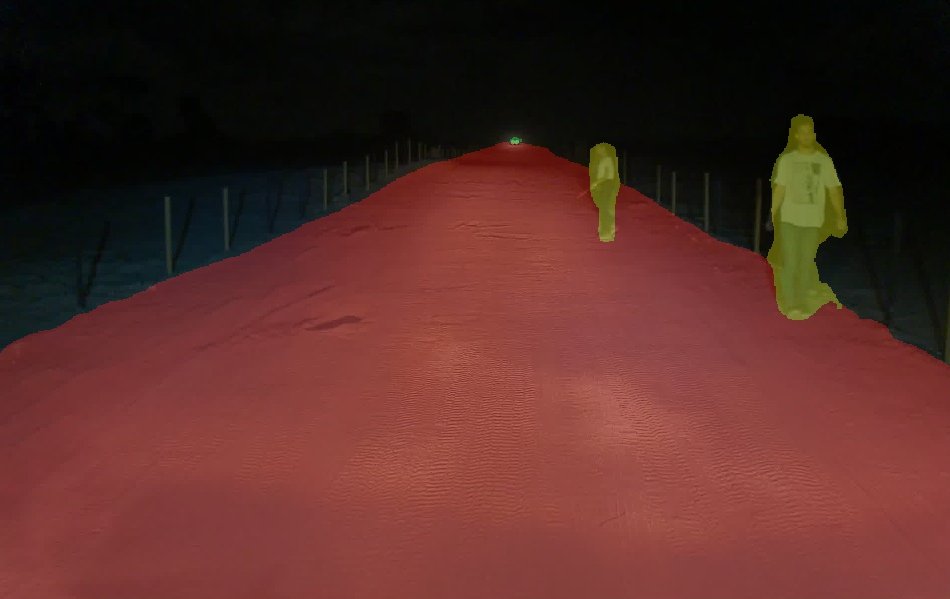}
	\end{subfigure}%
	\caption{Inference in nightly condition on the off-road track.}
	\label{fig:nightly}
	\fautor
\end{figure}

\begin{figure}[htb]
	\begin{center} 
		\includegraphics[width=\linewidth]{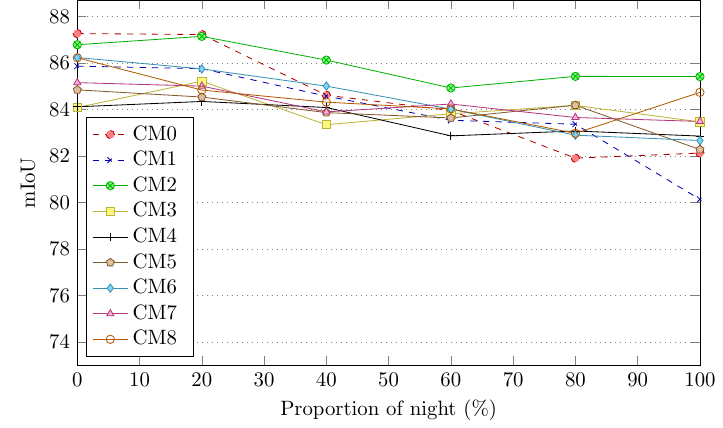}
	\end{center}
	\caption{Day vs. nightly condition evaluation. The axis x (\%) represents the proportion of nightly images in the evaluation, and the axis y ($mIoU$) represents the inference performance archived by each configuration of CMSNet.}
	\label{fig:day_night_condition}
	\fautor
\end{figure}

The subsets used in this evaluation also had the images collected in the off-road test track. However, the researchers have replaced dusty with nighty in the bad visibility images. \autoref{fig:nightly} shows images and their segmentation for this situation. \autoref{fig:day_night_condition} shows the graphic with the test results. The CM0 and CM1 (OS8 with GPP and SPP) were the most affected architectures by nightly conditions. CM0 has decreased performance by over 5 percentage points (pp), and CM1 has lost almost 6 pp. On the other hand, CM3 and CM4 (OS16 with GPP and SPP) have their performance decreased by just 1 pp. The best result in the night condition has been 85.42\% of $mIoU$.

\subsubsection{Nightly with dust}
\begin{figure}[htb]
	\captionsetup[subfigure]{font=scriptsize,labelformat=empty}
	\begin{subfigure}[b]{0.497\linewidth}
		\caption{Image}
		\includegraphics[width=\linewidth]{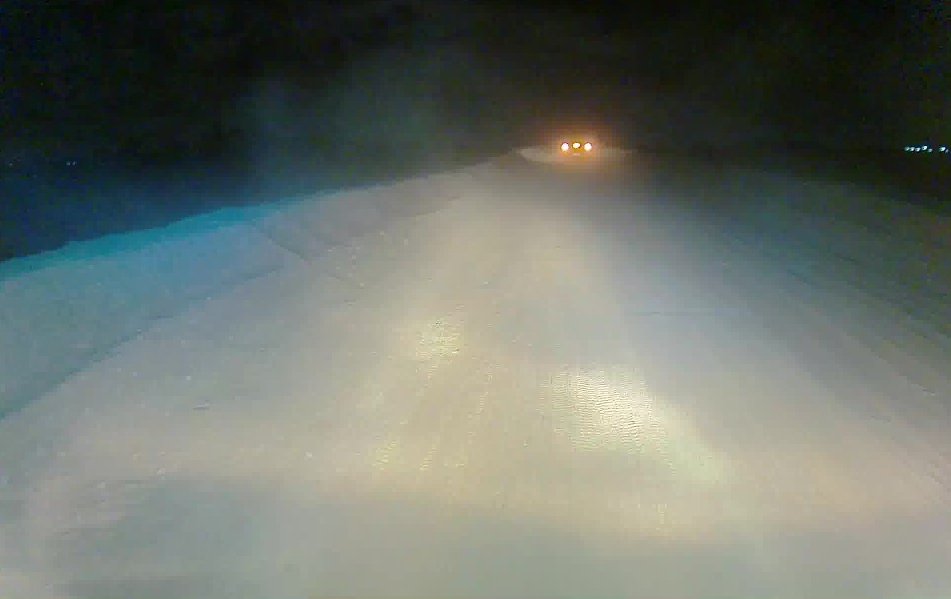}
	\end{subfigure}%
	\hfill
	\begin{subfigure}[b]{0.497\linewidth}
		\caption{Inference}
		\includegraphics[width=\linewidth]{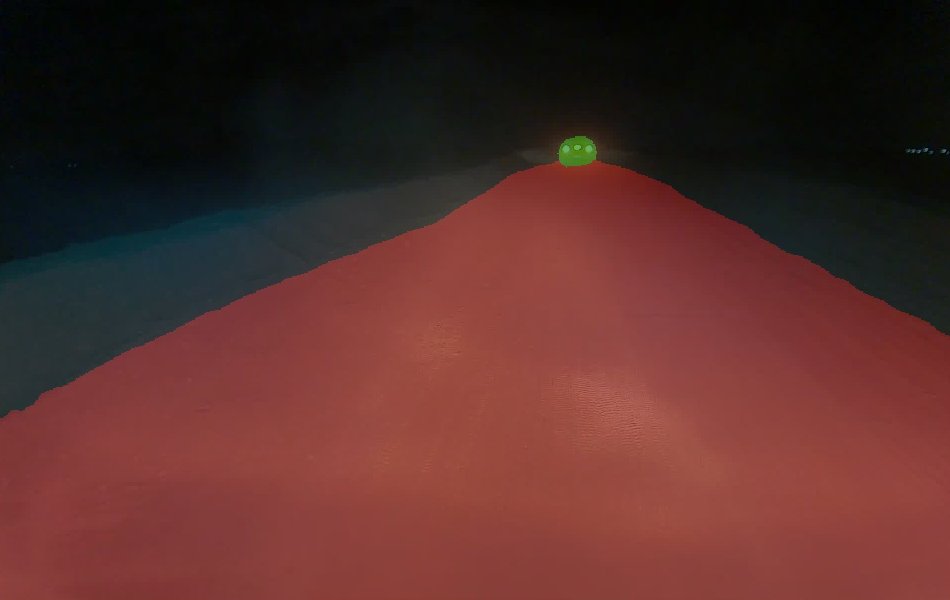}
	\end{subfigure}%
	\vspace{0.003\linewidth}
	\begin{subfigure}[b]{0.497\linewidth}
		\includegraphics[width=\linewidth]{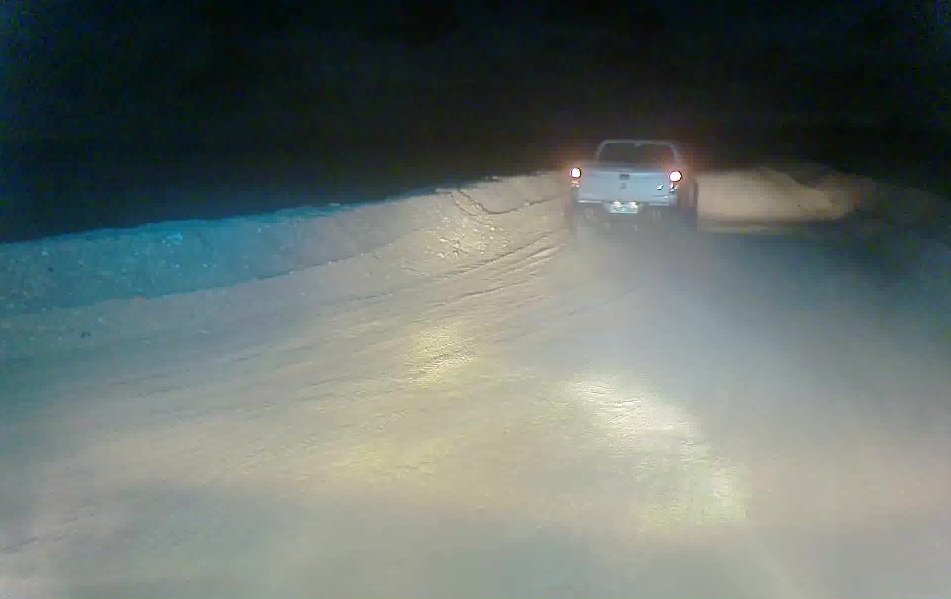}
	\end{subfigure}%
	\hfill
	\begin{subfigure}[b]{0.497\linewidth}
		\includegraphics[width=\linewidth]{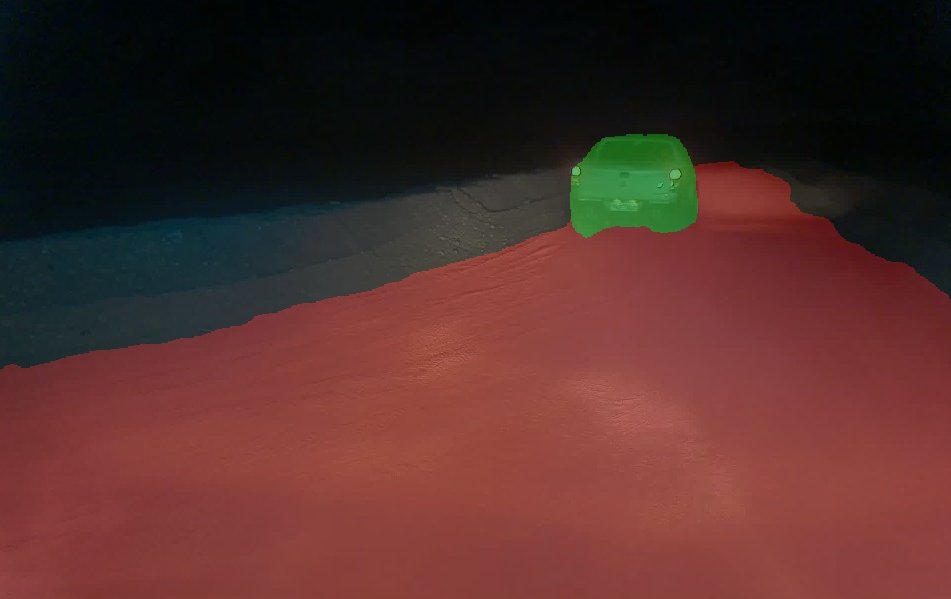}
	\end{subfigure}%
	\caption{Inference in  nightly with dust condition on the off-road track.}
	\label{fig:night_dusty}
	\fautor
\end{figure}

\begin{figure}[htb]
	\begin{center} 
		\includegraphics[width=\linewidth]{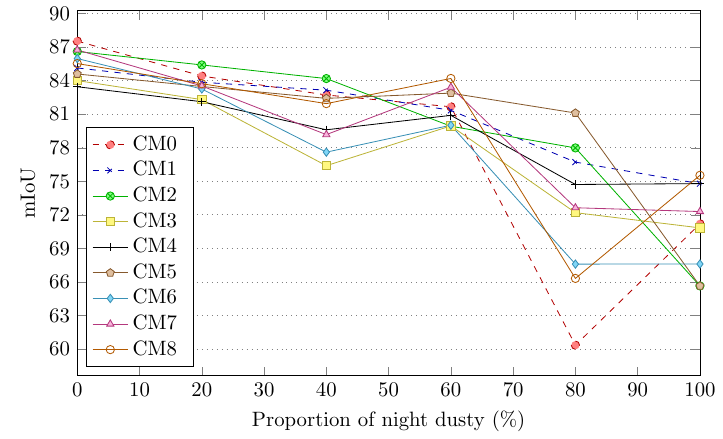}
	\end{center}
	\caption{Day vs. nightly with dust condition evaluation. The axis x (\%) represents the proportion of nightly dusty images in the evaluation, and the axis y ($mIoU$) represents the inference performance archived by each configuration of CMSNet.}
	\label{fig:day_night_dusty_condition}
	\fautor
\end{figure}

This test has mixed good quality images and images collected during the night with dust (Figures \ref{fig:daytime} and \ref{fig:night_dusty}). To generate the dust during the night, the researchers used the same strategy of having a pickup passing crossing in front of the vehicle. The inference quality degradation for that scenario is worse than the previous ones. The CM2 has lost about 21 pp and CM6 near to 19 pp, whereas the CM1 has lost 11 pp, CM4 has degraded about 9 pp and CM8 11,69 pp (\autoref{fig:day_night_dusty_condition}). In this scenario, the best inference results were for the CM1, CM4, and CM8 with about 75\% of $mIoU$.

\subsubsection{Rainy condition}
\begin{figure}[htb]
	\captionsetup[subfigure]{font=scriptsize,labelformat=empty}
	\begin{subfigure}[b]{0.497\linewidth}
		\caption{Daytime}
		\includegraphics[width=\linewidth]{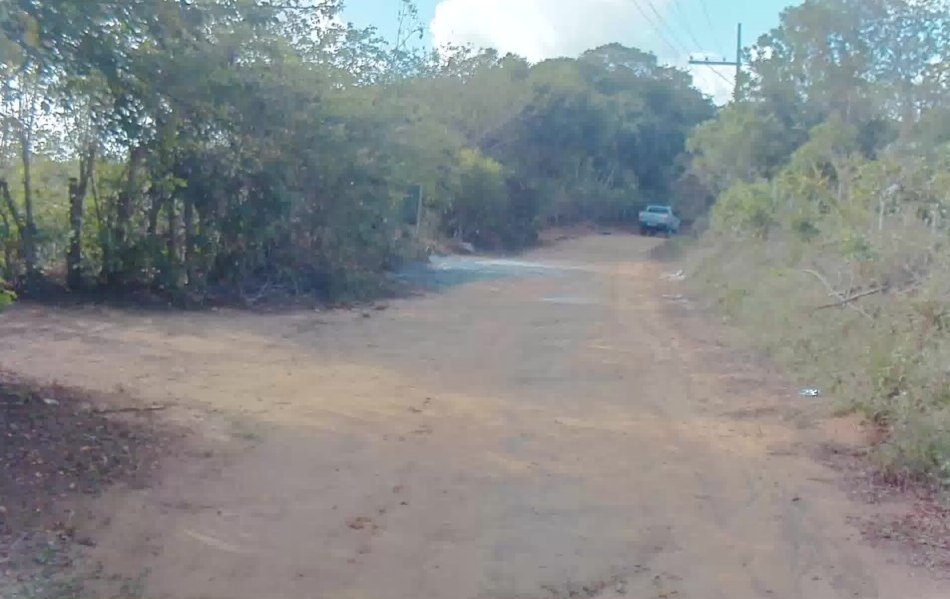}
	\end{subfigure}%
	\hfill
	\begin{subfigure}[b]{0.497\linewidth}
		\includegraphics[width=\linewidth]{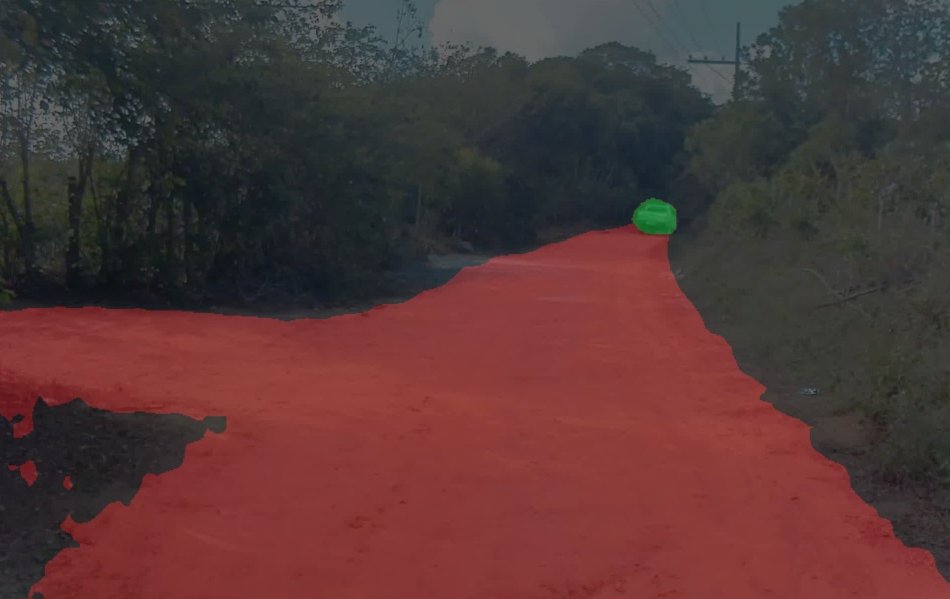}
	\end{subfigure}%
	\vspace{0.003\linewidth}
	\begin{subfigure}[b]{0.497\linewidth}
		\includegraphics[width=\linewidth]{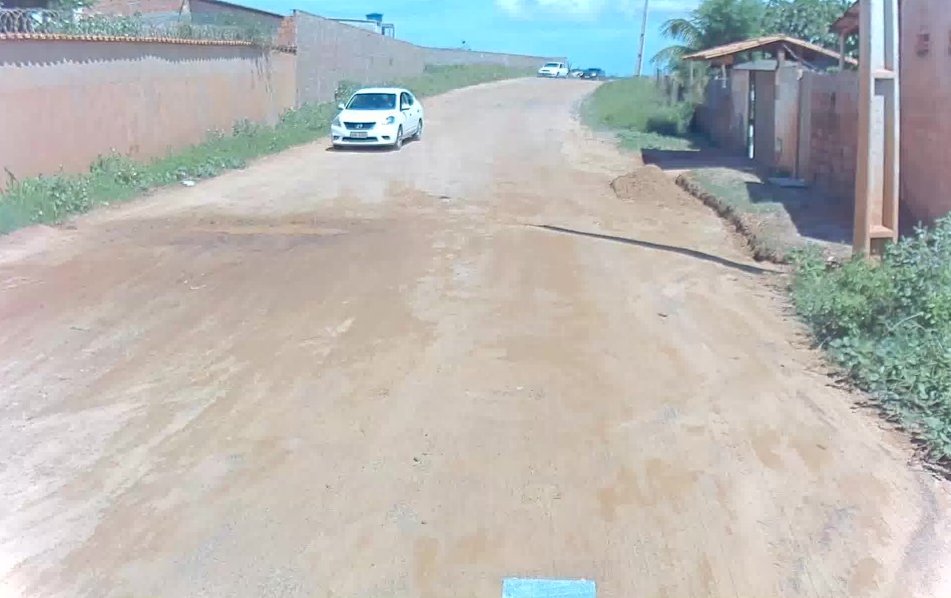}
	\end{subfigure}%
	\hfill
	\begin{subfigure}[b]{0.497\linewidth}
		\includegraphics[width=\linewidth]{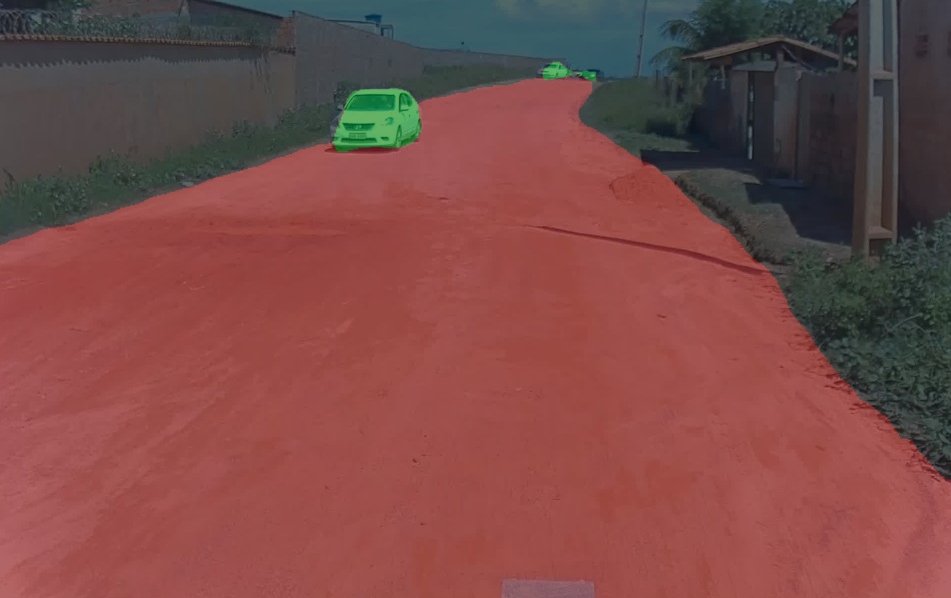}
	\end{subfigure}%
	\caption{Inference in daytime conditions on unpaved roads.}
	\label{fig:daytime2}
	\fautor
\end{figure}

\begin{figure}[htb]
	\captionsetup[subfigure]{font=scriptsize,labelformat=empty}
	\begin{subfigure}[b]{0.497\linewidth}
		\caption{Rainy}
		\includegraphics[width=\linewidth]{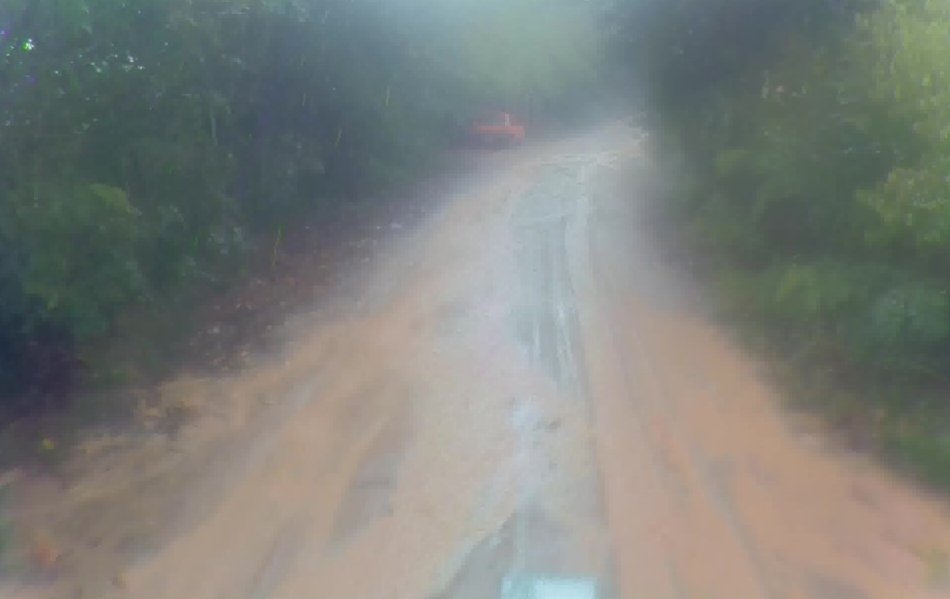}
	\end{subfigure}%
	\hfill
	\begin{subfigure}[b]{0.497\linewidth}
		\includegraphics[width=\linewidth]{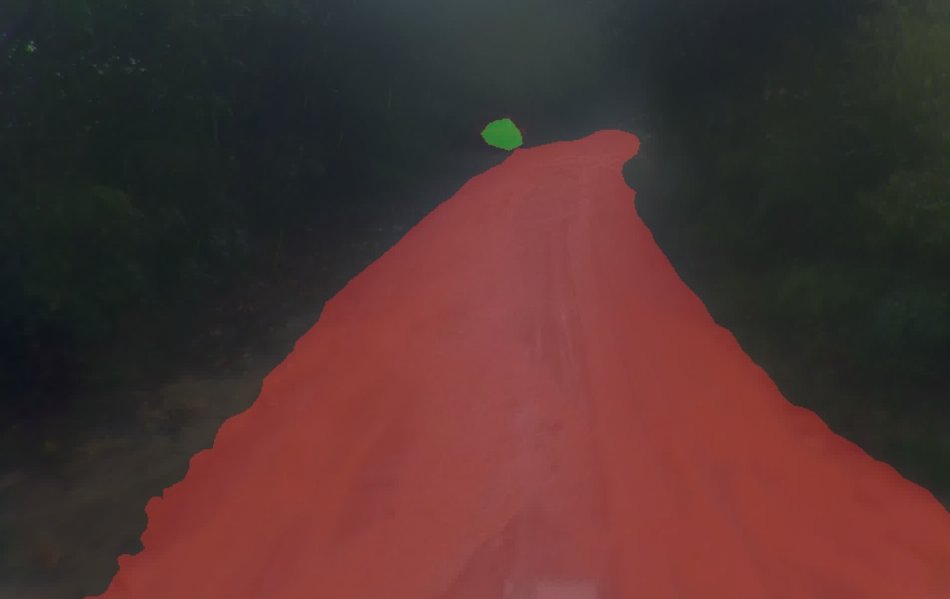}
	\end{subfigure}%
	\vspace{0.003\linewidth}
	\begin{subfigure}[b]{0.497\linewidth}
		\includegraphics[width=\linewidth]{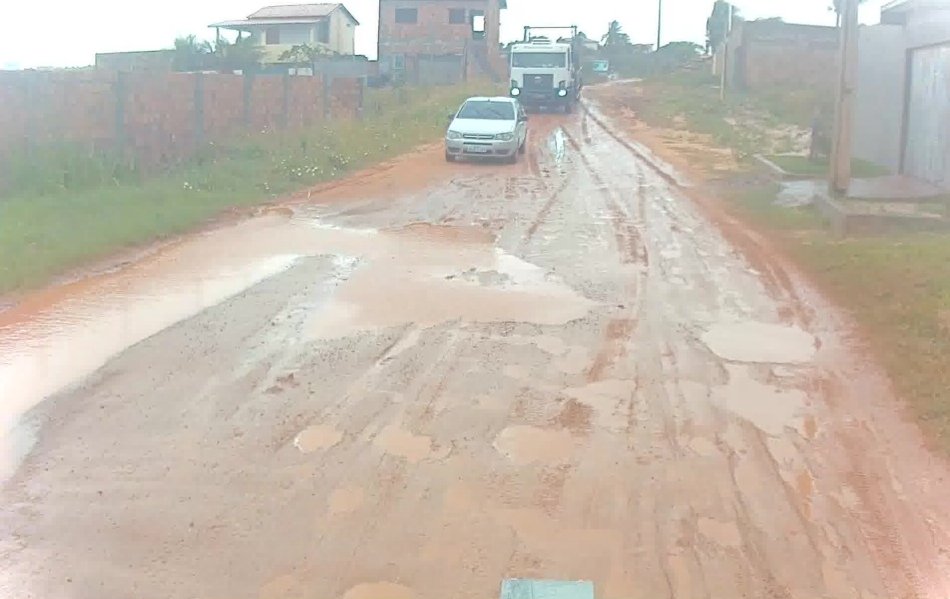}
	\end{subfigure}%
	\hfill
	\begin{subfigure}[b]{0.497\linewidth}
		\includegraphics[width=\linewidth]{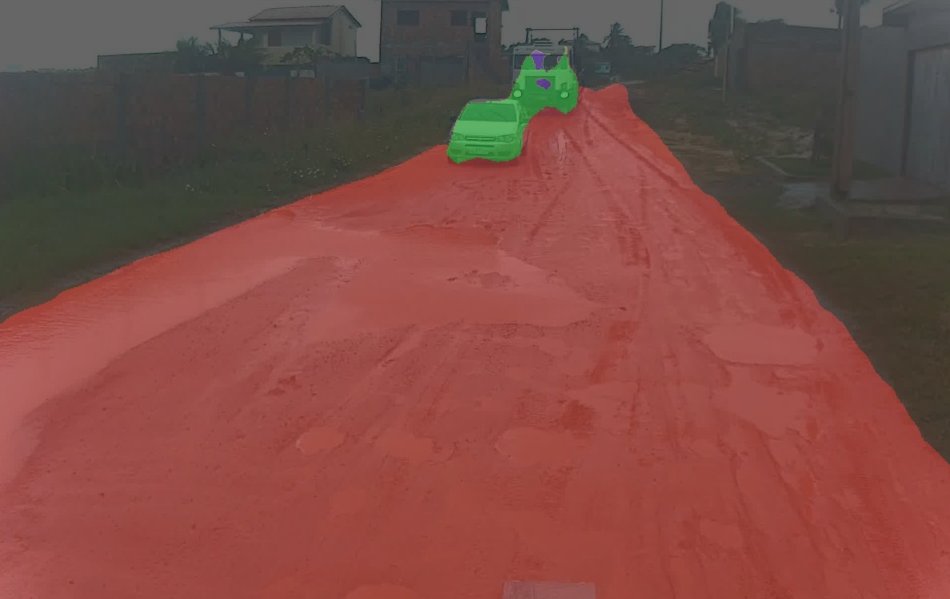}
	\end{subfigure}%
	\caption{Inference in rainy conditions on unpaved roads.}
	\label{fig:rainy}
	\fautor
\end{figure}

The subsets used in rainy tests were different from the previous ones. It has used good quality condition daytime images and bad condition images collected in unpaved roads in the metropolitan region of Salvador-BA (Figures \ref{fig:daytime2} and \ref{fig:rainy}). This strategy has been used to avoid getting the car stuck in the mud on the off-road test track.

\begin{figure}[htb]
	\begin{center} 
		\includegraphics[width=\linewidth]{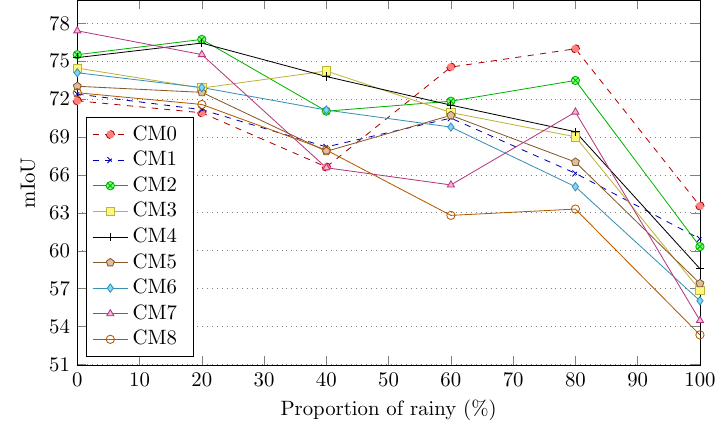}
	\end{center}
	\caption{Day vs. rainy condition evaluation. The axis x (\%) represents the proportion of rainy images in the evaluation, and the axis y ($mIoU$) represents the inference performance archived by each configuration of CMSNet.}
	\label{fig:day_rainy_condition}
	\fautor
\end{figure}

In the rainy condition scenario, inference degradation was even worse. We have the configuration CM7 with quality degradation of about 23 pp and CM0 with $mIoU$ degradation of 8 pp (\autoref{fig:day_rainy_condition}). The best inference result in this scenario, considering 100\% of daytime images, was near 77\%, and with 100\% rainy condition, was 63.55\%.

\subsection{CMSNet on synthetic impairments}
\label{subsection:cmsnet-on-synthetic-impairments}
In addition to the tests with low visibility images recorded in the real world, the researchers also have performed tests in adverse conditions with impairments generated synthetically. It has been created fog and noise. In both situations, the study used the whole dataset shown in the \autoref{tab:subset_composition}.

\subsubsection{Synthetic fog}
\begin{figure}[htb]
	\captionsetup[subfigure]{font=scriptsize,labelformat=empty}
	\begin{subfigure}[b]{0.497\linewidth}
		\caption{Foggy}
		\includegraphics[width=\linewidth]{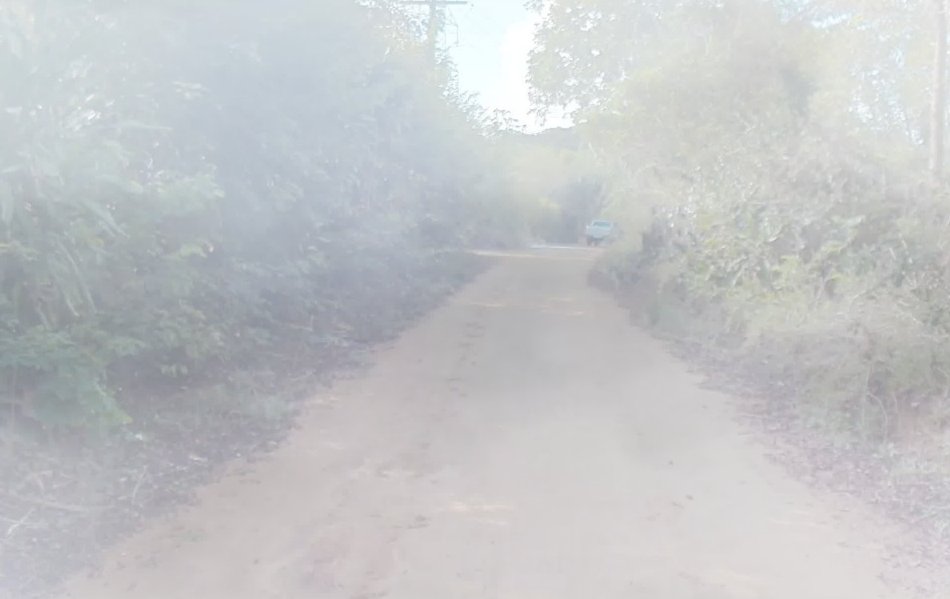}
	\end{subfigure}%
	\hfill
	\begin{subfigure}[b]{0.497\linewidth}
		\includegraphics[width=\linewidth]{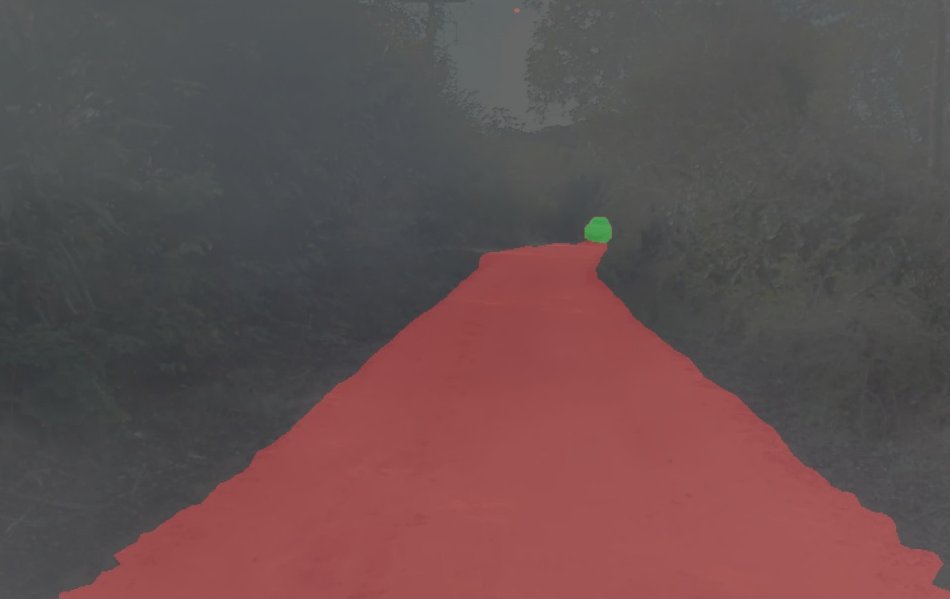}
	\end{subfigure}%
	\vspace{0.003\linewidth}
	\begin{subfigure}[b]{0.497\linewidth}
		\includegraphics[width=\linewidth]{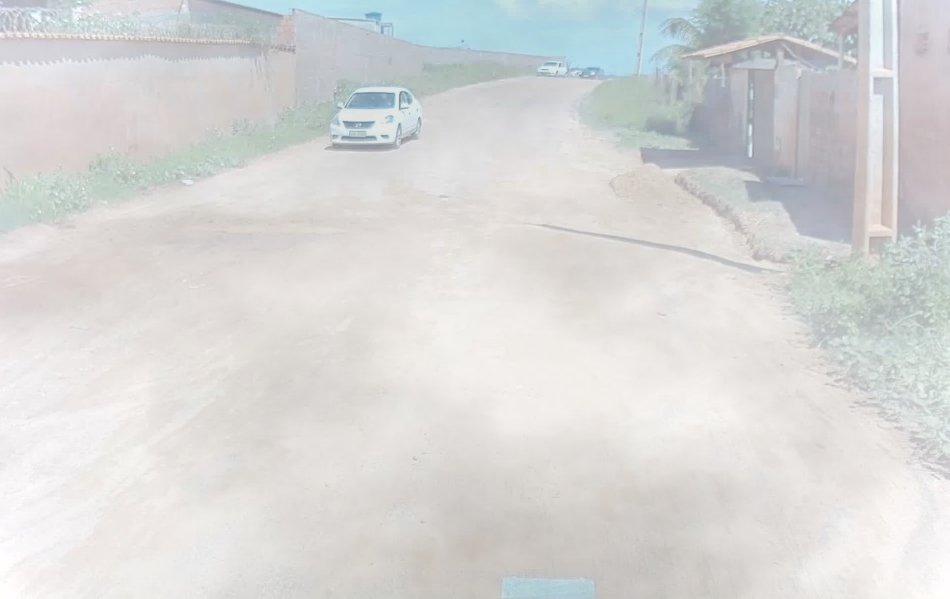}
	\end{subfigure}%
	\hfill
	\begin{subfigure}[b]{0.497\linewidth}
		\includegraphics[width=\linewidth]{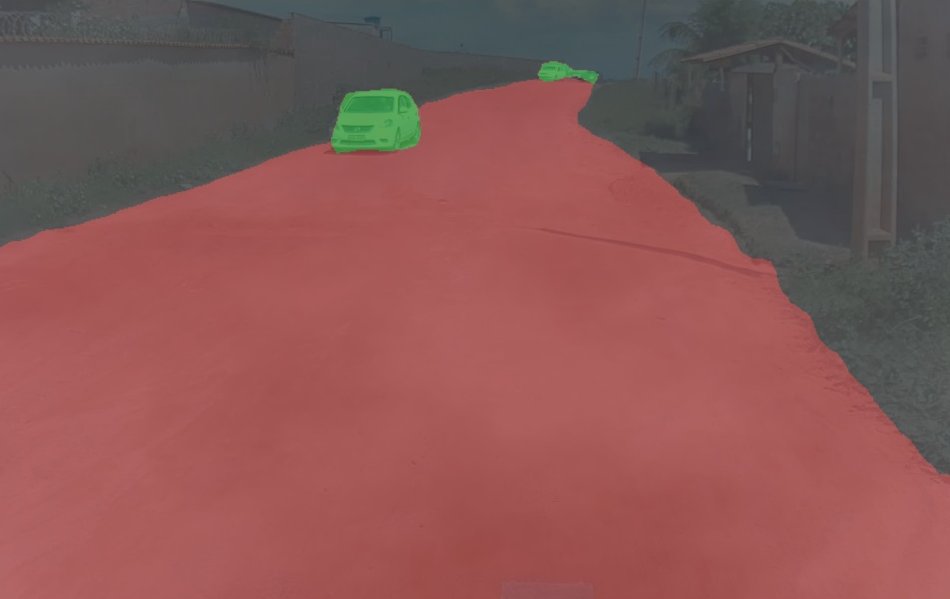}
	\end{subfigure}%
	\caption{Inference in foggy synthetically generated.}
	\label{fig:foggy}
	\fautor
\end{figure}

\begin{figure}[htb]
	\begin{center} 
		\includegraphics[width=\linewidth]{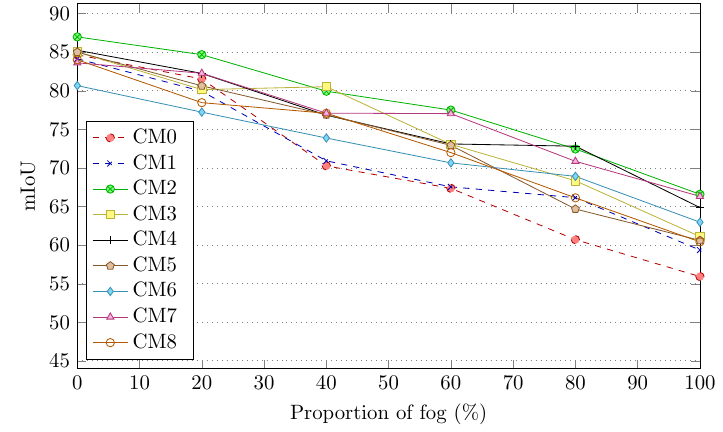}
	\end{center}
	\caption{Synthetic fog over the image. The axis x (\%) represents the proportion of foggy images in the evaluation, and the axis y ($mIoU$) represents the inference performance archived by each configuration of CMSNet.}
	\label{fig:fog_impairment}
	\fautor
\end{figure}

The researchers have used a strategy similar to the previous tests for the foggy condition. The evaluation started without fog in the images, so the proportion was changing by inserting it synthetically (going from 0\% until 100\%). This teste (\autoref{fig:foggy}) shows that the degradation of inference quality behaves like near a linear function. As it can be seen in \autoref{fig:fog_impairment}, the $mIoU$ reduced by about 29 pp for CM0 architecture (worst case) and decreased by 18 pp for CM6 (best situation). In this test, the best inference result was 86.98\% (CM2) of $mIoU$ for daytime while considering fog at 100\% the $mIoU$ was 66.59\% (CM2).  

\subsubsection{Synthetic noise}
\begin{figure}[htb]
	\captionsetup[subfigure]{font=scriptsize,labelformat=empty}
	\begin{subfigure}[b]{0.497\linewidth}
		\caption{Noise}
		\includegraphics[width=\linewidth]{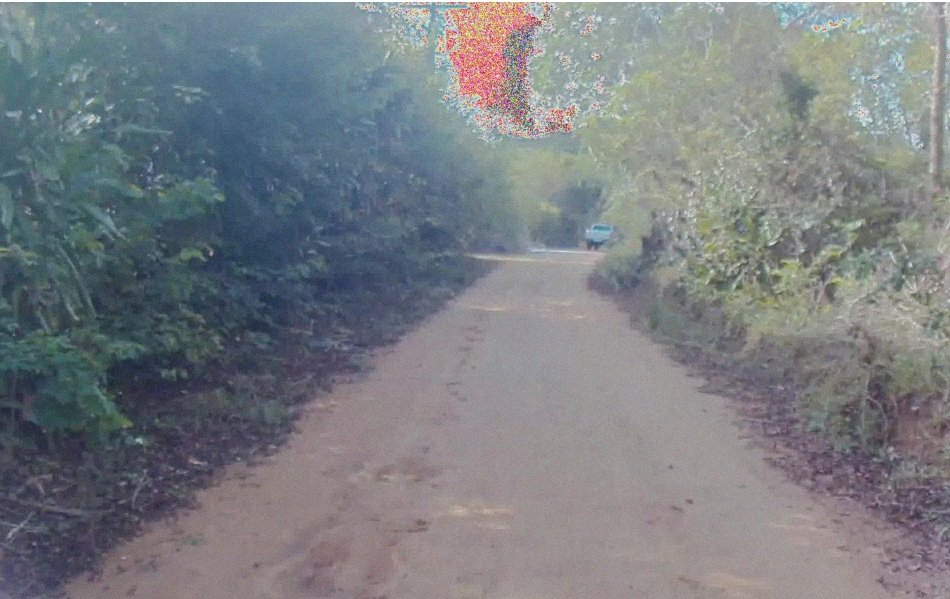}
	\end{subfigure}%
	\hfill
	\begin{subfigure}[b]{0.497\linewidth}
		\includegraphics[width=\linewidth]{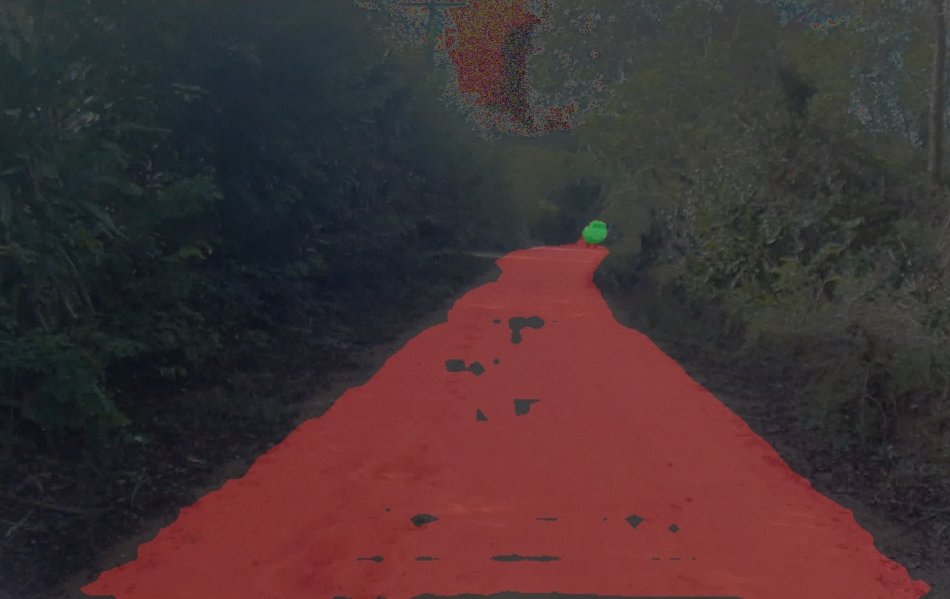}
	\end{subfigure}%
	\vspace{0.003\linewidth}
	\begin{subfigure}[b]{0.497\linewidth}
		\includegraphics[width=\linewidth]{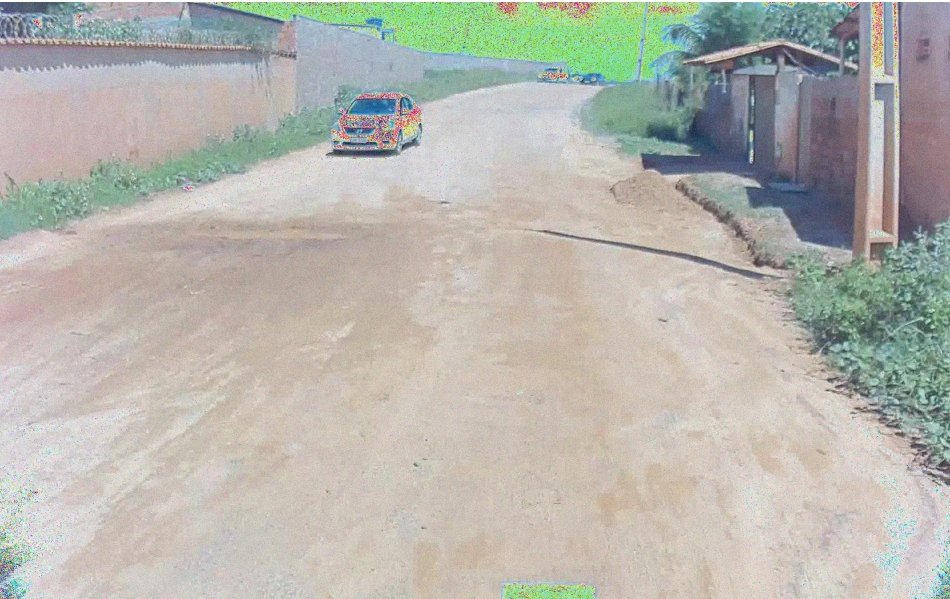}
	\end{subfigure}%
	\hfill
	\begin{subfigure}[b]{0.497\linewidth}
		\includegraphics[width=\linewidth]{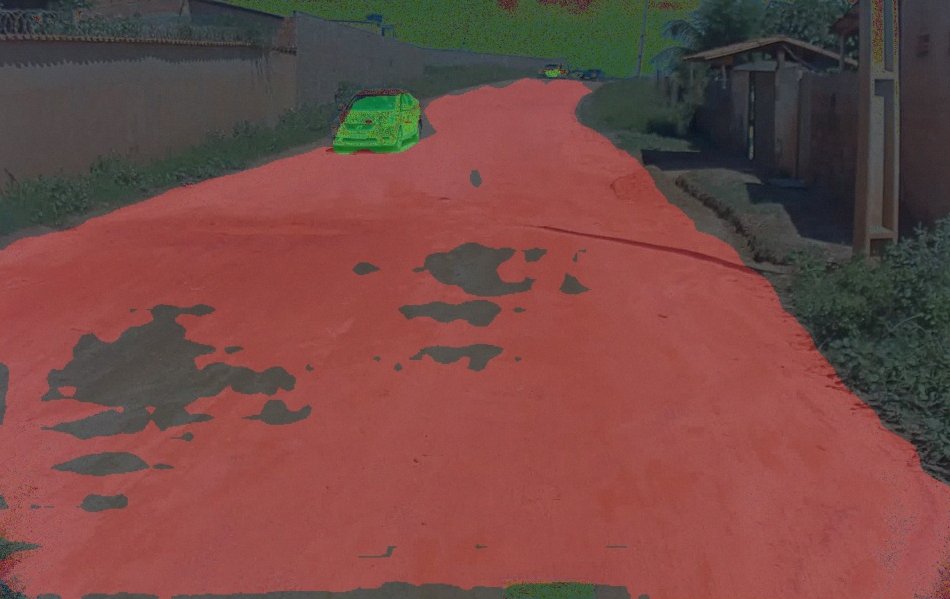}
	\end{subfigure}%
	\caption{Inference with noise synthetically generated.}
	\label{fig:noise}
	\fautor
\end{figure}

\begin{figure}[htb]
	\begin{center} 
		\includegraphics[width=\linewidth]{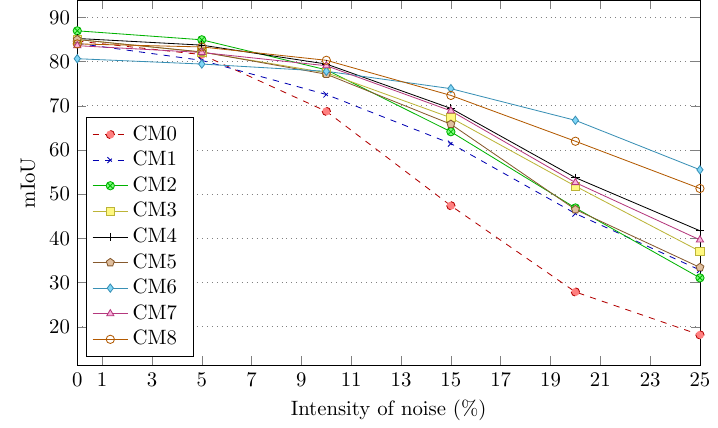}
	\end{center}
	\caption{Additive Gaussian noise over the image. The axis x (\%) represents the intensity proportion of synthetic noise over all the images in the evaluation, and the axis y ($mIoU$) represents the inference performance archived by each configuration of CMSNet.}
	\label{fig:noise_impairment}
	\fautor
\end{figure}

To compose the dataset with artificial noise, the researchers have used a different strategy from the previous tests. Instead of gradually replacing images without impairments with ones having the condition, they have increased the severity of the noise over image signal for all samples simultaneously. The inference started with 0\% of noise and scaled until 25\% of noise. Figures \ref{fig:noise} and \ref{fig:noise_impairment} show the result. The CM0 arrangement produced the worst degradation with mIoU 67 pp small, and the CM6 experienced the less intense degradation with a mIoU decrease of 25.14 pp. The CM6 also achieves the best inference result (55.53\% of mIoU) considering the most intense noise over the image signal.

\subsection{Comparing the adverse conditions results}
\label{subsection:comparing-the-adverse-conditions-results}
\autoref{table_conditions} shows the mIoU achieved by different configurations of CMSNet in diverse conditions of visibility, and \autoref{tab:degradation} shows the level of mIoU degradation achieved by each architecture on each scenario. Regarding the tests carried out on the off-road test track, the situation with night and dust has had the worst mIoU degradation in comparison with daytime images (\autoref{tab:degradation}), and the worst absolute mIoU for all architectures (\autoref{table_conditions}). On the other hand, the day dusty condition has had the best results related to inference quality degradation and absolute mIoU. Concerning the tests carried out with synthetic impairments, the fog has been less harmful than the noise. We also have noticed that rain has been more damaging to the inference quality than dust and night.

\begin{table*}[htb]
	\centering
	\scriptsize
	\addtolength{\leftskip} {-2cm}
	\addtolength{\rightskip}{-2cm}
	\caption{Comparison of mIoU for the evaluated methods on the different environmental conditions of our Kamino dataset during at daytime and nigth according to \autoref{tab:subset_composition}. ``All'' column is the averaged mIoU to a fully balanced set from the all the other subsets.}
	\label{table_conditions}
	\begin{tabular*}{\textwidth}{@{\extracolsep{\fill}} l c  c  c  c  c  c c c  c}
		\specialrule{.1em}{.05em}{.05em}
		\multirow{3}*{\bfseries Method} & \multicolumn{9}{c}{\bfseries mIoU (\%)}    \\
		\cline{2-10}
		& \multicolumn{4}{c}{\bfseries Off-road} & \multicolumn{2}{c}{\bfseries Unpaved}  & \multicolumn{3}{c}{\bfseries Synthetic (All images)}  \\
		\cline{2-5} \cline{6-7} \cline{8-10}
		& \bfseries Day & \bfseries Dusty & \bfseries Nightly & \bfseries Nightly\&Dust & \bfseries Day & \bfseries Rainy & \bfseries Day & \bfseries Foggy & \bfseries Noise \\
		\specialrule{.1em}{.05em}{.05em}
		CM0  & \bfseries 87.65 & \bfseries 85.60 & 82.13 & 71.20 & 71.87 & \bfseries 63.55 & 84.66 & 55.93 & 18.17 \\ 
		CM1  & 85.81 & 84.52 & 80.14 & \bfseries 74.79 & 72.40 & 60.94 & 84.15 & 59.36 & 32.84 \\ 
		CM2  & \bfseries 87.08 & 84.81 & \bfseries 85.42 & 65.65 & \bfseries 75.52 & 60.33 & \bfseries 86.98 & \bfseries 66.59 & 31.07 \\ 
		CM3  & 84.84 & 83.26 & \bfseries 83.47 & 70.84 & 74.46 & 56.88 & \bfseries 85.02 & 61.11 & 37.09 \\ 
		CM4  & 83.78 & 84.61 & 82.86 & \bfseries 74.80 & \bfseries 75.29 & 58.58 & \bfseries 85.25 & 64.87 & 41.73 \\ 
		CM5  & 83.63 & 84.15 & 82.28 & 65.65 & 73.02 & 57.41 & \bfseries 85.01 & 60.62 & 33.44 \\ 
		CM6  & 85.58 & \bfseries 85.96 & 82.67 & 67.61 & 74.09 & 56.04 & 80.67 & 62.97 & \bfseries 55.53 \\ 
		CM7  & 86.25 & 83.05 & \bfseries 83.49 & 72.30 & \bfseries 77.41 & 54.48 & 83.62 & \bfseries 66.32 & 39.67 \\ 
		CM8  & 84.72 & 83.52 & \bfseries 84.74 & \bfseries 75.54 & 72.51 & 53.34 & 84.02 & 60.42 & \bfseries 51.30 \\ 
		
		\specialrule{.1em}{.05em}{.05em}
	\end{tabular*}
\end{table*}

\begin{table}[htb]
	\centering
	\scriptsize
	\addtolength{\leftskip} {-2cm}
	\addtolength{\rightskip}{-2cm}
	\caption{Comparison of mIoU degradation for the evaluated methods on the different environmental conditions of our Kamino dataset during at daytime and nigth according to \autoref{tab:subset_composition}.}
	\label{tab:degradation}
	\begin{tabular*}{\textwidth}{@{\extracolsep{\fill}} l c  c  c  c  c  c}
		\specialrule{.1em}{.05em}{.05em}
		\multirow{2}*{\bfseries Method} & \multicolumn{6}{c}{\bfseries mIoU (\%)}   \\
		\cline{2-7}
		& \bfseries Dust & \bfseries Nightly & \bfseries Nightly\&Dust & \bfseries Rainy & \bfseries Foggy & \bfseries Noise \\
		\specialrule{.1em}{.05em}{.05em}
		CM0	& 1.67 & 5.14 & 16.07 &  \bfseries 8.32 & 28.73 & 66.49 \\
		CM1	& 1.35 & 5.73 & \bfseries 11.08 & \bfseries 11.46 & 24.79 & 51.31 \\
		CM2	& 1.98 & \bfseries 1.37 & 21.14 &  15.19 & \bfseries 20.39 & 55.91 \\
		CM3	& \bfseries 0.83 & \bfseries 0.62 & \bfseries 13.25 &  17.58 & 23.91 & 47.93 \\
		CM4	& \bfseries 0.00 & \bfseries 1.26 & \bfseries 9.32 &  16.71 & \bfseries 20.38 & 43.52 \\
		CM5	& \bfseries 0.70 & 2.57 & 19.20 &  15.61 & 24.39 & 51.57 \\ 
		CM6	& \bfseries 0.28 & 3.57 & 18.63 & 18.05 & \bfseries 17.70 & \bfseries 25.14 \\ 
		CM7	& 2.11 & 1.67 & \bfseries 12.86 & 22.93 & \bfseries 17.30 & 43.95 \\
		CM8	& 2.71 & 1.49 & \bfseries 10.69 & 19.17 & 23.60 & \bfseries 32.72 \\
		\specialrule{.1em}{.05em}{.05em}
	\end{tabular*}
	\fautor
\end{table}

The configurations CM3 and CM4 have had less mIoU degradation on the off-road tests. The architectures CM0 and CM1 have had the best results on raining testes, and CM6 and CM7 have performed better on synthetic impairments. However, CM3 and CM4 use output stride 16 and demand fewer parameters and MAC operations to carry out inference. 

\subsection{CMSNet results on DeepScene dataset}
\label{subsection:results-on-deepscene-dataset}
Besides comparing the configurations of CMSNet with themselves and with urban trained algorithms, the study also compared some architectures generated by the CMSNet with algorithms proposed in related works published in the last years. The solutions presented in  \citeonline{Semantic-Forested:Valada:2016}, and  \citeonline{Real-Time-Semantic-Off-Road:Maturana:2018} have been trained and compared between them in the DeepScene dataset  \cite{Semantic-Forested:Valada:2016}. Although this dataset does not have the magnitude of the one proposed in our work, it was the only possible way to compare the solutions' performance once those works had not published their source code, so this thesis study could not train them with our dataset.

The  \citeonline{Real-Time-Semantic-Off-Road:Maturana:2018} has presented two architectures: the FCN-based  \cite{Long:2015:FCN:ieeecvpr} cnns-fcn with CNN-S backbone for feature extraction, and the dark-fcn with Darknet's backbone. Those architectures were compared using the resolution $227\times227$ and $448\times448$. 

In the other hand,  \citeonline{Semantic-Forested:Valada:2016} has proposed the UpNet built from a VGG backbone  \cite{simonyan:2015:vgg}. The UpNet is an FCN similar architecture. However, there are some modifications in the last layer of VGG and at the number of upsampling steps. This thesis compared that architecture in the resolution $300\times300$. 

Regarding our CMSNet, the researcher has trained CMSNet-M0 with a resolution of $300\times300$ (CM0-300) and with resolution $448\time448$ (CM0-448). Also, the researcher have trained the configuration with GPP (\autoref{fig:global-pooling-module}) for output stride (OS) 16 (CM3-300 and CM3-448).

\autoref{tab:comparacao-avaliacao-metrica} shows the result $IoU$ per class and the $mIoU$. As it can be seen, the variations of architecture composed in CMSNet framework have reached better results than the networks proposed in  \citeonline{Real-Time-Semantic-Off-Road:Maturana:2018} (cnns-fcn-227, dark-fcn-448). The CMNet's variation (CM0-300, CM0-448, CM3-300, and CM3-448) have reaches 78.89\%, 80.94\%, 77.68\%, and 79.37\% of $mIoU$ against 58.51\%, and 60.61\% of  \citeonline{Real-Time-Semantic-Off-Road:Maturana:2018}.


\begin{table}[htb]
	\centering
	\scriptsize
	\addtolength{\leftskip} {-2cm}
	\addtolength{\rightskip}{-2cm}
	\caption{Results of the semantic segmentation on the categories of the DeepScene dataset.}
	\label{tab:comparacao-avaliacao-metrica}
	\begin{tabular*}{\textwidth}{@{\extracolsep{\fill}} l c  c  c  c  c c c c}
		\specialrule{.1em}{.05em}{.05em}
		\multirow{2}*{\bfseries Method} & \multicolumn{5}{c}{\bfseries IoU (\%)} & \multirow{2}*{\bfseries mIoU(\%)} & \multirow{2}*{\bfseries FPS} & \multirow{2}*{\bfseries StdDev} \\
		\cline{2-6}
		 & \bfseries Trail &  \bfseries Grass & \bfseries Vegetation & \bfseries Sky & \bfseries Obstacle  &  &  \\
		\specialrule{.1em}{.05em}{.05em}
		CM0-300      & 84.87 & 86.73 & 89.17 & 90.21 & 43.46 & \textbf{78.89} & \textbf{21.10} & 3.96\% \\
		CM0-448      & 86.70 & 87.72 & 89.78 & 91.06 & 49.42 & \textbf{80.94} &         16.07  & 2.96\% \\
		CM3-300      & 82.47 & 85.58 & 88.45 & 89.40 & 42.49 & \textbf{77.68} & \textbf{23.75} & 5.92\% \\
		CM3-448      & 84.69 & 87.06 & 89.46 & 90.30 & 45.35 & \textbf{79.37} & \textbf{21.33} & 4.66\% \\
		Upnet-300    & 85.03 & 86.78 & 90.90 & 90.39 & 45.31 & \textbf{79.68} &         20.09  & 9.47\% \\
		cnns-fcn-227 & 85.95 & 85.34 & 87.38 & 90.53 &  1.84 &         58.51  &          9.90  & 1.58\% \\
		dark-fcn-448 & 88.80 & 87.41 & 89.46 & 93.35 &  4.61 &         60.61  &         18.99  & 3.47\% \\
		\specialrule{.1em}{.05em}{.05em}
	\end{tabular*}
	\fautor
\end{table}

On the other hand, regarding the UpNet proposed in  \citeonline{Semantic-Forested:Valada:2016}, our results of $mIoU$ were approximately equivalents. CM0-300 and CM0-448 were better, CM3-300 was equal, and CM3-448 was slightly inferior.

We have implemented the evaluation of inference time for these architectures. In the \autoref{tab:comparacao-avaliacao-metrica} are shown the inference time results in a GTX 1060. Except for the CM0-448, all our proposed solutions are faster than the others. 

\subsection{Field Experiments and real-time embedded inference}

Although there has been a growth in the CNN application for vision algorithms, enabling increasingly accurate semantic segmentation, there is still a challenge of equalizing the demand for computational power since visual perception for autonomous vehicles needs to run in real-time. This study has ported the CM0 and CM3 to achieve real-time inference to embed them in a car and perform the field tests. The study has used the Drive PX2 hardware composed of ARM64 CPUs and CUDA cores. To carry out the reimplementation of our network, the researcher has used the framework TensorRT and C++/CUDA to remove, fuze, and customize some layers.  

\autoref{tab:inference_comparison} shows the results achieved in the embedded hardware Drive PX2. The study also has tested the optimized architectures in a GPU GTX 1080TI and has reached a significant increase in FPS compared with our simulation using Tensorflow. The CM0-TRT has achieved about 8 FPS in Drive PX2 and 40 FPS in the GTX 1080TI. It demands more MAC operations. On the other hand, the CM3-TRT has achieved 21 FPS in Drive PX2 and almost 100 FPS for GTX 1080TI.

\begin{table}[htb]
	\centering
	\scriptsize
	\addtolength{\leftskip} {-2cm}
	\addtolength{\rightskip}{-2cm}
	\caption{Inference time for optimized networks.}
	\label{tab:inference_comparison}
	\begin{tabular*}{\textwidth}{@{\extracolsep{\fill}} c l|  c  c}
		\specialrule{.1em}{.05em}{.05em}
		\bfseries Method & \bfseries Arquitecture & \bfseries FPS & \bfseries Std. \\
		\specialrule{.1em}{.05em}{.05em}
		\multirow{2}{*}{CM0-TRT}	& Drive PX2  &  7.92 & 0.06\% \\
		& GTX 1080TI & 40.47 & 1.42\% \\
		\hline
		\multirow{2}{*}{CM3-TRT}	& Drive PX2  & 21.19 & 0.17\%\\
		& GTX 1080TI & 99.09 & 5.74\% \\
		\hline
		CM0	& \multirow{2}{*}{GTX 1080TI}  & 24,37 & 2.87\%\\
		CM3	& 							  & 35,42 & 8.99\% \\
		\specialrule{.1em}{.05em}{.05em}
	\end{tabular*}

	\fautor
\end{table}

The optimized networks (CM0-TRT and CM3-TRT) have been capable of delivering better performance than their standard implementation and simulation on Tensorflow (CM0 and CM3). As can be seen in \autoref{tab:inference_comparison}, regarding the comparison on GTX 1080TI, the optimized version of CM0 almost has doubled the FPS and decreased the standard deviation $ \sigma $ (Std.) by the heaf. For the CM3, the inference speed has been more than double.

We have noted that the standard deviation $ \sigma $ for the embedded ARM64 platform has been smaller than for the x86\_64 hardware (GTX 1080TI). This indicates that, despite not having an FPS as high as the x86, the ARM platform delivers better predictability and stability for the system.

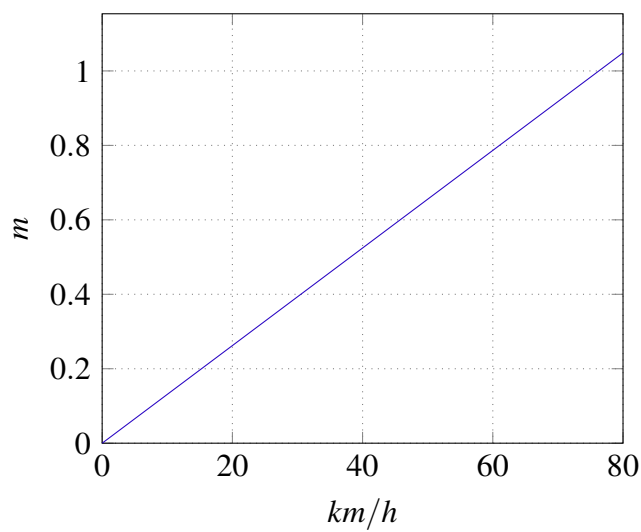
\begin{figure}[htb][!t] 
	\begin{center} 
		\begin{tikzpicture}
		\begin{axis}[xlabel=$km/h$,	ylabel=$m$, grid=major, xmin=0, xmax=80, ymin=0, 
		grid style={dotted,gray}]
		\addplot[color=blue, domain=0:80,samples=14]{
			(x/3.6)*((1/21.19))
		};
		\end{axis}
		\end{tikzpicture}
	\end{center}
	\caption{Distance until response with 21 FPS.} 
	\label{fig:distance-response-21-fps}
	\fautor
\end{figure}

\begin{figure}[htb]
	\captionsetup[subfigure]{font=scriptsize,labelformat=empty}
	\begin{subfigure}[b]{0.325\linewidth}
		\includegraphics[width=\linewidth]{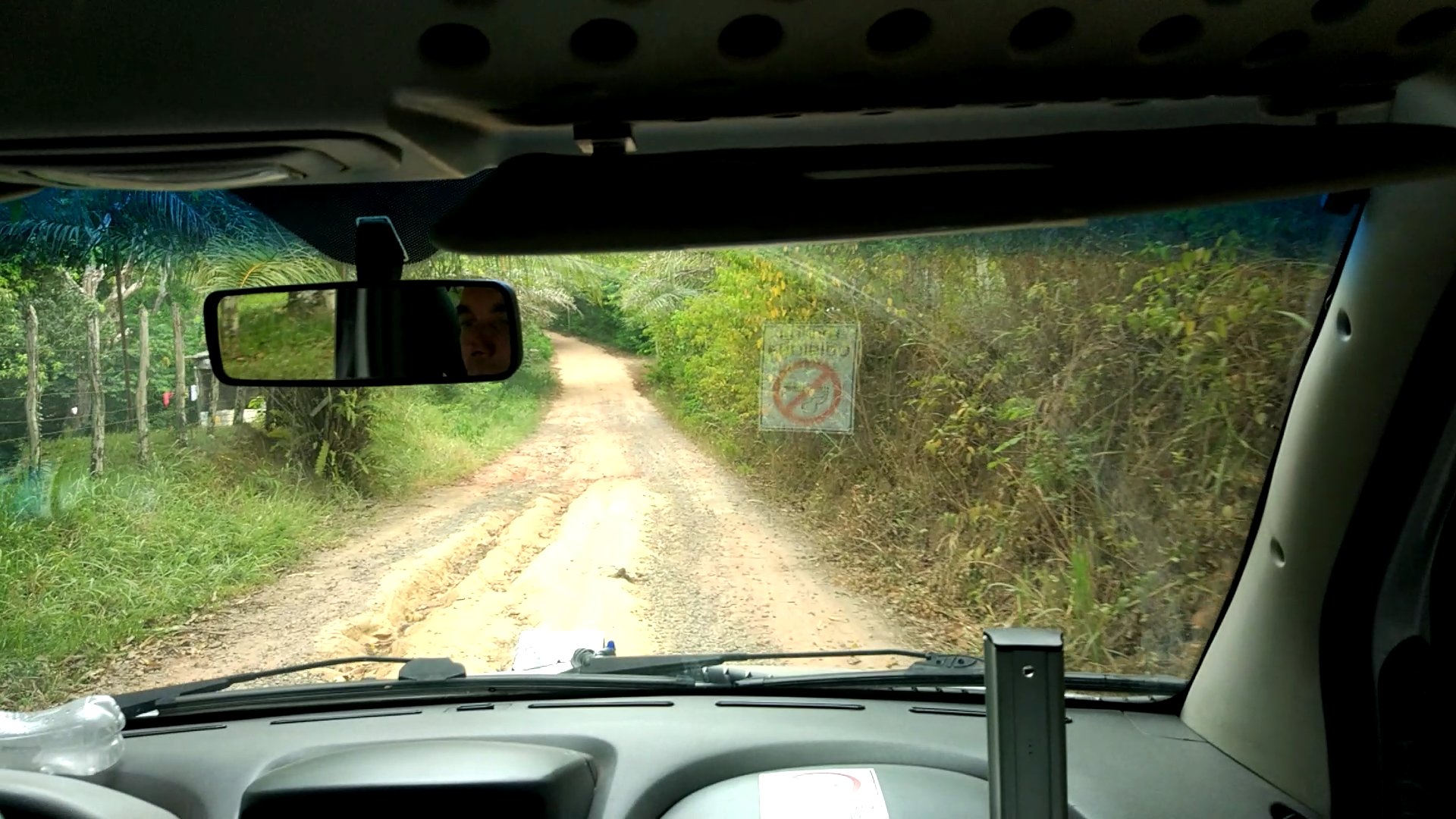}
	\end{subfigure}%
	\hfill
	\begin{subfigure}[b]{0.325\linewidth}
		\includegraphics[width=\linewidth]{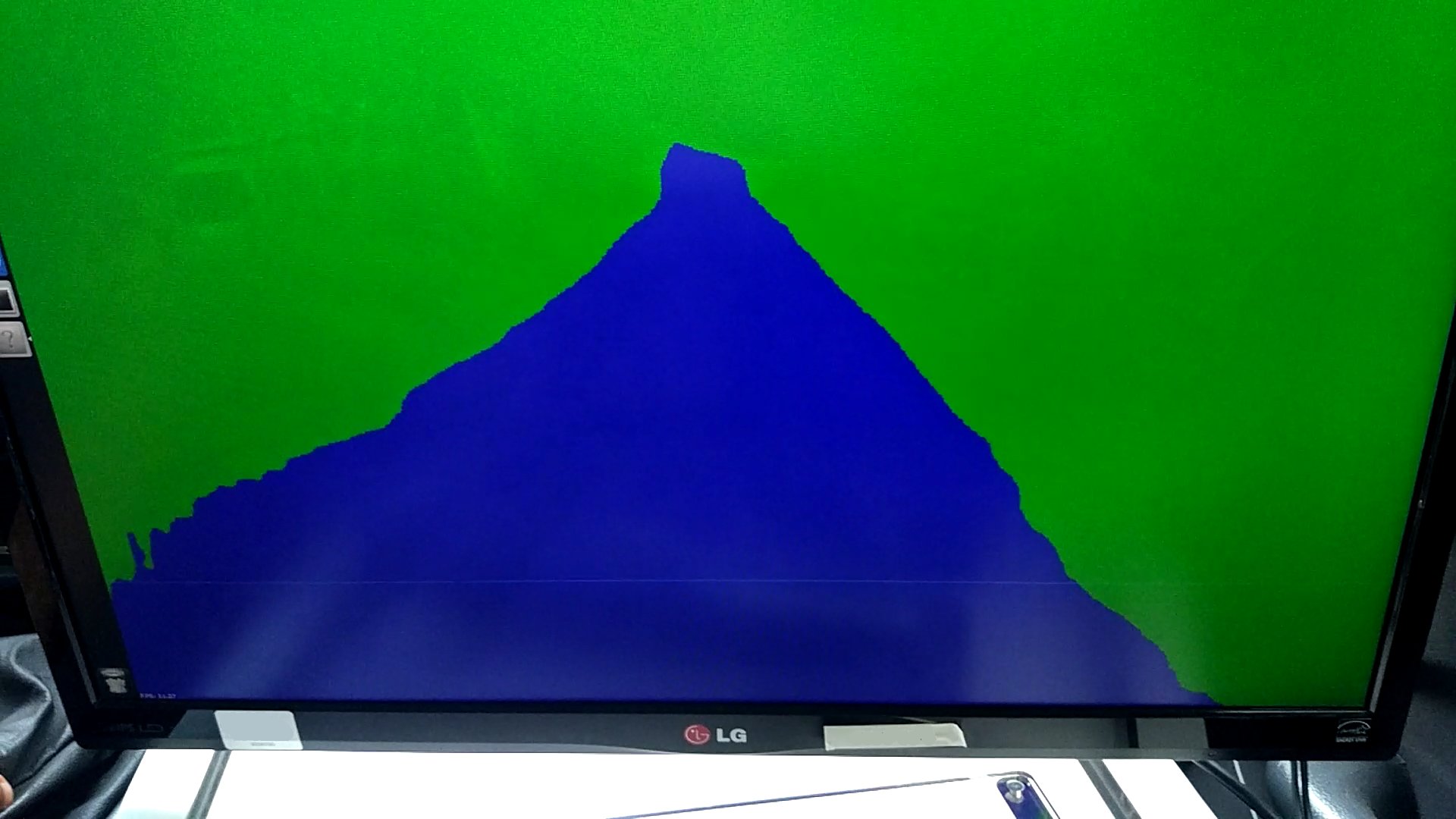}
	\end{subfigure}%
	\hfill
	\begin{subfigure}[b]{0.325\linewidth}
		\includegraphics[width=\linewidth]{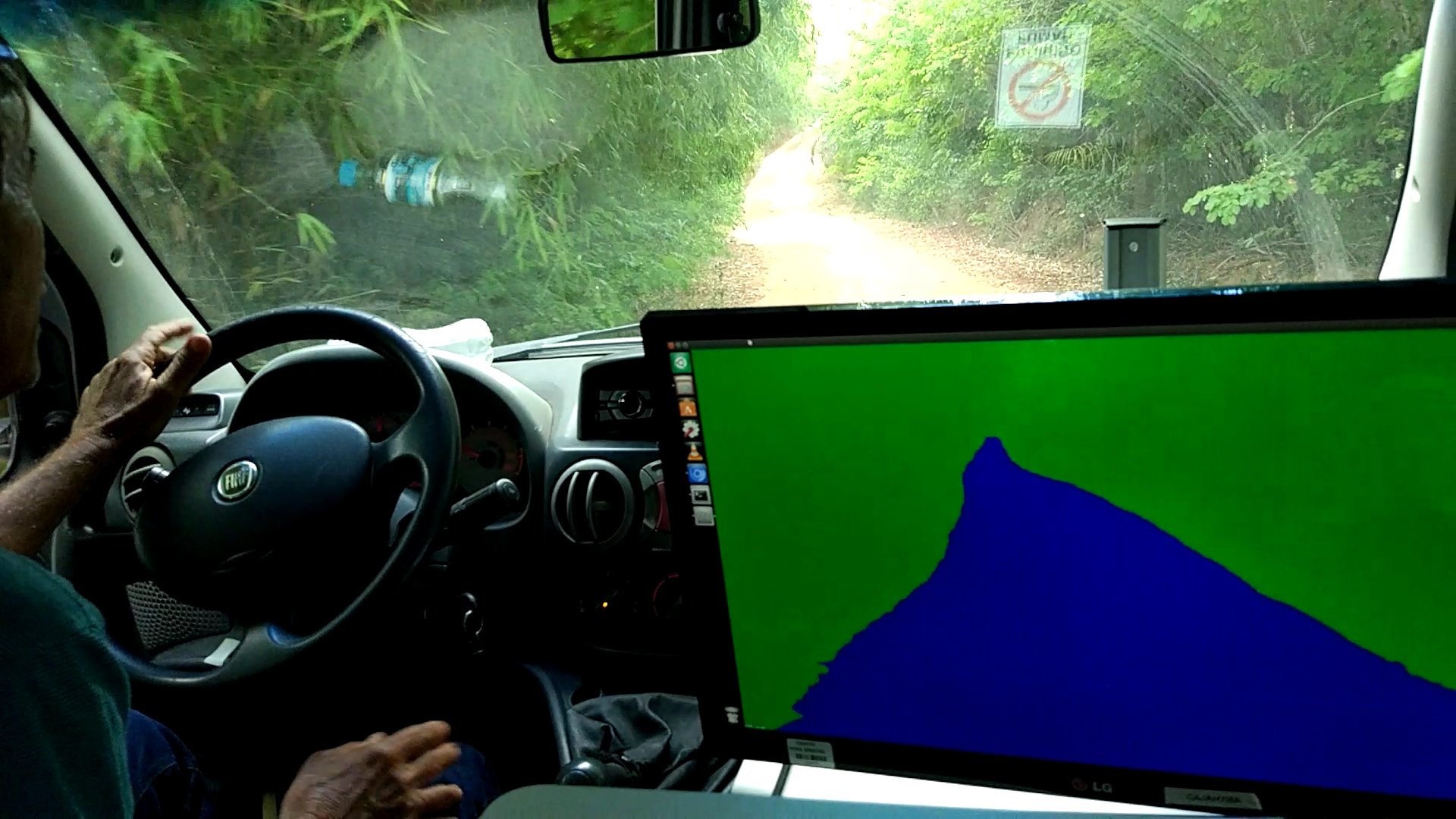}
	\end{subfigure}%
	\vspace{0.005\linewidth}
	
	\begin{subfigure}[b]{0.325\linewidth}
		\includegraphics[width=\linewidth]{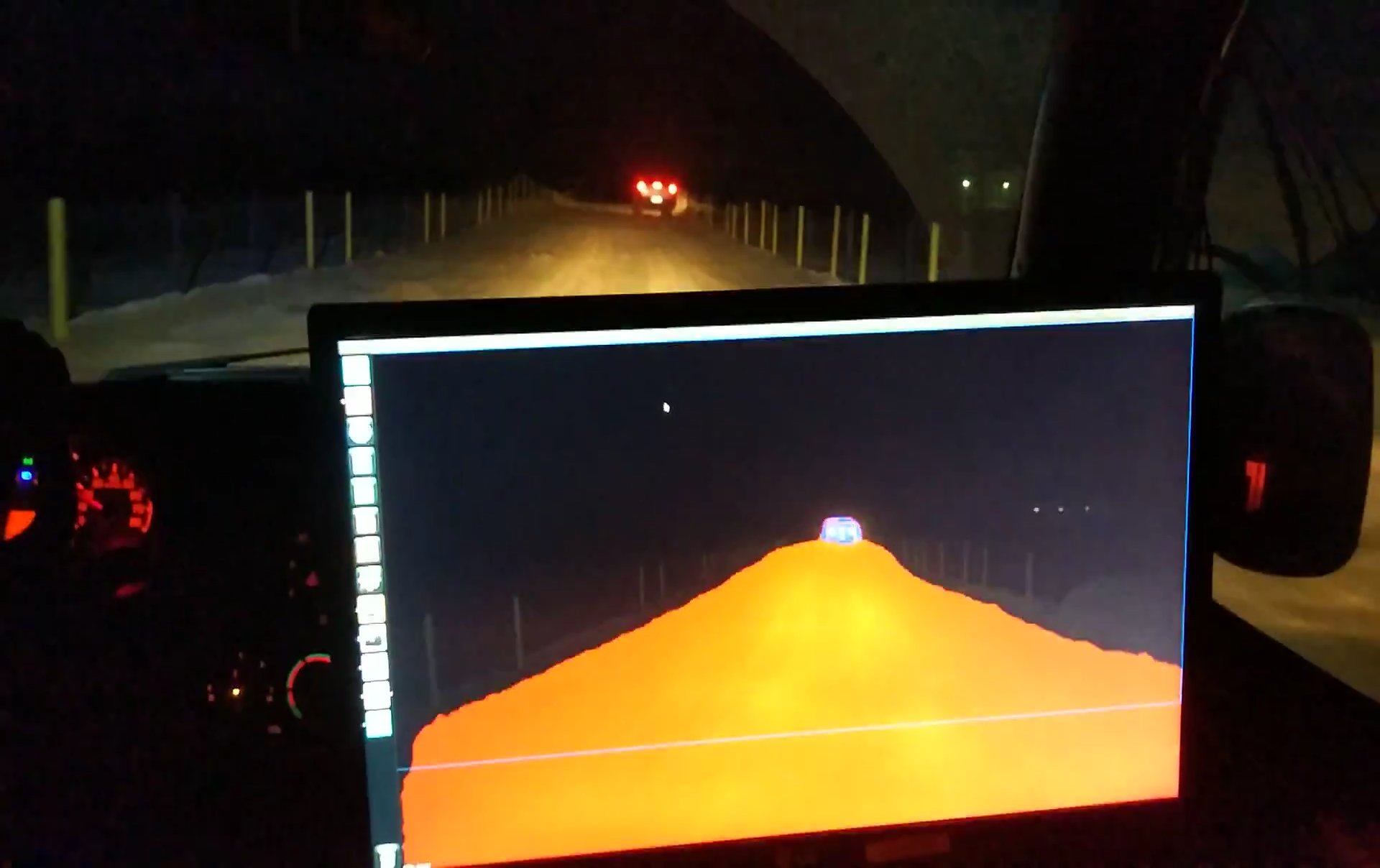}
	\end{subfigure}%
	\hfill
	\begin{subfigure}[b]{0.325\linewidth}
		\includegraphics[width=\linewidth]{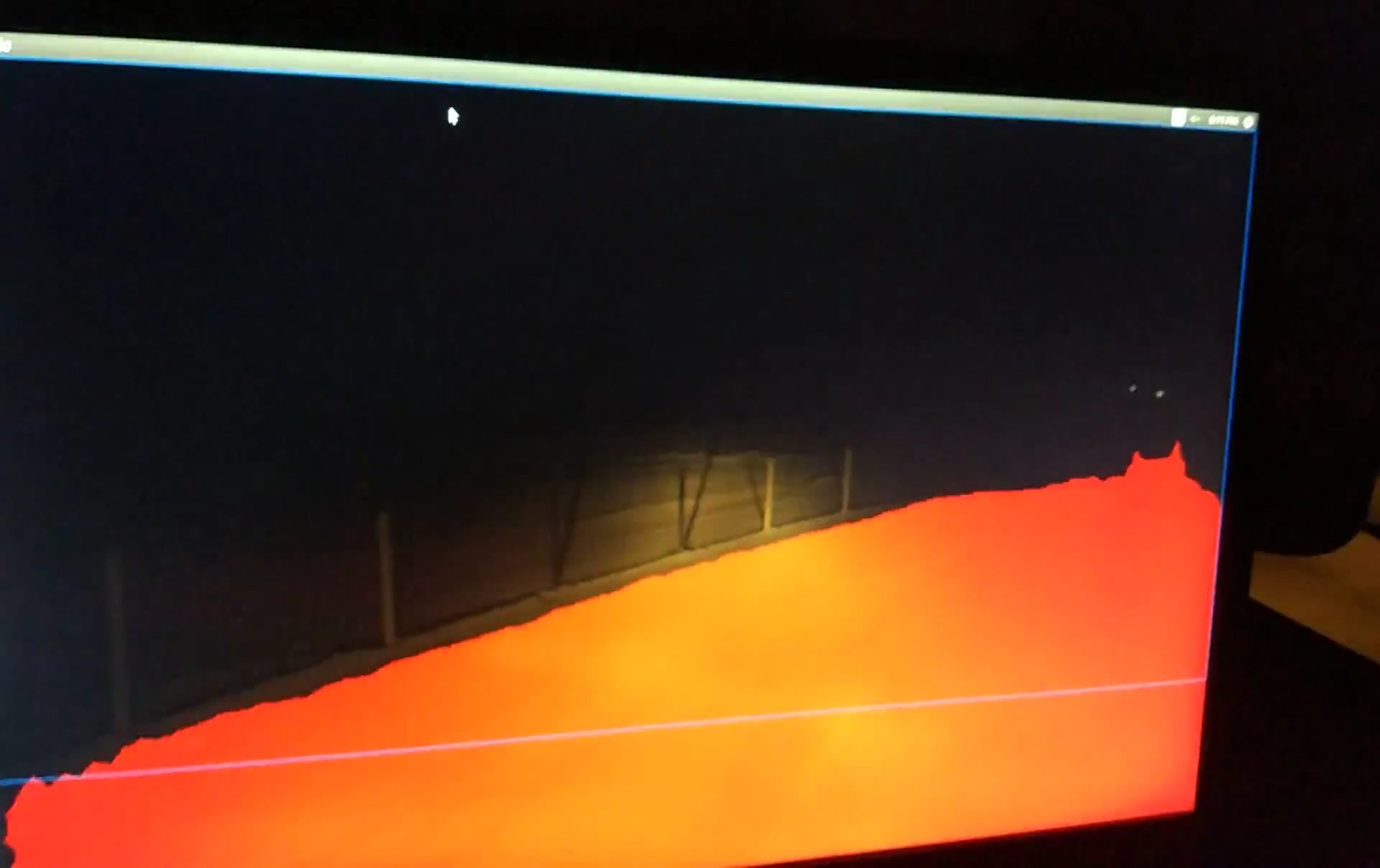}
	\end{subfigure}%
	\hfill
	\begin{subfigure}[b]{0.325\linewidth}
		\includegraphics[width=\linewidth]{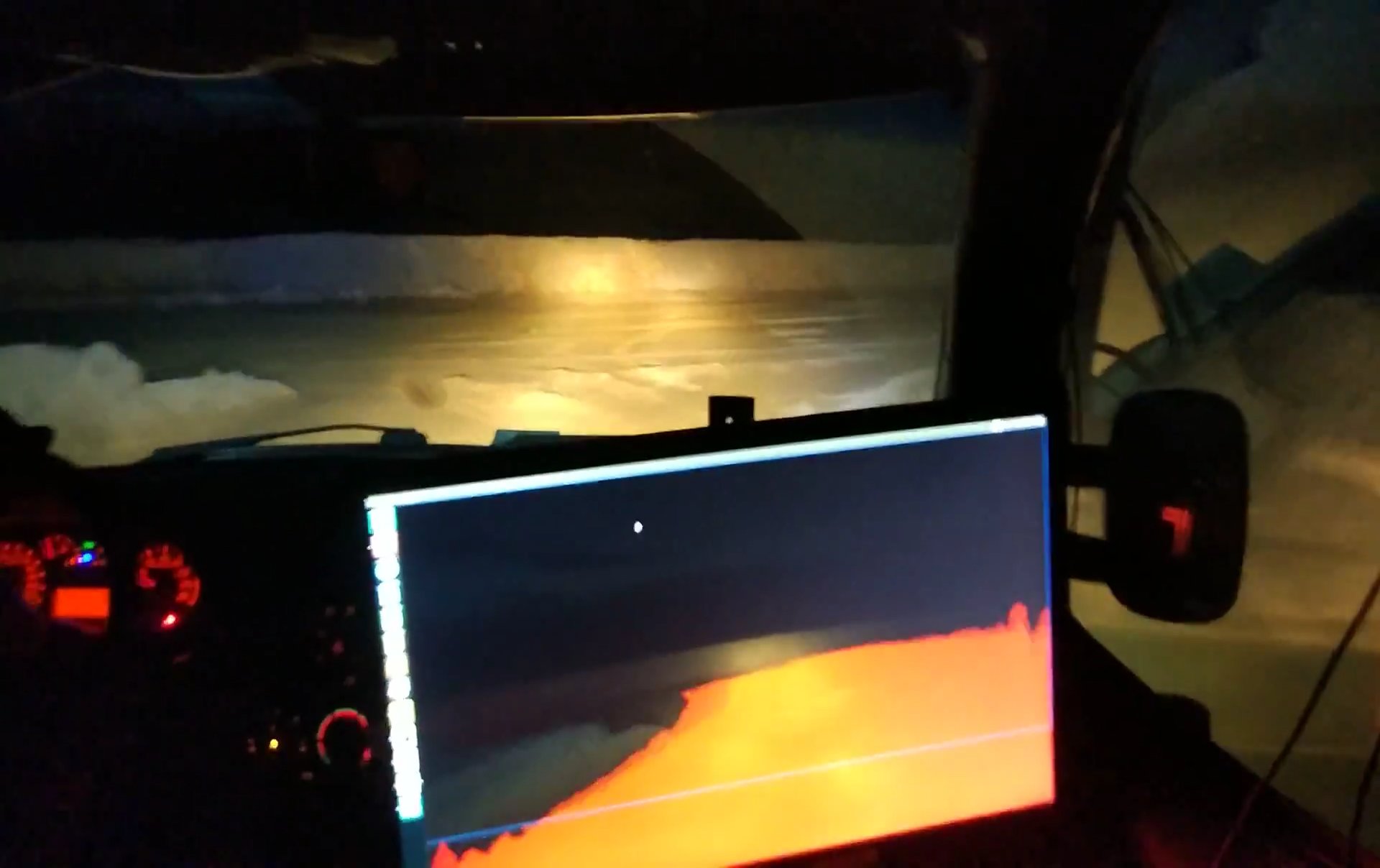}
	\end{subfigure}%

	\caption{Field tests carried out under different visibility conditions.}
	\label{fig:field_tests}
	\fautor
\end{figure}

\autoref{eq:distance-response-21-fps} and \autoref{fig:distance-response-21-fps} show a relationship between the velocity of the vehicle $V_{km/h}$ and distance $D_{m}$ traveled from the moment of the image capture and the processed information delivered. Considering the speed of 30 km/h with inference at 21 FPS on DRIVE PX 2, it is possible to have the information for decision making still 47 ms after the capture or only 39 cm from the event point. Using this approach, the researchers have obtained an acceptable response between what is perceived directly on the road and through the test monitor (\autoref{fig:field_tests}).

\begin{equation} \label{eq:distance-response-21-fps} 
D_{m}= \frac{V_{km/h}}{3.6*FPS}
\end{equation}

\section{Discussion}
\label{sec:discussion}
Considering the current scenario, our proposed Kamino\footnote{\url{https://github.com/Brazilian-Institute-of-Robotics/offroad\_dataset}}  dataset was an important step to help develop this work and answer the main research question. Besides, it should contribute to the research field by enabling future investigations on this kind of environment. There are datasets such as DeepScene~ \cite{Semantic-Forested:Valada:2016} having a small set of images recorded in a forest (136 labeled RGB images), and many others published focusing on the urban environment~ \cite{cordts:2016:cityscapes,complex-urban-dataset-jjeong-2019-ijrr,Sun:2020:CVPR:WaymoOpenDataset}. Also, there are in Brazil some works  \cite{BADUE:2021:IARA,BERRIEL:2017:IARA,BERRIEL:2017:IARA:2,Shinzato:2016:CaRINA} aimed at the urban paved regions or with data not publicly available. However, they didn't fit with the investigation carried out in this research.

The main question guiding this research was how visual perception modeled as a Deep Supervised Learning problem behaves on unpaved roads commonly found in developing countries such as Brazil and off-road industrial environments such as farming or open-pit mining. From the observed results, we saw (Figures \ref{fig:daytime}, \ref{fig:dusty}, \ref{fig:nightly}, \ref{fig:night_dusty}, \ref{fig:daytime2}, \ref{fig:rainy}, \ref{fig:foggy}, \ref{fig:noise}, and \ref{fig:field_tests}) that SL/DL visual perception can help ADAS e AVs to segment obstacles and traffic areas in off-road environments, even in bad visibility conditions. On the other hand, the tests depicted in the \autoref{fig:test_pspnet_deeplab_cityscape_unpaved_roads} and on the \autoref{tab:methods_eval} have shown that algorithms trained for well-paved urban environments~ \cite{Chen:2018:EncoderDecoderWA:ECCV:deeplabv3Plus, Zhao:2017:PSPNet:ieeecvpr, Long:2015:FCN:ieeecvpr} when applied to unpaved roads and off-road environments will not be efficient. The synthesis from those results is that Supervised DL perception proper trained will probably make ADAS and AVs capable operate in an off-road environment under adverse visibility conditions as required by SAE J3016 level 5  \cite{SocietyforAutomotiveEngineers2021}. Nevertheless, it is important to consider this kind of data at the moment of design, training, and test the system.

Unlike other works that have used just specific techniques for segmentation~ \cite{Chen:2018:EncoderDecoderWA:ECCV:deeplabv3Plus, Zhao:2017:PSPNet:ieeecvpr,Real-Time-Semantic-Off-Road:Maturana:2018,Semantic-Forested:Valada:2016,Long:2015:FCN:ieeecvpr}, this research has built a framework (CMSNet\footnote{\url{https://github.com/Brazilian-Institute-of-Robotics/autonomous\_perception}}) to make it possible to test different solutions for the problem. This study shows that the architectures composed by CSMNet could segment the traffic zone and obstacles in such environments quite well. The results showed that ASPP (CM2, CM5, and CM8) and SPP modules (CM1, CM4, and CM7) provide similar accuracy, with ASPP being slightly better (\autoref{tab:summary-testes-com-configuracoes-para-backbone-mobilenetv2}). The output stride controlled by using dilated convolution had a positive impact on the segmentation quality when combined with ASPP (CM2) but also has consequences on computer power demand (CM0, CM1, and CM2 on the \autoref{fig:inference_time}). On the other hand, the shortcut did not result in a relevant improvement in segmentation quality when used with another strategy like a pyramid of pooling and atrous convolution. However, it had increased the computer power demand. The researchers also have seen that the pyramid implemented with conventional convolution increased the number of parameters significantly (CM2, CM5, and CM8 on the \autoref{tab:summary-testes-com-configuracoes-para-backbone-mobilenetv2}) but did not provide relevant improvements on segmentation quality.

Another contribution of this research was to carry out experiments to verify how those DL algorithms have their segmentation capability affected by variations in visibility conditions such as rain, night, dust, fog (synthetic), and noise in different levels of severity. The researcher did not find other works doing such kinds of analysis. Works such ~ \citeonline{Semantic-Forested:Valada:2016} have pointed slightly qualitative observations related to adverse visibility conditions, but not quantitative ones. Other works ~ \cite{Chen:2018:EncoderDecoderWA:ECCV:deeplabv3Plus, Real-Time-Semantic-Off-Road:Maturana:2018,Zhao:2017:PSPNet:ieeecvpr} have not even evaluated those conditions. The test results showed that night and dust impact less the CMSNet segmentation than rain and fog. The researcher also has noted that the output stride of 8 performed better in rainy conditions, and the SPP module with OS 16 performed well in foggy. ASPP with both OS 8 and 16 went well in both visibility conditions. In a nutshell, the configuration with ASPP and OS 8 has shown the best inference quality overall, and the solution with GPP and OS 16 has the best performance in inference time. This result indicates that the visual perception can segment traffic zone and obstacles under a slightly rainy and foggy situation. However, in extreme visibility conditions, the CMSNet segmentation quality has been degraded (Figures  \ref{fig:day_dusty_condition}, \ref{fig:day_night_condition}, \ref{fig:day_night_dusty_condition}, \ref{fig:day_rainy_condition}, \ref{fig:fog_impairment} and \ref{fig:noise_impairment}).

The researcher also has observed that, due to the inherent characteristics of unpaved and off-road environments such as lower movement of vehicles and pedestrians, some classes were rare in the proposed dataset and, in some cases, were not sufficient for the test step. So, the experiments have not considered groups such as bus, motorcycle, animal, and bike. Future works may try to merge our proposed dataset with an urban dataset like Cytescape or another~\cite{cordts:2016:cityscapes,complex-urban-dataset-jjeong-2019-ijrr,Sun:2020:CVPR:WaymoOpenDataset} that includes more samples having these objects (obstacles), so make it significant for training and testing. It will also be interesting to try some innovative cost functions such as Focal Loss or Dice Loss to deal with the class imbalance in the images.

The last step in this work was to test the solution in the field (\autoref{fig:field_tests}). Since such DL algorithms have a high computational cost, the researcher has to convert some CMSNet architecture to make them suitable to embed in the ARM64 platform. The process involved fusing and implementing some layers in C++/CUDA. With this process, it has been achieved a lower standard deviation $ \sigma $ (\autoref{tab:inference_comparison}), which is better for real-time and provides predictability regarding the behavior of the system. The architecture with output stride 16 (CM3-TRT) performed better than OS 8 (CM0-TRT). The optimized version CM3 has triplicated the PFS in the same hardware, and the CM0 almost doubled the performance. It was possible to deliver 21 FPS with the embedded platform and achieve about 100 FPS with a GTX 1080TI. Despite the lower computational power available on the embedded platform, the system's stability has proved satisfactory. The standard deviation $ \sigma $ over the average time to perform each inference cycle was only 0.16\%  Drive PX2, while, in the  GTX 1080TI, that value was about 5\%.

In the proof of concept developed, the results were quite promising. Tests have shown that visual perception (focused on mass-produced sensors such as a camera) performs well in different terrain and visibility conditions. So, it will be possible to develop a very cost-competitive commercial version of the system using new low-cost ARM and RISCV based SoCs containing vector and matrix multiplication accelerators suitable for DL and those mass-production sensors.



\chapter{Conclusion}
\label{chapter:conclusion}
In this work, the researchers have conducted an empirical study applying the Supervised Learning theory with Deep Learning for vision-based perception in such situations and environments low exploited for other studies. Most of the studies regarding perception for ADAS and AVs are focused on the urban well-paved roads. However, our study verified that those trained for well-paved environments don’t perform well in off-road situations. The study has advanced by applying the theory in low exploited conditions and characterized accuracy of the perception inference in adverse visibility situations with different levels of severity.

The researchers have proposed a perception system for AVs and ADAS specialized in unpaved roads and off-road environments. The proposal focused on using Deep Supervised Learning with convolutional neural networks to perform the semantic segmentation of obstacles and areas of traffic on roads where there is no clear distinction between what is or not the track. Besides that, this work also has proposed a new dataset comprising almost 12,000 images exploring various aspects of off-road environments and unpaved roads commonly found in developing countries, including several conditions, such as rainy, nightly, and dusty.

To do that, the researchers have designed and built an off-road test track and assembled a hardware platform, including cameras and other sensors, to enable the appropriate conditions for creating a dataset and conducting tests and validation of the proposed system. The researchers also proposed a CMSNet framework to make it possible to create and test multiples architecture arrangements and find the most efficient ones to solve the segmentation problem in such conditions.  Besides that, some architectures have been ported to embedded hardware, and a backbone for features extraction focused on computational efficiency was selected. 

The results show that visual perception modeled as a Deep Supervised Learning problem, when properly trained, can help ADAS and AVs operate in off-road environments under adverse visibility conditions.  The architectures generated by CMSNet were capable of segment areas of traffic and obstacles with a high degree of accuracy. Architecture using ASSP was the best inference quality in such a situation, but the GPP module was more suitable for real-time embedded applications. With the embedded optimized version, it was possible to achieve about 99 FPS in a desktop GPU and 21 FPS in the ARM platform.

However, in extreme visibility conditions, the segmentation quality has been degraded. Besides, the results have shown that relevant data need to be considered at the moment of design, training, and test the system. The algorithms trained for well-paved urban environments when applied to unpaved roads and off-road environments were not efficient.

Although the proposed visual perception has achieved promising results, the researchers agree that more investigations need to be done before ADAS and AVs operate safely in off-road environments under extreme visibility conditions. Regarding the improvements, it is possible fusing the proposed off-road dataset with urban ones to increase the number of rare classes and make the training and tests more reliable. Maybe, trying some approaches combining thermal cameras and RGB ones can improve the results for extreme conditions. Furthermore, it also is possible to try cost functions for training as Focal Loss and Dice Loss to improve the segmentation of these rare classes. Other suggestions for future development are to try applying transformers-networks for semantic segmentation or even panoptic segmentation.

The developed system was a proof of concept prototyped to allow field tests and research development. The tests performed consisted of a restricted set of situations compared to the real world. Therefore, it is not possible to categorically state that CMSNet is effective in all types of existing non-uniform terrains. However, the results indicate that the approach is promising, so future studies and developments can help create a mature, safe and effective system to drive on all types of roads in the production environment.


\bookmarksetup{startatroot}%

\postextual%


\bibliography{references} 

\glsaddall%
\printglossaries%






 \begin{anexosenv}

\chapter{Annotation code style} 
\label{anexos:annotation-code-style}
	
	\begin{codigo}[caption={Annotation coding example in json.}, label={codigo:exemplo-de-codificacao-de-anotação-em-json}, language=json, breaklines=true, float=h]
		{
			"fillColor": [255, 0, 0, 128],
			"imageData": "image-hash",
			"flags": {},
			"shapes": [
			{"points": [[233,134],[568,78],...,[56,687]],
				"label": "road",},
			{"points": [[345,34],[34,58],...,[543,234]],
				"label": "car-0",},
			{"points": [[235,122],[34,453],...,[56,987]],
				"label": "person-0",},
			{"points": [[346,45],[568,124],...,[234,12]],
				"label": "person-1",}
			],
			"imagePath": "image_name.png",
			"lineColor": [255, 0, 0, 128],
		}
		
	\end{codigo}

 \end{anexosenv}

\end{document}